%% file: main.tex
\documentclass{article}

\PassOptionsToPackage{numbers, compress}{natbib}
\usepackage[main,preprint]{neurips_2026}

\usepackage[utf8]{inputenc}
\usepackage[T1]{fontenc}
\usepackage{hyperref}
\usepackage{url}
\usepackage{booktabs}
\usepackage{amsfonts}
\usepackage{amsmath}
\usepackage{nicefrac}
\usepackage{microtype}
\usepackage{xcolor}
\usepackage{textcomp}
\usepackage{graphicx}
\usepackage{placeins}
\usepackage{tikz}
\usepackage{fontawesome5}
\usetikzlibrary{positioning,arrows.meta,fit,backgrounds}
\definecolor{ablue}{HTML}{2A78D6}

\input{numbers.tex}
\input{facts.tex}
\input{appx_facts.tex}
\input{appx_numbers.tex}

\newcommand{\degC}{\,\textdegree{}C}

\title{Diagnosing the Transfer Gap of a Physics Prior Planner for HVAC Control}

\author{%
  Hironmoy Roy Rudra$^{\dagger}$ \quad
  Md.\ Moshiur Rahman$^{*}$ \quad
  Anik Fouzder$^{*}$ \\
  \textbf{Abu Nafis Mohammod Noor Rohan$^{*}$ \quad
  Nafis Fuyad$^{*}$} \\
  Computer Science and Engineering, BRAC University, Dhaka, Bangladesh \\
  \texttt{\{hironmoy.roy.rudra, md.moshiur.rahman2, anik.fouzder,} \\
  \texttt{mohammod.noor.rohan, nafis.fuyad\}@g.bracu.ac.bd} \\
  {\small $^{\dagger}$Corresponding author. \quad $^{*}$Equal contribution.}
}

\begin{document}

\maketitle

\begin{abstract}
HVAC controllers are usually tuned building by building, which is costly at scale. The Smart Buildings Challenge asks for one policy that runs unchanged on unseen buildings and climates, with no forecast or building identifier. We fit per zone thermal models on generated buildings, freeze their median as a prior and plan over it. On held out buildings the planner beats the official baseline and tuned rules. Its unpaired public score of \F{kaggle/first} was more than twice its held out score and a third above its shipped scenario score. On a harsher stress grid it degrades most in large rooms, where the prior overstates the effect of airflow. Four refinements, three selected on that grid, lower its grid score to \F{stress/s30/zc/score} but its public score only to \F{kaggle/zc}. Public scores instead track the grid's small rooms. A reimplemented public one step inversion scores over a third lower there. On eight scenarios it beats the planner given an approximate perfect model, so a better model alone does not close this gap. We report two simulator inconsistencies and two reward omissions.
\end{abstract}

\section{Introduction}

The NeurIPS 2026 Smart Buildings Challenge~\cite{goldfeder2026smart} asks for one control policy for unseen buildings and climates. Each building is simulated for thirty days at five minute resolution. The policy reads building metadata and live sensors, with no building identifier and no weather forecast. Its run time is capped at ten times that of a bang--bang controller.

The reward penalizes comfort regret with weight \WComfort, energy cost with weight \WEnergy{} and carbon emissions with weight \WCarbon. Comfort regret counts occupied zones only. It grows linearly up to \RegretLinK\,K outside the comfort band and quadratically beyond. The score is the mean absolute reward per step over all scenarios, and lower is better.

Our controller spends its learning budget on a physical model and its run time budget on planning. A per zone thermal model is fitted to many generated buildings. Its population median, the \emph{prior}, is used frozen on every new building. A rule layer handles schedules and safe defaults, and a planner searches actions against the prior. Its public leaderboard score was more than twice our held out estimate. A reimplemented public controller later outperformed it on every scenario of our stress grid over seven days. No single component is new. This paper contributes a measured account of where zero shot transfer failed on our grids and an evaluation design that exposes it. It also states what remains unexplained.

\begin{itemize}
  \item \textbf{Task Measurement:} Where the score of the official baselines goes, and two simulator inconsistencies and two reward omissions (Section~\ref{sec:task}).
  \item \textbf{Population Prior Planner:} A frozen prior planner that beats tuned rules in distribution but trails a one step inversion on the stress grid (Sections~\ref{sec:method} to~\ref{sec:shift}).
  \item \textbf{Transfer Gap Diagnosis:} A stress grid on which the planner degrades most in large rooms while public scores follow its small rooms (Section~\ref{sec:shift}).
  \item \textbf{Controlled Oracle Analyses:} What a perfect forecast and model add to this planner (Section~\ref{sec:oracle}).
\end{itemize}

\section{Related Work}

\textbf{Benchmark:} The challenge~\cite{goldfeder2026smart} builds on the Smart Buildings Control Suite~\cite{goldfeder2024smart}. Its simulator is a calibrated finite difference thermal model of buildings with many zones~\cite{goldfeder2023sim}. Other building control benchmarks include BOPTEST~\cite{blum2021boptest} and Sinergym~\cite{jimenez2021sinergym}.

\textbf{Comparable Competitions:} Winning entries in CityLearn and related competitions were structured controllers with few parameters, tuned by zeroth order search~\cite{khattar2023zoirl,nweye2023citylearn,garmendia2024chesca}.

\textbf{Model Based Control:} Model predictive control (MPC) is the established model based approach for buildings~\cite{drgona2020mpc,oldewurtel2012mpc}. With an adequate model it matches or outperforms model free reinforcement learning~\cite{arroyo2022comparison,arroyo2022rlmpc}.

\textbf{Grey Box Models and Transfer:} Low order resistance and capacitance models identify building heat dynamics from data~\cite{bacher2011}. Hierarchical calibration shares strength across buildings of one archetype~\cite{kristensen2018hierarchical}. Pretrained building models can transfer zero shot~\cite{raisch2026zeroshot,koch2026gems,pinto2022transfer}, and learned estimators can supply MPC parameters~\cite{raisch2026neural}.

\section{Task and Simulator}\label{sec:task}

\textbf{Plant:} Each air handling unit (AHU) mixes return air with outside air, with at most \RecircMaxPct\% recirculation. The AHU only cools, and its supply temperature is the minimum of mixed air and setpoint. Below about \SatInertC\degC{} outside, mixed air is colder than any allowed setpoint, so the setpoint has no effect. The supply sensor still reports it. All heating comes from zone reheat coils fed by a gas fired hot water loop. Fan power is airflow times static pressure over an efficiency of \FanEff. The exhaust fan moves only the outside air share. Cooling electricity is the coil load over a constant coefficient of performance of \CoolCOP. The simulator models sensible heat only, so humid outside air adds no latent load. The reward has no ventilation or freeze term, so an AHU may stay off during occupancy and an empty zone may freeze.

\textbf{Baseline Scores:} We reproduced the four official baselines on the thirty shipped scenarios (Table~\ref{tab:base}). We call the best of them, PID control with a hot water reset, the \emph{official baseline}.

\begin{table}[t]
  \caption{\textbf{Official baselines:} Scores on the thirty shipped scenarios, the sum of the comfort, energy and carbon terms. Lower is better. The first row is the official baseline. A station column is the mean over that station's six scenarios, and Denver is left out for space.}
  \label{tab:base}
  \centering\small
  \begin{tabular}{lrrrrrrrr}
    \toprule
    Controller & Score & Comfort & Energy & Carbon & Fairbanks & Phoenix & Singapore & Atlanta \\
    \midrule
    \TabBase
    \bottomrule
  \end{tabular}
\end{table}

\textbf{Billing Inconsistency:} Boiler gas is billed from the hot water setpoint, while the coils draw on the actual tank temperature. Holding the setpoint at its minimum while the tank is warm therefore lowers the billed cost without changing the delivered heat. On the stress grid, cycling the setpoint this way changes the score of the planner with economizer (Section~\ref{sec:refine}) by \CIpct{tank_vs_econ}. We call an action legitimate when it changes delivered heat, comfort or physical energy use. Gains from the reward omissions of this section pass this criterion but would not satisfy a building operator. No component of our controller targets this inconsistency, and the planner's rebuilt reward counts delivered reheat, not the billed setpoint. The rule layer idles the loop at its lowest setpoint, which may collect part of the credit. With gas billed at the tank temperature, our final configuration changes by \CIpct{zcfix_vs_zc} over seven days. The cycling controller then differs from it by \CIpct{tankfix_vs_zcfix}, so cycling is slightly worse. The inversion changes by \CIpct{s7/invfix_vs_inv} under the same correction, and its advantage over our final configuration becomes \CIpct{s7/invfix_vs_zcfix}.

\textbf{Holiday Inconsistency:} Occupancy follows the federal holiday calendar of the United States at every station, but the comfort schedule has no holidays. Both are simulator inconsistencies, while the missing ventilation and freeze terms are omissions of the reward.

\section{Method}\label{sec:method}

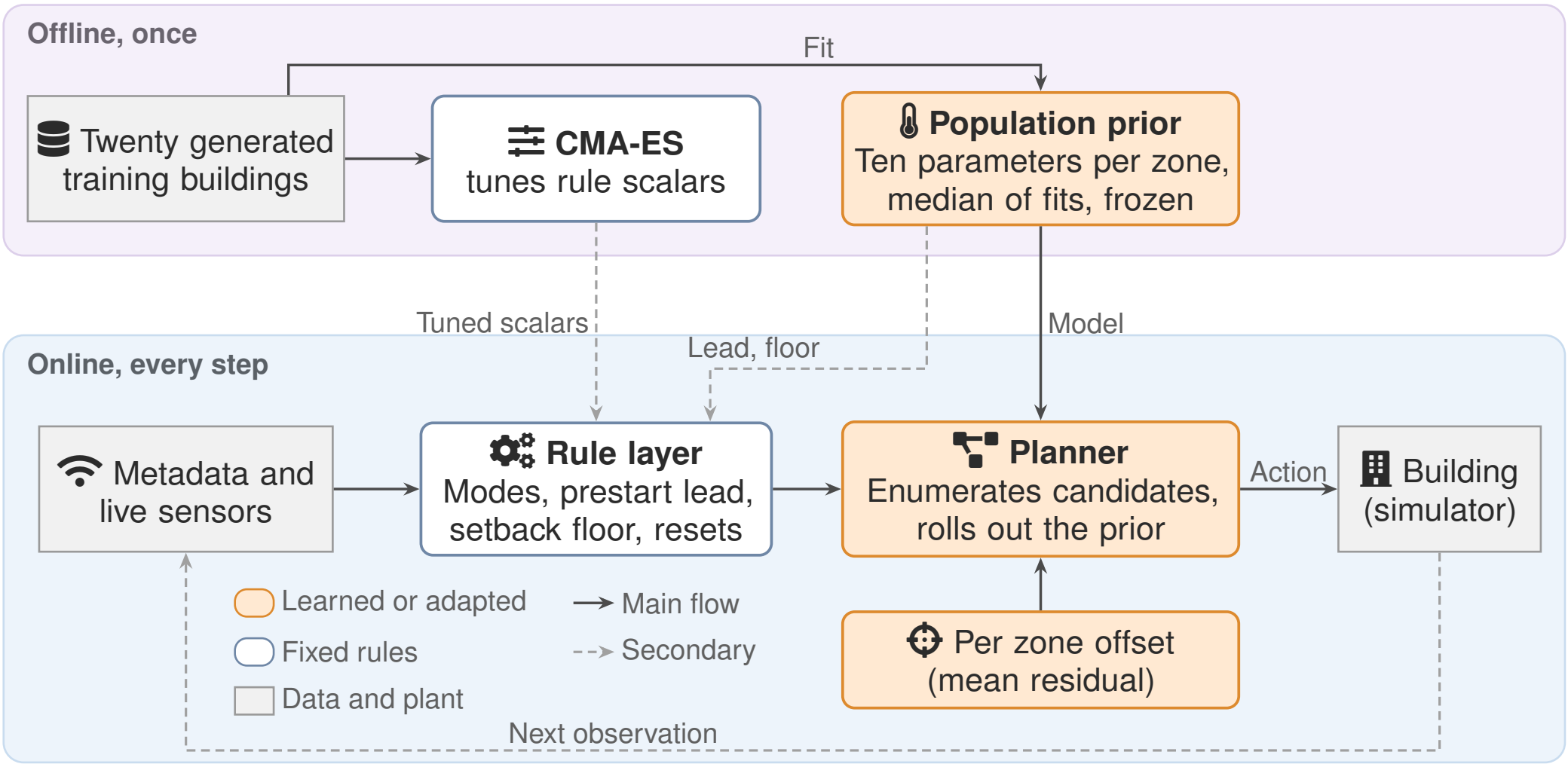
\begin{figure}[t]
  \centering
  \input{figs/overview.tex}
  \caption{\textbf{Method overview:} Offline, a per zone thermal model is fitted to twenty generated buildings and its population median is frozen. Online, the rule layer sets zone modes, a prestart lead and a setback floor. The planner rolls candidates out with the prior plus a per zone offset and applies the first step of the cheapest. Orange boxes are learned or adapted.}
  \label{fig:overview}
\end{figure}

Figure~\ref{fig:overview} shows the three parts of the controller.

\subsection{Population Thermal Prior}\label{sec:prior}

Each zone $i$ follows the model in Equation~\eqref{eq:zone}, which is linear in ten parameters and ignores neighboring zones.
\begin{align}
\Delta T_i = {} & a_i (T_\text{out} - T_i) + k_i (T_\text{m} - T_i) + c_i\, \text{occ} + \mathbf{s}_i^\top \boldsymbol{\phi}(\text{tod}) + d_i \nonumber \\
& + b_i u_i (T_\text{sa} - T_i) + b'_i u'_i (T'_\text{sa} - T_i) \nonumber \\
& + r_i v_i (T_\text{w} - T_i) + r'_i v'_i (T'_\text{w} - T_i) \label{eq:zone}
\end{align}
Here $\Delta T_i$ is the change of the zone temperature over one step. Its first line holds the envelope and a lag term for thermal mass. Occupancy, the time of day and a constant follow. Supply air enters in the second line and reheat in the third.

The airflow $u$ is the damper position times $\sqrt{P/P_0}$, the flow law of the simulator, and is zero when the AHU is off. Reheat is set by the valve position $v$. The supply temperature $T_\text{sa}$ is reconstructed as the minimum of mixed air and setpoint. Primed terms are one step delivery lags. The mass node $T_\text{m}$ is an exponential average of $T_i$ over two hours. The vector $\boldsymbol{\phi}$ holds the sine and cosine of the hour.

The prior is one parameter vector shared by every zone. It is the coordinate wise median of unconstrained ridge least squares fits. The fits cover \NPriorZones{} zones of twenty generated buildings in the shipped size range. Section~\ref{sec:oracle} shows that its reheat terms have the wrong structure. Tables~\ref{tab:main} and~\ref{tab:stress} use this form throughout.

\textbf{Online Identification:} We tested online identification with a Kalman filter. Each zone's ten parameters follow a random walk from the prior and update only on informative steps. A criterion fixed in advance required a \AcceptPct\% lower median error one hour ahead within \AcceptHours{} hours, on twenty held out buildings in unseen weather. The filter did not meet it, and in closed loop it changes the score by \CI{x3c_online_vs_frozen_pool_test} on the held out set. Closed loop data carry little information about ten coupled parameters~\cite{forssell1999closed}, so the prior stays frozen.

\subsection{Rule Layer}

A supervisor gives each zone an occupied, prestart or setback mode from the comfort schedule. Prestart leads come from a rollout of the prior, as in optimal start. The tuned rule controller uses a fixed lead instead. Targets are the band edges inset by a margin, not the band center. Supply air and hot water temperatures follow outdoor resets in place of the trim and respond logic of ASHRAE Guideline~36~\cite{ashrae2021g36,lu2024g36,deshpande2026g36}. CMA-ES~\cite{hansen2016cma} tunes its \TuneParamsRules{} scalar parameters, or \TuneParamsPrior{} under the planner. It runs \TuneGens{} generations of \TunePop{} candidates, a small budget, each scored over thirty days on twenty training buildings.

\subsection{Planner}

Every \ReplanMin{} minutes the planner enumerates candidate actions. The building gets one of three hot water settings. Each AHU is off or on, with a cooling or an idle supply temperature at low or high static pressure. Each zone is idle or receives half or full heating or cooling airflow.

Under the per zone model, zones are independent given the loop and AHU settings. The search therefore picks the best zone actions per AHU setting and the best AHU setting per loop setting. Each candidate is held for \HorizonMin{} minutes and rolled out with the prior at the current outside temperature. The hold equals the horizon, a form of move blocking. Each candidate is scored with the competition reward, rebuilt from the formulas of the simulator. The first step is applied and the search repeats~\cite{nagabandi2018mbrl}.

A horizon of \HorizonMin{} minutes cannot see a morning recovery of several hours. The planner therefore takes two terms from the rule layer as a terminal cost. The comfort band applies from each zone's prestart lead onward. At other unoccupied times, falling below a recoverable setback floor is penalized. Without these terms the planner let zones fall to \CoastLoC--\CoastHiC\degC{} on \NightC\degC{} nights.

Each zone also carries one adapted number, the exponential average over one hour of its one step prediction residuals. This is the disturbance estimate of offset free MPC~\cite{pannocchia2003offset,maeder2009offset}, without its guarantee under model mismatch. Unlike the ten parameters, this number is well conditioned under closed loop data. On the largest building, with \BigZones{} zones and \BigAHUs{} AHUs, the planner needs \RuntimeRatio{} times the run time of bang--bang control on an \TimingCPU{} processor.

\subsection{Refinements of the Planner Cost}\label{sec:refine}

Four refinements were adopted after the first public score. The first was selected on development buildings and the other three on the stress grid of Section~\ref{sec:protocol}. A change was kept if its interval excluded zero. The floor plan gain and the reheat refit of Table~\ref{tab:abl} were measured after the last planner submission.

\textbf{Deeper Setback:} The setback floor is removed and the prestart lead is doubled. On the fresh grid of Section~\ref{sec:protocol}, the lowest empty zone temperature has a median of \N{stress/rl_big/min_unocc/zc/median}\degC{} over scenarios. Its minimum is \N{stress/rl_big/min_unocc/zc/min}\degC. Restoring that floor costs \CIpct{s7/zcfloor_vs_zc} over seven days. A \SbFloorC\degC{} floor changes the planner's score by \CIpct{s7/zcf278_vs_zc} and the inversion's by \CIpct{s7/sb278_vs_inv}. The planner's floor is a cost term, not a limit, so a zone still reaches \N{stress/min_unocc_f278/min}\degC.

\textbf{Size Correction:} The airflow and reheat gains of the prior are halved (Section~\ref{sec:shift}).

\textbf{Economizer:} When outside air is cooler than return air, the planner may lower recirculation if the exhaust fan power costs less than the coil power saved.

\textbf{Zone Cooling Cost:} Each zone action is charged its own share of the cooling coil power. Charging the coil per AHU after the zone actions were chosen had biased zones toward overcooling.

\subsection{Evaluation Protocol}\label{sec:protocol}

The \emph{held out set} has twenty generated buildings in the shipped size range, with weather absent from development. The \emph{shipped set} has the thirty scenarios of the starter kit~\cite{sb2026starterkit}. The \emph{stress grid} has forty eight generated buildings over six weather stations, four seasons and two size ranges. Its stations are Atlanta, Denver, Fairbanks, Minneapolis, Phoenix and Singapore in January, April, July and October. Its size ranges are the shipped range and the default range of the generator, which allows larger rooms. Every building has its own seed, disjoint from training.

The stress grid was built after the first public score revealed the transfer gap. We ran fourteen planner configurations on it over seven days and kept three refinements, so their scores on it are in sample. We therefore repeated the comparison over seven days on a fresh grid with other seeds and weather periods at the same stations. It has \N{stress/rl_big/summ/zc/n} scenarios, because a bookkeeping error dropped two of forty eight before any controller ran. There the three refinements change the score by \CIpct{rlbig/zc_vs_c5}, against \CIpct{s7/zc_vs_c5} on the stress grid over seven days. Against the official baseline, the final configuration changes the score by \CIpct{rlbig/zc_vs_pid_water} on the fresh grid. Over thirty days on the stress grid it changes by \CIpct{s30/zc_vs_pid_water}. The held out set also informed the choice of planner variant, so it is not fully out of sample.

Intervals are paired percentile bootstrap intervals at \CILevel\% over scenarios, with \BootReps{} replicates. They are not adjusted for multiple comparisons. Percent changes divide the mean paired difference by the mean score of the reference, a ratio of means. The public leaderboard covers a rotating third of the hidden scenarios, so public scores are unpaired. They never select a configuration.

\section{Results}

\subsection{In Distribution Evaluation}\label{sec:indist}

The planner with offset, our initial configuration, has the lowest mean score on both in distribution sets (Table~\ref{tab:main}). On the held out set it changes the score by \CI{ship_vs_pid_pool_test} against the official baseline. Against the tuned rule controller the change is \CI{ship_vs_rules_pool_test}. The offset alone contributes \CI{offset_vs_none_pool_test}. Without the offset, the planner changes the score by \CI{rq1_mpc_vs_prior_pool_test} against rules that take their prestart leads from the same prior.

\begin{table}[t]
  \caption{\textbf{In distribution evaluation:} Twenty held out generated buildings in unseen weather on the left and the thirty shipped scenarios on the right, thirty days each. Tuned controllers had the same CMA-ES budget. Bold marks the best score of each block.}
  \label{tab:main}
  \centering\small
  \begin{tabular}{l rrrr rrrr}
    \toprule
    & \multicolumn{4}{c}{Held out buildings ($n=20$)} & \multicolumn{4}{c}{Shipped scenarios ($n=30$)} \\
    \cmidrule(lr){2-5}\cmidrule(lr){6-9}
    Controller & Score & Comfort & Energy & Carbon & Score & Comfort & Energy & Carbon \\
    \midrule
    \TabBoth
    \bottomrule
  \end{tabular}
\end{table}

\begin{figure}[t]
  \centering
  \includegraphics[width=0.8\linewidth]{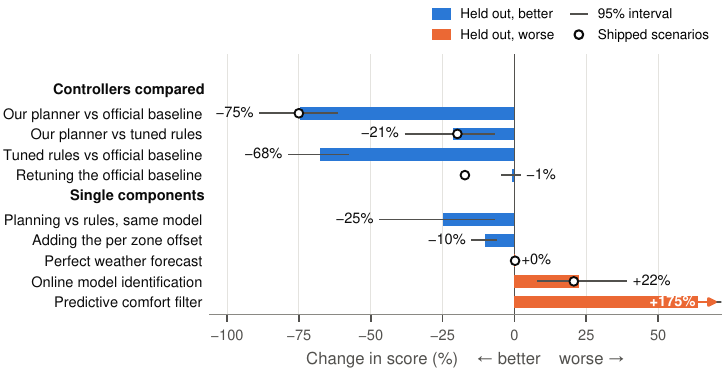}
  \caption{\textbf{Paired differences:} Change in score over thirty days relative to each reference. Bars show the held out buildings with \CILevel\% bootstrap intervals over scenarios, and circles show the shipped scenarios where run. Bars to the left are better, and the comfort filter bar is cut at the axis edge. Planning vs rules, same model uses rules whose prestart leads come from the prior.}
  \label{fig:forest}
\end{figure}

The official baseline was retuned with the same budget over its \TuneParamsPID{} parameters. It changes by \CI{pidtuned_vs_pid_pool_test} on the held out set and by \CI{pidtuned_vs_pid_shipped} on the shipped set. Against the official baseline, the tuned rule controller changes the score by \CI{rules_vs_pid_pool_test} on the held out set. The planner itself was designed by hand, so this equal budget covers tuning only. The tuned rules keep their setback floor, which the final configuration removes. On the stress grid over thirty days, our final configuration changes the score by \CIpct{s30/zc_vs_rules} against the tuned rule controller. On the fresh grid it changes by \CIpct{rlbig/zc_vs_rules}.

\begin{figure}[t]
  \centering
  \includegraphics[width=0.8\linewidth]{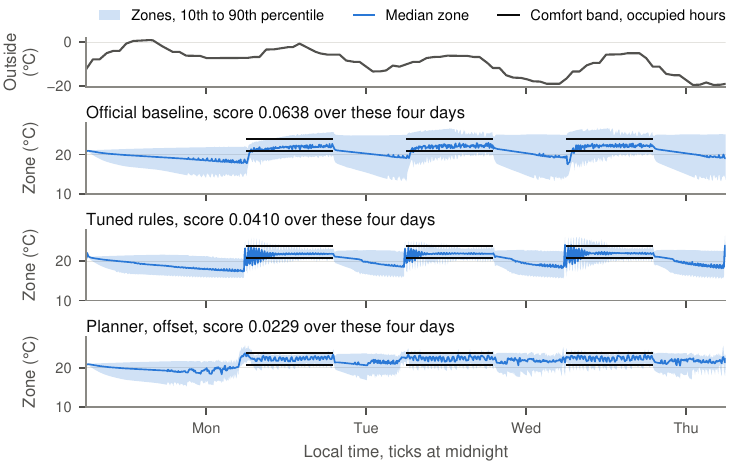}
  \caption{\textbf{Four days in deep cold:} A held out building in Fairbanks with rooms of shipped size. It is occupied on weekdays from \OccStartTime{} to \OccEndTime{} with a band of \BandLoC{} to \BandHiC\degC. The tuned rule controller cycles, and the planner preheats before opening.}
  \label{fig:traj}
\end{figure}

Figure~\ref{fig:traj} shows the three controllers on one building. Baseline comfort regret comes from zones spread across the band, likely because each reheat valve acts on its own zone at fixed AHU settings. The planner also removes cycling and the morning undershoot, because it scores the whole horizon.

\subsection{Transfer Gap and Stress Grid}\label{sec:shift}

The initial planner scored \F{kaggle/first} on the public leaderboard, against \F{pool_test/aether-mpc_off/score} on our held out set and \F{shipped/aether-mpc_off/score} on the shipped scenarios.

On the stress grid, the official baseline scores \F{stress/s30/pid_water/small} on small rooms and \F{stress/s30/pid_water/large} on large rooms. The initial planner scores \F{stress/s30/r1/small} on small and \F{stress/s30/r1/large} on large rooms, so it degrades more with room size than the baseline (Table~\ref{tab:stress}). Tuned rules also degrade, from \F{stress/s30/rules/small} to \F{stress/s30/rules/large}, and score close to the planner on large rooms. These size contrasts compare different buildings and carry no interval. The planner's small room score also exceeds its held out score, as does the baseline's, so weather or buildings explain part of the gap.

Two errors of the prior act in large rooms. Its airflow terms were fitted on small rooms. Every generated variable air volume (VAV) box has the same maximum airflow while capacitance grows with floor area. The fitted terms therefore overstate the effect of airflow in large rooms, where every controller has less airflow per unit of floor. This fixed box size is a property of the generator. The prior's reheat term ignores airflow (Section~\ref{sec:oracle}). Halving all airflow and reheat gains offsets the first error on average. We measured the closed loop error of one hour predictions, predicted minus simulated, on the eight scenarios of Section~\ref{sec:oracle}. In large rooms it is \N{stress/signed_err/large/fitted/heat}\,K in heating and \N{stress/signed_err/large/fitted/cool}\,K in cooling zones. Halving the gains reduces these errors to \N{stress/signed_err/large/halved/heat}\,K and \N{stress/signed_err/large/halved/cool}\,K.

Halving the gains changes the stress grid score by \CIpct{a05_vs_c5} over seven days and by \CIpct{s30/c6_vs_c5} over thirty days. Almost all of the change falls on large rooms, and small rooms worsen from \F{stress/s30/c5/small} to \F{stress/s30/c6/small}. The stress grid estimates \F{stress/s30/r1/score}, \F{stress/s30/c5/score} and \F{stress/s30/zc/score} for the initial, deeper setback and final configurations. Their public scores are \F{kaggle/first}, \F{kaggle/c5} and \F{kaggle/zc}. The grid predicts a far larger gain for the final configuration than the leaderboard shows. The public scores instead track the small room column, at about \PubSmallRatio{} times the small room score for all three. The hidden set thus appears to behave like rooms of shipped size, and the size shift does not explain the public gap.

\begin{table}[t]
  \caption{\textbf{Stress grid:} Forty eight scenarios, thirty days each. From the fourth row on, each row adds one change to the row above. Small and Large give the mean score per size range. Change is the paired change against the row above with its \CILevel\% interval. The last three changes were selected on this grid. The last row is a different controller, paired against the row above.}
  \label{tab:stress}
  \centering\small\setlength{\tabcolsep}{3.2pt}
  \begin{tabular}{lrrrrrrl}
    \toprule
    Controller & Score & Comfort & Energy & Carbon & Small & Large & Change \\
    \midrule
    \TabStress
    \bottomrule
  \end{tabular}
\end{table}

\textbf{One Step Inversion:} A public entrant~\cite{mors2026stopheating}, building on a public notebook by Daniel Wu~\cite{wu2026cpua}, replaces the planner by a one step inversion of an implicit zone model. Each zone receives a response gain from the floor plan, an online bias and the airflow that lands it on the band edge. We reimplemented these mechanisms on our parser and action writer in numpy only, without the entrant's learned residual ensemble and adaptive preheat. Paired against our final configuration over seven days, this controller changes the score by \CIpct{s7/inv_vs_zc} and wins all forty eight scenarios. It wins in small and large rooms alike, and most of its advantage is comfort regret (Table~\ref{tab:stress}). Over thirty days the change is \CIpct{s30/inv_vs_zc}, and on the fresh grid \CIpct{rlbig/inv_vs_zc}. Against the public entrant on the same grid it differs by \CIpct{s7/inv_vs_ctrlw5}. Our submission of the reimplementation scored \F{kaggle/inv} on the public leaderboard, against \F{kaggle/ctrlw5} reported for the entrant. Relative to our planner, it scores about the same fraction on the grid and in public. The release keeps the Apache 2.0 notices of both sources. The per zone gain alone, inside the planner, changes the score by \CIpct{s7/area1_vs_zc} (Table~\ref{tab:abl}). The inversion changes model, targets, actions and search at once, so this comparison does not isolate one of them. Given the small effect of the per zone gain, the comparison suggests that the candidates and long hold of the planner matter. We did not test them separately.

\begin{table}[t]
  \caption{\textbf{Single changes:} Paired change on the stress grid over seven days. The first three rows add changes in order. The inversion row modifies the inversion, and the billing row the planner with economizer. Other rows modify the final configuration.}
  \label{tab:abl}
  \centering\small
  \begin{tabular}{lr}
    \toprule
    Change & Paired change in score \\
    \midrule
    \TabAbl
    \bottomrule
  \end{tabular}
\end{table}

\subsection{Controlled Oracle Analyses}\label{sec:oracle}

\textbf{Weather:} Giving the planner the true future outside temperature changes the score by \CI{x3a_oracle_vs_persistence_pool_test} on the held out set.

\textbf{Perfect Model:} At each planning step the simulator is copied once per candidate, and each copy plays its candidate for \HorizonMin{} minutes from the current state. Each copy applies one zone option to every zone, so heat exchange between zones under differing options is approximated. The rollouts also contain the true future weather and occupancy, while energy cost still comes from the formulas of the planner. Each step needs about \RolloutsPerPlan{} rollouts, so we ran eight hand chosen scenarios that span the stations and sizes. Over seven days, the planner with this approximate perfect model scores \F{stress/bound/sim/score} against \F{stress/bound/model/score} with the prior. The paired change is \CIpct{bound}, with \Nw{stress/bound/ci/wins} of eight scenarios improving. The perfect model removes most of the comfort regret and raises the energy terms. On the same scenarios the inversion of Section~\ref{sec:shift} has an approximate model and no future information. Against the planner with the perfect model it changes the score by \CIpct{bound_inv_sim} and wins all eight. A better model alone thus does not close the gap to the inversion on these scenarios. The planner keeps a cost tuned with the prior, so the oracle may understate what a better model could add. The gain from the perfect model and the further gain of the inversion differ by \CIpct{bound_diff}, relative to the prior planner's score. The interval includes zero, so which step is larger is not established.

The same comparison shows where the prior fails. On a winter morning, full reheat for \HorizonMin{} minutes warms one building by \SimHeatK\,K, while the prior predicts \ModelHeatK\,K. The simulator delivers reheat through supply air heated up to \DischargeMaxC\degC, so the delivered heat scales with airflow. In the prior, reheat scales with valve position only. This test isolates the valve term, which the prior understates. Closed loop heating in Section~\ref{sec:shift} acts mostly through the airflow terms, which it overstates. A refit moves reheat into the airflow terms through the box discharge temperature~\cite{goyal2012model}. It drops the valve term and refits all parameters on the same buildings. It helps only with the halving kept (Table~\ref{tab:abl}). Without the halving, the effects of airflow and reheat are both overstated in large rooms.

\textbf{Predictive Comfort Filter, Not Adopted:} A filter~\cite{wabersich2021psf,alshiekh2018shield,xu2025shield} overrides a zone with full corrective action when a one hour rollout of the prior leaves the band. It changes the score by \CI{x4_guard_pool_test} on the held out set. A filter on a better model was not tested.

\section{Limitations and Recommendations}

The size shift is supported on our grids only. The public scores track the grid's small rooms, so the cause of the public gap is not established. Every generated scenario uses a gas boiler, although the kit simulator also models a heat pump plant.

\textbf{Recommendations to the Organizers:} Gas could be billed at the actual tank temperature (Section~\ref{sec:task}). Occupied zones could need a ventilation minimum~\cite{ashrae2021g36}. On the seven day stress grid, our planner kept the AHU off in \N{stress/ahu_off/planner/pooled}\% of occupied zone steps and the inversion in \N{stress/ahu_off/inv/pooled}\%. The cost of forcing it on was not measured. Empty zones could need a freeze limit~\cite{ashrae2021g36}. The comfort schedule could include holidays. Released training buildings could span the full room size range. A fixed public subset would allow paired comparisons.

\section{Conclusion}

A planner over a frozen population prior outperforms tuned rules in distribution. Its unpaired public score was more than twice our estimate. On our grid it degrades most in large rooms, yet its public scores track the grid's small rooms. Room size thus accounts for much of the grid gap but not the public one. A reimplemented one step inversion~\cite{mors2026stopheating} scores over a third below the planner, and a better model alone does not close this gap on eight scenarios.

\bibliographystyle{plainnat}
\bibliography{refs}

\input{appendix}

\end{document}

%% file: numbers.tex
\providecommand{\pending}{\textcolor{red}{\textbf{[TBD]}}}
\newcommand{\N}[1]{\ifcsname N:#1\endcsname\csname N:#1\endcsname\else\pending\fi}
\newcommand{\F}[1]{\ifcsname F:#1\endcsname\csname F:#1\endcsname\else\pending\fi}
\newcommand{\CI}[1]{\ifcsname CI:#1\endcsname\csname CI:#1\endcsname\else\pending\fi}
\newcommand{\FF}[1]{\ifcsname FF:#1\endcsname\csname FF:#1\endcsname\else\pending\fi}
\newcommand{\CIpct}[1]{\ifcsname CIpct:#1\endcsname\csname CIpct:#1\endcsname\else\pending\fi}
\newcommand{\Nw}[1]{\ifcsname W:#1\endcsname\csname W:#1\endcsname\else\pending\fi}
\expandafter\def\csname N:baselines_shipped/pid_water/n\endcsname{30}
\expandafter\def\csname N:baselines_shipped/pid_water/score\endcsname{0.05496}
\expandafter\def\csname F:baselines_shipped/pid_water/score\endcsname{0.0550}
\expandafter\def\csname FF:baselines_shipped/pid_water/score\endcsname{0.05496}
\expandafter\def\csname N:baselines_shipped/pid_water/comfort\endcsname{0.03489}
\expandafter\def\csname F:baselines_shipped/pid_water/comfort\endcsname{0.0349}
\expandafter\def\csname FF:baselines_shipped/pid_water/comfort\endcsname{0.03489}
\expandafter\def\csname N:baselines_shipped/pid_water/energy\endcsname{0.0115}
\expandafter\def\csname F:baselines_shipped/pid_water/energy\endcsname{0.0115}
\expandafter\def\csname FF:baselines_shipped/pid_water/energy\endcsname{0.01150}
\expandafter\def\csname N:baselines_shipped/pid_water/carbon\endcsname{0.00857}
\expandafter\def\csname F:baselines_shipped/pid_water/carbon\endcsname{0.0086}
\expandafter\def\csname FF:baselines_shipped/pid_water/carbon\endcsname{0.00857}
\expandafter\def\csname N:baselines_shipped/pid_water/by_station/phoenix\endcsname{0.08001}
\expandafter\def\csname F:baselines_shipped/pid_water/by_station/phoenix\endcsname{0.0800}
\expandafter\def\csname FF:baselines_shipped/pid_water/by_station/phoenix\endcsname{0.08001}
\expandafter\def\csname N:baselines_shipped/pid_water/by_station/fairbanks\endcsname{0.07906}
\expandafter\def\csname F:baselines_shipped/pid_water/by_station/fairbanks\endcsname{0.0791}
\expandafter\def\csname FF:baselines_shipped/pid_water/by_station/fairbanks\endcsname{0.07906}
\expandafter\def\csname N:baselines_shipped/pid_water/by_station/denver\endcsname{0.03794}
\expandafter\def\csname F:baselines_shipped/pid_water/by_station/denver\endcsname{0.0379}
\expandafter\def\csname FF:baselines_shipped/pid_water/by_station/denver\endcsname{0.03794}
\expandafter\def\csname N:baselines_shipped/pid_water/by_station/singapore\endcsname{0.04714}
\expandafter\def\csname F:baselines_shipped/pid_water/by_station/singapore\endcsname{0.0471}
\expandafter\def\csname FF:baselines_shipped/pid_water/by_station/singapore\endcsname{0.04714}
\expandafter\def\csname N:baselines_shipped/pid_water/by_station/atlanta\endcsname{0.03064}
\expandafter\def\csname F:baselines_shipped/pid_water/by_station/atlanta\endcsname{0.0306}
\expandafter\def\csname FF:baselines_shipped/pid_water/by_station/atlanta\endcsname{0.03064}
\expandafter\def\csname N:baselines_shipped/pid_water/worst\endcsname{building_train_1/fairbanks_1}
\expandafter\def\csname N:baselines_shipped/pid/n\endcsname{30}
\expandafter\def\csname N:baselines_shipped/pid/score\endcsname{0.06782}
\expandafter\def\csname F:baselines_shipped/pid/score\endcsname{0.0678}
\expandafter\def\csname FF:baselines_shipped/pid/score\endcsname{0.06782}
\expandafter\def\csname N:baselines_shipped/pid/comfort\endcsname{0.04213}
\expandafter\def\csname F:baselines_shipped/pid/comfort\endcsname{0.0421}
\expandafter\def\csname FF:baselines_shipped/pid/comfort\endcsname{0.04213}
\expandafter\def\csname N:baselines_shipped/pid/energy\endcsname{0.01434}
\expandafter\def\csname F:baselines_shipped/pid/energy\endcsname{0.0143}
\expandafter\def\csname FF:baselines_shipped/pid/energy\endcsname{0.01434}
\expandafter\def\csname N:baselines_shipped/pid/carbon\endcsname{0.01135}
\expandafter\def\csname F:baselines_shipped/pid/carbon\endcsname{0.0114}
\expandafter\def\csname FF:baselines_shipped/pid/carbon\endcsname{0.01135}
\expandafter\def\csname N:baselines_shipped/pid/by_station/phoenix\endcsname{0.09546}
\expandafter\def\csname F:baselines_shipped/pid/by_station/phoenix\endcsname{0.0955}
\expandafter\def\csname FF:baselines_shipped/pid/by_station/phoenix\endcsname{0.09546}
\expandafter\def\csname N:baselines_shipped/pid/by_station/fairbanks\endcsname{0.0759}
\expandafter\def\csname F:baselines_shipped/pid/by_station/fairbanks\endcsname{0.0759}
\expandafter\def\csname FF:baselines_shipped/pid/by_station/fairbanks\endcsname{0.07590}
\expandafter\def\csname N:baselines_shipped/pid/by_station/denver\endcsname{0.05064}
\expandafter\def\csname F:baselines_shipped/pid/by_station/denver\endcsname{0.0506}
\expandafter\def\csname FF:baselines_shipped/pid/by_station/denver\endcsname{0.05064}
\expandafter\def\csname N:baselines_shipped/pid/by_station/singapore\endcsname{0.07392}
\expandafter\def\csname F:baselines_shipped/pid/by_station/singapore\endcsname{0.0739}
\expandafter\def\csname FF:baselines_shipped/pid/by_station/singapore\endcsname{0.07392}
\expandafter\def\csname N:baselines_shipped/pid/by_station/atlanta\endcsname{0.04319}
\expandafter\def\csname F:baselines_shipped/pid/by_station/atlanta\endcsname{0.0432}
\expandafter\def\csname FF:baselines_shipped/pid/by_station/atlanta\endcsname{0.04319}
\expandafter\def\csname N:baselines_shipped/pid/worst\endcsname{building_train_8/phoenix_4}
\expandafter\def\csname N:baselines_shipped/fixed_setpoint/n\endcsname{30}
\expandafter\def\csname N:baselines_shipped/fixed_setpoint/score\endcsname{0.0655}
\expandafter\def\csname F:baselines_shipped/fixed_setpoint/score\endcsname{0.0655}
\expandafter\def\csname FF:baselines_shipped/fixed_setpoint/score\endcsname{0.06550}
\expandafter\def\csname N:baselines_shipped/fixed_setpoint/comfort\endcsname{0.03367}
\expandafter\def\csname F:baselines_shipped/fixed_setpoint/comfort\endcsname{0.0337}
\expandafter\def\csname FF:baselines_shipped/fixed_setpoint/comfort\endcsname{0.03367}
\expandafter\def\csname N:baselines_shipped/fixed_setpoint/energy\endcsname{0.01758}
\expandafter\def\csname F:baselines_shipped/fixed_setpoint/energy\endcsname{0.0176}
\expandafter\def\csname FF:baselines_shipped/fixed_setpoint/energy\endcsname{0.01758}
\expandafter\def\csname N:baselines_shipped/fixed_setpoint/carbon\endcsname{0.01425}
\expandafter\def\csname F:baselines_shipped/fixed_setpoint/carbon\endcsname{0.0143}
\expandafter\def\csname FF:baselines_shipped/fixed_setpoint/carbon\endcsname{0.01425}
\expandafter\def\csname N:baselines_shipped/fixed_setpoint/by_station/phoenix\endcsname{0.11615}
\expandafter\def\csname F:baselines_shipped/fixed_setpoint/by_station/phoenix\endcsname{0.1162}
\expandafter\def\csname FF:baselines_shipped/fixed_setpoint/by_station/phoenix\endcsname{0.11615}
\expandafter\def\csname N:baselines_shipped/fixed_setpoint/by_station/fairbanks\endcsname{0.07586}
\expandafter\def\csname F:baselines_shipped/fixed_setpoint/by_station/fairbanks\endcsname{0.0759}
\expandafter\def\csname FF:baselines_shipped/fixed_setpoint/by_station/fairbanks\endcsname{0.07586}
\expandafter\def\csname N:baselines_shipped/fixed_setpoint/by_station/denver\endcsname{0.07721}
\expandafter\def\csname F:baselines_shipped/fixed_setpoint/by_station/denver\endcsname{0.0772}
\expandafter\def\csname FF:baselines_shipped/fixed_setpoint/by_station/denver\endcsname{0.07721}
\expandafter\def\csname N:baselines_shipped/fixed_setpoint/by_station/singapore\endcsname{0.02533}
\expandafter\def\csname F:baselines_shipped/fixed_setpoint/by_station/singapore\endcsname{0.0253}
\expandafter\def\csname FF:baselines_shipped/fixed_setpoint/by_station/singapore\endcsname{0.02533}
\expandafter\def\csname N:baselines_shipped/fixed_setpoint/by_station/atlanta\endcsname{0.03296}
\expandafter\def\csname F:baselines_shipped/fixed_setpoint/by_station/atlanta\endcsname{0.0330}
\expandafter\def\csname FF:baselines_shipped/fixed_setpoint/by_station/atlanta\endcsname{0.03296}
\expandafter\def\csname N:baselines_shipped/fixed_setpoint/worst\endcsname{building_train_8/phoenix_4}
\expandafter\def\csname N:baselines_shipped/bang_bang/n\endcsname{30}
\expandafter\def\csname N:baselines_shipped/bang_bang/score\endcsname{0.22453}
\expandafter\def\csname F:baselines_shipped/bang_bang/score\endcsname{0.2245}
\expandafter\def\csname FF:baselines_shipped/bang_bang/score\endcsname{0.22453}
\expandafter\def\csname N:baselines_shipped/bang_bang/comfort\endcsname{0.11476}
\expandafter\def\csname F:baselines_shipped/bang_bang/comfort\endcsname{0.1148}
\expandafter\def\csname FF:baselines_shipped/bang_bang/comfort\endcsname{0.11476}
\expandafter\def\csname N:baselines_shipped/bang_bang/energy\endcsname{0.06068}
\expandafter\def\csname F:baselines_shipped/bang_bang/energy\endcsname{0.0607}
\expandafter\def\csname FF:baselines_shipped/bang_bang/energy\endcsname{0.06068}
\expandafter\def\csname N:baselines_shipped/bang_bang/carbon\endcsname{0.04909}
\expandafter\def\csname F:baselines_shipped/bang_bang/carbon\endcsname{0.0491}
\expandafter\def\csname FF:baselines_shipped/bang_bang/carbon\endcsname{0.04909}
\expandafter\def\csname N:baselines_shipped/bang_bang/by_station/phoenix\endcsname{0.23673}
\expandafter\def\csname F:baselines_shipped/bang_bang/by_station/phoenix\endcsname{0.2367}
\expandafter\def\csname FF:baselines_shipped/bang_bang/by_station/phoenix\endcsname{0.23673}
\expandafter\def\csname N:baselines_shipped/bang_bang/by_station/fairbanks\endcsname{0.19959}
\expandafter\def\csname F:baselines_shipped/bang_bang/by_station/fairbanks\endcsname{0.1996}
\expandafter\def\csname FF:baselines_shipped/bang_bang/by_station/fairbanks\endcsname{0.19959}
\expandafter\def\csname N:baselines_shipped/bang_bang/by_station/denver\endcsname{0.183}
\expandafter\def\csname F:baselines_shipped/bang_bang/by_station/denver\endcsname{0.1830}
\expandafter\def\csname FF:baselines_shipped/bang_bang/by_station/denver\endcsname{0.18300}
\expandafter\def\csname N:baselines_shipped/bang_bang/by_station/singapore\endcsname{0.27183}
\expandafter\def\csname F:baselines_shipped/bang_bang/by_station/singapore\endcsname{0.2718}
\expandafter\def\csname FF:baselines_shipped/bang_bang/by_station/singapore\endcsname{0.27183}
\expandafter\def\csname N:baselines_shipped/bang_bang/by_station/atlanta\endcsname{0.23151}
\expandafter\def\csname F:baselines_shipped/bang_bang/by_station/atlanta\endcsname{0.2315}
\expandafter\def\csname FF:baselines_shipped/bang_bang/by_station/atlanta\endcsname{0.23151}
\expandafter\def\csname N:baselines_shipped/bang_bang/worst\endcsname{building_train_8/phoenix_4}
\expandafter\def\csname N:subset10/aether-rules/n\endcsname{10}
\expandafter\def\csname W:subset10/aether-rules/n\endcsname{ten}
\expandafter\def\csname N:subset10/aether-rules/score\endcsname{0.03059}
\expandafter\def\csname F:subset10/aether-rules/score\endcsname{0.0306}
\expandafter\def\csname FF:subset10/aether-rules/score\endcsname{0.03059}
\expandafter\def\csname N:subset10/aether-rules/comfort\endcsname{0.00573}
\expandafter\def\csname F:subset10/aether-rules/comfort\endcsname{0.0057}
\expandafter\def\csname FF:subset10/aether-rules/comfort\endcsname{0.00573}
\expandafter\def\csname N:subset10/aether-rules/energy\endcsname{0.01211}
\expandafter\def\csname F:subset10/aether-rules/energy\endcsname{0.0121}
\expandafter\def\csname FF:subset10/aether-rules/energy\endcsname{0.01211}
\expandafter\def\csname N:subset10/aether-rules/carbon\endcsname{0.01275}
\expandafter\def\csname F:subset10/aether-rules/carbon\endcsname{0.0127}
\expandafter\def\csname FF:subset10/aether-rules/carbon\endcsname{0.01275}
\expandafter\def\csname N:subset10/aether-rules/by_station/denver\endcsname{0.019}
\expandafter\def\csname F:subset10/aether-rules/by_station/denver\endcsname{0.0190}
\expandafter\def\csname FF:subset10/aether-rules/by_station/denver\endcsname{0.01900}
\expandafter\def\csname N:subset10/aether-rules/by_station/singapore\endcsname{0.01218}
\expandafter\def\csname F:subset10/aether-rules/by_station/singapore\endcsname{0.0122}
\expandafter\def\csname FF:subset10/aether-rules/by_station/singapore\endcsname{0.01218}
\expandafter\def\csname N:subset10/aether-rules/by_station/phoenix\endcsname{0.02635}
\expandafter\def\csname F:subset10/aether-rules/by_station/phoenix\endcsname{0.0263}
\expandafter\def\csname FF:subset10/aether-rules/by_station/phoenix\endcsname{0.02635}
\expandafter\def\csname N:subset10/aether-rules/by_station/fairbanks\endcsname{0.07654}
\expandafter\def\csname F:subset10/aether-rules/by_station/fairbanks\endcsname{0.0765}
\expandafter\def\csname FF:subset10/aether-rules/by_station/fairbanks\endcsname{0.07654}
\expandafter\def\csname N:subset10/aether-rules/by_station/atlanta\endcsname{0.01887}
\expandafter\def\csname F:subset10/aether-rules/by_station/atlanta\endcsname{0.0189}
\expandafter\def\csname FF:subset10/aether-rules/by_station/atlanta\endcsname{0.01887}
\expandafter\def\csname N:subset10/aether-rules/worst\endcsname{building_train_1/fairbanks_1}
\expandafter\def\csname N:subset10/aether-prior/n\endcsname{9}
\expandafter\def\csname W:subset10/aether-prior/n\endcsname{nine}
\expandafter\def\csname N:subset10/aether-prior/score\endcsname{0.03306}
\expandafter\def\csname F:subset10/aether-prior/score\endcsname{0.0331}
\expandafter\def\csname FF:subset10/aether-prior/score\endcsname{0.03306}
\expandafter\def\csname N:subset10/aether-prior/comfort\endcsname{0.0068}
\expandafter\def\csname F:subset10/aether-prior/comfort\endcsname{0.0068}
\expandafter\def\csname FF:subset10/aether-prior/comfort\endcsname{0.00680}
\expandafter\def\csname N:subset10/aether-prior/energy\endcsname{0.0126}
\expandafter\def\csname F:subset10/aether-prior/energy\endcsname{0.0126}
\expandafter\def\csname FF:subset10/aether-prior/energy\endcsname{0.01260}
\expandafter\def\csname N:subset10/aether-prior/carbon\endcsname{0.01366}
\expandafter\def\csname F:subset10/aether-prior/carbon\endcsname{0.0137}
\expandafter\def\csname FF:subset10/aether-prior/carbon\endcsname{0.01366}
\expandafter\def\csname N:subset10/aether-prior/by_station/denver\endcsname{0.01883}
\expandafter\def\csname F:subset10/aether-prior/by_station/denver\endcsname{0.0188}
\expandafter\def\csname FF:subset10/aether-prior/by_station/denver\endcsname{0.01883}
\expandafter\def\csname N:subset10/aether-prior/by_station/singapore\endcsname{0.01602}
\expandafter\def\csname F:subset10/aether-prior/by_station/singapore\endcsname{0.0160}
\expandafter\def\csname FF:subset10/aether-prior/by_station/singapore\endcsname{0.01602}
\expandafter\def\csname N:subset10/aether-prior/by_station/phoenix\endcsname{0.02629}
\expandafter\def\csname F:subset10/aether-prior/by_station/phoenix\endcsname{0.0263}
\expandafter\def\csname FF:subset10/aether-prior/by_station/phoenix\endcsname{0.02629}
\expandafter\def\csname N:subset10/aether-prior/by_station/fairbanks\endcsname{0.07679}
\expandafter\def\csname F:subset10/aether-prior/by_station/fairbanks\endcsname{0.0768}
\expandafter\def\csname FF:subset10/aether-prior/by_station/fairbanks\endcsname{0.07679}
\expandafter\def\csname N:subset10/aether-prior/by_station/atlanta\endcsname{0.01887}
\expandafter\def\csname F:subset10/aether-prior/by_station/atlanta\endcsname{0.0189}
\expandafter\def\csname FF:subset10/aether-prior/by_station/atlanta\endcsname{0.01887}
\expandafter\def\csname N:subset10/aether-prior/worst\endcsname{building_train_1/fairbanks_1}
\expandafter\def\csname N:subset10/aether-mpc/n\endcsname{10}
\expandafter\def\csname W:subset10/aether-mpc/n\endcsname{ten}
\expandafter\def\csname N:subset10/aether-mpc/score\endcsname{0.02422}
\expandafter\def\csname F:subset10/aether-mpc/score\endcsname{0.0242}
\expandafter\def\csname FF:subset10/aether-mpc/score\endcsname{0.02422}
\expandafter\def\csname N:subset10/aether-mpc/comfort\endcsname{0.01454}
\expandafter\def\csname F:subset10/aether-mpc/comfort\endcsname{0.0145}
\expandafter\def\csname FF:subset10/aether-mpc/comfort\endcsname{0.01454}
\expandafter\def\csname N:subset10/aether-mpc/energy\endcsname{0.00527}
\expandafter\def\csname F:subset10/aether-mpc/energy\endcsname{0.0053}
\expandafter\def\csname FF:subset10/aether-mpc/energy\endcsname{0.00527}
\expandafter\def\csname N:subset10/aether-mpc/carbon\endcsname{0.00441}
\expandafter\def\csname F:subset10/aether-mpc/carbon\endcsname{0.0044}
\expandafter\def\csname FF:subset10/aether-mpc/carbon\endcsname{0.00441}
\expandafter\def\csname N:subset10/aether-mpc/by_station/denver\endcsname{0.00913}
\expandafter\def\csname F:subset10/aether-mpc/by_station/denver\endcsname{0.0091}
\expandafter\def\csname FF:subset10/aether-mpc/by_station/denver\endcsname{0.00913}
\expandafter\def\csname N:subset10/aether-mpc/by_station/singapore\endcsname{0.00756}
\expandafter\def\csname F:subset10/aether-mpc/by_station/singapore\endcsname{0.0076}
\expandafter\def\csname FF:subset10/aether-mpc/by_station/singapore\endcsname{0.00756}
\expandafter\def\csname N:subset10/aether-mpc/by_station/phoenix\endcsname{0.01867}
\expandafter\def\csname F:subset10/aether-mpc/by_station/phoenix\endcsname{0.0187}
\expandafter\def\csname FF:subset10/aether-mpc/by_station/phoenix\endcsname{0.01867}
\expandafter\def\csname N:subset10/aether-mpc/by_station/fairbanks\endcsname{0.07375}
\expandafter\def\csname F:subset10/aether-mpc/by_station/fairbanks\endcsname{0.0737}
\expandafter\def\csname FF:subset10/aether-mpc/by_station/fairbanks\endcsname{0.07375}
\expandafter\def\csname N:subset10/aether-mpc/by_station/atlanta\endcsname{0.01198}
\expandafter\def\csname F:subset10/aether-mpc/by_station/atlanta\endcsname{0.0120}
\expandafter\def\csname FF:subset10/aether-mpc/by_station/atlanta\endcsname{0.01198}
\expandafter\def\csname N:subset10/aether-mpc/worst\endcsname{building_train_1/fairbanks_1}
\expandafter\def\csname N:subset10/pid_water/n\endcsname{10}
\expandafter\def\csname W:subset10/pid_water/n\endcsname{ten}
\expandafter\def\csname N:subset10/pid_water/score\endcsname{0.05941}
\expandafter\def\csname F:subset10/pid_water/score\endcsname{0.0594}
\expandafter\def\csname FF:subset10/pid_water/score\endcsname{0.05941}
\expandafter\def\csname N:subset10_value_proxy/aether-mpc/n\endcsname{10}
\expandafter\def\csname W:subset10_value_proxy/aether-mpc/n\endcsname{ten}
\expandafter\def\csname N:subset10_value_proxy/aether-mpc/score\endcsname{0.01758}
\expandafter\def\csname F:subset10_value_proxy/aether-mpc/score\endcsname{0.0176}
\expandafter\def\csname FF:subset10_value_proxy/aether-mpc/score\endcsname{0.01758}
\expandafter\def\csname N:subset10_value_proxy/aether-mpc/comfort\endcsname{0.00636}
\expandafter\def\csname F:subset10_value_proxy/aether-mpc/comfort\endcsname{0.0064}
\expandafter\def\csname FF:subset10_value_proxy/aether-mpc/comfort\endcsname{0.00636}
\expandafter\def\csname N:subset10_value_proxy/aether-mpc/energy\endcsname{0.00577}
\expandafter\def\csname F:subset10_value_proxy/aether-mpc/energy\endcsname{0.0058}
\expandafter\def\csname FF:subset10_value_proxy/aether-mpc/energy\endcsname{0.00577}
\expandafter\def\csname N:subset10_value_proxy/aether-mpc/carbon\endcsname{0.00545}
\expandafter\def\csname F:subset10_value_proxy/aether-mpc/carbon\endcsname{0.0054}
\expandafter\def\csname FF:subset10_value_proxy/aether-mpc/carbon\endcsname{0.00545}
\expandafter\def\csname N:subset10_value_proxy/aether-mpc/by_station/denver\endcsname{0.00883}
\expandafter\def\csname F:subset10_value_proxy/aether-mpc/by_station/denver\endcsname{0.0088}
\expandafter\def\csname FF:subset10_value_proxy/aether-mpc/by_station/denver\endcsname{0.00883}
\expandafter\def\csname N:subset10_value_proxy/aether-mpc/by_station/singapore\endcsname{0.00759}
\expandafter\def\csname F:subset10_value_proxy/aether-mpc/by_station/singapore\endcsname{0.0076}
\expandafter\def\csname FF:subset10_value_proxy/aether-mpc/by_station/singapore\endcsname{0.00759}
\expandafter\def\csname N:subset10_value_proxy/aether-mpc/by_station/phoenix\endcsname{0.01854}
\expandafter\def\csname F:subset10_value_proxy/aether-mpc/by_station/phoenix\endcsname{0.0185}
\expandafter\def\csname FF:subset10_value_proxy/aether-mpc/by_station/phoenix\endcsname{0.01854}
\expandafter\def\csname N:subset10_value_proxy/aether-mpc/by_station/fairbanks\endcsname{0.04134}
\expandafter\def\csname F:subset10_value_proxy/aether-mpc/by_station/fairbanks\endcsname{0.0413}
\expandafter\def\csname FF:subset10_value_proxy/aether-mpc/by_station/fairbanks\endcsname{0.04134}
\expandafter\def\csname N:subset10_value_proxy/aether-mpc/by_station/atlanta\endcsname{0.01161}
\expandafter\def\csname F:subset10_value_proxy/aether-mpc/by_station/atlanta\endcsname{0.0116}
\expandafter\def\csname FF:subset10_value_proxy/aether-mpc/by_station/atlanta\endcsname{0.01161}
\expandafter\def\csname N:subset10_value_proxy/aether-mpc/worst\endcsname{building_train_1/fairbanks_1}
\expandafter\def\csname N:offset3/aether-mpc_off/n\endcsname{3}
\expandafter\def\csname W:offset3/aether-mpc_off/n\endcsname{three}
\expandafter\def\csname N:offset3/aether-mpc_off/score\endcsname{0.03772}
\expandafter\def\csname F:offset3/aether-mpc_off/score\endcsname{0.0377}
\expandafter\def\csname FF:offset3/aether-mpc_off/score\endcsname{0.03772}
\expandafter\def\csname N:offset3/aether-mpc_off/comfort\endcsname{0.01257}
\expandafter\def\csname F:offset3/aether-mpc_off/comfort\endcsname{0.0126}
\expandafter\def\csname FF:offset3/aether-mpc_off/comfort\endcsname{0.01257}
\expandafter\def\csname N:offset3/aether-mpc_off/energy\endcsname{0.01034}
\expandafter\def\csname F:offset3/aether-mpc_off/energy\endcsname{0.0103}
\expandafter\def\csname FF:offset3/aether-mpc_off/energy\endcsname{0.01034}
\expandafter\def\csname N:offset3/aether-mpc_off/carbon\endcsname{0.01482}
\expandafter\def\csname F:offset3/aether-mpc_off/carbon\endcsname{0.0148}
\expandafter\def\csname FF:offset3/aether-mpc_off/carbon\endcsname{0.01482}
\expandafter\def\csname N:offset3/aether-mpc_off/by_station/atlanta\endcsname{0.01113}
\expandafter\def\csname F:offset3/aether-mpc_off/by_station/atlanta\endcsname{0.0111}
\expandafter\def\csname FF:offset3/aether-mpc_off/by_station/atlanta\endcsname{0.01113}
\expandafter\def\csname N:offset3/aether-mpc_off/by_station/phoenix\endcsname{0.02412}
\expandafter\def\csname F:offset3/aether-mpc_off/by_station/phoenix\endcsname{0.0241}
\expandafter\def\csname FF:offset3/aether-mpc_off/by_station/phoenix\endcsname{0.02412}
\expandafter\def\csname N:offset3/aether-mpc_off/by_station/fairbanks\endcsname{0.07793}
\expandafter\def\csname F:offset3/aether-mpc_off/by_station/fairbanks\endcsname{0.0779}
\expandafter\def\csname FF:offset3/aether-mpc_off/by_station/fairbanks\endcsname{0.07793}
\expandafter\def\csname N:offset3/aether-mpc_off/worst\endcsname{building_train_1/fairbanks_1}
\expandafter\def\csname N:idcheck/prior\endcsname{0.372}
\expandafter\def\csname F:idcheck/prior\endcsname{0.3720}
\expandafter\def\csname FF:idcheck/prior\endcsname{0.37200}
\expandafter\def\csname N:idcheck/online_24h\endcsname{0.393}
\expandafter\def\csname F:idcheck/online_24h\endcsname{0.3930}
\expandafter\def\csname FF:idcheck/online_24h\endcsname{0.39300}
\expandafter\def\csname N:idcheck/online_48h\endcsname{0.401}
\expandafter\def\csname F:idcheck/online_48h\endcsname{0.4010}
\expandafter\def\csname FF:idcheck/online_48h\endcsname{0.40100}
\expandafter\def\csname N:idcheck/online_120h\endcsname{0.383}
\expandafter\def\csname F:idcheck/online_120h\endcsname{0.3830}
\expandafter\def\csname FF:idcheck/online_120h\endcsname{0.38300}
\expandafter\def\csname N:idcheck/trusted_end\endcsname{0.852}
\expandafter\def\csname F:idcheck/trusted_end\endcsname{0.8520}
\expandafter\def\csname FF:idcheck/trusted_end\endcsname{0.85200}
\expandafter\def\csname N:idcheck/improvement_48h_pct\endcsname{$-7.8$}
\expandafter\def\csname F:idcheck/improvement_48h_pct\endcsname{-7.8000}
\expandafter\def\csname FF:idcheck/improvement_48h_pct\endcsname{-7.80000}
\expandafter\def\csname N:idcheck/error_change_48h_pct\endcsname{$+7.8$}
\expandafter\def\csname F:idcheck/error_change_48h_pct\endcsname{7.8000}
\expandafter\def\csname FF:idcheck/error_change_48h_pct\endcsname{7.80000}
\expandafter\def\csname N:stress/s30/pid_water/n\endcsname{48}
\expandafter\def\csname N:stress/s30/pid_water/score\endcsname{0.0508}
\expandafter\def\csname F:stress/s30/pid_water/score\endcsname{0.0508}
\expandafter\def\csname FF:stress/s30/pid_water/score\endcsname{0.05080}
\expandafter\def\csname N:stress/s30/pid_water/comfort\endcsname{0.0352}
\expandafter\def\csname F:stress/s30/pid_water/comfort\endcsname{0.0352}
\expandafter\def\csname FF:stress/s30/pid_water/comfort\endcsname{0.03520}
\expandafter\def\csname N:stress/s30/pid_water/energy\endcsname{0.00937}
\expandafter\def\csname F:stress/s30/pid_water/energy\endcsname{0.0094}
\expandafter\def\csname FF:stress/s30/pid_water/energy\endcsname{0.00937}
\expandafter\def\csname N:stress/s30/pid_water/carbon\endcsname{0.00622}
\expandafter\def\csname F:stress/s30/pid_water/carbon\endcsname{0.0062}
\expandafter\def\csname FF:stress/s30/pid_water/carbon\endcsname{0.00622}
\expandafter\def\csname N:stress/s30/pid_water/small\endcsname{0.04556}
\expandafter\def\csname F:stress/s30/pid_water/small\endcsname{0.0456}
\expandafter\def\csname FF:stress/s30/pid_water/small\endcsname{0.04556}
\expandafter\def\csname N:stress/s30/pid_water/large\endcsname{0.05603}
\expandafter\def\csname F:stress/s30/pid_water/large\endcsname{0.0560}
\expandafter\def\csname FF:stress/s30/pid_water/large\endcsname{0.05603}
\expandafter\def\csname N:stress/s30/rules/n\endcsname{48}
\expandafter\def\csname N:stress/s30/rules/score\endcsname{0.0206}
\expandafter\def\csname F:stress/s30/rules/score\endcsname{0.0206}
\expandafter\def\csname FF:stress/s30/rules/score\endcsname{0.02060}
\expandafter\def\csname N:stress/s30/rules/comfort\endcsname{0.00793}
\expandafter\def\csname F:stress/s30/rules/comfort\endcsname{0.0079}
\expandafter\def\csname FF:stress/s30/rules/comfort\endcsname{0.00793}
\expandafter\def\csname N:stress/s30/rules/energy\endcsname{0.00638}
\expandafter\def\csname F:stress/s30/rules/energy\endcsname{0.0064}
\expandafter\def\csname FF:stress/s30/rules/energy\endcsname{0.00638}
\expandafter\def\csname N:stress/s30/rules/carbon\endcsname{0.00629}
\expandafter\def\csname F:stress/s30/rules/carbon\endcsname{0.0063}
\expandafter\def\csname FF:stress/s30/rules/carbon\endcsname{0.00629}
\expandafter\def\csname N:stress/s30/rules/small\endcsname{0.01514}
\expandafter\def\csname F:stress/s30/rules/small\endcsname{0.0151}
\expandafter\def\csname FF:stress/s30/rules/small\endcsname{0.01514}
\expandafter\def\csname N:stress/s30/rules/large\endcsname{0.02605}
\expandafter\def\csname F:stress/s30/rules/large\endcsname{0.0261}
\expandafter\def\csname FF:stress/s30/rules/large\endcsname{0.02605}
\expandafter\def\csname N:stress/s30/r1/n\endcsname{48}
\expandafter\def\csname N:stress/s30/r1/score\endcsname{0.01889}
\expandafter\def\csname F:stress/s30/r1/score\endcsname{0.0189}
\expandafter\def\csname FF:stress/s30/r1/score\endcsname{0.01889}
\expandafter\def\csname N:stress/s30/r1/comfort\endcsname{0.00823}
\expandafter\def\csname F:stress/s30/r1/comfort\endcsname{0.0082}
\expandafter\def\csname FF:stress/s30/r1/comfort\endcsname{0.00823}
\expandafter\def\csname N:stress/s30/r1/energy\endcsname{0.00588}
\expandafter\def\csname F:stress/s30/r1/energy\endcsname{0.0059}
\expandafter\def\csname FF:stress/s30/r1/energy\endcsname{0.00588}
\expandafter\def\csname N:stress/s30/r1/carbon\endcsname{0.00478}
\expandafter\def\csname F:stress/s30/r1/carbon\endcsname{0.0048}
\expandafter\def\csname FF:stress/s30/r1/carbon\endcsname{0.00478}
\expandafter\def\csname N:stress/s30/r1/small\endcsname{0.01207}
\expandafter\def\csname F:stress/s30/r1/small\endcsname{0.0121}
\expandafter\def\csname FF:stress/s30/r1/small\endcsname{0.01207}
\expandafter\def\csname N:stress/s30/r1/large\endcsname{0.0257}
\expandafter\def\csname F:stress/s30/r1/large\endcsname{0.0257}
\expandafter\def\csname FF:stress/s30/r1/large\endcsname{0.02570}
\expandafter\def\csname N:stress/s30/c5/n\endcsname{48}
\expandafter\def\csname N:stress/s30/c5/score\endcsname{0.01773}
\expandafter\def\csname F:stress/s30/c5/score\endcsname{0.0177}
\expandafter\def\csname FF:stress/s30/c5/score\endcsname{0.01773}
\expandafter\def\csname N:stress/s30/c5/comfort\endcsname{0.00754}
\expandafter\def\csname F:stress/s30/c5/comfort\endcsname{0.0075}
\expandafter\def\csname FF:stress/s30/c5/comfort\endcsname{0.00754}
\expandafter\def\csname N:stress/s30/c5/energy\endcsname{0.00586}
\expandafter\def\csname F:stress/s30/c5/energy\endcsname{0.0059}
\expandafter\def\csname FF:stress/s30/c5/energy\endcsname{0.00586}
\expandafter\def\csname N:stress/s30/c5/carbon\endcsname{0.00434}
\expandafter\def\csname F:stress/s30/c5/carbon\endcsname{0.0043}
\expandafter\def\csname FF:stress/s30/c5/carbon\endcsname{0.00434}
\expandafter\def\csname N:stress/s30/c5/small\endcsname{0.01079}
\expandafter\def\csname F:stress/s30/c5/small\endcsname{0.0108}
\expandafter\def\csname FF:stress/s30/c5/small\endcsname{0.01079}
\expandafter\def\csname N:stress/s30/c5/large\endcsname{0.02467}
\expandafter\def\csname F:stress/s30/c5/large\endcsname{0.0247}
\expandafter\def\csname FF:stress/s30/c5/large\endcsname{0.02467}
\expandafter\def\csname N:stress/s30/c6/n\endcsname{48}
\expandafter\def\csname N:stress/s30/c6/score\endcsname{0.01476}
\expandafter\def\csname F:stress/s30/c6/score\endcsname{0.0148}
\expandafter\def\csname FF:stress/s30/c6/score\endcsname{0.01476}
\expandafter\def\csname N:stress/s30/c6/comfort\endcsname{0.00404}
\expandafter\def\csname F:stress/s30/c6/comfort\endcsname{0.0040}
\expandafter\def\csname FF:stress/s30/c6/comfort\endcsname{0.00404}
\expandafter\def\csname N:stress/s30/c6/energy\endcsname{0.00609}
\expandafter\def\csname F:stress/s30/c6/energy\endcsname{0.0061}
\expandafter\def\csname FF:stress/s30/c6/energy\endcsname{0.00609}
\expandafter\def\csname N:stress/s30/c6/carbon\endcsname{0.00463}
\expandafter\def\csname F:stress/s30/c6/carbon\endcsname{0.0046}
\expandafter\def\csname FF:stress/s30/c6/carbon\endcsname{0.00463}
\expandafter\def\csname N:stress/s30/c6/small\endcsname{0.0111}
\expandafter\def\csname F:stress/s30/c6/small\endcsname{0.0111}
\expandafter\def\csname FF:stress/s30/c6/small\endcsname{0.01110}
\expandafter\def\csname N:stress/s30/c6/large\endcsname{0.01843}
\expandafter\def\csname F:stress/s30/c6/large\endcsname{0.0184}
\expandafter\def\csname FF:stress/s30/c6/large\endcsname{0.01843}
\expandafter\def\csname N:stress/s30/econ/n\endcsname{48}
\expandafter\def\csname N:stress/s30/econ/score\endcsname{0.01434}
\expandafter\def\csname F:stress/s30/econ/score\endcsname{0.0143}
\expandafter\def\csname FF:stress/s30/econ/score\endcsname{0.01434}
\expandafter\def\csname N:stress/s30/econ/comfort\endcsname{0.00402}
\expandafter\def\csname F:stress/s30/econ/comfort\endcsname{0.0040}
\expandafter\def\csname FF:stress/s30/econ/comfort\endcsname{0.00402}
\expandafter\def\csname N:stress/s30/econ/energy\endcsname{0.00576}
\expandafter\def\csname F:stress/s30/econ/energy\endcsname{0.0058}
\expandafter\def\csname FF:stress/s30/econ/energy\endcsname{0.00576}
\expandafter\def\csname N:stress/s30/econ/carbon\endcsname{0.00456}
\expandafter\def\csname F:stress/s30/econ/carbon\endcsname{0.0046}
\expandafter\def\csname FF:stress/s30/econ/carbon\endcsname{0.00456}
\expandafter\def\csname N:stress/s30/econ/small\endcsname{0.01088}
\expandafter\def\csname F:stress/s30/econ/small\endcsname{0.0109}
\expandafter\def\csname FF:stress/s30/econ/small\endcsname{0.01088}
\expandafter\def\csname N:stress/s30/econ/large\endcsname{0.01781}
\expandafter\def\csname F:stress/s30/econ/large\endcsname{0.0178}
\expandafter\def\csname FF:stress/s30/econ/large\endcsname{0.01781}
\expandafter\def\csname N:stress/s30/zc/n\endcsname{48}
\expandafter\def\csname N:stress/s30/zc/score\endcsname{0.01424}
\expandafter\def\csname F:stress/s30/zc/score\endcsname{0.0142}
\expandafter\def\csname FF:stress/s30/zc/score\endcsname{0.01424}
\expandafter\def\csname N:stress/s30/zc/comfort\endcsname{0.00407}
\expandafter\def\csname F:stress/s30/zc/comfort\endcsname{0.0041}
\expandafter\def\csname FF:stress/s30/zc/comfort\endcsname{0.00407}
\expandafter\def\csname N:stress/s30/zc/energy\endcsname{0.00564}
\expandafter\def\csname F:stress/s30/zc/energy\endcsname{0.0056}
\expandafter\def\csname FF:stress/s30/zc/energy\endcsname{0.00564}
\expandafter\def\csname N:stress/s30/zc/carbon\endcsname{0.00453}
\expandafter\def\csname F:stress/s30/zc/carbon\endcsname{0.0045}
\expandafter\def\csname FF:stress/s30/zc/carbon\endcsname{0.00453}
\expandafter\def\csname N:stress/s30/zc/small\endcsname{0.01071}
\expandafter\def\csname F:stress/s30/zc/small\endcsname{0.0107}
\expandafter\def\csname FF:stress/s30/zc/small\endcsname{0.01071}
\expandafter\def\csname N:stress/s30/zc/large\endcsname{0.01778}
\expandafter\def\csname F:stress/s30/zc/large\endcsname{0.0178}
\expandafter\def\csname FF:stress/s30/zc/large\endcsname{0.01778}
\expandafter\def\csname N:stress/s30/hol9zc/n\endcsname{48}
\expandafter\def\csname N:stress/s30/hol9zc/score\endcsname{0.01423}
\expandafter\def\csname F:stress/s30/hol9zc/score\endcsname{0.0142}
\expandafter\def\csname FF:stress/s30/hol9zc/score\endcsname{0.01423}
\expandafter\def\csname N:stress/s30/hol9zc/comfort\endcsname{0.00414}
\expandafter\def\csname F:stress/s30/hol9zc/comfort\endcsname{0.0041}
\expandafter\def\csname FF:stress/s30/hol9zc/comfort\endcsname{0.00414}
\expandafter\def\csname N:stress/s30/hol9zc/energy\endcsname{0.0056}
\expandafter\def\csname F:stress/s30/hol9zc/energy\endcsname{0.0056}
\expandafter\def\csname FF:stress/s30/hol9zc/energy\endcsname{0.00560}
\expandafter\def\csname N:stress/s30/hol9zc/carbon\endcsname{0.00449}
\expandafter\def\csname F:stress/s30/hol9zc/carbon\endcsname{0.0045}
\expandafter\def\csname FF:stress/s30/hol9zc/carbon\endcsname{0.00449}
\expandafter\def\csname N:stress/s30/hol9zc/small\endcsname{0.01066}
\expandafter\def\csname F:stress/s30/hol9zc/small\endcsname{0.0107}
\expandafter\def\csname FF:stress/s30/hol9zc/small\endcsname{0.01066}
\expandafter\def\csname N:stress/s30/hol9zc/large\endcsname{0.01781}
\expandafter\def\csname F:stress/s30/hol9zc/large\endcsname{0.0178}
\expandafter\def\csname FF:stress/s30/hol9zc/large\endcsname{0.01781}
\expandafter\def\csname N:stress/s30/inv/n\endcsname{48}
\expandafter\def\csname N:stress/s30/inv/score\endcsname{0.00901}
\expandafter\def\csname F:stress/s30/inv/score\endcsname{0.0090}
\expandafter\def\csname FF:stress/s30/inv/score\endcsname{0.00901}
\expandafter\def\csname N:stress/s30/inv/comfort\endcsname{0.00051}
\expandafter\def\csname F:stress/s30/inv/comfort\endcsname{0.0005}
\expandafter\def\csname FF:stress/s30/inv/comfort\endcsname{0.00051}
\expandafter\def\csname N:stress/s30/inv/energy\endcsname{0.00478}
\expandafter\def\csname F:stress/s30/inv/energy\endcsname{0.0048}
\expandafter\def\csname FF:stress/s30/inv/energy\endcsname{0.00478}
\expandafter\def\csname N:stress/s30/inv/carbon\endcsname{0.00372}
\expandafter\def\csname F:stress/s30/inv/carbon\endcsname{0.0037}
\expandafter\def\csname FF:stress/s30/inv/carbon\endcsname{0.00372}
\expandafter\def\csname N:stress/s30/inv/small\endcsname{0.00643}
\expandafter\def\csname F:stress/s30/inv/small\endcsname{0.0064}
\expandafter\def\csname FF:stress/s30/inv/small\endcsname{0.00643}
\expandafter\def\csname N:stress/s30/inv/large\endcsname{0.01158}
\expandafter\def\csname F:stress/s30/inv/large\endcsname{0.0116}
\expandafter\def\csname FF:stress/s30/inv/large\endcsname{0.01158}
\expandafter\def\csname N:stress/s30_ci/zc_vs_pid_water/n\endcsname{48}
\expandafter\def\csname N:stress/s30_ci/zc_vs_pid_water/mean\endcsname{$-0.03655$}
\expandafter\def\csname F:stress/s30_ci/zc_vs_pid_water/mean\endcsname{-0.0365}
\expandafter\def\csname FF:stress/s30_ci/zc_vs_pid_water/mean\endcsname{-0.03655}
\expandafter\def\csname N:stress/s30_ci/zc_vs_pid_water/lo\endcsname{$-0.04344$}
\expandafter\def\csname F:stress/s30_ci/zc_vs_pid_water/lo\endcsname{-0.0434}
\expandafter\def\csname FF:stress/s30_ci/zc_vs_pid_water/lo\endcsname{-0.04344}
\expandafter\def\csname N:stress/s30_ci/zc_vs_pid_water/hi\endcsname{$-0.03025$}
\expandafter\def\csname F:stress/s30_ci/zc_vs_pid_water/hi\endcsname{-0.0302}
\expandafter\def\csname FF:stress/s30_ci/zc_vs_pid_water/hi\endcsname{-0.03025}
\expandafter\def\csname N:stress/s30_ci/zc_vs_pid_water/pct\endcsname{$-72.0$}
\expandafter\def\csname F:stress/s30_ci/zc_vs_pid_water/pct\endcsname{-72.0000}
\expandafter\def\csname FF:stress/s30_ci/zc_vs_pid_water/pct\endcsname{-72.00000}
\expandafter\def\csname N:stress/s30_ci/zc_vs_pid_water/pct_lo\endcsname{$-85.5$}
\expandafter\def\csname F:stress/s30_ci/zc_vs_pid_water/pct_lo\endcsname{-85.5000}
\expandafter\def\csname FF:stress/s30_ci/zc_vs_pid_water/pct_lo\endcsname{-85.50000}
\expandafter\def\csname N:stress/s30_ci/zc_vs_pid_water/pct_hi\endcsname{$-59.6$}
\expandafter\def\csname F:stress/s30_ci/zc_vs_pid_water/pct_hi\endcsname{-59.6000}
\expandafter\def\csname FF:stress/s30_ci/zc_vs_pid_water/pct_hi\endcsname{-59.60000}
\expandafter\def\csname N:stress/s30_ci/zc_vs_pid_water/by_station_pct/fairbanks\endcsname{$-72.5$}
\expandafter\def\csname F:stress/s30_ci/zc_vs_pid_water/by_station_pct/fairbanks\endcsname{-72.5000}
\expandafter\def\csname FF:stress/s30_ci/zc_vs_pid_water/by_station_pct/fairbanks\endcsname{-72.50000}
\expandafter\def\csname N:stress/s30_ci/zc_vs_pid_water/by_station_pct/denver\endcsname{$-77.0$}
\expandafter\def\csname F:stress/s30_ci/zc_vs_pid_water/by_station_pct/denver\endcsname{-77.0000}
\expandafter\def\csname FF:stress/s30_ci/zc_vs_pid_water/by_station_pct/denver\endcsname{-77.00000}
\expandafter\def\csname N:stress/s30_ci/zc_vs_pid_water/by_station_pct/atlanta\endcsname{$-77.9$}
\expandafter\def\csname F:stress/s30_ci/zc_vs_pid_water/by_station_pct/atlanta\endcsname{-77.9000}
\expandafter\def\csname FF:stress/s30_ci/zc_vs_pid_water/by_station_pct/atlanta\endcsname{-77.90000}
\expandafter\def\csname N:stress/s30_ci/zc_vs_pid_water/by_station_pct/phoenix\endcsname{$-73.8$}
\expandafter\def\csname F:stress/s30_ci/zc_vs_pid_water/by_station_pct/phoenix\endcsname{-73.8000}
\expandafter\def\csname FF:stress/s30_ci/zc_vs_pid_water/by_station_pct/phoenix\endcsname{-73.80000}
\expandafter\def\csname N:stress/s30_ci/zc_vs_pid_water/by_station_pct/singapore\endcsname{$-71.6$}
\expandafter\def\csname F:stress/s30_ci/zc_vs_pid_water/by_station_pct/singapore\endcsname{-71.6000}
\expandafter\def\csname FF:stress/s30_ci/zc_vs_pid_water/by_station_pct/singapore\endcsname{-71.60000}
\expandafter\def\csname N:stress/s30_ci/zc_vs_pid_water/by_station_pct/minneapolis\endcsname{$-72.1$}
\expandafter\def\csname F:stress/s30_ci/zc_vs_pid_water/by_station_pct/minneapolis\endcsname{-72.1000}
\expandafter\def\csname FF:stress/s30_ci/zc_vs_pid_water/by_station_pct/minneapolis\endcsname{-72.10000}
\expandafter\def\csname N:stress/s30_ci/zc_vs_pid_water/wins\endcsname{48}
\expandafter\def\csname N:stress/s30_ci/zc_vs_rules/n\endcsname{48}
\expandafter\def\csname N:stress/s30_ci/zc_vs_rules/mean\endcsname{$-0.00635$}
\expandafter\def\csname F:stress/s30_ci/zc_vs_rules/mean\endcsname{-0.0063}
\expandafter\def\csname FF:stress/s30_ci/zc_vs_rules/mean\endcsname{-0.00635}
\expandafter\def\csname N:stress/s30_ci/zc_vs_rules/lo\endcsname{$-0.00855$}
\expandafter\def\csname F:stress/s30_ci/zc_vs_rules/lo\endcsname{-0.0086}
\expandafter\def\csname FF:stress/s30_ci/zc_vs_rules/lo\endcsname{-0.00855}
\expandafter\def\csname N:stress/s30_ci/zc_vs_rules/hi\endcsname{$-0.00418$}
\expandafter\def\csname F:stress/s30_ci/zc_vs_rules/hi\endcsname{-0.0042}
\expandafter\def\csname FF:stress/s30_ci/zc_vs_rules/hi\endcsname{-0.00418}
\expandafter\def\csname N:stress/s30_ci/zc_vs_rules/pct\endcsname{$-30.8$}
\expandafter\def\csname F:stress/s30_ci/zc_vs_rules/pct\endcsname{-30.8000}
\expandafter\def\csname FF:stress/s30_ci/zc_vs_rules/pct\endcsname{-30.80000}
\expandafter\def\csname N:stress/s30_ci/zc_vs_rules/pct_lo\endcsname{$-41.5$}
\expandafter\def\csname F:stress/s30_ci/zc_vs_rules/pct_lo\endcsname{-41.5000}
\expandafter\def\csname FF:stress/s30_ci/zc_vs_rules/pct_lo\endcsname{-41.50000}
\expandafter\def\csname N:stress/s30_ci/zc_vs_rules/pct_hi\endcsname{$-20.3$}
\expandafter\def\csname F:stress/s30_ci/zc_vs_rules/pct_hi\endcsname{-20.3000}
\expandafter\def\csname FF:stress/s30_ci/zc_vs_rules/pct_hi\endcsname{-20.30000}
\expandafter\def\csname N:stress/s30_ci/zc_vs_rules/by_station_pct/fairbanks\endcsname{$-16.5$}
\expandafter\def\csname F:stress/s30_ci/zc_vs_rules/by_station_pct/fairbanks\endcsname{-16.5000}
\expandafter\def\csname FF:stress/s30_ci/zc_vs_rules/by_station_pct/fairbanks\endcsname{-16.50000}
\expandafter\def\csname N:stress/s30_ci/zc_vs_rules/by_station_pct/denver\endcsname{$-28.1$}
\expandafter\def\csname F:stress/s30_ci/zc_vs_rules/by_station_pct/denver\endcsname{-28.1000}
\expandafter\def\csname FF:stress/s30_ci/zc_vs_rules/by_station_pct/denver\endcsname{-28.10000}
\expandafter\def\csname N:stress/s30_ci/zc_vs_rules/by_station_pct/atlanta\endcsname{$-31.0$}
\expandafter\def\csname F:stress/s30_ci/zc_vs_rules/by_station_pct/atlanta\endcsname{-31.0000}
\expandafter\def\csname FF:stress/s30_ci/zc_vs_rules/by_station_pct/atlanta\endcsname{-31.00000}
\expandafter\def\csname N:stress/s30_ci/zc_vs_rules/by_station_pct/phoenix\endcsname{$-36.3$}
\expandafter\def\csname F:stress/s30_ci/zc_vs_rules/by_station_pct/phoenix\endcsname{-36.3000}
\expandafter\def\csname FF:stress/s30_ci/zc_vs_rules/by_station_pct/phoenix\endcsname{-36.30000}
\expandafter\def\csname N:stress/s30_ci/zc_vs_rules/by_station_pct/singapore\endcsname{$-16.7$}
\expandafter\def\csname F:stress/s30_ci/zc_vs_rules/by_station_pct/singapore\endcsname{-16.7000}
\expandafter\def\csname FF:stress/s30_ci/zc_vs_rules/by_station_pct/singapore\endcsname{-16.70000}
\expandafter\def\csname N:stress/s30_ci/zc_vs_rules/by_station_pct/minneapolis\endcsname{$-34.7$}
\expandafter\def\csname F:stress/s30_ci/zc_vs_rules/by_station_pct/minneapolis\endcsname{-34.7000}
\expandafter\def\csname FF:stress/s30_ci/zc_vs_rules/by_station_pct/minneapolis\endcsname{-34.70000}
\expandafter\def\csname N:stress/s30_ci/zc_vs_rules/wins\endcsname{41}
\expandafter\def\csname N:stress/s30_ci/zc_vs_r1/n\endcsname{48}
\expandafter\def\csname N:stress/s30_ci/zc_vs_r1/mean\endcsname{$-0.00464$}
\expandafter\def\csname F:stress/s30_ci/zc_vs_r1/mean\endcsname{-0.0046}
\expandafter\def\csname FF:stress/s30_ci/zc_vs_r1/mean\endcsname{-0.00464}
\expandafter\def\csname N:stress/s30_ci/zc_vs_r1/lo\endcsname{$-0.00726$}
\expandafter\def\csname F:stress/s30_ci/zc_vs_r1/lo\endcsname{-0.0073}
\expandafter\def\csname FF:stress/s30_ci/zc_vs_r1/lo\endcsname{-0.00726}
\expandafter\def\csname N:stress/s30_ci/zc_vs_r1/hi\endcsname{$-0.00255$}
\expandafter\def\csname F:stress/s30_ci/zc_vs_r1/hi\endcsname{-0.0026}
\expandafter\def\csname FF:stress/s30_ci/zc_vs_r1/hi\endcsname{-0.00255}
\expandafter\def\csname N:stress/s30_ci/zc_vs_r1/pct\endcsname{$-24.6$}
\expandafter\def\csname F:stress/s30_ci/zc_vs_r1/pct\endcsname{-24.6000}
\expandafter\def\csname FF:stress/s30_ci/zc_vs_r1/pct\endcsname{-24.60000}
\expandafter\def\csname N:stress/s30_ci/zc_vs_r1/pct_lo\endcsname{$-38.4$}
\expandafter\def\csname F:stress/s30_ci/zc_vs_r1/pct_lo\endcsname{-38.4000}
\expandafter\def\csname FF:stress/s30_ci/zc_vs_r1/pct_lo\endcsname{-38.40000}
\expandafter\def\csname N:stress/s30_ci/zc_vs_r1/pct_hi\endcsname{$-13.5$}
\expandafter\def\csname F:stress/s30_ci/zc_vs_r1/pct_hi\endcsname{-13.5000}
\expandafter\def\csname FF:stress/s30_ci/zc_vs_r1/pct_hi\endcsname{-13.50000}
\expandafter\def\csname N:stress/s30_ci/zc_vs_r1/by_station_pct/fairbanks\endcsname{$-9.6$}
\expandafter\def\csname F:stress/s30_ci/zc_vs_r1/by_station_pct/fairbanks\endcsname{-9.6000}
\expandafter\def\csname FF:stress/s30_ci/zc_vs_r1/by_station_pct/fairbanks\endcsname{-9.60000}
\expandafter\def\csname N:stress/s30_ci/zc_vs_r1/by_station_pct/denver\endcsname{$-12.7$}
\expandafter\def\csname F:stress/s30_ci/zc_vs_r1/by_station_pct/denver\endcsname{-12.7000}
\expandafter\def\csname FF:stress/s30_ci/zc_vs_r1/by_station_pct/denver\endcsname{-12.70000}
\expandafter\def\csname N:stress/s30_ci/zc_vs_r1/by_station_pct/atlanta\endcsname{$-24.5$}
\expandafter\def\csname F:stress/s30_ci/zc_vs_r1/by_station_pct/atlanta\endcsname{-24.5000}
\expandafter\def\csname FF:stress/s30_ci/zc_vs_r1/by_station_pct/atlanta\endcsname{-24.50000}
\expandafter\def\csname N:stress/s30_ci/zc_vs_r1/by_station_pct/phoenix\endcsname{$-28.4$}
\expandafter\def\csname F:stress/s30_ci/zc_vs_r1/by_station_pct/phoenix\endcsname{-28.4000}
\expandafter\def\csname FF:stress/s30_ci/zc_vs_r1/by_station_pct/phoenix\endcsname{-28.40000}
\expandafter\def\csname N:stress/s30_ci/zc_vs_r1/by_station_pct/singapore\endcsname{$-19.5$}
\expandafter\def\csname F:stress/s30_ci/zc_vs_r1/by_station_pct/singapore\endcsname{-19.5000}
\expandafter\def\csname FF:stress/s30_ci/zc_vs_r1/by_station_pct/singapore\endcsname{-19.50000}
\expandafter\def\csname N:stress/s30_ci/zc_vs_r1/by_station_pct/minneapolis\endcsname{$-22.0$}
\expandafter\def\csname F:stress/s30_ci/zc_vs_r1/by_station_pct/minneapolis\endcsname{-22.0000}
\expandafter\def\csname FF:stress/s30_ci/zc_vs_r1/by_station_pct/minneapolis\endcsname{-22.00000}
\expandafter\def\csname N:stress/s30_ci/zc_vs_r1/wins\endcsname{46}
\expandafter\def\csname N:stress/s30_ci/c6_vs_c5/n\endcsname{48}
\expandafter\def\csname N:stress/s30_ci/c6_vs_c5/mean\endcsname{$-0.00297$}
\expandafter\def\csname F:stress/s30_ci/c6_vs_c5/mean\endcsname{-0.0030}
\expandafter\def\csname FF:stress/s30_ci/c6_vs_c5/mean\endcsname{-0.00297}
\expandafter\def\csname N:stress/s30_ci/c6_vs_c5/lo\endcsname{$-0.00548$}
\expandafter\def\csname F:stress/s30_ci/c6_vs_c5/lo\endcsname{-0.0055}
\expandafter\def\csname FF:stress/s30_ci/c6_vs_c5/lo\endcsname{-0.00548}
\expandafter\def\csname N:stress/s30_ci/c6_vs_c5/hi\endcsname{$-0.00103$}
\expandafter\def\csname F:stress/s30_ci/c6_vs_c5/hi\endcsname{-0.0010}
\expandafter\def\csname FF:stress/s30_ci/c6_vs_c5/hi\endcsname{-0.00103}
\expandafter\def\csname N:stress/s30_ci/c6_vs_c5/pct\endcsname{$-16.7$}
\expandafter\def\csname F:stress/s30_ci/c6_vs_c5/pct\endcsname{-16.7000}
\expandafter\def\csname FF:stress/s30_ci/c6_vs_c5/pct\endcsname{-16.70000}
\expandafter\def\csname N:stress/s30_ci/c6_vs_c5/pct_lo\endcsname{$-30.9$}
\expandafter\def\csname F:stress/s30_ci/c6_vs_c5/pct_lo\endcsname{-30.9000}
\expandafter\def\csname FF:stress/s30_ci/c6_vs_c5/pct_lo\endcsname{-30.90000}
\expandafter\def\csname N:stress/s30_ci/c6_vs_c5/pct_hi\endcsname{$-5.8$}
\expandafter\def\csname F:stress/s30_ci/c6_vs_c5/pct_hi\endcsname{-5.8000}
\expandafter\def\csname FF:stress/s30_ci/c6_vs_c5/pct_hi\endcsname{-5.80000}
\expandafter\def\csname N:stress/s30_ci/c6_vs_c5/by_station_pct/fairbanks\endcsname{$+4.7$}
\expandafter\def\csname F:stress/s30_ci/c6_vs_c5/by_station_pct/fairbanks\endcsname{4.7000}
\expandafter\def\csname FF:stress/s30_ci/c6_vs_c5/by_station_pct/fairbanks\endcsname{4.70000}
\expandafter\def\csname N:stress/s30_ci/c6_vs_c5/by_station_pct/denver\endcsname{$-0.5$}
\expandafter\def\csname F:stress/s30_ci/c6_vs_c5/by_station_pct/denver\endcsname{-0.5000}
\expandafter\def\csname FF:stress/s30_ci/c6_vs_c5/by_station_pct/denver\endcsname{-0.50000}
\expandafter\def\csname N:stress/s30_ci/c6_vs_c5/by_station_pct/atlanta\endcsname{$-11.7$}
\expandafter\def\csname F:stress/s30_ci/c6_vs_c5/by_station_pct/atlanta\endcsname{-11.7000}
\expandafter\def\csname FF:stress/s30_ci/c6_vs_c5/by_station_pct/atlanta\endcsname{-11.70000}
\expandafter\def\csname N:stress/s30_ci/c6_vs_c5/by_station_pct/phoenix\endcsname{$-19.1$}
\expandafter\def\csname F:stress/s30_ci/c6_vs_c5/by_station_pct/phoenix\endcsname{-19.1000}
\expandafter\def\csname FF:stress/s30_ci/c6_vs_c5/by_station_pct/phoenix\endcsname{-19.10000}
\expandafter\def\csname N:stress/s30_ci/c6_vs_c5/by_station_pct/singapore\endcsname{$-11.0$}
\expandafter\def\csname F:stress/s30_ci/c6_vs_c5/by_station_pct/singapore\endcsname{-11.0000}
\expandafter\def\csname FF:stress/s30_ci/c6_vs_c5/by_station_pct/singapore\endcsname{-11.00000}
\expandafter\def\csname N:stress/s30_ci/c6_vs_c5/by_station_pct/minneapolis\endcsname{$-9.5$}
\expandafter\def\csname F:stress/s30_ci/c6_vs_c5/by_station_pct/minneapolis\endcsname{-9.5000}
\expandafter\def\csname FF:stress/s30_ci/c6_vs_c5/by_station_pct/minneapolis\endcsname{-9.50000}
\expandafter\def\csname N:stress/s30_ci/c6_vs_c5/wins\endcsname{30}
\expandafter\def\csname N:stress/s30_ci/econ_vs_c6/n\endcsname{48}
\expandafter\def\csname N:stress/s30_ci/econ_vs_c6/mean\endcsname{$-0.00042$}
\expandafter\def\csname F:stress/s30_ci/econ_vs_c6/mean\endcsname{-0.0004}
\expandafter\def\csname FF:stress/s30_ci/econ_vs_c6/mean\endcsname{-0.00042}
\expandafter\def\csname N:stress/s30_ci/econ_vs_c6/lo\endcsname{$-0.00059$}
\expandafter\def\csname F:stress/s30_ci/econ_vs_c6/lo\endcsname{-0.0006}
\expandafter\def\csname FF:stress/s30_ci/econ_vs_c6/lo\endcsname{-0.00059}
\expandafter\def\csname N:stress/s30_ci/econ_vs_c6/hi\endcsname{$-0.00027$}
\expandafter\def\csname F:stress/s30_ci/econ_vs_c6/hi\endcsname{-0.0003}
\expandafter\def\csname FF:stress/s30_ci/econ_vs_c6/hi\endcsname{-0.00027}
\expandafter\def\csname N:stress/s30_ci/econ_vs_c6/pct\endcsname{$-2.8$}
\expandafter\def\csname F:stress/s30_ci/econ_vs_c6/pct\endcsname{-2.8000}
\expandafter\def\csname FF:stress/s30_ci/econ_vs_c6/pct\endcsname{-2.80000}
\expandafter\def\csname N:stress/s30_ci/econ_vs_c6/pct_lo\endcsname{$-4.0$}
\expandafter\def\csname F:stress/s30_ci/econ_vs_c6/pct_lo\endcsname{-4.0000}
\expandafter\def\csname FF:stress/s30_ci/econ_vs_c6/pct_lo\endcsname{-4.00000}
\expandafter\def\csname N:stress/s30_ci/econ_vs_c6/pct_hi\endcsname{$-1.8$}
\expandafter\def\csname F:stress/s30_ci/econ_vs_c6/pct_hi\endcsname{-1.8000}
\expandafter\def\csname FF:stress/s30_ci/econ_vs_c6/pct_hi\endcsname{-1.80000}
\expandafter\def\csname N:stress/s30_ci/econ_vs_c6/by_station_pct/fairbanks\endcsname{$-4.1$}
\expandafter\def\csname F:stress/s30_ci/econ_vs_c6/by_station_pct/fairbanks\endcsname{-4.1000}
\expandafter\def\csname FF:stress/s30_ci/econ_vs_c6/by_station_pct/fairbanks\endcsname{-4.10000}
\expandafter\def\csname N:stress/s30_ci/econ_vs_c6/by_station_pct/denver\endcsname{$-5.4$}
\expandafter\def\csname F:stress/s30_ci/econ_vs_c6/by_station_pct/denver\endcsname{-5.4000}
\expandafter\def\csname FF:stress/s30_ci/econ_vs_c6/by_station_pct/denver\endcsname{-5.40000}
\expandafter\def\csname N:stress/s30_ci/econ_vs_c6/by_station_pct/atlanta\endcsname{$-10.5$}
\expandafter\def\csname F:stress/s30_ci/econ_vs_c6/by_station_pct/atlanta\endcsname{-10.5000}
\expandafter\def\csname FF:stress/s30_ci/econ_vs_c6/by_station_pct/atlanta\endcsname{-10.50000}
\expandafter\def\csname N:stress/s30_ci/econ_vs_c6/by_station_pct/phoenix\endcsname{$-5.6$}
\expandafter\def\csname F:stress/s30_ci/econ_vs_c6/by_station_pct/phoenix\endcsname{-5.6000}
\expandafter\def\csname FF:stress/s30_ci/econ_vs_c6/by_station_pct/phoenix\endcsname{-5.60000}
\expandafter\def\csname N:stress/s30_ci/econ_vs_c6/by_station_pct/singapore\endcsname{$+0.1$}
\expandafter\def\csname F:stress/s30_ci/econ_vs_c6/by_station_pct/singapore\endcsname{0.1000}
\expandafter\def\csname FF:stress/s30_ci/econ_vs_c6/by_station_pct/singapore\endcsname{0.10000}
\expandafter\def\csname N:stress/s30_ci/econ_vs_c6/by_station_pct/minneapolis\endcsname{$-6.2$}
\expandafter\def\csname F:stress/s30_ci/econ_vs_c6/by_station_pct/minneapolis\endcsname{-6.2000}
\expandafter\def\csname FF:stress/s30_ci/econ_vs_c6/by_station_pct/minneapolis\endcsname{-6.20000}
\expandafter\def\csname N:stress/s30_ci/econ_vs_c6/wins\endcsname{37}
\expandafter\def\csname N:stress/s30_ci/zc_vs_econ/n\endcsname{48}
\expandafter\def\csname N:stress/s30_ci/zc_vs_econ/mean\endcsname{$-0.0001$}
\expandafter\def\csname F:stress/s30_ci/zc_vs_econ/mean\endcsname{-0.0001}
\expandafter\def\csname FF:stress/s30_ci/zc_vs_econ/mean\endcsname{-0.00010}
\expandafter\def\csname N:stress/s30_ci/zc_vs_econ/lo\endcsname{$-0.00016$}
\expandafter\def\csname F:stress/s30_ci/zc_vs_econ/lo\endcsname{-0.0002}
\expandafter\def\csname FF:stress/s30_ci/zc_vs_econ/lo\endcsname{-0.00016}
\expandafter\def\csname N:stress/s30_ci/zc_vs_econ/hi\endcsname{$-4e-05$}
\expandafter\def\csname F:stress/s30_ci/zc_vs_econ/hi\endcsname{-0.0000}
\expandafter\def\csname FF:stress/s30_ci/zc_vs_econ/hi\endcsname{-0.00004}
\expandafter\def\csname N:stress/s30_ci/zc_vs_econ/pct\endcsname{$-0.7$}
\expandafter\def\csname F:stress/s30_ci/zc_vs_econ/pct\endcsname{-0.7000}
\expandafter\def\csname FF:stress/s30_ci/zc_vs_econ/pct\endcsname{-0.70000}
\expandafter\def\csname N:stress/s30_ci/zc_vs_econ/pct_lo\endcsname{$-1.1$}
\expandafter\def\csname F:stress/s30_ci/zc_vs_econ/pct_lo\endcsname{-1.1000}
\expandafter\def\csname FF:stress/s30_ci/zc_vs_econ/pct_lo\endcsname{-1.10000}
\expandafter\def\csname N:stress/s30_ci/zc_vs_econ/pct_hi\endcsname{$-0.3$}
\expandafter\def\csname F:stress/s30_ci/zc_vs_econ/pct_hi\endcsname{-0.3000}
\expandafter\def\csname FF:stress/s30_ci/zc_vs_econ/pct_hi\endcsname{-0.30000}
\expandafter\def\csname N:stress/s30_ci/zc_vs_econ/by_station_pct/fairbanks\endcsname{$-1.2$}
\expandafter\def\csname F:stress/s30_ci/zc_vs_econ/by_station_pct/fairbanks\endcsname{-1.2000}
\expandafter\def\csname FF:stress/s30_ci/zc_vs_econ/by_station_pct/fairbanks\endcsname{-1.20000}
\expandafter\def\csname N:stress/s30_ci/zc_vs_econ/by_station_pct/denver\endcsname{$-1.9$}
\expandafter\def\csname F:stress/s30_ci/zc_vs_econ/by_station_pct/denver\endcsname{-1.9000}
\expandafter\def\csname FF:stress/s30_ci/zc_vs_econ/by_station_pct/denver\endcsname{-1.90000}
\expandafter\def\csname N:stress/s30_ci/zc_vs_econ/by_station_pct/atlanta\endcsname{$-2.1$}
\expandafter\def\csname F:stress/s30_ci/zc_vs_econ/by_station_pct/atlanta\endcsname{-2.1000}
\expandafter\def\csname FF:stress/s30_ci/zc_vs_econ/by_station_pct/atlanta\endcsname{-2.10000}
\expandafter\def\csname N:stress/s30_ci/zc_vs_econ/by_station_pct/phoenix\endcsname{$-1.5$}
\expandafter\def\csname F:stress/s30_ci/zc_vs_econ/by_station_pct/phoenix\endcsname{-1.5000}
\expandafter\def\csname FF:stress/s30_ci/zc_vs_econ/by_station_pct/phoenix\endcsname{-1.50000}
\expandafter\def\csname N:stress/s30_ci/zc_vs_econ/by_station_pct/singapore\endcsname{$-3.5$}
\expandafter\def\csname F:stress/s30_ci/zc_vs_econ/by_station_pct/singapore\endcsname{-3.5000}
\expandafter\def\csname FF:stress/s30_ci/zc_vs_econ/by_station_pct/singapore\endcsname{-3.50000}
\expandafter\def\csname N:stress/s30_ci/zc_vs_econ/by_station_pct/minneapolis\endcsname{$-0.9$}
\expandafter\def\csname F:stress/s30_ci/zc_vs_econ/by_station_pct/minneapolis\endcsname{-0.9000}
\expandafter\def\csname FF:stress/s30_ci/zc_vs_econ/by_station_pct/minneapolis\endcsname{-0.90000}
\expandafter\def\csname N:stress/s30_ci/zc_vs_econ/wins\endcsname{36}
\expandafter\def\csname N:stress/s30_ci/hol9zc_vs_zc/n\endcsname{48}
\expandafter\def\csname N:stress/s30_ci/hol9zc_vs_zc/mean\endcsname{$-1e-05$}
\expandafter\def\csname F:stress/s30_ci/hol9zc_vs_zc/mean\endcsname{-0.0000}
\expandafter\def\csname FF:stress/s30_ci/hol9zc_vs_zc/mean\endcsname{-0.00001}
\expandafter\def\csname N:stress/s30_ci/hol9zc_vs_zc/lo\endcsname{$-6e-05$}
\expandafter\def\csname F:stress/s30_ci/hol9zc_vs_zc/lo\endcsname{-0.0001}
\expandafter\def\csname FF:stress/s30_ci/hol9zc_vs_zc/lo\endcsname{-0.00006}
\expandafter\def\csname N:stress/s30_ci/hol9zc_vs_zc/hi\endcsname{5e-05}
\expandafter\def\csname F:stress/s30_ci/hol9zc_vs_zc/hi\endcsname{0.0001}
\expandafter\def\csname FF:stress/s30_ci/hol9zc_vs_zc/hi\endcsname{0.00005}
\expandafter\def\csname N:stress/s30_ci/hol9zc_vs_zc/pct\endcsname{$-0.1$}
\expandafter\def\csname F:stress/s30_ci/hol9zc_vs_zc/pct\endcsname{-0.1000}
\expandafter\def\csname FF:stress/s30_ci/hol9zc_vs_zc/pct\endcsname{-0.10000}
\expandafter\def\csname N:stress/s30_ci/hol9zc_vs_zc/pct_lo\endcsname{$-0.4$}
\expandafter\def\csname F:stress/s30_ci/hol9zc_vs_zc/pct_lo\endcsname{-0.4000}
\expandafter\def\csname FF:stress/s30_ci/hol9zc_vs_zc/pct_lo\endcsname{-0.40000}
\expandafter\def\csname N:stress/s30_ci/hol9zc_vs_zc/pct_hi\endcsname{$+0.4$}
\expandafter\def\csname F:stress/s30_ci/hol9zc_vs_zc/pct_hi\endcsname{0.4000}
\expandafter\def\csname FF:stress/s30_ci/hol9zc_vs_zc/pct_hi\endcsname{0.40000}
\expandafter\def\csname N:stress/s30_ci/hol9zc_vs_zc/by_station_pct/fairbanks\endcsname{$-0.9$}
\expandafter\def\csname F:stress/s30_ci/hol9zc_vs_zc/by_station_pct/fairbanks\endcsname{-0.9000}
\expandafter\def\csname FF:stress/s30_ci/hol9zc_vs_zc/by_station_pct/fairbanks\endcsname{-0.90000}
\expandafter\def\csname N:stress/s30_ci/hol9zc_vs_zc/by_station_pct/denver\endcsname{$-0.4$}
\expandafter\def\csname F:stress/s30_ci/hol9zc_vs_zc/by_station_pct/denver\endcsname{-0.4000}
\expandafter\def\csname FF:stress/s30_ci/hol9zc_vs_zc/by_station_pct/denver\endcsname{-0.40000}
\expandafter\def\csname N:stress/s30_ci/hol9zc_vs_zc/by_station_pct/atlanta\endcsname{$-0.5$}
\expandafter\def\csname F:stress/s30_ci/hol9zc_vs_zc/by_station_pct/atlanta\endcsname{-0.5000}
\expandafter\def\csname FF:stress/s30_ci/hol9zc_vs_zc/by_station_pct/atlanta\endcsname{-0.50000}
\expandafter\def\csname N:stress/s30_ci/hol9zc_vs_zc/by_station_pct/phoenix\endcsname{$-0.3$}
\expandafter\def\csname F:stress/s30_ci/hol9zc_vs_zc/by_station_pct/phoenix\endcsname{-0.3000}
\expandafter\def\csname FF:stress/s30_ci/hol9zc_vs_zc/by_station_pct/phoenix\endcsname{-0.30000}
\expandafter\def\csname N:stress/s30_ci/hol9zc_vs_zc/by_station_pct/singapore\endcsname{$-1.0$}
\expandafter\def\csname F:stress/s30_ci/hol9zc_vs_zc/by_station_pct/singapore\endcsname{-1.0000}
\expandafter\def\csname FF:stress/s30_ci/hol9zc_vs_zc/by_station_pct/singapore\endcsname{-1.00000}
\expandafter\def\csname N:stress/s30_ci/hol9zc_vs_zc/by_station_pct/minneapolis\endcsname{$-0.3$}
\expandafter\def\csname F:stress/s30_ci/hol9zc_vs_zc/by_station_pct/minneapolis\endcsname{-0.3000}
\expandafter\def\csname FF:stress/s30_ci/hol9zc_vs_zc/by_station_pct/minneapolis\endcsname{-0.30000}
\expandafter\def\csname N:stress/s30_ci/hol9zc_vs_zc/wins\endcsname{27}
\expandafter\def\csname N:stress/s30_ci/c5_vs_r1/n\endcsname{48}
\expandafter\def\csname N:stress/s30_ci/c5_vs_r1/mean\endcsname{$-0.00116$}
\expandafter\def\csname F:stress/s30_ci/c5_vs_r1/mean\endcsname{-0.0012}
\expandafter\def\csname FF:stress/s30_ci/c5_vs_r1/mean\endcsname{-0.00116}
\expandafter\def\csname N:stress/s30_ci/c5_vs_r1/lo\endcsname{$-0.00172$}
\expandafter\def\csname F:stress/s30_ci/c5_vs_r1/lo\endcsname{-0.0017}
\expandafter\def\csname FF:stress/s30_ci/c5_vs_r1/lo\endcsname{-0.00172}
\expandafter\def\csname N:stress/s30_ci/c5_vs_r1/hi\endcsname{$-0.0006$}
\expandafter\def\csname F:stress/s30_ci/c5_vs_r1/hi\endcsname{-0.0006}
\expandafter\def\csname FF:stress/s30_ci/c5_vs_r1/hi\endcsname{-0.00060}
\expandafter\def\csname N:stress/s30_ci/c5_vs_r1/pct\endcsname{$-6.1$}
\expandafter\def\csname F:stress/s30_ci/c5_vs_r1/pct\endcsname{-6.1000}
\expandafter\def\csname FF:stress/s30_ci/c5_vs_r1/pct\endcsname{-6.10000}
\expandafter\def\csname N:stress/s30_ci/c5_vs_r1/pct_lo\endcsname{$-9.1$}
\expandafter\def\csname F:stress/s30_ci/c5_vs_r1/pct_lo\endcsname{-9.1000}
\expandafter\def\csname FF:stress/s30_ci/c5_vs_r1/pct_lo\endcsname{-9.10000}
\expandafter\def\csname N:stress/s30_ci/c5_vs_r1/pct_hi\endcsname{$-3.2$}
\expandafter\def\csname F:stress/s30_ci/c5_vs_r1/pct_hi\endcsname{-3.2000}
\expandafter\def\csname FF:stress/s30_ci/c5_vs_r1/pct_hi\endcsname{-3.20000}
\expandafter\def\csname N:stress/s30_ci/c5_vs_r1/by_station_pct/fairbanks\endcsname{$-8.1$}
\expandafter\def\csname F:stress/s30_ci/c5_vs_r1/by_station_pct/fairbanks\endcsname{-8.1000}
\expandafter\def\csname FF:stress/s30_ci/c5_vs_r1/by_station_pct/fairbanks\endcsname{-8.10000}
\expandafter\def\csname N:stress/s30_ci/c5_vs_r1/by_station_pct/denver\endcsname{$-5.1$}
\expandafter\def\csname F:stress/s30_ci/c5_vs_r1/by_station_pct/denver\endcsname{-5.1000}
\expandafter\def\csname FF:stress/s30_ci/c5_vs_r1/by_station_pct/denver\endcsname{-5.10000}
\expandafter\def\csname N:stress/s30_ci/c5_vs_r1/by_station_pct/atlanta\endcsname{$-2.9$}
\expandafter\def\csname F:stress/s30_ci/c5_vs_r1/by_station_pct/atlanta\endcsname{-2.9000}
\expandafter\def\csname FF:stress/s30_ci/c5_vs_r1/by_station_pct/atlanta\endcsname{-2.90000}
\expandafter\def\csname N:stress/s30_ci/c5_vs_r1/by_station_pct/phoenix\endcsname{$-4.3$}
\expandafter\def\csname F:stress/s30_ci/c5_vs_r1/by_station_pct/phoenix\endcsname{-4.3000}
\expandafter\def\csname FF:stress/s30_ci/c5_vs_r1/by_station_pct/phoenix\endcsname{-4.30000}
\expandafter\def\csname N:stress/s30_ci/c5_vs_r1/by_station_pct/singapore\endcsname{$-6.3$}
\expandafter\def\csname F:stress/s30_ci/c5_vs_r1/by_station_pct/singapore\endcsname{-6.3000}
\expandafter\def\csname FF:stress/s30_ci/c5_vs_r1/by_station_pct/singapore\endcsname{-6.30000}
\expandafter\def\csname N:stress/s30_ci/c5_vs_r1/by_station_pct/minneapolis\endcsname{$-7.5$}
\expandafter\def\csname F:stress/s30_ci/c5_vs_r1/by_station_pct/minneapolis\endcsname{-7.5000}
\expandafter\def\csname FF:stress/s30_ci/c5_vs_r1/by_station_pct/minneapolis\endcsname{-7.50000}
\expandafter\def\csname N:stress/s30_ci/c5_vs_r1/wins\endcsname{38}
\expandafter\def\csname N:stress/s30_ci/inv_vs_zc/n\endcsname{48}
\expandafter\def\csname N:stress/s30_ci/inv_vs_zc/mean\endcsname{$-0.00524$}
\expandafter\def\csname F:stress/s30_ci/inv_vs_zc/mean\endcsname{-0.0052}
\expandafter\def\csname FF:stress/s30_ci/inv_vs_zc/mean\endcsname{-0.00524}
\expandafter\def\csname N:stress/s30_ci/inv_vs_zc/lo\endcsname{$-0.00703$}
\expandafter\def\csname F:stress/s30_ci/inv_vs_zc/lo\endcsname{-0.0070}
\expandafter\def\csname FF:stress/s30_ci/inv_vs_zc/lo\endcsname{-0.00703}
\expandafter\def\csname N:stress/s30_ci/inv_vs_zc/hi\endcsname{$-0.00363$}
\expandafter\def\csname F:stress/s30_ci/inv_vs_zc/hi\endcsname{-0.0036}
\expandafter\def\csname FF:stress/s30_ci/inv_vs_zc/hi\endcsname{-0.00363}
\expandafter\def\csname N:stress/s30_ci/inv_vs_zc/pct\endcsname{$-36.8$}
\expandafter\def\csname F:stress/s30_ci/inv_vs_zc/pct\endcsname{-36.8000}
\expandafter\def\csname FF:stress/s30_ci/inv_vs_zc/pct\endcsname{-36.80000}
\expandafter\def\csname N:stress/s30_ci/inv_vs_zc/pct_lo\endcsname{$-49.3$}
\expandafter\def\csname F:stress/s30_ci/inv_vs_zc/pct_lo\endcsname{-49.3000}
\expandafter\def\csname FF:stress/s30_ci/inv_vs_zc/pct_lo\endcsname{-49.30000}
\expandafter\def\csname N:stress/s30_ci/inv_vs_zc/pct_hi\endcsname{$-25.5$}
\expandafter\def\csname F:stress/s30_ci/inv_vs_zc/pct_hi\endcsname{-25.5000}
\expandafter\def\csname FF:stress/s30_ci/inv_vs_zc/pct_hi\endcsname{-25.50000}
\expandafter\def\csname N:stress/s30_ci/inv_vs_zc/by_station_pct/fairbanks\endcsname{$-43.9$}
\expandafter\def\csname F:stress/s30_ci/inv_vs_zc/by_station_pct/fairbanks\endcsname{-43.9000}
\expandafter\def\csname FF:stress/s30_ci/inv_vs_zc/by_station_pct/fairbanks\endcsname{-43.90000}
\expandafter\def\csname N:stress/s30_ci/inv_vs_zc/by_station_pct/denver\endcsname{$-43.8$}
\expandafter\def\csname F:stress/s30_ci/inv_vs_zc/by_station_pct/denver\endcsname{-43.8000}
\expandafter\def\csname FF:stress/s30_ci/inv_vs_zc/by_station_pct/denver\endcsname{-43.80000}
\expandafter\def\csname N:stress/s30_ci/inv_vs_zc/by_station_pct/atlanta\endcsname{$-31.1$}
\expandafter\def\csname F:stress/s30_ci/inv_vs_zc/by_station_pct/atlanta\endcsname{-31.1000}
\expandafter\def\csname FF:stress/s30_ci/inv_vs_zc/by_station_pct/atlanta\endcsname{-31.10000}
\expandafter\def\csname N:stress/s30_ci/inv_vs_zc/by_station_pct/phoenix\endcsname{$-27.4$}
\expandafter\def\csname F:stress/s30_ci/inv_vs_zc/by_station_pct/phoenix\endcsname{-27.4000}
\expandafter\def\csname FF:stress/s30_ci/inv_vs_zc/by_station_pct/phoenix\endcsname{-27.40000}
\expandafter\def\csname N:stress/s30_ci/inv_vs_zc/by_station_pct/singapore\endcsname{$-21.2$}
\expandafter\def\csname F:stress/s30_ci/inv_vs_zc/by_station_pct/singapore\endcsname{-21.2000}
\expandafter\def\csname FF:stress/s30_ci/inv_vs_zc/by_station_pct/singapore\endcsname{-21.20000}
\expandafter\def\csname N:stress/s30_ci/inv_vs_zc/by_station_pct/minneapolis\endcsname{$-39.7$}
\expandafter\def\csname F:stress/s30_ci/inv_vs_zc/by_station_pct/minneapolis\endcsname{-39.7000}
\expandafter\def\csname FF:stress/s30_ci/inv_vs_zc/by_station_pct/minneapolis\endcsname{-39.70000}
\expandafter\def\csname N:stress/s30_ci/inv_vs_zc/wins\endcsname{48}
\expandafter\def\csname N:stress/s7/c5/n\endcsname{48}
\expandafter\def\csname N:stress/s7/c5/score\endcsname{0.01591}
\expandafter\def\csname F:stress/s7/c5/score\endcsname{0.0159}
\expandafter\def\csname FF:stress/s7/c5/score\endcsname{0.01591}
\expandafter\def\csname N:stress/s7/c5/comfort\endcsname{0.00639}
\expandafter\def\csname F:stress/s7/c5/comfort\endcsname{0.0064}
\expandafter\def\csname FF:stress/s7/c5/comfort\endcsname{0.00639}
\expandafter\def\csname N:stress/s7/c5/energy\endcsname{0.00549}
\expandafter\def\csname F:stress/s7/c5/energy\endcsname{0.0055}
\expandafter\def\csname FF:stress/s7/c5/energy\endcsname{0.00549}
\expandafter\def\csname N:stress/s7/c5/carbon\endcsname{0.00403}
\expandafter\def\csname F:stress/s7/c5/carbon\endcsname{0.0040}
\expandafter\def\csname FF:stress/s7/c5/carbon\endcsname{0.00403}
\expandafter\def\csname N:stress/s7/c5/small\endcsname{0.0101}
\expandafter\def\csname F:stress/s7/c5/small\endcsname{0.0101}
\expandafter\def\csname FF:stress/s7/c5/small\endcsname{0.01010}
\expandafter\def\csname N:stress/s7/c5/large\endcsname{0.02173}
\expandafter\def\csname F:stress/s7/c5/large\endcsname{0.0217}
\expandafter\def\csname FF:stress/s7/c5/large\endcsname{0.02173}
\expandafter\def\csname N:stress/s7/a05/n\endcsname{48}
\expandafter\def\csname N:stress/s7/a05/score\endcsname{0.01349}
\expandafter\def\csname F:stress/s7/a05/score\endcsname{0.0135}
\expandafter\def\csname FF:stress/s7/a05/score\endcsname{0.01349}
\expandafter\def\csname N:stress/s7/a05/comfort\endcsname{0.0035}
\expandafter\def\csname F:stress/s7/a05/comfort\endcsname{0.0035}
\expandafter\def\csname FF:stress/s7/a05/comfort\endcsname{0.00350}
\expandafter\def\csname N:stress/s7/a05/energy\endcsname{0.00571}
\expandafter\def\csname F:stress/s7/a05/energy\endcsname{0.0057}
\expandafter\def\csname FF:stress/s7/a05/energy\endcsname{0.00571}
\expandafter\def\csname N:stress/s7/a05/carbon\endcsname{0.00428}
\expandafter\def\csname F:stress/s7/a05/carbon\endcsname{0.0043}
\expandafter\def\csname FF:stress/s7/a05/carbon\endcsname{0.00428}
\expandafter\def\csname N:stress/s7/a05/small\endcsname{0.01045}
\expandafter\def\csname F:stress/s7/a05/small\endcsname{0.0104}
\expandafter\def\csname FF:stress/s7/a05/small\endcsname{0.01045}
\expandafter\def\csname N:stress/s7/a05/large\endcsname{0.01654}
\expandafter\def\csname F:stress/s7/a05/large\endcsname{0.0165}
\expandafter\def\csname FF:stress/s7/a05/large\endcsname{0.01654}
\expandafter\def\csname N:stress/s7/econ/n\endcsname{48}
\expandafter\def\csname N:stress/s7/econ/score\endcsname{0.01312}
\expandafter\def\csname F:stress/s7/econ/score\endcsname{0.0131}
\expandafter\def\csname FF:stress/s7/econ/score\endcsname{0.01312}
\expandafter\def\csname N:stress/s7/econ/comfort\endcsname{0.00349}
\expandafter\def\csname F:stress/s7/econ/comfort\endcsname{0.0035}
\expandafter\def\csname FF:stress/s7/econ/comfort\endcsname{0.00349}
\expandafter\def\csname N:stress/s7/econ/energy\endcsname{0.00541}
\expandafter\def\csname F:stress/s7/econ/energy\endcsname{0.0054}
\expandafter\def\csname FF:stress/s7/econ/energy\endcsname{0.00541}
\expandafter\def\csname N:stress/s7/econ/carbon\endcsname{0.00422}
\expandafter\def\csname F:stress/s7/econ/carbon\endcsname{0.0042}
\expandafter\def\csname FF:stress/s7/econ/carbon\endcsname{0.00422}
\expandafter\def\csname N:stress/s7/econ/small\endcsname{0.01024}
\expandafter\def\csname F:stress/s7/econ/small\endcsname{0.0102}
\expandafter\def\csname FF:stress/s7/econ/small\endcsname{0.01024}
\expandafter\def\csname N:stress/s7/econ/large\endcsname{0.01599}
\expandafter\def\csname F:stress/s7/econ/large\endcsname{0.0160}
\expandafter\def\csname FF:stress/s7/econ/large\endcsname{0.01599}
\expandafter\def\csname N:stress/s7/zc/n\endcsname{48}
\expandafter\def\csname N:stress/s7/zc/score\endcsname{0.013}
\expandafter\def\csname F:stress/s7/zc/score\endcsname{0.0130}
\expandafter\def\csname FF:stress/s7/zc/score\endcsname{0.01300}
\expandafter\def\csname N:stress/s7/zc/comfort\endcsname{0.00353}
\expandafter\def\csname F:stress/s7/zc/comfort\endcsname{0.0035}
\expandafter\def\csname FF:stress/s7/zc/comfort\endcsname{0.00353}
\expandafter\def\csname N:stress/s7/zc/energy\endcsname{0.00528}
\expandafter\def\csname F:stress/s7/zc/energy\endcsname{0.0053}
\expandafter\def\csname FF:stress/s7/zc/energy\endcsname{0.00528}
\expandafter\def\csname N:stress/s7/zc/carbon\endcsname{0.00419}
\expandafter\def\csname F:stress/s7/zc/carbon\endcsname{0.0042}
\expandafter\def\csname FF:stress/s7/zc/carbon\endcsname{0.00419}
\expandafter\def\csname N:stress/s7/zc/small\endcsname{0.01007}
\expandafter\def\csname F:stress/s7/zc/small\endcsname{0.0101}
\expandafter\def\csname FF:stress/s7/zc/small\endcsname{0.01007}
\expandafter\def\csname N:stress/s7/zc/large\endcsname{0.01592}
\expandafter\def\csname F:stress/s7/zc/large\endcsname{0.0159}
\expandafter\def\csname FF:stress/s7/zc/large\endcsname{0.01592}
\expandafter\def\csname N:stress/s7/hol9/n\endcsname{48}
\expandafter\def\csname N:stress/s7/hol9/score\endcsname{0.01309}
\expandafter\def\csname F:stress/s7/hol9/score\endcsname{0.0131}
\expandafter\def\csname FF:stress/s7/hol9/score\endcsname{0.01309}
\expandafter\def\csname N:stress/s7/hol9/comfort\endcsname{0.00354}
\expandafter\def\csname F:stress/s7/hol9/comfort\endcsname{0.0035}
\expandafter\def\csname FF:stress/s7/hol9/comfort\endcsname{0.00354}
\expandafter\def\csname N:stress/s7/hol9/energy\endcsname{0.00536}
\expandafter\def\csname F:stress/s7/hol9/energy\endcsname{0.0054}
\expandafter\def\csname FF:stress/s7/hol9/energy\endcsname{0.00536}
\expandafter\def\csname N:stress/s7/hol9/carbon\endcsname{0.00418}
\expandafter\def\csname F:stress/s7/hol9/carbon\endcsname{0.0042}
\expandafter\def\csname FF:stress/s7/hol9/carbon\endcsname{0.00418}
\expandafter\def\csname N:stress/s7/hol9/small\endcsname{0.01018}
\expandafter\def\csname F:stress/s7/hol9/small\endcsname{0.0102}
\expandafter\def\csname FF:stress/s7/hol9/small\endcsname{0.01018}
\expandafter\def\csname N:stress/s7/hol9/large\endcsname{0.016}
\expandafter\def\csname F:stress/s7/hol9/large\endcsname{0.0160}
\expandafter\def\csname FF:stress/s7/hol9/large\endcsname{0.01600}
\expandafter\def\csname N:stress/s7/hol9zc/n\endcsname{48}
\expandafter\def\csname N:stress/s7/hol9zc/score\endcsname{0.01298}
\expandafter\def\csname F:stress/s7/hol9zc/score\endcsname{0.0130}
\expandafter\def\csname FF:stress/s7/hol9zc/score\endcsname{0.01298}
\expandafter\def\csname N:stress/s7/hol9zc/comfort\endcsname{0.00359}
\expandafter\def\csname F:stress/s7/hol9zc/comfort\endcsname{0.0036}
\expandafter\def\csname FF:stress/s7/hol9zc/comfort\endcsname{0.00359}
\expandafter\def\csname N:stress/s7/hol9zc/energy\endcsname{0.00524}
\expandafter\def\csname F:stress/s7/hol9zc/energy\endcsname{0.0052}
\expandafter\def\csname FF:stress/s7/hol9zc/energy\endcsname{0.00524}
\expandafter\def\csname N:stress/s7/hol9zc/carbon\endcsname{0.00415}
\expandafter\def\csname F:stress/s7/hol9zc/carbon\endcsname{0.0042}
\expandafter\def\csname FF:stress/s7/hol9zc/carbon\endcsname{0.00415}
\expandafter\def\csname N:stress/s7/hol9zc/small\endcsname{0.01002}
\expandafter\def\csname F:stress/s7/hol9zc/small\endcsname{0.0100}
\expandafter\def\csname FF:stress/s7/hol9zc/small\endcsname{0.01002}
\expandafter\def\csname N:stress/s7/hol9zc/large\endcsname{0.01594}
\expandafter\def\csname F:stress/s7/hol9zc/large\endcsname{0.0159}
\expandafter\def\csname FF:stress/s7/hol9zc/large\endcsname{0.01594}
\expandafter\def\csname N:stress/s7/ls1/n\endcsname{48}
\expandafter\def\csname N:stress/s7/ls1/score\endcsname{0.01379}
\expandafter\def\csname F:stress/s7/ls1/score\endcsname{0.0138}
\expandafter\def\csname FF:stress/s7/ls1/score\endcsname{0.01379}
\expandafter\def\csname N:stress/s7/ls1/comfort\endcsname{0.00401}
\expandafter\def\csname F:stress/s7/ls1/comfort\endcsname{0.0040}
\expandafter\def\csname FF:stress/s7/ls1/comfort\endcsname{0.00401}
\expandafter\def\csname N:stress/s7/ls1/energy\endcsname{0.00564}
\expandafter\def\csname F:stress/s7/ls1/energy\endcsname{0.0056}
\expandafter\def\csname FF:stress/s7/ls1/energy\endcsname{0.00564}
\expandafter\def\csname N:stress/s7/ls1/carbon\endcsname{0.00414}
\expandafter\def\csname F:stress/s7/ls1/carbon\endcsname{0.0041}
\expandafter\def\csname FF:stress/s7/ls1/carbon\endcsname{0.00414}
\expandafter\def\csname N:stress/s7/ls1/small\endcsname{0.01043}
\expandafter\def\csname F:stress/s7/ls1/small\endcsname{0.0104}
\expandafter\def\csname FF:stress/s7/ls1/small\endcsname{0.01043}
\expandafter\def\csname N:stress/s7/ls1/large\endcsname{0.01714}
\expandafter\def\csname F:stress/s7/ls1/large\endcsname{0.0171}
\expandafter\def\csname FF:stress/s7/ls1/large\endcsname{0.01714}
\expandafter\def\csname N:stress/s7/tank/n\endcsname{48}
\expandafter\def\csname N:stress/s7/tank/score\endcsname{0.01038}
\expandafter\def\csname F:stress/s7/tank/score\endcsname{0.0104}
\expandafter\def\csname FF:stress/s7/tank/score\endcsname{0.01038}
\expandafter\def\csname N:stress/s7/tank/comfort\endcsname{0.00349}
\expandafter\def\csname F:stress/s7/tank/comfort\endcsname{0.0035}
\expandafter\def\csname FF:stress/s7/tank/comfort\endcsname{0.00349}
\expandafter\def\csname N:stress/s7/tank/energy\endcsname{0.00469}
\expandafter\def\csname F:stress/s7/tank/energy\endcsname{0.0047}
\expandafter\def\csname FF:stress/s7/tank/energy\endcsname{0.00469}
\expandafter\def\csname N:stress/s7/tank/carbon\endcsname{0.00221}
\expandafter\def\csname F:stress/s7/tank/carbon\endcsname{0.0022}
\expandafter\def\csname FF:stress/s7/tank/carbon\endcsname{0.00221}
\expandafter\def\csname N:stress/s7/tank/small\endcsname{0.00769}
\expandafter\def\csname F:stress/s7/tank/small\endcsname{0.0077}
\expandafter\def\csname FF:stress/s7/tank/small\endcsname{0.00769}
\expandafter\def\csname N:stress/s7/tank/large\endcsname{0.01308}
\expandafter\def\csname F:stress/s7/tank/large\endcsname{0.0131}
\expandafter\def\csname FF:stress/s7/tank/large\endcsname{0.01308}
\expandafter\def\csname N:stress/s7/zcfix/n\endcsname{48}
\expandafter\def\csname N:stress/s7/zcfix/score\endcsname{0.01325}
\expandafter\def\csname F:stress/s7/zcfix/score\endcsname{0.0132}
\expandafter\def\csname FF:stress/s7/zcfix/score\endcsname{0.01325}
\expandafter\def\csname N:stress/s7/zcfix/comfort\endcsname{0.00353}
\expandafter\def\csname F:stress/s7/zcfix/comfort\endcsname{0.0035}
\expandafter\def\csname FF:stress/s7/zcfix/comfort\endcsname{0.00353}
\expandafter\def\csname N:stress/s7/zcfix/energy\endcsname{0.00534}
\expandafter\def\csname F:stress/s7/zcfix/energy\endcsname{0.0053}
\expandafter\def\csname FF:stress/s7/zcfix/energy\endcsname{0.00534}
\expandafter\def\csname N:stress/s7/zcfix/carbon\endcsname{0.00438}
\expandafter\def\csname F:stress/s7/zcfix/carbon\endcsname{0.0044}
\expandafter\def\csname FF:stress/s7/zcfix/carbon\endcsname{0.00438}
\expandafter\def\csname N:stress/s7/zcfix/small\endcsname{0.0104}
\expandafter\def\csname F:stress/s7/zcfix/small\endcsname{0.0104}
\expandafter\def\csname FF:stress/s7/zcfix/small\endcsname{0.01040}
\expandafter\def\csname N:stress/s7/zcfix/large\endcsname{0.0161}
\expandafter\def\csname F:stress/s7/zcfix/large\endcsname{0.0161}
\expandafter\def\csname FF:stress/s7/zcfix/large\endcsname{0.01610}
\expandafter\def\csname N:stress/s7/tankfix/n\endcsname{48}
\expandafter\def\csname N:stress/s7/tankfix/score\endcsname{0.01336}
\expandafter\def\csname F:stress/s7/tankfix/score\endcsname{0.0134}
\expandafter\def\csname FF:stress/s7/tankfix/score\endcsname{0.01336}
\expandafter\def\csname N:stress/s7/tankfix/comfort\endcsname{0.00349}
\expandafter\def\csname F:stress/s7/tankfix/comfort\endcsname{0.0035}
\expandafter\def\csname FF:stress/s7/tankfix/comfort\endcsname{0.00349}
\expandafter\def\csname N:stress/s7/tankfix/energy\endcsname{0.00547}
\expandafter\def\csname F:stress/s7/tankfix/energy\endcsname{0.0055}
\expandafter\def\csname FF:stress/s7/tankfix/energy\endcsname{0.00547}
\expandafter\def\csname N:stress/s7/tankfix/carbon\endcsname{0.0044}
\expandafter\def\csname F:stress/s7/tankfix/carbon\endcsname{0.0044}
\expandafter\def\csname FF:stress/s7/tankfix/carbon\endcsname{0.00440}
\expandafter\def\csname N:stress/s7/tankfix/small\endcsname{0.01056}
\expandafter\def\csname F:stress/s7/tankfix/small\endcsname{0.0106}
\expandafter\def\csname FF:stress/s7/tankfix/small\endcsname{0.01056}
\expandafter\def\csname N:stress/s7/tankfix/large\endcsname{0.01616}
\expandafter\def\csname F:stress/s7/tankfix/large\endcsname{0.0162}
\expandafter\def\csname FF:stress/s7/tankfix/large\endcsname{0.01616}
\expandafter\def\csname N:stress/s7/area1/n\endcsname{48}
\expandafter\def\csname N:stress/s7/area1/score\endcsname{0.01249}
\expandafter\def\csname F:stress/s7/area1/score\endcsname{0.0125}
\expandafter\def\csname FF:stress/s7/area1/score\endcsname{0.01249}
\expandafter\def\csname N:stress/s7/area1/comfort\endcsname{0.00298}
\expandafter\def\csname F:stress/s7/area1/comfort\endcsname{0.0030}
\expandafter\def\csname FF:stress/s7/area1/comfort\endcsname{0.00298}
\expandafter\def\csname N:stress/s7/area1/energy\endcsname{0.00536}
\expandafter\def\csname F:stress/s7/area1/energy\endcsname{0.0054}
\expandafter\def\csname FF:stress/s7/area1/energy\endcsname{0.00536}
\expandafter\def\csname N:stress/s7/area1/carbon\endcsname{0.00415}
\expandafter\def\csname F:stress/s7/area1/carbon\endcsname{0.0042}
\expandafter\def\csname FF:stress/s7/area1/carbon\endcsname{0.00415}
\expandafter\def\csname N:stress/s7/area1/small\endcsname{0.00968}
\expandafter\def\csname F:stress/s7/area1/small\endcsname{0.0097}
\expandafter\def\csname FF:stress/s7/area1/small\endcsname{0.00968}
\expandafter\def\csname N:stress/s7/area1/large\endcsname{0.01531}
\expandafter\def\csname F:stress/s7/area1/large\endcsname{0.0153}
\expandafter\def\csname FF:stress/s7/area1/large\endcsname{0.01531}
\expandafter\def\csname N:stress/s7/zcfloor/n\endcsname{48}
\expandafter\def\csname N:stress/s7/zcfloor/score\endcsname{0.01451}
\expandafter\def\csname F:stress/s7/zcfloor/score\endcsname{0.0145}
\expandafter\def\csname FF:stress/s7/zcfloor/score\endcsname{0.01451}
\expandafter\def\csname N:stress/s7/zcfloor/comfort\endcsname{0.00346}
\expandafter\def\csname F:stress/s7/zcfloor/comfort\endcsname{0.0035}
\expandafter\def\csname FF:stress/s7/zcfloor/comfort\endcsname{0.00346}
\expandafter\def\csname N:stress/s7/zcfloor/energy\endcsname{0.00575}
\expandafter\def\csname F:stress/s7/zcfloor/energy\endcsname{0.0057}
\expandafter\def\csname FF:stress/s7/zcfloor/energy\endcsname{0.00575}
\expandafter\def\csname N:stress/s7/zcfloor/carbon\endcsname{0.0053}
\expandafter\def\csname F:stress/s7/zcfloor/carbon\endcsname{0.0053}
\expandafter\def\csname FF:stress/s7/zcfloor/carbon\endcsname{0.00530}
\expandafter\def\csname N:stress/s7/zcfloor/small\endcsname{0.01156}
\expandafter\def\csname F:stress/s7/zcfloor/small\endcsname{0.0116}
\expandafter\def\csname FF:stress/s7/zcfloor/small\endcsname{0.01156}
\expandafter\def\csname N:stress/s7/zcfloor/large\endcsname{0.01747}
\expandafter\def\csname F:stress/s7/zcfloor/large\endcsname{0.0175}
\expandafter\def\csname FF:stress/s7/zcfloor/large\endcsname{0.01747}
\expandafter\def\csname N:stress/s7/ra1/n\endcsname{48}
\expandafter\def\csname N:stress/s7/ra1/score\endcsname{0.01219}
\expandafter\def\csname F:stress/s7/ra1/score\endcsname{0.0122}
\expandafter\def\csname FF:stress/s7/ra1/score\endcsname{0.01219}
\expandafter\def\csname N:stress/s7/ra1/comfort\endcsname{0.00327}
\expandafter\def\csname F:stress/s7/ra1/comfort\endcsname{0.0033}
\expandafter\def\csname FF:stress/s7/ra1/comfort\endcsname{0.00327}
\expandafter\def\csname N:stress/s7/ra1/energy\endcsname{0.00519}
\expandafter\def\csname F:stress/s7/ra1/energy\endcsname{0.0052}
\expandafter\def\csname FF:stress/s7/ra1/energy\endcsname{0.00519}
\expandafter\def\csname N:stress/s7/ra1/carbon\endcsname{0.00373}
\expandafter\def\csname F:stress/s7/ra1/carbon\endcsname{0.0037}
\expandafter\def\csname FF:stress/s7/ra1/carbon\endcsname{0.00373}
\expandafter\def\csname N:stress/s7/ra1/small\endcsname{0.00934}
\expandafter\def\csname F:stress/s7/ra1/small\endcsname{0.0093}
\expandafter\def\csname FF:stress/s7/ra1/small\endcsname{0.00934}
\expandafter\def\csname N:stress/s7/ra1/large\endcsname{0.01504}
\expandafter\def\csname F:stress/s7/ra1/large\endcsname{0.0150}
\expandafter\def\csname FF:stress/s7/ra1/large\endcsname{0.01504}
\expandafter\def\csname N:stress/s7/ra1s1/n\endcsname{48}
\expandafter\def\csname N:stress/s7/ra1s1/score\endcsname{0.01778}
\expandafter\def\csname F:stress/s7/ra1s1/score\endcsname{0.0178}
\expandafter\def\csname FF:stress/s7/ra1s1/score\endcsname{0.01778}
\expandafter\def\csname N:stress/s7/ra1s1/comfort\endcsname{0.00912}
\expandafter\def\csname F:stress/s7/ra1s1/comfort\endcsname{0.0091}
\expandafter\def\csname FF:stress/s7/ra1s1/comfort\endcsname{0.00912}
\expandafter\def\csname N:stress/s7/ra1s1/energy\endcsname{0.00498}
\expandafter\def\csname F:stress/s7/ra1s1/energy\endcsname{0.0050}
\expandafter\def\csname FF:stress/s7/ra1s1/energy\endcsname{0.00498}
\expandafter\def\csname N:stress/s7/ra1s1/carbon\endcsname{0.00368}
\expandafter\def\csname F:stress/s7/ra1s1/carbon\endcsname{0.0037}
\expandafter\def\csname FF:stress/s7/ra1s1/carbon\endcsname{0.00368}
\expandafter\def\csname N:stress/s7/ra1s1/small\endcsname{0.01117}
\expandafter\def\csname F:stress/s7/ra1s1/small\endcsname{0.0112}
\expandafter\def\csname FF:stress/s7/ra1s1/small\endcsname{0.01117}
\expandafter\def\csname N:stress/s7/ra1s1/large\endcsname{0.02439}
\expandafter\def\csname F:stress/s7/ra1s1/large\endcsname{0.0244}
\expandafter\def\csname FF:stress/s7/ra1s1/large\endcsname{0.02439}
\expandafter\def\csname N:stress/s7/zcf278/n\endcsname{48}
\expandafter\def\csname N:stress/s7/zcf278/score\endcsname{0.01314}
\expandafter\def\csname F:stress/s7/zcf278/score\endcsname{0.0131}
\expandafter\def\csname FF:stress/s7/zcf278/score\endcsname{0.01314}
\expandafter\def\csname N:stress/s7/zcf278/comfort\endcsname{0.00353}
\expandafter\def\csname F:stress/s7/zcf278/comfort\endcsname{0.0035}
\expandafter\def\csname FF:stress/s7/zcf278/comfort\endcsname{0.00353}
\expandafter\def\csname N:stress/s7/zcf278/energy\endcsname{0.00532}
\expandafter\def\csname F:stress/s7/zcf278/energy\endcsname{0.0053}
\expandafter\def\csname FF:stress/s7/zcf278/energy\endcsname{0.00532}
\expandafter\def\csname N:stress/s7/zcf278/carbon\endcsname{0.00428}
\expandafter\def\csname F:stress/s7/zcf278/carbon\endcsname{0.0043}
\expandafter\def\csname FF:stress/s7/zcf278/carbon\endcsname{0.00428}
\expandafter\def\csname N:stress/s7/zcf278/small\endcsname{0.01022}
\expandafter\def\csname F:stress/s7/zcf278/small\endcsname{0.0102}
\expandafter\def\csname FF:stress/s7/zcf278/small\endcsname{0.01022}
\expandafter\def\csname N:stress/s7/zcf278/large\endcsname{0.01605}
\expandafter\def\csname F:stress/s7/zcf278/large\endcsname{0.0160}
\expandafter\def\csname FF:stress/s7/zcf278/large\endcsname{0.01605}
\expandafter\def\csname N:stress/s7/inv/n\endcsname{48}
\expandafter\def\csname N:stress/s7/inv/score\endcsname{0.00813}
\expandafter\def\csname F:stress/s7/inv/score\endcsname{0.0081}
\expandafter\def\csname FF:stress/s7/inv/score\endcsname{0.00813}
\expandafter\def\csname N:stress/s7/inv/comfort\endcsname{0.00028}
\expandafter\def\csname F:stress/s7/inv/comfort\endcsname{0.0003}
\expandafter\def\csname FF:stress/s7/inv/comfort\endcsname{0.00028}
\expandafter\def\csname N:stress/s7/inv/energy\endcsname{0.0044}
\expandafter\def\csname F:stress/s7/inv/energy\endcsname{0.0044}
\expandafter\def\csname FF:stress/s7/inv/energy\endcsname{0.00440}
\expandafter\def\csname N:stress/s7/inv/carbon\endcsname{0.00345}
\expandafter\def\csname F:stress/s7/inv/carbon\endcsname{0.0034}
\expandafter\def\csname FF:stress/s7/inv/carbon\endcsname{0.00345}
\expandafter\def\csname N:stress/s7/inv/small\endcsname{0.00592}
\expandafter\def\csname F:stress/s7/inv/small\endcsname{0.0059}
\expandafter\def\csname FF:stress/s7/inv/small\endcsname{0.00592}
\expandafter\def\csname N:stress/s7/inv/large\endcsname{0.01033}
\expandafter\def\csname F:stress/s7/inv/large\endcsname{0.0103}
\expandafter\def\csname FF:stress/s7/inv/large\endcsname{0.01033}
\expandafter\def\csname N:stress/s7/ctrlw5/n\endcsname{48}
\expandafter\def\csname N:stress/s7/ctrlw5/score\endcsname{0.00805}
\expandafter\def\csname F:stress/s7/ctrlw5/score\endcsname{0.0080}
\expandafter\def\csname FF:stress/s7/ctrlw5/score\endcsname{0.00805}
\expandafter\def\csname N:stress/s7/ctrlw5/comfort\endcsname{0.00025}
\expandafter\def\csname F:stress/s7/ctrlw5/comfort\endcsname{0.0003}
\expandafter\def\csname FF:stress/s7/ctrlw5/comfort\endcsname{0.00025}
\expandafter\def\csname N:stress/s7/ctrlw5/energy\endcsname{0.00438}
\expandafter\def\csname F:stress/s7/ctrlw5/energy\endcsname{0.0044}
\expandafter\def\csname FF:stress/s7/ctrlw5/energy\endcsname{0.00438}
\expandafter\def\csname N:stress/s7/ctrlw5/carbon\endcsname{0.00343}
\expandafter\def\csname F:stress/s7/ctrlw5/carbon\endcsname{0.0034}
\expandafter\def\csname FF:stress/s7/ctrlw5/carbon\endcsname{0.00343}
\expandafter\def\csname N:stress/s7/ctrlw5/small\endcsname{0.00588}
\expandafter\def\csname F:stress/s7/ctrlw5/small\endcsname{0.0059}
\expandafter\def\csname FF:stress/s7/ctrlw5/small\endcsname{0.00588}
\expandafter\def\csname N:stress/s7/ctrlw5/large\endcsname{0.01022}
\expandafter\def\csname F:stress/s7/ctrlw5/large\endcsname{0.0102}
\expandafter\def\csname FF:stress/s7/ctrlw5/large\endcsname{0.01022}
\expandafter\def\csname N:stress/s7/sb278/n\endcsname{48}
\expandafter\def\csname N:stress/s7/sb278/score\endcsname{0.00822}
\expandafter\def\csname F:stress/s7/sb278/score\endcsname{0.0082}
\expandafter\def\csname FF:stress/s7/sb278/score\endcsname{0.00822}
\expandafter\def\csname N:stress/s7/sb278/comfort\endcsname{0.00027}
\expandafter\def\csname F:stress/s7/sb278/comfort\endcsname{0.0003}
\expandafter\def\csname FF:stress/s7/sb278/comfort\endcsname{0.00027}
\expandafter\def\csname N:stress/s7/sb278/energy\endcsname{0.00443}
\expandafter\def\csname F:stress/s7/sb278/energy\endcsname{0.0044}
\expandafter\def\csname FF:stress/s7/sb278/energy\endcsname{0.00443}
\expandafter\def\csname N:stress/s7/sb278/carbon\endcsname{0.00351}
\expandafter\def\csname F:stress/s7/sb278/carbon\endcsname{0.0035}
\expandafter\def\csname FF:stress/s7/sb278/carbon\endcsname{0.00351}
\expandafter\def\csname N:stress/s7/sb278/small\endcsname{0.00607}
\expandafter\def\csname F:stress/s7/sb278/small\endcsname{0.0061}
\expandafter\def\csname FF:stress/s7/sb278/small\endcsname{0.00607}
\expandafter\def\csname N:stress/s7/sb278/large\endcsname{0.01037}
\expandafter\def\csname F:stress/s7/sb278/large\endcsname{0.0104}
\expandafter\def\csname FF:stress/s7/sb278/large\endcsname{0.01037}
\expandafter\def\csname N:stress/s7/invfix/n\endcsname{48}
\expandafter\def\csname N:stress/s7/invfix/score\endcsname{0.00806}
\expandafter\def\csname F:stress/s7/invfix/score\endcsname{0.0081}
\expandafter\def\csname FF:stress/s7/invfix/score\endcsname{0.00806}
\expandafter\def\csname N:stress/s7/invfix/comfort\endcsname{0.00028}
\expandafter\def\csname F:stress/s7/invfix/comfort\endcsname{0.0003}
\expandafter\def\csname FF:stress/s7/invfix/comfort\endcsname{0.00028}
\expandafter\def\csname N:stress/s7/invfix/energy\endcsname{0.00439}
\expandafter\def\csname F:stress/s7/invfix/energy\endcsname{0.0044}
\expandafter\def\csname FF:stress/s7/invfix/energy\endcsname{0.00439}
\expandafter\def\csname N:stress/s7/invfix/carbon\endcsname{0.0034}
\expandafter\def\csname F:stress/s7/invfix/carbon\endcsname{0.0034}
\expandafter\def\csname FF:stress/s7/invfix/carbon\endcsname{0.00340}
\expandafter\def\csname N:stress/s7/invfix/small\endcsname{0.00586}
\expandafter\def\csname F:stress/s7/invfix/small\endcsname{0.0059}
\expandafter\def\csname FF:stress/s7/invfix/small\endcsname{0.00586}
\expandafter\def\csname N:stress/s7/invfix/large\endcsname{0.01027}
\expandafter\def\csname F:stress/s7/invfix/large\endcsname{0.0103}
\expandafter\def\csname FF:stress/s7/invfix/large\endcsname{0.01027}
\expandafter\def\csname N:stress/s7_ci/a05_vs_c5/n\endcsname{48}
\expandafter\def\csname N:stress/s7_ci/a05_vs_c5/mean\endcsname{$-0.00242$}
\expandafter\def\csname F:stress/s7_ci/a05_vs_c5/mean\endcsname{-0.0024}
\expandafter\def\csname FF:stress/s7_ci/a05_vs_c5/mean\endcsname{-0.00242}
\expandafter\def\csname N:stress/s7_ci/a05_vs_c5/lo\endcsname{$-0.00464$}
\expandafter\def\csname F:stress/s7_ci/a05_vs_c5/lo\endcsname{-0.0046}
\expandafter\def\csname FF:stress/s7_ci/a05_vs_c5/lo\endcsname{-0.00464}
\expandafter\def\csname N:stress/s7_ci/a05_vs_c5/hi\endcsname{$-0.00069$}
\expandafter\def\csname F:stress/s7_ci/a05_vs_c5/hi\endcsname{-0.0007}
\expandafter\def\csname FF:stress/s7_ci/a05_vs_c5/hi\endcsname{-0.00069}
\expandafter\def\csname N:stress/s7_ci/a05_vs_c5/pct\endcsname{$-15.2$}
\expandafter\def\csname F:stress/s7_ci/a05_vs_c5/pct\endcsname{-15.2000}
\expandafter\def\csname FF:stress/s7_ci/a05_vs_c5/pct\endcsname{-15.20000}
\expandafter\def\csname N:stress/s7_ci/a05_vs_c5/pct_lo\endcsname{$-29.2$}
\expandafter\def\csname F:stress/s7_ci/a05_vs_c5/pct_lo\endcsname{-29.2000}
\expandafter\def\csname FF:stress/s7_ci/a05_vs_c5/pct_lo\endcsname{-29.20000}
\expandafter\def\csname N:stress/s7_ci/a05_vs_c5/pct_hi\endcsname{$-4.3$}
\expandafter\def\csname F:stress/s7_ci/a05_vs_c5/pct_hi\endcsname{-4.3000}
\expandafter\def\csname FF:stress/s7_ci/a05_vs_c5/pct_hi\endcsname{-4.30000}
\expandafter\def\csname N:stress/s7_ci/a05_vs_c5/by_station_pct/fairbanks\endcsname{$+1.7$}
\expandafter\def\csname F:stress/s7_ci/a05_vs_c5/by_station_pct/fairbanks\endcsname{1.7000}
\expandafter\def\csname FF:stress/s7_ci/a05_vs_c5/by_station_pct/fairbanks\endcsname{1.70000}
\expandafter\def\csname N:stress/s7_ci/a05_vs_c5/by_station_pct/denver\endcsname{$+2.2$}
\expandafter\def\csname F:stress/s7_ci/a05_vs_c5/by_station_pct/denver\endcsname{2.2000}
\expandafter\def\csname FF:stress/s7_ci/a05_vs_c5/by_station_pct/denver\endcsname{2.20000}
\expandafter\def\csname N:stress/s7_ci/a05_vs_c5/by_station_pct/atlanta\endcsname{$-8.7$}
\expandafter\def\csname F:stress/s7_ci/a05_vs_c5/by_station_pct/atlanta\endcsname{-8.7000}
\expandafter\def\csname FF:stress/s7_ci/a05_vs_c5/by_station_pct/atlanta\endcsname{-8.70000}
\expandafter\def\csname N:stress/s7_ci/a05_vs_c5/by_station_pct/phoenix\endcsname{$-17.6$}
\expandafter\def\csname F:stress/s7_ci/a05_vs_c5/by_station_pct/phoenix\endcsname{-17.6000}
\expandafter\def\csname FF:stress/s7_ci/a05_vs_c5/by_station_pct/phoenix\endcsname{-17.60000}
\expandafter\def\csname N:stress/s7_ci/a05_vs_c5/by_station_pct/singapore\endcsname{$-7.1$}
\expandafter\def\csname F:stress/s7_ci/a05_vs_c5/by_station_pct/singapore\endcsname{-7.1000}
\expandafter\def\csname FF:stress/s7_ci/a05_vs_c5/by_station_pct/singapore\endcsname{-7.10000}
\expandafter\def\csname N:stress/s7_ci/a05_vs_c5/by_station_pct/minneapolis\endcsname{$-8.0$}
\expandafter\def\csname F:stress/s7_ci/a05_vs_c5/by_station_pct/minneapolis\endcsname{-8.0000}
\expandafter\def\csname FF:stress/s7_ci/a05_vs_c5/by_station_pct/minneapolis\endcsname{-8.00000}
\expandafter\def\csname N:stress/s7_ci/a05_vs_c5/wins\endcsname{29}
\expandafter\def\csname N:stress/s7_ci/econ_vs_a05/n\endcsname{48}
\expandafter\def\csname N:stress/s7_ci/econ_vs_a05/mean\endcsname{$-0.00038$}
\expandafter\def\csname F:stress/s7_ci/econ_vs_a05/mean\endcsname{-0.0004}
\expandafter\def\csname FF:stress/s7_ci/econ_vs_a05/mean\endcsname{-0.00038}
\expandafter\def\csname N:stress/s7_ci/econ_vs_a05/lo\endcsname{$-0.00055$}
\expandafter\def\csname F:stress/s7_ci/econ_vs_a05/lo\endcsname{-0.0006}
\expandafter\def\csname FF:stress/s7_ci/econ_vs_a05/lo\endcsname{-0.00055}
\expandafter\def\csname N:stress/s7_ci/econ_vs_a05/hi\endcsname{$-0.00023$}
\expandafter\def\csname F:stress/s7_ci/econ_vs_a05/hi\endcsname{-0.0002}
\expandafter\def\csname FF:stress/s7_ci/econ_vs_a05/hi\endcsname{-0.00023}
\expandafter\def\csname N:stress/s7_ci/econ_vs_a05/pct\endcsname{$-2.8$}
\expandafter\def\csname F:stress/s7_ci/econ_vs_a05/pct\endcsname{-2.8000}
\expandafter\def\csname FF:stress/s7_ci/econ_vs_a05/pct\endcsname{-2.80000}
\expandafter\def\csname N:stress/s7_ci/econ_vs_a05/pct_lo\endcsname{$-4.0$}
\expandafter\def\csname F:stress/s7_ci/econ_vs_a05/pct_lo\endcsname{-4.0000}
\expandafter\def\csname FF:stress/s7_ci/econ_vs_a05/pct_lo\endcsname{-4.00000}
\expandafter\def\csname N:stress/s7_ci/econ_vs_a05/pct_hi\endcsname{$-1.7$}
\expandafter\def\csname F:stress/s7_ci/econ_vs_a05/pct_hi\endcsname{-1.7000}
\expandafter\def\csname FF:stress/s7_ci/econ_vs_a05/pct_hi\endcsname{-1.70000}
\expandafter\def\csname N:stress/s7_ci/econ_vs_a05/by_station_pct/fairbanks\endcsname{$-4.0$}
\expandafter\def\csname F:stress/s7_ci/econ_vs_a05/by_station_pct/fairbanks\endcsname{-4.0000}
\expandafter\def\csname FF:stress/s7_ci/econ_vs_a05/by_station_pct/fairbanks\endcsname{-4.00000}
\expandafter\def\csname N:stress/s7_ci/econ_vs_a05/by_station_pct/denver\endcsname{$-5.5$}
\expandafter\def\csname F:stress/s7_ci/econ_vs_a05/by_station_pct/denver\endcsname{-5.5000}
\expandafter\def\csname FF:stress/s7_ci/econ_vs_a05/by_station_pct/denver\endcsname{-5.50000}
\expandafter\def\csname N:stress/s7_ci/econ_vs_a05/by_station_pct/atlanta\endcsname{$-11.1$}
\expandafter\def\csname F:stress/s7_ci/econ_vs_a05/by_station_pct/atlanta\endcsname{-11.1000}
\expandafter\def\csname FF:stress/s7_ci/econ_vs_a05/by_station_pct/atlanta\endcsname{-11.10000}
\expandafter\def\csname N:stress/s7_ci/econ_vs_a05/by_station_pct/phoenix\endcsname{$-4.9$}
\expandafter\def\csname F:stress/s7_ci/econ_vs_a05/by_station_pct/phoenix\endcsname{-4.9000}
\expandafter\def\csname FF:stress/s7_ci/econ_vs_a05/by_station_pct/phoenix\endcsname{-4.90000}
\expandafter\def\csname N:stress/s7_ci/econ_vs_a05/by_station_pct/singapore\endcsname{$0.0$}
\expandafter\def\csname F:stress/s7_ci/econ_vs_a05/by_station_pct/singapore\endcsname{0.0000}
\expandafter\def\csname FF:stress/s7_ci/econ_vs_a05/by_station_pct/singapore\endcsname{0.00000}
\expandafter\def\csname N:stress/s7_ci/econ_vs_a05/by_station_pct/minneapolis\endcsname{$-7.5$}
\expandafter\def\csname F:stress/s7_ci/econ_vs_a05/by_station_pct/minneapolis\endcsname{-7.5000}
\expandafter\def\csname FF:stress/s7_ci/econ_vs_a05/by_station_pct/minneapolis\endcsname{-7.50000}
\expandafter\def\csname N:stress/s7_ci/econ_vs_a05/wins\endcsname{33}
\expandafter\def\csname N:stress/s7_ci/zc_vs_econ/n\endcsname{48}
\expandafter\def\csname N:stress/s7_ci/zc_vs_econ/mean\endcsname{$-0.00012$}
\expandafter\def\csname F:stress/s7_ci/zc_vs_econ/mean\endcsname{-0.0001}
\expandafter\def\csname FF:stress/s7_ci/zc_vs_econ/mean\endcsname{-0.00012}
\expandafter\def\csname N:stress/s7_ci/zc_vs_econ/lo\endcsname{$-0.00018$}
\expandafter\def\csname F:stress/s7_ci/zc_vs_econ/lo\endcsname{-0.0002}
\expandafter\def\csname FF:stress/s7_ci/zc_vs_econ/lo\endcsname{-0.00018}
\expandafter\def\csname N:stress/s7_ci/zc_vs_econ/hi\endcsname{$-6e-05$}
\expandafter\def\csname F:stress/s7_ci/zc_vs_econ/hi\endcsname{-0.0001}
\expandafter\def\csname FF:stress/s7_ci/zc_vs_econ/hi\endcsname{-0.00006}
\expandafter\def\csname N:stress/s7_ci/zc_vs_econ/pct\endcsname{$-0.9$}
\expandafter\def\csname F:stress/s7_ci/zc_vs_econ/pct\endcsname{-0.9000}
\expandafter\def\csname FF:stress/s7_ci/zc_vs_econ/pct\endcsname{-0.90000}
\expandafter\def\csname N:stress/s7_ci/zc_vs_econ/pct_lo\endcsname{$-1.3$}
\expandafter\def\csname F:stress/s7_ci/zc_vs_econ/pct_lo\endcsname{-1.3000}
\expandafter\def\csname FF:stress/s7_ci/zc_vs_econ/pct_lo\endcsname{-1.30000}
\expandafter\def\csname N:stress/s7_ci/zc_vs_econ/pct_hi\endcsname{$-0.5$}
\expandafter\def\csname F:stress/s7_ci/zc_vs_econ/pct_hi\endcsname{-0.5000}
\expandafter\def\csname FF:stress/s7_ci/zc_vs_econ/pct_hi\endcsname{-0.50000}
\expandafter\def\csname N:stress/s7_ci/zc_vs_econ/by_station_pct/fairbanks\endcsname{$-1.4$}
\expandafter\def\csname F:stress/s7_ci/zc_vs_econ/by_station_pct/fairbanks\endcsname{-1.4000}
\expandafter\def\csname FF:stress/s7_ci/zc_vs_econ/by_station_pct/fairbanks\endcsname{-1.40000}
\expandafter\def\csname N:stress/s7_ci/zc_vs_econ/by_station_pct/denver\endcsname{$-2.6$}
\expandafter\def\csname F:stress/s7_ci/zc_vs_econ/by_station_pct/denver\endcsname{-2.6000}
\expandafter\def\csname FF:stress/s7_ci/zc_vs_econ/by_station_pct/denver\endcsname{-2.60000}
\expandafter\def\csname N:stress/s7_ci/zc_vs_econ/by_station_pct/atlanta\endcsname{$-2.9$}
\expandafter\def\csname F:stress/s7_ci/zc_vs_econ/by_station_pct/atlanta\endcsname{-2.9000}
\expandafter\def\csname FF:stress/s7_ci/zc_vs_econ/by_station_pct/atlanta\endcsname{-2.90000}
\expandafter\def\csname N:stress/s7_ci/zc_vs_econ/by_station_pct/phoenix\endcsname{$-1.8$}
\expandafter\def\csname F:stress/s7_ci/zc_vs_econ/by_station_pct/phoenix\endcsname{-1.8000}
\expandafter\def\csname FF:stress/s7_ci/zc_vs_econ/by_station_pct/phoenix\endcsname{-1.80000}
\expandafter\def\csname N:stress/s7_ci/zc_vs_econ/by_station_pct/singapore\endcsname{$-4.0$}
\expandafter\def\csname F:stress/s7_ci/zc_vs_econ/by_station_pct/singapore\endcsname{-4.0000}
\expandafter\def\csname FF:stress/s7_ci/zc_vs_econ/by_station_pct/singapore\endcsname{-4.00000}
\expandafter\def\csname N:stress/s7_ci/zc_vs_econ/by_station_pct/minneapolis\endcsname{$-1.2$}
\expandafter\def\csname F:stress/s7_ci/zc_vs_econ/by_station_pct/minneapolis\endcsname{-1.2000}
\expandafter\def\csname FF:stress/s7_ci/zc_vs_econ/by_station_pct/minneapolis\endcsname{-1.20000}
\expandafter\def\csname N:stress/s7_ci/zc_vs_econ/wins\endcsname{37}
\expandafter\def\csname N:stress/s7_ci/hol9_vs_econ/n\endcsname{48}
\expandafter\def\csname N:stress/s7_ci/hol9_vs_econ/mean\endcsname{$-3e-05$}
\expandafter\def\csname F:stress/s7_ci/hol9_vs_econ/mean\endcsname{-0.0000}
\expandafter\def\csname FF:stress/s7_ci/hol9_vs_econ/mean\endcsname{-0.00003}
\expandafter\def\csname N:stress/s7_ci/hol9_vs_econ/lo\endcsname{$-8e-05$}
\expandafter\def\csname F:stress/s7_ci/hol9_vs_econ/lo\endcsname{-0.0001}
\expandafter\def\csname FF:stress/s7_ci/hol9_vs_econ/lo\endcsname{-0.00008}
\expandafter\def\csname N:stress/s7_ci/hol9_vs_econ/hi\endcsname{3e-05}
\expandafter\def\csname F:stress/s7_ci/hol9_vs_econ/hi\endcsname{0.0000}
\expandafter\def\csname FF:stress/s7_ci/hol9_vs_econ/hi\endcsname{0.00003}
\expandafter\def\csname N:stress/s7_ci/hol9_vs_econ/pct\endcsname{$-0.2$}
\expandafter\def\csname F:stress/s7_ci/hol9_vs_econ/pct\endcsname{-0.2000}
\expandafter\def\csname FF:stress/s7_ci/hol9_vs_econ/pct\endcsname{-0.20000}
\expandafter\def\csname N:stress/s7_ci/hol9_vs_econ/pct_lo\endcsname{$-0.6$}
\expandafter\def\csname F:stress/s7_ci/hol9_vs_econ/pct_lo\endcsname{-0.6000}
\expandafter\def\csname FF:stress/s7_ci/hol9_vs_econ/pct_lo\endcsname{-0.60000}
\expandafter\def\csname N:stress/s7_ci/hol9_vs_econ/pct_hi\endcsname{$+0.3$}
\expandafter\def\csname F:stress/s7_ci/hol9_vs_econ/pct_hi\endcsname{0.3000}
\expandafter\def\csname FF:stress/s7_ci/hol9_vs_econ/pct_hi\endcsname{0.30000}
\expandafter\def\csname N:stress/s7_ci/hol9_vs_econ/by_station_pct/fairbanks\endcsname{$-1.3$}
\expandafter\def\csname F:stress/s7_ci/hol9_vs_econ/by_station_pct/fairbanks\endcsname{-1.3000}
\expandafter\def\csname FF:stress/s7_ci/hol9_vs_econ/by_station_pct/fairbanks\endcsname{-1.30000}
\expandafter\def\csname N:stress/s7_ci/hol9_vs_econ/by_station_pct/denver\endcsname{$-2.8$}
\expandafter\def\csname F:stress/s7_ci/hol9_vs_econ/by_station_pct/denver\endcsname{-2.8000}
\expandafter\def\csname FF:stress/s7_ci/hol9_vs_econ/by_station_pct/denver\endcsname{-2.80000}
\expandafter\def\csname N:stress/s7_ci/hol9_vs_econ/by_station_pct/atlanta\endcsname{$0.0$}
\expandafter\def\csname F:stress/s7_ci/hol9_vs_econ/by_station_pct/atlanta\endcsname{0.0000}
\expandafter\def\csname FF:stress/s7_ci/hol9_vs_econ/by_station_pct/atlanta\endcsname{0.00000}
\expandafter\def\csname N:stress/s7_ci/hol9_vs_econ/by_station_pct/phoenix\endcsname{$0.0$}
\expandafter\def\csname F:stress/s7_ci/hol9_vs_econ/by_station_pct/phoenix\endcsname{-0.0000}
\expandafter\def\csname FF:stress/s7_ci/hol9_vs_econ/by_station_pct/phoenix\endcsname{-0.00000}
\expandafter\def\csname N:stress/s7_ci/hol9_vs_econ/by_station_pct/singapore\endcsname{$0.0$}
\expandafter\def\csname F:stress/s7_ci/hol9_vs_econ/by_station_pct/singapore\endcsname{0.0000}
\expandafter\def\csname FF:stress/s7_ci/hol9_vs_econ/by_station_pct/singapore\endcsname{0.00000}
\expandafter\def\csname N:stress/s7_ci/hol9_vs_econ/by_station_pct/minneapolis\endcsname{$0.0$}
\expandafter\def\csname F:stress/s7_ci/hol9_vs_econ/by_station_pct/minneapolis\endcsname{0.0000}
\expandafter\def\csname FF:stress/s7_ci/hol9_vs_econ/by_station_pct/minneapolis\endcsname{0.00000}
\expandafter\def\csname N:stress/s7_ci/hol9_vs_econ/wins\endcsname{8}
\expandafter\def\csname W:stress/s7_ci/hol9_vs_econ/wins\endcsname{eight}
\expandafter\def\csname N:stress/s7_ci/hol9zc_vs_econ/n\endcsname{48}
\expandafter\def\csname N:stress/s7_ci/hol9zc_vs_econ/mean\endcsname{$-0.00014$}
\expandafter\def\csname F:stress/s7_ci/hol9zc_vs_econ/mean\endcsname{-0.0001}
\expandafter\def\csname FF:stress/s7_ci/hol9zc_vs_econ/mean\endcsname{-0.00014}
\expandafter\def\csname N:stress/s7_ci/hol9zc_vs_econ/lo\endcsname{$-0.00022$}
\expandafter\def\csname F:stress/s7_ci/hol9zc_vs_econ/lo\endcsname{-0.0002}
\expandafter\def\csname FF:stress/s7_ci/hol9zc_vs_econ/lo\endcsname{-0.00022}
\expandafter\def\csname N:stress/s7_ci/hol9zc_vs_econ/hi\endcsname{$-4e-05$}
\expandafter\def\csname F:stress/s7_ci/hol9zc_vs_econ/hi\endcsname{-0.0000}
\expandafter\def\csname FF:stress/s7_ci/hol9zc_vs_econ/hi\endcsname{-0.00004}
\expandafter\def\csname N:stress/s7_ci/hol9zc_vs_econ/pct\endcsname{$-1.0$}
\expandafter\def\csname F:stress/s7_ci/hol9zc_vs_econ/pct\endcsname{-1.0000}
\expandafter\def\csname FF:stress/s7_ci/hol9zc_vs_econ/pct\endcsname{-1.00000}
\expandafter\def\csname N:stress/s7_ci/hol9zc_vs_econ/pct_lo\endcsname{$-1.7$}
\expandafter\def\csname F:stress/s7_ci/hol9zc_vs_econ/pct_lo\endcsname{-1.7000}
\expandafter\def\csname FF:stress/s7_ci/hol9zc_vs_econ/pct_lo\endcsname{-1.70000}
\expandafter\def\csname N:stress/s7_ci/hol9zc_vs_econ/pct_hi\endcsname{$-0.3$}
\expandafter\def\csname F:stress/s7_ci/hol9zc_vs_econ/pct_hi\endcsname{-0.3000}
\expandafter\def\csname FF:stress/s7_ci/hol9zc_vs_econ/pct_hi\endcsname{-0.30000}
\expandafter\def\csname N:stress/s7_ci/hol9zc_vs_econ/by_station_pct/fairbanks\endcsname{$-2.4$}
\expandafter\def\csname F:stress/s7_ci/hol9zc_vs_econ/by_station_pct/fairbanks\endcsname{-2.4000}
\expandafter\def\csname FF:stress/s7_ci/hol9zc_vs_econ/by_station_pct/fairbanks\endcsname{-2.40000}
\expandafter\def\csname N:stress/s7_ci/hol9zc_vs_econ/by_station_pct/denver\endcsname{$-5.1$}
\expandafter\def\csname F:stress/s7_ci/hol9zc_vs_econ/by_station_pct/denver\endcsname{-5.1000}
\expandafter\def\csname FF:stress/s7_ci/hol9zc_vs_econ/by_station_pct/denver\endcsname{-5.10000}
\expandafter\def\csname N:stress/s7_ci/hol9zc_vs_econ/by_station_pct/atlanta\endcsname{$-2.9$}
\expandafter\def\csname F:stress/s7_ci/hol9zc_vs_econ/by_station_pct/atlanta\endcsname{-2.9000}
\expandafter\def\csname FF:stress/s7_ci/hol9zc_vs_econ/by_station_pct/atlanta\endcsname{-2.90000}
\expandafter\def\csname N:stress/s7_ci/hol9zc_vs_econ/by_station_pct/phoenix\endcsname{$-1.6$}
\expandafter\def\csname F:stress/s7_ci/hol9zc_vs_econ/by_station_pct/phoenix\endcsname{-1.6000}
\expandafter\def\csname FF:stress/s7_ci/hol9zc_vs_econ/by_station_pct/phoenix\endcsname{-1.60000}
\expandafter\def\csname N:stress/s7_ci/hol9zc_vs_econ/by_station_pct/singapore\endcsname{$-4.0$}
\expandafter\def\csname F:stress/s7_ci/hol9zc_vs_econ/by_station_pct/singapore\endcsname{-4.0000}
\expandafter\def\csname FF:stress/s7_ci/hol9zc_vs_econ/by_station_pct/singapore\endcsname{-4.00000}
\expandafter\def\csname N:stress/s7_ci/hol9zc_vs_econ/by_station_pct/minneapolis\endcsname{$-1.2$}
\expandafter\def\csname F:stress/s7_ci/hol9zc_vs_econ/by_station_pct/minneapolis\endcsname{-1.2000}
\expandafter\def\csname FF:stress/s7_ci/hol9zc_vs_econ/by_station_pct/minneapolis\endcsname{-1.20000}
\expandafter\def\csname N:stress/s7_ci/hol9zc_vs_econ/wins\endcsname{38}
\expandafter\def\csname N:stress/s7_ci/ls1_vs_a05/n\endcsname{48}
\expandafter\def\csname N:stress/s7_ci/ls1_vs_a05/mean\endcsname{0.00029}
\expandafter\def\csname F:stress/s7_ci/ls1_vs_a05/mean\endcsname{0.0003}
\expandafter\def\csname FF:stress/s7_ci/ls1_vs_a05/mean\endcsname{0.00029}
\expandafter\def\csname N:stress/s7_ci/ls1_vs_a05/lo\endcsname{2e-05}
\expandafter\def\csname F:stress/s7_ci/ls1_vs_a05/lo\endcsname{0.0000}
\expandafter\def\csname FF:stress/s7_ci/ls1_vs_a05/lo\endcsname{0.00002}
\expandafter\def\csname N:stress/s7_ci/ls1_vs_a05/hi\endcsname{0.00066}
\expandafter\def\csname F:stress/s7_ci/ls1_vs_a05/hi\endcsname{0.0007}
\expandafter\def\csname FF:stress/s7_ci/ls1_vs_a05/hi\endcsname{0.00066}
\expandafter\def\csname N:stress/s7_ci/ls1_vs_a05/pct\endcsname{$+2.2$}
\expandafter\def\csname F:stress/s7_ci/ls1_vs_a05/pct\endcsname{2.2000}
\expandafter\def\csname FF:stress/s7_ci/ls1_vs_a05/pct\endcsname{2.20000}
\expandafter\def\csname N:stress/s7_ci/ls1_vs_a05/pct_lo\endcsname{$+0.1$}
\expandafter\def\csname F:stress/s7_ci/ls1_vs_a05/pct_lo\endcsname{0.1000}
\expandafter\def\csname FF:stress/s7_ci/ls1_vs_a05/pct_lo\endcsname{0.10000}
\expandafter\def\csname N:stress/s7_ci/ls1_vs_a05/pct_hi\endcsname{$+4.9$}
\expandafter\def\csname F:stress/s7_ci/ls1_vs_a05/pct_hi\endcsname{4.9000}
\expandafter\def\csname FF:stress/s7_ci/ls1_vs_a05/pct_hi\endcsname{4.90000}
\expandafter\def\csname N:stress/s7_ci/ls1_vs_a05/by_station_pct/fairbanks\endcsname{$+1.9$}
\expandafter\def\csname F:stress/s7_ci/ls1_vs_a05/by_station_pct/fairbanks\endcsname{1.9000}
\expandafter\def\csname FF:stress/s7_ci/ls1_vs_a05/by_station_pct/fairbanks\endcsname{1.90000}
\expandafter\def\csname N:stress/s7_ci/ls1_vs_a05/by_station_pct/denver\endcsname{$+0.5$}
\expandafter\def\csname F:stress/s7_ci/ls1_vs_a05/by_station_pct/denver\endcsname{0.5000}
\expandafter\def\csname FF:stress/s7_ci/ls1_vs_a05/by_station_pct/denver\endcsname{0.50000}
\expandafter\def\csname N:stress/s7_ci/ls1_vs_a05/by_station_pct/atlanta\endcsname{$+0.4$}
\expandafter\def\csname F:stress/s7_ci/ls1_vs_a05/by_station_pct/atlanta\endcsname{0.4000}
\expandafter\def\csname FF:stress/s7_ci/ls1_vs_a05/by_station_pct/atlanta\endcsname{0.40000}
\expandafter\def\csname N:stress/s7_ci/ls1_vs_a05/by_station_pct/phoenix\endcsname{$+0.8$}
\expandafter\def\csname F:stress/s7_ci/ls1_vs_a05/by_station_pct/phoenix\endcsname{0.8000}
\expandafter\def\csname FF:stress/s7_ci/ls1_vs_a05/by_station_pct/phoenix\endcsname{0.80000}
\expandafter\def\csname N:stress/s7_ci/ls1_vs_a05/by_station_pct/singapore\endcsname{$0.0$}
\expandafter\def\csname F:stress/s7_ci/ls1_vs_a05/by_station_pct/singapore\endcsname{0.0000}
\expandafter\def\csname FF:stress/s7_ci/ls1_vs_a05/by_station_pct/singapore\endcsname{0.00000}
\expandafter\def\csname N:stress/s7_ci/ls1_vs_a05/by_station_pct/minneapolis\endcsname{$+2.0$}
\expandafter\def\csname F:stress/s7_ci/ls1_vs_a05/by_station_pct/minneapolis\endcsname{2.0000}
\expandafter\def\csname FF:stress/s7_ci/ls1_vs_a05/by_station_pct/minneapolis\endcsname{2.00000}
\expandafter\def\csname N:stress/s7_ci/ls1_vs_a05/wins\endcsname{21}
\expandafter\def\csname N:stress/s7_ci/tank_vs_a05/n\endcsname{48}
\expandafter\def\csname N:stress/s7_ci/tank_vs_a05/mean\endcsname{$-0.00311$}
\expandafter\def\csname F:stress/s7_ci/tank_vs_a05/mean\endcsname{-0.0031}
\expandafter\def\csname FF:stress/s7_ci/tank_vs_a05/mean\endcsname{-0.00311}
\expandafter\def\csname N:stress/s7_ci/tank_vs_a05/lo\endcsname{$-0.00481$}
\expandafter\def\csname F:stress/s7_ci/tank_vs_a05/lo\endcsname{-0.0048}
\expandafter\def\csname FF:stress/s7_ci/tank_vs_a05/lo\endcsname{-0.00481}
\expandafter\def\csname N:stress/s7_ci/tank_vs_a05/hi\endcsname{$-0.00167$}
\expandafter\def\csname F:stress/s7_ci/tank_vs_a05/hi\endcsname{-0.0017}
\expandafter\def\csname FF:stress/s7_ci/tank_vs_a05/hi\endcsname{-0.00167}
\expandafter\def\csname N:stress/s7_ci/tank_vs_a05/pct\endcsname{$-23.1$}
\expandafter\def\csname F:stress/s7_ci/tank_vs_a05/pct\endcsname{-23.1000}
\expandafter\def\csname FF:stress/s7_ci/tank_vs_a05/pct\endcsname{-23.10000}
\expandafter\def\csname N:stress/s7_ci/tank_vs_a05/pct_lo\endcsname{$-35.6$}
\expandafter\def\csname F:stress/s7_ci/tank_vs_a05/pct_lo\endcsname{-35.6000}
\expandafter\def\csname FF:stress/s7_ci/tank_vs_a05/pct_lo\endcsname{-35.60000}
\expandafter\def\csname N:stress/s7_ci/tank_vs_a05/pct_hi\endcsname{$-12.4$}
\expandafter\def\csname F:stress/s7_ci/tank_vs_a05/pct_hi\endcsname{-12.4000}
\expandafter\def\csname FF:stress/s7_ci/tank_vs_a05/pct_hi\endcsname{-12.40000}
\expandafter\def\csname N:stress/s7_ci/tank_vs_a05/by_station_pct/fairbanks\endcsname{$-28.5$}
\expandafter\def\csname F:stress/s7_ci/tank_vs_a05/by_station_pct/fairbanks\endcsname{-28.5000}
\expandafter\def\csname FF:stress/s7_ci/tank_vs_a05/by_station_pct/fairbanks\endcsname{-28.50000}
\expandafter\def\csname N:stress/s7_ci/tank_vs_a05/by_station_pct/denver\endcsname{$-22.3$}
\expandafter\def\csname F:stress/s7_ci/tank_vs_a05/by_station_pct/denver\endcsname{-22.3000}
\expandafter\def\csname FF:stress/s7_ci/tank_vs_a05/by_station_pct/denver\endcsname{-22.30000}
\expandafter\def\csname N:stress/s7_ci/tank_vs_a05/by_station_pct/atlanta\endcsname{$-14.5$}
\expandafter\def\csname F:stress/s7_ci/tank_vs_a05/by_station_pct/atlanta\endcsname{-14.5000}
\expandafter\def\csname FF:stress/s7_ci/tank_vs_a05/by_station_pct/atlanta\endcsname{-14.50000}
\expandafter\def\csname N:stress/s7_ci/tank_vs_a05/by_station_pct/phoenix\endcsname{$-6.0$}
\expandafter\def\csname F:stress/s7_ci/tank_vs_a05/by_station_pct/phoenix\endcsname{-6.0000}
\expandafter\def\csname FF:stress/s7_ci/tank_vs_a05/by_station_pct/phoenix\endcsname{-6.00000}
\expandafter\def\csname N:stress/s7_ci/tank_vs_a05/by_station_pct/singapore\endcsname{$0.0$}
\expandafter\def\csname F:stress/s7_ci/tank_vs_a05/by_station_pct/singapore\endcsname{0.0000}
\expandafter\def\csname FF:stress/s7_ci/tank_vs_a05/by_station_pct/singapore\endcsname{0.00000}
\expandafter\def\csname N:stress/s7_ci/tank_vs_a05/by_station_pct/minneapolis\endcsname{$-25.1$}
\expandafter\def\csname F:stress/s7_ci/tank_vs_a05/by_station_pct/minneapolis\endcsname{-25.1000}
\expandafter\def\csname FF:stress/s7_ci/tank_vs_a05/by_station_pct/minneapolis\endcsname{-25.10000}
\expandafter\def\csname N:stress/s7_ci/tank_vs_a05/wins\endcsname{37}
\expandafter\def\csname N:stress/s7_ci/tank_vs_econ/n\endcsname{48}
\expandafter\def\csname N:stress/s7_ci/tank_vs_econ/mean\endcsname{$-0.00273$}
\expandafter\def\csname F:stress/s7_ci/tank_vs_econ/mean\endcsname{-0.0027}
\expandafter\def\csname FF:stress/s7_ci/tank_vs_econ/mean\endcsname{-0.00273}
\expandafter\def\csname N:stress/s7_ci/tank_vs_econ/lo\endcsname{$-0.00446$}
\expandafter\def\csname F:stress/s7_ci/tank_vs_econ/lo\endcsname{-0.0045}
\expandafter\def\csname FF:stress/s7_ci/tank_vs_econ/lo\endcsname{-0.00446}
\expandafter\def\csname N:stress/s7_ci/tank_vs_econ/hi\endcsname{$-0.00126$}
\expandafter\def\csname F:stress/s7_ci/tank_vs_econ/hi\endcsname{-0.0013}
\expandafter\def\csname FF:stress/s7_ci/tank_vs_econ/hi\endcsname{-0.00126}
\expandafter\def\csname N:stress/s7_ci/tank_vs_econ/pct\endcsname{$-20.8$}
\expandafter\def\csname F:stress/s7_ci/tank_vs_econ/pct\endcsname{-20.8000}
\expandafter\def\csname FF:stress/s7_ci/tank_vs_econ/pct\endcsname{-20.80000}
\expandafter\def\csname N:stress/s7_ci/tank_vs_econ/pct_lo\endcsname{$-34.0$}
\expandafter\def\csname F:stress/s7_ci/tank_vs_econ/pct_lo\endcsname{-34.0000}
\expandafter\def\csname FF:stress/s7_ci/tank_vs_econ/pct_lo\endcsname{-34.00000}
\expandafter\def\csname N:stress/s7_ci/tank_vs_econ/pct_hi\endcsname{$-9.6$}
\expandafter\def\csname F:stress/s7_ci/tank_vs_econ/pct_hi\endcsname{-9.6000}
\expandafter\def\csname FF:stress/s7_ci/tank_vs_econ/pct_hi\endcsname{-9.60000}
\expandafter\def\csname N:stress/s7_ci/tank_vs_econ/by_station_pct/fairbanks\endcsname{$-24.7$}
\expandafter\def\csname F:stress/s7_ci/tank_vs_econ/by_station_pct/fairbanks\endcsname{-24.7000}
\expandafter\def\csname FF:stress/s7_ci/tank_vs_econ/by_station_pct/fairbanks\endcsname{-24.70000}
\expandafter\def\csname N:stress/s7_ci/tank_vs_econ/by_station_pct/denver\endcsname{$-17.4$}
\expandafter\def\csname F:stress/s7_ci/tank_vs_econ/by_station_pct/denver\endcsname{-17.4000}
\expandafter\def\csname FF:stress/s7_ci/tank_vs_econ/by_station_pct/denver\endcsname{-17.40000}
\expandafter\def\csname N:stress/s7_ci/tank_vs_econ/by_station_pct/atlanta\endcsname{$-4.3$}
\expandafter\def\csname F:stress/s7_ci/tank_vs_econ/by_station_pct/atlanta\endcsname{-4.3000}
\expandafter\def\csname FF:stress/s7_ci/tank_vs_econ/by_station_pct/atlanta\endcsname{-4.30000}
\expandafter\def\csname N:stress/s7_ci/tank_vs_econ/by_station_pct/phoenix\endcsname{$-1.3$}
\expandafter\def\csname F:stress/s7_ci/tank_vs_econ/by_station_pct/phoenix\endcsname{-1.3000}
\expandafter\def\csname FF:stress/s7_ci/tank_vs_econ/by_station_pct/phoenix\endcsname{-1.30000}
\expandafter\def\csname N:stress/s7_ci/tank_vs_econ/by_station_pct/singapore\endcsname{$0.0$}
\expandafter\def\csname F:stress/s7_ci/tank_vs_econ/by_station_pct/singapore\endcsname{0.0000}
\expandafter\def\csname FF:stress/s7_ci/tank_vs_econ/by_station_pct/singapore\endcsname{0.00000}
\expandafter\def\csname N:stress/s7_ci/tank_vs_econ/by_station_pct/minneapolis\endcsname{$-17.7$}
\expandafter\def\csname F:stress/s7_ci/tank_vs_econ/by_station_pct/minneapolis\endcsname{-17.7000}
\expandafter\def\csname FF:stress/s7_ci/tank_vs_econ/by_station_pct/minneapolis\endcsname{-17.70000}
\expandafter\def\csname N:stress/s7_ci/tank_vs_econ/wins\endcsname{28}
\expandafter\def\csname N:stress/s7_ci/zcfix_vs_zc/n\endcsname{48}
\expandafter\def\csname N:stress/s7_ci/zcfix_vs_zc/mean\endcsname{0.00025}
\expandafter\def\csname F:stress/s7_ci/zcfix_vs_zc/mean\endcsname{0.0003}
\expandafter\def\csname FF:stress/s7_ci/zcfix_vs_zc/mean\endcsname{0.00025}
\expandafter\def\csname N:stress/s7_ci/zcfix_vs_zc/lo\endcsname{0.00013}
\expandafter\def\csname F:stress/s7_ci/zcfix_vs_zc/lo\endcsname{0.0001}
\expandafter\def\csname FF:stress/s7_ci/zcfix_vs_zc/lo\endcsname{0.00013}
\expandafter\def\csname N:stress/s7_ci/zcfix_vs_zc/hi\endcsname{0.00039}
\expandafter\def\csname F:stress/s7_ci/zcfix_vs_zc/hi\endcsname{0.0004}
\expandafter\def\csname FF:stress/s7_ci/zcfix_vs_zc/hi\endcsname{0.00039}
\expandafter\def\csname N:stress/s7_ci/zcfix_vs_zc/pct\endcsname{$+2.0$}
\expandafter\def\csname F:stress/s7_ci/zcfix_vs_zc/pct\endcsname{2.0000}
\expandafter\def\csname FF:stress/s7_ci/zcfix_vs_zc/pct\endcsname{2.00000}
\expandafter\def\csname N:stress/s7_ci/zcfix_vs_zc/pct_lo\endcsname{$+1.0$}
\expandafter\def\csname F:stress/s7_ci/zcfix_vs_zc/pct_lo\endcsname{1.0000}
\expandafter\def\csname FF:stress/s7_ci/zcfix_vs_zc/pct_lo\endcsname{1.00000}
\expandafter\def\csname N:stress/s7_ci/zcfix_vs_zc/pct_hi\endcsname{$+3.0$}
\expandafter\def\csname F:stress/s7_ci/zcfix_vs_zc/pct_hi\endcsname{3.0000}
\expandafter\def\csname FF:stress/s7_ci/zcfix_vs_zc/pct_hi\endcsname{3.00000}
\expandafter\def\csname N:stress/s7_ci/zcfix_vs_zc/by_station_pct/fairbanks\endcsname{$+3.5$}
\expandafter\def\csname F:stress/s7_ci/zcfix_vs_zc/by_station_pct/fairbanks\endcsname{3.5000}
\expandafter\def\csname FF:stress/s7_ci/zcfix_vs_zc/by_station_pct/fairbanks\endcsname{3.50000}
\expandafter\def\csname N:stress/s7_ci/zcfix_vs_zc/by_station_pct/denver\endcsname{$+6.0$}
\expandafter\def\csname F:stress/s7_ci/zcfix_vs_zc/by_station_pct/denver\endcsname{6.0000}
\expandafter\def\csname FF:stress/s7_ci/zcfix_vs_zc/by_station_pct/denver\endcsname{6.00000}
\expandafter\def\csname N:stress/s7_ci/zcfix_vs_zc/by_station_pct/atlanta\endcsname{$+2.7$}
\expandafter\def\csname F:stress/s7_ci/zcfix_vs_zc/by_station_pct/atlanta\endcsname{2.7000}
\expandafter\def\csname FF:stress/s7_ci/zcfix_vs_zc/by_station_pct/atlanta\endcsname{2.70000}
\expandafter\def\csname N:stress/s7_ci/zcfix_vs_zc/by_station_pct/phoenix\endcsname{$+3.3$}
\expandafter\def\csname F:stress/s7_ci/zcfix_vs_zc/by_station_pct/phoenix\endcsname{3.3000}
\expandafter\def\csname FF:stress/s7_ci/zcfix_vs_zc/by_station_pct/phoenix\endcsname{3.30000}
\expandafter\def\csname N:stress/s7_ci/zcfix_vs_zc/by_station_pct/singapore\endcsname{$0.0$}
\expandafter\def\csname F:stress/s7_ci/zcfix_vs_zc/by_station_pct/singapore\endcsname{0.0000}
\expandafter\def\csname FF:stress/s7_ci/zcfix_vs_zc/by_station_pct/singapore\endcsname{0.00000}
\expandafter\def\csname N:stress/s7_ci/zcfix_vs_zc/by_station_pct/minneapolis\endcsname{$+2.1$}
\expandafter\def\csname F:stress/s7_ci/zcfix_vs_zc/by_station_pct/minneapolis\endcsname{2.1000}
\expandafter\def\csname FF:stress/s7_ci/zcfix_vs_zc/by_station_pct/minneapolis\endcsname{2.10000}
\expandafter\def\csname N:stress/s7_ci/zcfix_vs_zc/wins\endcsname{13}
\expandafter\def\csname N:stress/s7_ci/tankfix_vs_zcfix/n\endcsname{48}
\expandafter\def\csname N:stress/s7_ci/tankfix_vs_zcfix/mean\endcsname{0.00011}
\expandafter\def\csname F:stress/s7_ci/tankfix_vs_zcfix/mean\endcsname{0.0001}
\expandafter\def\csname FF:stress/s7_ci/tankfix_vs_zcfix/mean\endcsname{0.00011}
\expandafter\def\csname N:stress/s7_ci/tankfix_vs_zcfix/lo\endcsname{5e-05}
\expandafter\def\csname F:stress/s7_ci/tankfix_vs_zcfix/lo\endcsname{0.0001}
\expandafter\def\csname FF:stress/s7_ci/tankfix_vs_zcfix/lo\endcsname{0.00005}
\expandafter\def\csname N:stress/s7_ci/tankfix_vs_zcfix/hi\endcsname{0.00017}
\expandafter\def\csname F:stress/s7_ci/tankfix_vs_zcfix/hi\endcsname{0.0002}
\expandafter\def\csname FF:stress/s7_ci/tankfix_vs_zcfix/hi\endcsname{0.00017}
\expandafter\def\csname N:stress/s7_ci/tankfix_vs_zcfix/pct\endcsname{$+0.8$}
\expandafter\def\csname F:stress/s7_ci/tankfix_vs_zcfix/pct\endcsname{0.8000}
\expandafter\def\csname FF:stress/s7_ci/tankfix_vs_zcfix/pct\endcsname{0.80000}
\expandafter\def\csname N:stress/s7_ci/tankfix_vs_zcfix/pct_lo\endcsname{$+0.4$}
\expandafter\def\csname F:stress/s7_ci/tankfix_vs_zcfix/pct_lo\endcsname{0.4000}
\expandafter\def\csname FF:stress/s7_ci/tankfix_vs_zcfix/pct_lo\endcsname{0.40000}
\expandafter\def\csname N:stress/s7_ci/tankfix_vs_zcfix/pct_hi\endcsname{$+1.3$}
\expandafter\def\csname F:stress/s7_ci/tankfix_vs_zcfix/pct_hi\endcsname{1.3000}
\expandafter\def\csname FF:stress/s7_ci/tankfix_vs_zcfix/pct_hi\endcsname{1.30000}
\expandafter\def\csname N:stress/s7_ci/tankfix_vs_zcfix/by_station_pct/fairbanks\endcsname{$+1.1$}
\expandafter\def\csname F:stress/s7_ci/tankfix_vs_zcfix/by_station_pct/fairbanks\endcsname{1.1000}
\expandafter\def\csname FF:stress/s7_ci/tankfix_vs_zcfix/by_station_pct/fairbanks\endcsname{1.10000}
\expandafter\def\csname N:stress/s7_ci/tankfix_vs_zcfix/by_station_pct/denver\endcsname{$+2.5$}
\expandafter\def\csname F:stress/s7_ci/tankfix_vs_zcfix/by_station_pct/denver\endcsname{2.5000}
\expandafter\def\csname FF:stress/s7_ci/tankfix_vs_zcfix/by_station_pct/denver\endcsname{2.50000}
\expandafter\def\csname N:stress/s7_ci/tankfix_vs_zcfix/by_station_pct/atlanta\endcsname{$+3.0$}
\expandafter\def\csname F:stress/s7_ci/tankfix_vs_zcfix/by_station_pct/atlanta\endcsname{3.0000}
\expandafter\def\csname FF:stress/s7_ci/tankfix_vs_zcfix/by_station_pct/atlanta\endcsname{3.00000}
\expandafter\def\csname N:stress/s7_ci/tankfix_vs_zcfix/by_station_pct/phoenix\endcsname{$+1.8$}
\expandafter\def\csname F:stress/s7_ci/tankfix_vs_zcfix/by_station_pct/phoenix\endcsname{1.8000}
\expandafter\def\csname FF:stress/s7_ci/tankfix_vs_zcfix/by_station_pct/phoenix\endcsname{1.80000}
\expandafter\def\csname N:stress/s7_ci/tankfix_vs_zcfix/by_station_pct/singapore\endcsname{$+4.3$}
\expandafter\def\csname F:stress/s7_ci/tankfix_vs_zcfix/by_station_pct/singapore\endcsname{4.3000}
\expandafter\def\csname FF:stress/s7_ci/tankfix_vs_zcfix/by_station_pct/singapore\endcsname{4.30000}
\expandafter\def\csname N:stress/s7_ci/tankfix_vs_zcfix/by_station_pct/minneapolis\endcsname{$+1.5$}
\expandafter\def\csname F:stress/s7_ci/tankfix_vs_zcfix/by_station_pct/minneapolis\endcsname{1.5000}
\expandafter\def\csname FF:stress/s7_ci/tankfix_vs_zcfix/by_station_pct/minneapolis\endcsname{1.50000}
\expandafter\def\csname N:stress/s7_ci/tankfix_vs_zcfix/wins\endcsname{15}
\expandafter\def\csname N:stress/s7_ci/zc_vs_c5/n\endcsname{48}
\expandafter\def\csname N:stress/s7_ci/zc_vs_c5/mean\endcsname{$-0.00292$}
\expandafter\def\csname F:stress/s7_ci/zc_vs_c5/mean\endcsname{-0.0029}
\expandafter\def\csname FF:stress/s7_ci/zc_vs_c5/mean\endcsname{-0.00292}
\expandafter\def\csname N:stress/s7_ci/zc_vs_c5/lo\endcsname{$-0.0051$}
\expandafter\def\csname F:stress/s7_ci/zc_vs_c5/lo\endcsname{-0.0051}
\expandafter\def\csname FF:stress/s7_ci/zc_vs_c5/lo\endcsname{-0.00510}
\expandafter\def\csname N:stress/s7_ci/zc_vs_c5/hi\endcsname{$-0.0012$}
\expandafter\def\csname F:stress/s7_ci/zc_vs_c5/hi\endcsname{-0.0012}
\expandafter\def\csname FF:stress/s7_ci/zc_vs_c5/hi\endcsname{-0.00120}
\expandafter\def\csname N:stress/s7_ci/zc_vs_c5/pct\endcsname{$-18.3$}
\expandafter\def\csname F:stress/s7_ci/zc_vs_c5/pct\endcsname{-18.3000}
\expandafter\def\csname FF:stress/s7_ci/zc_vs_c5/pct\endcsname{-18.30000}
\expandafter\def\csname N:stress/s7_ci/zc_vs_c5/pct_lo\endcsname{$-32.0$}
\expandafter\def\csname F:stress/s7_ci/zc_vs_c5/pct_lo\endcsname{-32.0000}
\expandafter\def\csname FF:stress/s7_ci/zc_vs_c5/pct_lo\endcsname{-32.00000}
\expandafter\def\csname N:stress/s7_ci/zc_vs_c5/pct_hi\endcsname{$-7.5$}
\expandafter\def\csname F:stress/s7_ci/zc_vs_c5/pct_hi\endcsname{-7.5000}
\expandafter\def\csname FF:stress/s7_ci/zc_vs_c5/pct_hi\endcsname{-7.50000}
\expandafter\def\csname N:stress/s7_ci/zc_vs_c5/by_station_pct/fairbanks\endcsname{$-3.8$}
\expandafter\def\csname F:stress/s7_ci/zc_vs_c5/by_station_pct/fairbanks\endcsname{-3.8000}
\expandafter\def\csname FF:stress/s7_ci/zc_vs_c5/by_station_pct/fairbanks\endcsname{-3.80000}
\expandafter\def\csname N:stress/s7_ci/zc_vs_c5/by_station_pct/denver\endcsname{$-5.8$}
\expandafter\def\csname F:stress/s7_ci/zc_vs_c5/by_station_pct/denver\endcsname{-5.8000}
\expandafter\def\csname FF:stress/s7_ci/zc_vs_c5/by_station_pct/denver\endcsname{-5.80000}
\expandafter\def\csname N:stress/s7_ci/zc_vs_c5/by_station_pct/atlanta\endcsname{$-20.7$}
\expandafter\def\csname F:stress/s7_ci/zc_vs_c5/by_station_pct/atlanta\endcsname{-20.7000}
\expandafter\def\csname FF:stress/s7_ci/zc_vs_c5/by_station_pct/atlanta\endcsname{-20.70000}
\expandafter\def\csname N:stress/s7_ci/zc_vs_c5/by_station_pct/phoenix\endcsname{$-23.7$}
\expandafter\def\csname F:stress/s7_ci/zc_vs_c5/by_station_pct/phoenix\endcsname{-23.7000}
\expandafter\def\csname FF:stress/s7_ci/zc_vs_c5/by_station_pct/phoenix\endcsname{-23.70000}
\expandafter\def\csname N:stress/s7_ci/zc_vs_c5/by_station_pct/singapore\endcsname{$-11.3$}
\expandafter\def\csname F:stress/s7_ci/zc_vs_c5/by_station_pct/singapore\endcsname{-11.3000}
\expandafter\def\csname FF:stress/s7_ci/zc_vs_c5/by_station_pct/singapore\endcsname{-11.30000}
\expandafter\def\csname N:stress/s7_ci/zc_vs_c5/by_station_pct/minneapolis\endcsname{$-15.1$}
\expandafter\def\csname F:stress/s7_ci/zc_vs_c5/by_station_pct/minneapolis\endcsname{-15.1000}
\expandafter\def\csname FF:stress/s7_ci/zc_vs_c5/by_station_pct/minneapolis\endcsname{-15.10000}
\expandafter\def\csname N:stress/s7_ci/zc_vs_c5/wins\endcsname{40}
\expandafter\def\csname N:stress/s7_ci/area1_vs_zc/n\endcsname{48}
\expandafter\def\csname N:stress/s7_ci/area1_vs_zc/mean\endcsname{$-0.0005$}
\expandafter\def\csname F:stress/s7_ci/area1_vs_zc/mean\endcsname{-0.0005}
\expandafter\def\csname FF:stress/s7_ci/area1_vs_zc/mean\endcsname{-0.00050}
\expandafter\def\csname N:stress/s7_ci/area1_vs_zc/lo\endcsname{$-0.00096$}
\expandafter\def\csname F:stress/s7_ci/area1_vs_zc/lo\endcsname{-0.0010}
\expandafter\def\csname FF:stress/s7_ci/area1_vs_zc/lo\endcsname{-0.00096}
\expandafter\def\csname N:stress/s7_ci/area1_vs_zc/hi\endcsname{$-9e-05$}
\expandafter\def\csname F:stress/s7_ci/area1_vs_zc/hi\endcsname{-0.0001}
\expandafter\def\csname FF:stress/s7_ci/area1_vs_zc/hi\endcsname{-0.00009}
\expandafter\def\csname N:stress/s7_ci/area1_vs_zc/pct\endcsname{$-3.9$}
\expandafter\def\csname F:stress/s7_ci/area1_vs_zc/pct\endcsname{-3.9000}
\expandafter\def\csname FF:stress/s7_ci/area1_vs_zc/pct\endcsname{-3.90000}
\expandafter\def\csname N:stress/s7_ci/area1_vs_zc/pct_lo\endcsname{$-7.4$}
\expandafter\def\csname F:stress/s7_ci/area1_vs_zc/pct_lo\endcsname{-7.4000}
\expandafter\def\csname FF:stress/s7_ci/area1_vs_zc/pct_lo\endcsname{-7.40000}
\expandafter\def\csname N:stress/s7_ci/area1_vs_zc/pct_hi\endcsname{$-0.7$}
\expandafter\def\csname F:stress/s7_ci/area1_vs_zc/pct_hi\endcsname{-0.7000}
\expandafter\def\csname FF:stress/s7_ci/area1_vs_zc/pct_hi\endcsname{-0.70000}
\expandafter\def\csname N:stress/s7_ci/area1_vs_zc/by_station_pct/fairbanks\endcsname{$-4.8$}
\expandafter\def\csname F:stress/s7_ci/area1_vs_zc/by_station_pct/fairbanks\endcsname{-4.8000}
\expandafter\def\csname FF:stress/s7_ci/area1_vs_zc/by_station_pct/fairbanks\endcsname{-4.80000}
\expandafter\def\csname N:stress/s7_ci/area1_vs_zc/by_station_pct/denver\endcsname{$-8.7$}
\expandafter\def\csname F:stress/s7_ci/area1_vs_zc/by_station_pct/denver\endcsname{-8.7000}
\expandafter\def\csname FF:stress/s7_ci/area1_vs_zc/by_station_pct/denver\endcsname{-8.70000}
\expandafter\def\csname N:stress/s7_ci/area1_vs_zc/by_station_pct/atlanta\endcsname{$-5.7$}
\expandafter\def\csname F:stress/s7_ci/area1_vs_zc/by_station_pct/atlanta\endcsname{-5.7000}
\expandafter\def\csname FF:stress/s7_ci/area1_vs_zc/by_station_pct/atlanta\endcsname{-5.70000}
\expandafter\def\csname N:stress/s7_ci/area1_vs_zc/by_station_pct/phoenix\endcsname{$+6.0$}
\expandafter\def\csname F:stress/s7_ci/area1_vs_zc/by_station_pct/phoenix\endcsname{6.0000}
\expandafter\def\csname FF:stress/s7_ci/area1_vs_zc/by_station_pct/phoenix\endcsname{6.00000}
\expandafter\def\csname N:stress/s7_ci/area1_vs_zc/by_station_pct/singapore\endcsname{$+2.3$}
\expandafter\def\csname F:stress/s7_ci/area1_vs_zc/by_station_pct/singapore\endcsname{2.3000}
\expandafter\def\csname FF:stress/s7_ci/area1_vs_zc/by_station_pct/singapore\endcsname{2.30000}
\expandafter\def\csname N:stress/s7_ci/area1_vs_zc/by_station_pct/minneapolis\endcsname{$-4.3$}
\expandafter\def\csname F:stress/s7_ci/area1_vs_zc/by_station_pct/minneapolis\endcsname{-4.3000}
\expandafter\def\csname FF:stress/s7_ci/area1_vs_zc/by_station_pct/minneapolis\endcsname{-4.30000}
\expandafter\def\csname N:stress/s7_ci/area1_vs_zc/wins\endcsname{26}
\expandafter\def\csname N:stress/s7_ci/inv_vs_zc/n\endcsname{48}
\expandafter\def\csname N:stress/s7_ci/inv_vs_zc/mean\endcsname{$-0.00487$}
\expandafter\def\csname F:stress/s7_ci/inv_vs_zc/mean\endcsname{-0.0049}
\expandafter\def\csname FF:stress/s7_ci/inv_vs_zc/mean\endcsname{-0.00487}
\expandafter\def\csname N:stress/s7_ci/inv_vs_zc/lo\endcsname{$-0.00666$}
\expandafter\def\csname F:stress/s7_ci/inv_vs_zc/lo\endcsname{-0.0067}
\expandafter\def\csname FF:stress/s7_ci/inv_vs_zc/lo\endcsname{-0.00666}
\expandafter\def\csname N:stress/s7_ci/inv_vs_zc/hi\endcsname{$-0.00327$}
\expandafter\def\csname F:stress/s7_ci/inv_vs_zc/hi\endcsname{-0.0033}
\expandafter\def\csname FF:stress/s7_ci/inv_vs_zc/hi\endcsname{-0.00327}
\expandafter\def\csname N:stress/s7_ci/inv_vs_zc/pct\endcsname{$-37.5$}
\expandafter\def\csname F:stress/s7_ci/inv_vs_zc/pct\endcsname{-37.5000}
\expandafter\def\csname FF:stress/s7_ci/inv_vs_zc/pct\endcsname{-37.50000}
\expandafter\def\csname N:stress/s7_ci/inv_vs_zc/pct_lo\endcsname{$-51.2$}
\expandafter\def\csname F:stress/s7_ci/inv_vs_zc/pct_lo\endcsname{-51.2000}
\expandafter\def\csname FF:stress/s7_ci/inv_vs_zc/pct_lo\endcsname{-51.20000}
\expandafter\def\csname N:stress/s7_ci/inv_vs_zc/pct_hi\endcsname{$-25.2$}
\expandafter\def\csname F:stress/s7_ci/inv_vs_zc/pct_hi\endcsname{-25.2000}
\expandafter\def\csname FF:stress/s7_ci/inv_vs_zc/pct_hi\endcsname{-25.20000}
\expandafter\def\csname N:stress/s7_ci/inv_vs_zc/by_station_pct/fairbanks\endcsname{$-45.2$}
\expandafter\def\csname F:stress/s7_ci/inv_vs_zc/by_station_pct/fairbanks\endcsname{-45.2000}
\expandafter\def\csname FF:stress/s7_ci/inv_vs_zc/by_station_pct/fairbanks\endcsname{-45.20000}
\expandafter\def\csname N:stress/s7_ci/inv_vs_zc/by_station_pct/denver\endcsname{$-43.9$}
\expandafter\def\csname F:stress/s7_ci/inv_vs_zc/by_station_pct/denver\endcsname{-43.9000}
\expandafter\def\csname FF:stress/s7_ci/inv_vs_zc/by_station_pct/denver\endcsname{-43.90000}
\expandafter\def\csname N:stress/s7_ci/inv_vs_zc/by_station_pct/atlanta\endcsname{$-27.9$}
\expandafter\def\csname F:stress/s7_ci/inv_vs_zc/by_station_pct/atlanta\endcsname{-27.9000}
\expandafter\def\csname FF:stress/s7_ci/inv_vs_zc/by_station_pct/atlanta\endcsname{-27.90000}
\expandafter\def\csname N:stress/s7_ci/inv_vs_zc/by_station_pct/phoenix\endcsname{$-26.8$}
\expandafter\def\csname F:stress/s7_ci/inv_vs_zc/by_station_pct/phoenix\endcsname{-26.8000}
\expandafter\def\csname FF:stress/s7_ci/inv_vs_zc/by_station_pct/phoenix\endcsname{-26.80000}
\expandafter\def\csname N:stress/s7_ci/inv_vs_zc/by_station_pct/singapore\endcsname{$-21.2$}
\expandafter\def\csname F:stress/s7_ci/inv_vs_zc/by_station_pct/singapore\endcsname{-21.2000}
\expandafter\def\csname FF:stress/s7_ci/inv_vs_zc/by_station_pct/singapore\endcsname{-21.20000}
\expandafter\def\csname N:stress/s7_ci/inv_vs_zc/by_station_pct/minneapolis\endcsname{$-38.6$}
\expandafter\def\csname F:stress/s7_ci/inv_vs_zc/by_station_pct/minneapolis\endcsname{-38.6000}
\expandafter\def\csname FF:stress/s7_ci/inv_vs_zc/by_station_pct/minneapolis\endcsname{-38.60000}
\expandafter\def\csname N:stress/s7_ci/inv_vs_zc/wins\endcsname{48}
\expandafter\def\csname N:stress/s7_ci/inv_vs_ctrlw5/n\endcsname{48}
\expandafter\def\csname N:stress/s7_ci/inv_vs_ctrlw5/mean\endcsname{8e-05}
\expandafter\def\csname F:stress/s7_ci/inv_vs_ctrlw5/mean\endcsname{0.0001}
\expandafter\def\csname FF:stress/s7_ci/inv_vs_ctrlw5/mean\endcsname{0.00008}
\expandafter\def\csname N:stress/s7_ci/inv_vs_ctrlw5/lo\endcsname{4e-05}
\expandafter\def\csname F:stress/s7_ci/inv_vs_ctrlw5/lo\endcsname{0.0000}
\expandafter\def\csname FF:stress/s7_ci/inv_vs_ctrlw5/lo\endcsname{0.00004}
\expandafter\def\csname N:stress/s7_ci/inv_vs_ctrlw5/hi\endcsname{0.00012}
\expandafter\def\csname F:stress/s7_ci/inv_vs_ctrlw5/hi\endcsname{0.0001}
\expandafter\def\csname FF:stress/s7_ci/inv_vs_ctrlw5/hi\endcsname{0.00012}
\expandafter\def\csname N:stress/s7_ci/inv_vs_ctrlw5/pct\endcsname{$+1.0$}
\expandafter\def\csname F:stress/s7_ci/inv_vs_ctrlw5/pct\endcsname{1.0000}
\expandafter\def\csname FF:stress/s7_ci/inv_vs_ctrlw5/pct\endcsname{1.00000}
\expandafter\def\csname N:stress/s7_ci/inv_vs_ctrlw5/pct_lo\endcsname{$+0.4$}
\expandafter\def\csname F:stress/s7_ci/inv_vs_ctrlw5/pct_lo\endcsname{0.4000}
\expandafter\def\csname FF:stress/s7_ci/inv_vs_ctrlw5/pct_lo\endcsname{0.40000}
\expandafter\def\csname N:stress/s7_ci/inv_vs_ctrlw5/pct_hi\endcsname{$+1.5$}
\expandafter\def\csname F:stress/s7_ci/inv_vs_ctrlw5/pct_hi\endcsname{1.5000}
\expandafter\def\csname FF:stress/s7_ci/inv_vs_ctrlw5/pct_hi\endcsname{1.50000}
\expandafter\def\csname N:stress/s7_ci/inv_vs_ctrlw5/by_station_pct/fairbanks\endcsname{$+1.9$}
\expandafter\def\csname F:stress/s7_ci/inv_vs_ctrlw5/by_station_pct/fairbanks\endcsname{1.9000}
\expandafter\def\csname FF:stress/s7_ci/inv_vs_ctrlw5/by_station_pct/fairbanks\endcsname{1.90000}
\expandafter\def\csname N:stress/s7_ci/inv_vs_ctrlw5/by_station_pct/denver\endcsname{$+1.8$}
\expandafter\def\csname F:stress/s7_ci/inv_vs_ctrlw5/by_station_pct/denver\endcsname{1.8000}
\expandafter\def\csname FF:stress/s7_ci/inv_vs_ctrlw5/by_station_pct/denver\endcsname{1.80000}
\expandafter\def\csname N:stress/s7_ci/inv_vs_ctrlw5/by_station_pct/atlanta\endcsname{$+1.9$}
\expandafter\def\csname F:stress/s7_ci/inv_vs_ctrlw5/by_station_pct/atlanta\endcsname{1.9000}
\expandafter\def\csname FF:stress/s7_ci/inv_vs_ctrlw5/by_station_pct/atlanta\endcsname{1.90000}
\expandafter\def\csname N:stress/s7_ci/inv_vs_ctrlw5/by_station_pct/phoenix\endcsname{$+0.7$}
\expandafter\def\csname F:stress/s7_ci/inv_vs_ctrlw5/by_station_pct/phoenix\endcsname{0.7000}
\expandafter\def\csname FF:stress/s7_ci/inv_vs_ctrlw5/by_station_pct/phoenix\endcsname{0.70000}
\expandafter\def\csname N:stress/s7_ci/inv_vs_ctrlw5/by_station_pct/singapore\endcsname{$-0.8$}
\expandafter\def\csname F:stress/s7_ci/inv_vs_ctrlw5/by_station_pct/singapore\endcsname{-0.8000}
\expandafter\def\csname FF:stress/s7_ci/inv_vs_ctrlw5/by_station_pct/singapore\endcsname{-0.80000}
\expandafter\def\csname N:stress/s7_ci/inv_vs_ctrlw5/by_station_pct/minneapolis\endcsname{$+1.5$}
\expandafter\def\csname F:stress/s7_ci/inv_vs_ctrlw5/by_station_pct/minneapolis\endcsname{1.5000}
\expandafter\def\csname FF:stress/s7_ci/inv_vs_ctrlw5/by_station_pct/minneapolis\endcsname{1.50000}
\expandafter\def\csname N:stress/s7_ci/inv_vs_ctrlw5/wins\endcsname{12}
\expandafter\def\csname W:stress/s7_ci/inv_vs_ctrlw5/wins\endcsname{twelve}
\expandafter\def\csname N:stress/s7_ci/ctrlw5_vs_zc/n\endcsname{48}
\expandafter\def\csname N:stress/s7_ci/ctrlw5_vs_zc/mean\endcsname{$-0.00495$}
\expandafter\def\csname F:stress/s7_ci/ctrlw5_vs_zc/mean\endcsname{-0.0050}
\expandafter\def\csname FF:stress/s7_ci/ctrlw5_vs_zc/mean\endcsname{-0.00495}
\expandafter\def\csname N:stress/s7_ci/ctrlw5_vs_zc/lo\endcsname{$-0.00677$}
\expandafter\def\csname F:stress/s7_ci/ctrlw5_vs_zc/lo\endcsname{-0.0068}
\expandafter\def\csname FF:stress/s7_ci/ctrlw5_vs_zc/lo\endcsname{-0.00677}
\expandafter\def\csname N:stress/s7_ci/ctrlw5_vs_zc/hi\endcsname{$-0.00332$}
\expandafter\def\csname F:stress/s7_ci/ctrlw5_vs_zc/hi\endcsname{-0.0033}
\expandafter\def\csname FF:stress/s7_ci/ctrlw5_vs_zc/hi\endcsname{-0.00332}
\expandafter\def\csname N:stress/s7_ci/ctrlw5_vs_zc/pct\endcsname{$-38.1$}
\expandafter\def\csname F:stress/s7_ci/ctrlw5_vs_zc/pct\endcsname{-38.1000}
\expandafter\def\csname FF:stress/s7_ci/ctrlw5_vs_zc/pct\endcsname{-38.10000}
\expandafter\def\csname N:stress/s7_ci/ctrlw5_vs_zc/pct_lo\endcsname{$-52.1$}
\expandafter\def\csname F:stress/s7_ci/ctrlw5_vs_zc/pct_lo\endcsname{-52.1000}
\expandafter\def\csname FF:stress/s7_ci/ctrlw5_vs_zc/pct_lo\endcsname{-52.10000}
\expandafter\def\csname N:stress/s7_ci/ctrlw5_vs_zc/pct_hi\endcsname{$-25.6$}
\expandafter\def\csname F:stress/s7_ci/ctrlw5_vs_zc/pct_hi\endcsname{-25.6000}
\expandafter\def\csname FF:stress/s7_ci/ctrlw5_vs_zc/pct_hi\endcsname{-25.60000}
\expandafter\def\csname N:stress/s7_ci/ctrlw5_vs_zc/by_station_pct/fairbanks\endcsname{$-45.9$}
\expandafter\def\csname F:stress/s7_ci/ctrlw5_vs_zc/by_station_pct/fairbanks\endcsname{-45.9000}
\expandafter\def\csname FF:stress/s7_ci/ctrlw5_vs_zc/by_station_pct/fairbanks\endcsname{-45.90000}
\expandafter\def\csname N:stress/s7_ci/ctrlw5_vs_zc/by_station_pct/denver\endcsname{$-44.6$}
\expandafter\def\csname F:stress/s7_ci/ctrlw5_vs_zc/by_station_pct/denver\endcsname{-44.6000}
\expandafter\def\csname FF:stress/s7_ci/ctrlw5_vs_zc/by_station_pct/denver\endcsname{-44.60000}
\expandafter\def\csname N:stress/s7_ci/ctrlw5_vs_zc/by_station_pct/atlanta\endcsname{$-29.0$}
\expandafter\def\csname F:stress/s7_ci/ctrlw5_vs_zc/by_station_pct/atlanta\endcsname{-29.0000}
\expandafter\def\csname FF:stress/s7_ci/ctrlw5_vs_zc/by_station_pct/atlanta\endcsname{-29.00000}
\expandafter\def\csname N:stress/s7_ci/ctrlw5_vs_zc/by_station_pct/phoenix\endcsname{$-27.2$}
\expandafter\def\csname F:stress/s7_ci/ctrlw5_vs_zc/by_station_pct/phoenix\endcsname{-27.2000}
\expandafter\def\csname FF:stress/s7_ci/ctrlw5_vs_zc/by_station_pct/phoenix\endcsname{-27.20000}
\expandafter\def\csname N:stress/s7_ci/ctrlw5_vs_zc/by_station_pct/singapore\endcsname{$-20.5$}
\expandafter\def\csname F:stress/s7_ci/ctrlw5_vs_zc/by_station_pct/singapore\endcsname{-20.5000}
\expandafter\def\csname FF:stress/s7_ci/ctrlw5_vs_zc/by_station_pct/singapore\endcsname{-20.50000}
\expandafter\def\csname N:stress/s7_ci/ctrlw5_vs_zc/by_station_pct/minneapolis\endcsname{$-39.5$}
\expandafter\def\csname F:stress/s7_ci/ctrlw5_vs_zc/by_station_pct/minneapolis\endcsname{-39.5000}
\expandafter\def\csname FF:stress/s7_ci/ctrlw5_vs_zc/by_station_pct/minneapolis\endcsname{-39.50000}
\expandafter\def\csname N:stress/s7_ci/ctrlw5_vs_zc/wins\endcsname{48}
\expandafter\def\csname N:stress/s7_ci/zcfloor_vs_zc/n\endcsname{48}
\expandafter\def\csname N:stress/s7_ci/zcfloor_vs_zc/mean\endcsname{0.00152}
\expandafter\def\csname F:stress/s7_ci/zcfloor_vs_zc/mean\endcsname{0.0015}
\expandafter\def\csname FF:stress/s7_ci/zcfloor_vs_zc/mean\endcsname{0.00152}
\expandafter\def\csname N:stress/s7_ci/zcfloor_vs_zc/lo\endcsname{0.00054}
\expandafter\def\csname F:stress/s7_ci/zcfloor_vs_zc/lo\endcsname{0.0005}
\expandafter\def\csname FF:stress/s7_ci/zcfloor_vs_zc/lo\endcsname{0.00054}
\expandafter\def\csname N:stress/s7_ci/zcfloor_vs_zc/hi\endcsname{0.00274}
\expandafter\def\csname F:stress/s7_ci/zcfloor_vs_zc/hi\endcsname{0.0027}
\expandafter\def\csname FF:stress/s7_ci/zcfloor_vs_zc/hi\endcsname{0.00274}
\expandafter\def\csname N:stress/s7_ci/zcfloor_vs_zc/pct\endcsname{$+11.7$}
\expandafter\def\csname F:stress/s7_ci/zcfloor_vs_zc/pct\endcsname{11.7000}
\expandafter\def\csname FF:stress/s7_ci/zcfloor_vs_zc/pct\endcsname{11.70000}
\expandafter\def\csname N:stress/s7_ci/zcfloor_vs_zc/pct_lo\endcsname{$+4.2$}
\expandafter\def\csname F:stress/s7_ci/zcfloor_vs_zc/pct_lo\endcsname{4.2000}
\expandafter\def\csname FF:stress/s7_ci/zcfloor_vs_zc/pct_lo\endcsname{4.20000}
\expandafter\def\csname N:stress/s7_ci/zcfloor_vs_zc/pct_hi\endcsname{$+21.1$}
\expandafter\def\csname F:stress/s7_ci/zcfloor_vs_zc/pct_hi\endcsname{21.1000}
\expandafter\def\csname FF:stress/s7_ci/zcfloor_vs_zc/pct_hi\endcsname{21.10000}
\expandafter\def\csname N:stress/s7_ci/zcfloor_vs_zc/by_station_pct/fairbanks\endcsname{$+12.9$}
\expandafter\def\csname F:stress/s7_ci/zcfloor_vs_zc/by_station_pct/fairbanks\endcsname{12.9000}
\expandafter\def\csname FF:stress/s7_ci/zcfloor_vs_zc/by_station_pct/fairbanks\endcsname{12.90000}
\expandafter\def\csname N:stress/s7_ci/zcfloor_vs_zc/by_station_pct/denver\endcsname{$+7.1$}
\expandafter\def\csname F:stress/s7_ci/zcfloor_vs_zc/by_station_pct/denver\endcsname{7.1000}
\expandafter\def\csname FF:stress/s7_ci/zcfloor_vs_zc/by_station_pct/denver\endcsname{7.10000}
\expandafter\def\csname N:stress/s7_ci/zcfloor_vs_zc/by_station_pct/atlanta\endcsname{$+1.3$}
\expandafter\def\csname F:stress/s7_ci/zcfloor_vs_zc/by_station_pct/atlanta\endcsname{1.3000}
\expandafter\def\csname FF:stress/s7_ci/zcfloor_vs_zc/by_station_pct/atlanta\endcsname{1.30000}
\expandafter\def\csname N:stress/s7_ci/zcfloor_vs_zc/by_station_pct/phoenix\endcsname{$+3.8$}
\expandafter\def\csname F:stress/s7_ci/zcfloor_vs_zc/by_station_pct/phoenix\endcsname{3.8000}
\expandafter\def\csname FF:stress/s7_ci/zcfloor_vs_zc/by_station_pct/phoenix\endcsname{3.80000}
\expandafter\def\csname N:stress/s7_ci/zcfloor_vs_zc/by_station_pct/singapore\endcsname{$0.0$}
\expandafter\def\csname F:stress/s7_ci/zcfloor_vs_zc/by_station_pct/singapore\endcsname{0.0000}
\expandafter\def\csname FF:stress/s7_ci/zcfloor_vs_zc/by_station_pct/singapore\endcsname{0.00000}
\expandafter\def\csname N:stress/s7_ci/zcfloor_vs_zc/by_station_pct/minneapolis\endcsname{$+12.7$}
\expandafter\def\csname F:stress/s7_ci/zcfloor_vs_zc/by_station_pct/minneapolis\endcsname{12.7000}
\expandafter\def\csname FF:stress/s7_ci/zcfloor_vs_zc/by_station_pct/minneapolis\endcsname{12.70000}
\expandafter\def\csname N:stress/s7_ci/zcfloor_vs_zc/wins\endcsname{1}
\expandafter\def\csname W:stress/s7_ci/zcfloor_vs_zc/wins\endcsname{one}
\expandafter\def\csname N:stress/s7_ci/sb278_vs_inv/n\endcsname{48}
\expandafter\def\csname N:stress/s7_ci/sb278_vs_inv/mean\endcsname{9e-05}
\expandafter\def\csname F:stress/s7_ci/sb278_vs_inv/mean\endcsname{0.0001}
\expandafter\def\csname FF:stress/s7_ci/sb278_vs_inv/mean\endcsname{0.00009}
\expandafter\def\csname N:stress/s7_ci/sb278_vs_inv/lo\endcsname{$-1e-05$}
\expandafter\def\csname F:stress/s7_ci/sb278_vs_inv/lo\endcsname{-0.0000}
\expandafter\def\csname FF:stress/s7_ci/sb278_vs_inv/lo\endcsname{-0.00001}
\expandafter\def\csname N:stress/s7_ci/sb278_vs_inv/hi\endcsname{0.00026}
\expandafter\def\csname F:stress/s7_ci/sb278_vs_inv/hi\endcsname{0.0003}
\expandafter\def\csname FF:stress/s7_ci/sb278_vs_inv/hi\endcsname{0.00026}
\expandafter\def\csname N:stress/s7_ci/sb278_vs_inv/pct\endcsname{$+1.1$}
\expandafter\def\csname F:stress/s7_ci/sb278_vs_inv/pct\endcsname{1.1000}
\expandafter\def\csname FF:stress/s7_ci/sb278_vs_inv/pct\endcsname{1.10000}
\expandafter\def\csname N:stress/s7_ci/sb278_vs_inv/pct_lo\endcsname{$-0.2$}
\expandafter\def\csname F:stress/s7_ci/sb278_vs_inv/pct_lo\endcsname{-0.2000}
\expandafter\def\csname FF:stress/s7_ci/sb278_vs_inv/pct_lo\endcsname{-0.20000}
\expandafter\def\csname N:stress/s7_ci/sb278_vs_inv/pct_hi\endcsname{$+3.2$}
\expandafter\def\csname F:stress/s7_ci/sb278_vs_inv/pct_hi\endcsname{3.2000}
\expandafter\def\csname FF:stress/s7_ci/sb278_vs_inv/pct_hi\endcsname{3.20000}
\expandafter\def\csname N:stress/s7_ci/sb278_vs_inv/by_station_pct/fairbanks\endcsname{$+2.2$}
\expandafter\def\csname F:stress/s7_ci/sb278_vs_inv/by_station_pct/fairbanks\endcsname{2.2000}
\expandafter\def\csname FF:stress/s7_ci/sb278_vs_inv/by_station_pct/fairbanks\endcsname{2.20000}
\expandafter\def\csname N:stress/s7_ci/sb278_vs_inv/by_station_pct/denver\endcsname{$+0.1$}
\expandafter\def\csname F:stress/s7_ci/sb278_vs_inv/by_station_pct/denver\endcsname{0.1000}
\expandafter\def\csname FF:stress/s7_ci/sb278_vs_inv/by_station_pct/denver\endcsname{0.10000}
\expandafter\def\csname N:stress/s7_ci/sb278_vs_inv/by_station_pct/atlanta\endcsname{$0.0$}
\expandafter\def\csname F:stress/s7_ci/sb278_vs_inv/by_station_pct/atlanta\endcsname{0.0000}
\expandafter\def\csname FF:stress/s7_ci/sb278_vs_inv/by_station_pct/atlanta\endcsname{0.00000}
\expandafter\def\csname N:stress/s7_ci/sb278_vs_inv/by_station_pct/phoenix\endcsname{$0.0$}
\expandafter\def\csname F:stress/s7_ci/sb278_vs_inv/by_station_pct/phoenix\endcsname{0.0000}
\expandafter\def\csname FF:stress/s7_ci/sb278_vs_inv/by_station_pct/phoenix\endcsname{0.00000}
\expandafter\def\csname N:stress/s7_ci/sb278_vs_inv/by_station_pct/singapore\endcsname{$0.0$}
\expandafter\def\csname F:stress/s7_ci/sb278_vs_inv/by_station_pct/singapore\endcsname{0.0000}
\expandafter\def\csname FF:stress/s7_ci/sb278_vs_inv/by_station_pct/singapore\endcsname{0.00000}
\expandafter\def\csname N:stress/s7_ci/sb278_vs_inv/by_station_pct/minneapolis\endcsname{$+0.2$}
\expandafter\def\csname F:stress/s7_ci/sb278_vs_inv/by_station_pct/minneapolis\endcsname{0.2000}
\expandafter\def\csname FF:stress/s7_ci/sb278_vs_inv/by_station_pct/minneapolis\endcsname{0.20000}
\expandafter\def\csname N:stress/s7_ci/sb278_vs_inv/wins\endcsname{2}
\expandafter\def\csname W:stress/s7_ci/sb278_vs_inv/wins\endcsname{two}
\expandafter\def\csname N:stress/s7_ci/ra1_vs_zc/n\endcsname{48}
\expandafter\def\csname N:stress/s7_ci/ra1_vs_zc/mean\endcsname{$-0.0008$}
\expandafter\def\csname F:stress/s7_ci/ra1_vs_zc/mean\endcsname{-0.0008}
\expandafter\def\csname FF:stress/s7_ci/ra1_vs_zc/mean\endcsname{-0.00080}
\expandafter\def\csname N:stress/s7_ci/ra1_vs_zc/lo\endcsname{$-0.00136$}
\expandafter\def\csname F:stress/s7_ci/ra1_vs_zc/lo\endcsname{-0.0014}
\expandafter\def\csname FF:stress/s7_ci/ra1_vs_zc/lo\endcsname{-0.00136}
\expandafter\def\csname N:stress/s7_ci/ra1_vs_zc/hi\endcsname{$-0.00033$}
\expandafter\def\csname F:stress/s7_ci/ra1_vs_zc/hi\endcsname{-0.0003}
\expandafter\def\csname FF:stress/s7_ci/ra1_vs_zc/hi\endcsname{-0.00033}
\expandafter\def\csname N:stress/s7_ci/ra1_vs_zc/pct\endcsname{$-6.2$}
\expandafter\def\csname F:stress/s7_ci/ra1_vs_zc/pct\endcsname{-6.2000}
\expandafter\def\csname FF:stress/s7_ci/ra1_vs_zc/pct\endcsname{-6.20000}
\expandafter\def\csname N:stress/s7_ci/ra1_vs_zc/pct_lo\endcsname{$-10.5$}
\expandafter\def\csname F:stress/s7_ci/ra1_vs_zc/pct_lo\endcsname{-10.5000}
\expandafter\def\csname FF:stress/s7_ci/ra1_vs_zc/pct_lo\endcsname{-10.50000}
\expandafter\def\csname N:stress/s7_ci/ra1_vs_zc/pct_hi\endcsname{$-2.5$}
\expandafter\def\csname F:stress/s7_ci/ra1_vs_zc/pct_hi\endcsname{-2.5000}
\expandafter\def\csname FF:stress/s7_ci/ra1_vs_zc/pct_hi\endcsname{-2.50000}
\expandafter\def\csname N:stress/s7_ci/ra1_vs_zc/by_station_pct/fairbanks\endcsname{$-3.4$}
\expandafter\def\csname F:stress/s7_ci/ra1_vs_zc/by_station_pct/fairbanks\endcsname{-3.4000}
\expandafter\def\csname FF:stress/s7_ci/ra1_vs_zc/by_station_pct/fairbanks\endcsname{-3.40000}
\expandafter\def\csname N:stress/s7_ci/ra1_vs_zc/by_station_pct/denver\endcsname{$-4.8$}
\expandafter\def\csname F:stress/s7_ci/ra1_vs_zc/by_station_pct/denver\endcsname{-4.8000}
\expandafter\def\csname FF:stress/s7_ci/ra1_vs_zc/by_station_pct/denver\endcsname{-4.80000}
\expandafter\def\csname N:stress/s7_ci/ra1_vs_zc/by_station_pct/atlanta\endcsname{$-1.3$}
\expandafter\def\csname F:stress/s7_ci/ra1_vs_zc/by_station_pct/atlanta\endcsname{-1.3000}
\expandafter\def\csname FF:stress/s7_ci/ra1_vs_zc/by_station_pct/atlanta\endcsname{-1.30000}
\expandafter\def\csname N:stress/s7_ci/ra1_vs_zc/by_station_pct/phoenix\endcsname{$+2.2$}
\expandafter\def\csname F:stress/s7_ci/ra1_vs_zc/by_station_pct/phoenix\endcsname{2.2000}
\expandafter\def\csname FF:stress/s7_ci/ra1_vs_zc/by_station_pct/phoenix\endcsname{2.20000}
\expandafter\def\csname N:stress/s7_ci/ra1_vs_zc/by_station_pct/singapore\endcsname{$+0.1$}
\expandafter\def\csname F:stress/s7_ci/ra1_vs_zc/by_station_pct/singapore\endcsname{0.1000}
\expandafter\def\csname FF:stress/s7_ci/ra1_vs_zc/by_station_pct/singapore\endcsname{0.10000}
\expandafter\def\csname N:stress/s7_ci/ra1_vs_zc/by_station_pct/minneapolis\endcsname{$-7.6$}
\expandafter\def\csname F:stress/s7_ci/ra1_vs_zc/by_station_pct/minneapolis\endcsname{-7.6000}
\expandafter\def\csname FF:stress/s7_ci/ra1_vs_zc/by_station_pct/minneapolis\endcsname{-7.60000}
\expandafter\def\csname N:stress/s7_ci/ra1_vs_zc/wins\endcsname{30}
\expandafter\def\csname N:stress/s7_ci/ra1s1_vs_zc/n\endcsname{48}
\expandafter\def\csname N:stress/s7_ci/ra1s1_vs_zc/mean\endcsname{0.00478}
\expandafter\def\csname F:stress/s7_ci/ra1s1_vs_zc/mean\endcsname{0.0048}
\expandafter\def\csname FF:stress/s7_ci/ra1s1_vs_zc/mean\endcsname{0.00478}
\expandafter\def\csname N:stress/s7_ci/ra1s1_vs_zc/lo\endcsname{0.00272}
\expandafter\def\csname F:stress/s7_ci/ra1s1_vs_zc/lo\endcsname{0.0027}
\expandafter\def\csname FF:stress/s7_ci/ra1s1_vs_zc/lo\endcsname{0.00272}
\expandafter\def\csname N:stress/s7_ci/ra1s1_vs_zc/hi\endcsname{0.00728}
\expandafter\def\csname F:stress/s7_ci/ra1s1_vs_zc/hi\endcsname{0.0073}
\expandafter\def\csname FF:stress/s7_ci/ra1s1_vs_zc/hi\endcsname{0.00728}
\expandafter\def\csname N:stress/s7_ci/ra1s1_vs_zc/pct\endcsname{$+36.8$}
\expandafter\def\csname F:stress/s7_ci/ra1s1_vs_zc/pct\endcsname{36.8000}
\expandafter\def\csname FF:stress/s7_ci/ra1s1_vs_zc/pct\endcsname{36.80000}
\expandafter\def\csname N:stress/s7_ci/ra1s1_vs_zc/pct_lo\endcsname{$+20.9$}
\expandafter\def\csname F:stress/s7_ci/ra1s1_vs_zc/pct_lo\endcsname{20.9000}
\expandafter\def\csname FF:stress/s7_ci/ra1s1_vs_zc/pct_lo\endcsname{20.90000}
\expandafter\def\csname N:stress/s7_ci/ra1s1_vs_zc/pct_hi\endcsname{$+56.0$}
\expandafter\def\csname F:stress/s7_ci/ra1s1_vs_zc/pct_hi\endcsname{56.0000}
\expandafter\def\csname FF:stress/s7_ci/ra1s1_vs_zc/pct_hi\endcsname{56.00000}
\expandafter\def\csname N:stress/s7_ci/ra1s1_vs_zc/by_station_pct/fairbanks\endcsname{$+15.0$}
\expandafter\def\csname F:stress/s7_ci/ra1s1_vs_zc/by_station_pct/fairbanks\endcsname{15.0000}
\expandafter\def\csname FF:stress/s7_ci/ra1s1_vs_zc/by_station_pct/fairbanks\endcsname{15.00000}
\expandafter\def\csname N:stress/s7_ci/ra1s1_vs_zc/by_station_pct/denver\endcsname{$+22.8$}
\expandafter\def\csname F:stress/s7_ci/ra1s1_vs_zc/by_station_pct/denver\endcsname{22.8000}
\expandafter\def\csname FF:stress/s7_ci/ra1s1_vs_zc/by_station_pct/denver\endcsname{22.80000}
\expandafter\def\csname N:stress/s7_ci/ra1s1_vs_zc/by_station_pct/atlanta\endcsname{$+35.6$}
\expandafter\def\csname F:stress/s7_ci/ra1s1_vs_zc/by_station_pct/atlanta\endcsname{35.6000}
\expandafter\def\csname FF:stress/s7_ci/ra1s1_vs_zc/by_station_pct/atlanta\endcsname{35.60000}
\expandafter\def\csname N:stress/s7_ci/ra1s1_vs_zc/by_station_pct/phoenix\endcsname{$+41.8$}
\expandafter\def\csname F:stress/s7_ci/ra1s1_vs_zc/by_station_pct/phoenix\endcsname{41.8000}
\expandafter\def\csname FF:stress/s7_ci/ra1s1_vs_zc/by_station_pct/phoenix\endcsname{41.80000}
\expandafter\def\csname N:stress/s7_ci/ra1s1_vs_zc/by_station_pct/singapore\endcsname{$+23.9$}
\expandafter\def\csname F:stress/s7_ci/ra1s1_vs_zc/by_station_pct/singapore\endcsname{23.9000}
\expandafter\def\csname FF:stress/s7_ci/ra1s1_vs_zc/by_station_pct/singapore\endcsname{23.90000}
\expandafter\def\csname N:stress/s7_ci/ra1s1_vs_zc/by_station_pct/minneapolis\endcsname{$+40.6$}
\expandafter\def\csname F:stress/s7_ci/ra1s1_vs_zc/by_station_pct/minneapolis\endcsname{40.6000}
\expandafter\def\csname FF:stress/s7_ci/ra1s1_vs_zc/by_station_pct/minneapolis\endcsname{40.60000}
\expandafter\def\csname N:stress/s7_ci/ra1s1_vs_zc/wins\endcsname{2}
\expandafter\def\csname W:stress/s7_ci/ra1s1_vs_zc/wins\endcsname{two}
\expandafter\def\csname N:stress/s7_ci/zcf278_vs_zc/n\endcsname{48}
\expandafter\def\csname N:stress/s7_ci/zcf278_vs_zc/mean\endcsname{0.00014}
\expandafter\def\csname F:stress/s7_ci/zcf278_vs_zc/mean\endcsname{0.0001}
\expandafter\def\csname FF:stress/s7_ci/zcf278_vs_zc/mean\endcsname{0.00014}
\expandafter\def\csname N:stress/s7_ci/zcf278_vs_zc/lo\endcsname{1e-05}
\expandafter\def\csname F:stress/s7_ci/zcf278_vs_zc/lo\endcsname{0.0000}
\expandafter\def\csname FF:stress/s7_ci/zcf278_vs_zc/lo\endcsname{0.00001}
\expandafter\def\csname N:stress/s7_ci/zcf278_vs_zc/hi\endcsname{0.00033}
\expandafter\def\csname F:stress/s7_ci/zcf278_vs_zc/hi\endcsname{0.0003}
\expandafter\def\csname FF:stress/s7_ci/zcf278_vs_zc/hi\endcsname{0.00033}
\expandafter\def\csname N:stress/s7_ci/zcf278_vs_zc/pct\endcsname{$+1.1$}
\expandafter\def\csname F:stress/s7_ci/zcf278_vs_zc/pct\endcsname{1.1000}
\expandafter\def\csname FF:stress/s7_ci/zcf278_vs_zc/pct\endcsname{1.10000}
\expandafter\def\csname N:stress/s7_ci/zcf278_vs_zc/pct_lo\endcsname{$+0.1$}
\expandafter\def\csname F:stress/s7_ci/zcf278_vs_zc/pct_lo\endcsname{0.1000}
\expandafter\def\csname FF:stress/s7_ci/zcf278_vs_zc/pct_lo\endcsname{0.10000}
\expandafter\def\csname N:stress/s7_ci/zcf278_vs_zc/pct_hi\endcsname{$+2.6$}
\expandafter\def\csname F:stress/s7_ci/zcf278_vs_zc/pct_hi\endcsname{2.6000}
\expandafter\def\csname FF:stress/s7_ci/zcf278_vs_zc/pct_hi\endcsname{2.60000}
\expandafter\def\csname N:stress/s7_ci/zcf278_vs_zc/by_station_pct/fairbanks\endcsname{$+1.6$}
\expandafter\def\csname F:stress/s7_ci/zcf278_vs_zc/by_station_pct/fairbanks\endcsname{1.6000}
\expandafter\def\csname FF:stress/s7_ci/zcf278_vs_zc/by_station_pct/fairbanks\endcsname{1.60000}
\expandafter\def\csname N:stress/s7_ci/zcf278_vs_zc/by_station_pct/denver\endcsname{$0.0$}
\expandafter\def\csname F:stress/s7_ci/zcf278_vs_zc/by_station_pct/denver\endcsname{0.0000}
\expandafter\def\csname FF:stress/s7_ci/zcf278_vs_zc/by_station_pct/denver\endcsname{0.00000}
\expandafter\def\csname N:stress/s7_ci/zcf278_vs_zc/by_station_pct/atlanta\endcsname{$0.0$}
\expandafter\def\csname F:stress/s7_ci/zcf278_vs_zc/by_station_pct/atlanta\endcsname{0.0000}
\expandafter\def\csname FF:stress/s7_ci/zcf278_vs_zc/by_station_pct/atlanta\endcsname{0.00000}
\expandafter\def\csname N:stress/s7_ci/zcf278_vs_zc/by_station_pct/phoenix\endcsname{$0.0$}
\expandafter\def\csname F:stress/s7_ci/zcf278_vs_zc/by_station_pct/phoenix\endcsname{0.0000}
\expandafter\def\csname FF:stress/s7_ci/zcf278_vs_zc/by_station_pct/phoenix\endcsname{0.00000}
\expandafter\def\csname N:stress/s7_ci/zcf278_vs_zc/by_station_pct/singapore\endcsname{$0.0$}
\expandafter\def\csname F:stress/s7_ci/zcf278_vs_zc/by_station_pct/singapore\endcsname{0.0000}
\expandafter\def\csname FF:stress/s7_ci/zcf278_vs_zc/by_station_pct/singapore\endcsname{0.00000}
\expandafter\def\csname N:stress/s7_ci/zcf278_vs_zc/by_station_pct/minneapolis\endcsname{$+0.4$}
\expandafter\def\csname F:stress/s7_ci/zcf278_vs_zc/by_station_pct/minneapolis\endcsname{0.4000}
\expandafter\def\csname FF:stress/s7_ci/zcf278_vs_zc/by_station_pct/minneapolis\endcsname{0.40000}
\expandafter\def\csname N:stress/s7_ci/zcf278_vs_zc/wins\endcsname{0}
\expandafter\def\csname W:stress/s7_ci/zcf278_vs_zc/wins\endcsname{zero}
\expandafter\def\csname N:stress/s7_ci/invfix_vs_inv/n\endcsname{48}
\expandafter\def\csname N:stress/s7_ci/invfix_vs_inv/mean\endcsname{$-6e-05$}
\expandafter\def\csname F:stress/s7_ci/invfix_vs_inv/mean\endcsname{-0.0001}
\expandafter\def\csname FF:stress/s7_ci/invfix_vs_inv/mean\endcsname{-0.00006}
\expandafter\def\csname N:stress/s7_ci/invfix_vs_inv/lo\endcsname{$-0.0001$}
\expandafter\def\csname F:stress/s7_ci/invfix_vs_inv/lo\endcsname{-0.0001}
\expandafter\def\csname FF:stress/s7_ci/invfix_vs_inv/lo\endcsname{-0.00010}
\expandafter\def\csname N:stress/s7_ci/invfix_vs_inv/hi\endcsname{$-3e-05$}
\expandafter\def\csname F:stress/s7_ci/invfix_vs_inv/hi\endcsname{-0.0000}
\expandafter\def\csname FF:stress/s7_ci/invfix_vs_inv/hi\endcsname{-0.00003}
\expandafter\def\csname N:stress/s7_ci/invfix_vs_inv/pct\endcsname{$-0.8$}
\expandafter\def\csname F:stress/s7_ci/invfix_vs_inv/pct\endcsname{-0.8000}
\expandafter\def\csname FF:stress/s7_ci/invfix_vs_inv/pct\endcsname{-0.80000}
\expandafter\def\csname N:stress/s7_ci/invfix_vs_inv/pct_lo\endcsname{$-1.3$}
\expandafter\def\csname F:stress/s7_ci/invfix_vs_inv/pct_lo\endcsname{-1.3000}
\expandafter\def\csname FF:stress/s7_ci/invfix_vs_inv/pct_lo\endcsname{-1.30000}
\expandafter\def\csname N:stress/s7_ci/invfix_vs_inv/pct_hi\endcsname{$-0.4$}
\expandafter\def\csname F:stress/s7_ci/invfix_vs_inv/pct_hi\endcsname{-0.4000}
\expandafter\def\csname FF:stress/s7_ci/invfix_vs_inv/pct_hi\endcsname{-0.40000}
\expandafter\def\csname N:stress/s7_ci/invfix_vs_inv/by_station_pct/fairbanks\endcsname{$-1.0$}
\expandafter\def\csname F:stress/s7_ci/invfix_vs_inv/by_station_pct/fairbanks\endcsname{-1.0000}
\expandafter\def\csname FF:stress/s7_ci/invfix_vs_inv/by_station_pct/fairbanks\endcsname{-1.00000}
\expandafter\def\csname N:stress/s7_ci/invfix_vs_inv/by_station_pct/denver\endcsname{$-0.8$}
\expandafter\def\csname F:stress/s7_ci/invfix_vs_inv/by_station_pct/denver\endcsname{-0.8000}
\expandafter\def\csname FF:stress/s7_ci/invfix_vs_inv/by_station_pct/denver\endcsname{-0.80000}
\expandafter\def\csname N:stress/s7_ci/invfix_vs_inv/by_station_pct/atlanta\endcsname{$-0.2$}
\expandafter\def\csname F:stress/s7_ci/invfix_vs_inv/by_station_pct/atlanta\endcsname{-0.2000}
\expandafter\def\csname FF:stress/s7_ci/invfix_vs_inv/by_station_pct/atlanta\endcsname{-0.20000}
\expandafter\def\csname N:stress/s7_ci/invfix_vs_inv/by_station_pct/phoenix\endcsname{$-0.1$}
\expandafter\def\csname F:stress/s7_ci/invfix_vs_inv/by_station_pct/phoenix\endcsname{-0.1000}
\expandafter\def\csname FF:stress/s7_ci/invfix_vs_inv/by_station_pct/phoenix\endcsname{-0.10000}
\expandafter\def\csname N:stress/s7_ci/invfix_vs_inv/by_station_pct/singapore\endcsname{$0.0$}
\expandafter\def\csname F:stress/s7_ci/invfix_vs_inv/by_station_pct/singapore\endcsname{0.0000}
\expandafter\def\csname FF:stress/s7_ci/invfix_vs_inv/by_station_pct/singapore\endcsname{0.00000}
\expandafter\def\csname N:stress/s7_ci/invfix_vs_inv/by_station_pct/minneapolis\endcsname{$-0.8$}
\expandafter\def\csname F:stress/s7_ci/invfix_vs_inv/by_station_pct/minneapolis\endcsname{-0.8000}
\expandafter\def\csname FF:stress/s7_ci/invfix_vs_inv/by_station_pct/minneapolis\endcsname{-0.80000}
\expandafter\def\csname N:stress/s7_ci/invfix_vs_inv/wins\endcsname{29}
\expandafter\def\csname N:stress/s7_ci/invfix_vs_zcfix/n\endcsname{48}
\expandafter\def\csname N:stress/s7_ci/invfix_vs_zcfix/mean\endcsname{$-0.00519$}
\expandafter\def\csname F:stress/s7_ci/invfix_vs_zcfix/mean\endcsname{-0.0052}
\expandafter\def\csname FF:stress/s7_ci/invfix_vs_zcfix/mean\endcsname{-0.00519}
\expandafter\def\csname N:stress/s7_ci/invfix_vs_zcfix/lo\endcsname{$-0.00702$}
\expandafter\def\csname F:stress/s7_ci/invfix_vs_zcfix/lo\endcsname{-0.0070}
\expandafter\def\csname FF:stress/s7_ci/invfix_vs_zcfix/lo\endcsname{-0.00702}
\expandafter\def\csname N:stress/s7_ci/invfix_vs_zcfix/hi\endcsname{$-0.00352$}
\expandafter\def\csname F:stress/s7_ci/invfix_vs_zcfix/hi\endcsname{-0.0035}
\expandafter\def\csname FF:stress/s7_ci/invfix_vs_zcfix/hi\endcsname{-0.00352}
\expandafter\def\csname N:stress/s7_ci/invfix_vs_zcfix/pct\endcsname{$-39.2$}
\expandafter\def\csname F:stress/s7_ci/invfix_vs_zcfix/pct\endcsname{-39.2000}
\expandafter\def\csname FF:stress/s7_ci/invfix_vs_zcfix/pct\endcsname{-39.20000}
\expandafter\def\csname N:stress/s7_ci/invfix_vs_zcfix/pct_lo\endcsname{$-53.0$}
\expandafter\def\csname F:stress/s7_ci/invfix_vs_zcfix/pct_lo\endcsname{-53.0000}
\expandafter\def\csname FF:stress/s7_ci/invfix_vs_zcfix/pct_lo\endcsname{-53.00000}
\expandafter\def\csname N:stress/s7_ci/invfix_vs_zcfix/pct_hi\endcsname{$-26.6$}
\expandafter\def\csname F:stress/s7_ci/invfix_vs_zcfix/pct_hi\endcsname{-26.6000}
\expandafter\def\csname FF:stress/s7_ci/invfix_vs_zcfix/pct_hi\endcsname{-26.60000}
\expandafter\def\csname N:stress/s7_ci/invfix_vs_zcfix/by_station_pct/fairbanks\endcsname{$-46.9$}
\expandafter\def\csname F:stress/s7_ci/invfix_vs_zcfix/by_station_pct/fairbanks\endcsname{-46.9000}
\expandafter\def\csname FF:stress/s7_ci/invfix_vs_zcfix/by_station_pct/fairbanks\endcsname{-46.90000}
\expandafter\def\csname N:stress/s7_ci/invfix_vs_zcfix/by_station_pct/denver\endcsname{$-47.0$}
\expandafter\def\csname F:stress/s7_ci/invfix_vs_zcfix/by_station_pct/denver\endcsname{-47.0000}
\expandafter\def\csname FF:stress/s7_ci/invfix_vs_zcfix/by_station_pct/denver\endcsname{-47.00000}
\expandafter\def\csname N:stress/s7_ci/invfix_vs_zcfix/by_station_pct/atlanta\endcsname{$-29.5$}
\expandafter\def\csname F:stress/s7_ci/invfix_vs_zcfix/by_station_pct/atlanta\endcsname{-29.5000}
\expandafter\def\csname FF:stress/s7_ci/invfix_vs_zcfix/by_station_pct/atlanta\endcsname{-29.50000}
\expandafter\def\csname N:stress/s7_ci/invfix_vs_zcfix/by_station_pct/phoenix\endcsname{$-28.5$}
\expandafter\def\csname F:stress/s7_ci/invfix_vs_zcfix/by_station_pct/phoenix\endcsname{-28.5000}
\expandafter\def\csname FF:stress/s7_ci/invfix_vs_zcfix/by_station_pct/phoenix\endcsname{-28.50000}
\expandafter\def\csname N:stress/s7_ci/invfix_vs_zcfix/by_station_pct/singapore\endcsname{$-21.2$}
\expandafter\def\csname F:stress/s7_ci/invfix_vs_zcfix/by_station_pct/singapore\endcsname{-21.2000}
\expandafter\def\csname FF:stress/s7_ci/invfix_vs_zcfix/by_station_pct/singapore\endcsname{-21.20000}
\expandafter\def\csname N:stress/s7_ci/invfix_vs_zcfix/by_station_pct/minneapolis\endcsname{$-40.1$}
\expandafter\def\csname F:stress/s7_ci/invfix_vs_zcfix/by_station_pct/minneapolis\endcsname{-40.1000}
\expandafter\def\csname FF:stress/s7_ci/invfix_vs_zcfix/by_station_pct/minneapolis\endcsname{-40.10000}
\expandafter\def\csname N:stress/s7_ci/invfix_vs_zcfix/wins\endcsname{48}
\expandafter\def\csname N:stress/bound/model/n\endcsname{8}
\expandafter\def\csname W:stress/bound/model/n\endcsname{eight}
\expandafter\def\csname N:stress/bound/model/score\endcsname{0.02323}
\expandafter\def\csname F:stress/bound/model/score\endcsname{0.0232}
\expandafter\def\csname FF:stress/bound/model/score\endcsname{0.02323}
\expandafter\def\csname N:stress/bound/model/comfort\endcsname{0.00936}
\expandafter\def\csname F:stress/bound/model/comfort\endcsname{0.0094}
\expandafter\def\csname FF:stress/bound/model/comfort\endcsname{0.00936}
\expandafter\def\csname N:stress/bound/model/energy\endcsname{0.00735}
\expandafter\def\csname F:stress/bound/model/energy\endcsname{0.0073}
\expandafter\def\csname FF:stress/bound/model/energy\endcsname{0.00735}
\expandafter\def\csname N:stress/bound/model/carbon\endcsname{0.00651}
\expandafter\def\csname F:stress/bound/model/carbon\endcsname{0.0065}
\expandafter\def\csname FF:stress/bound/model/carbon\endcsname{0.00651}
\expandafter\def\csname N:stress/bound/model/small\endcsname{0.00909}
\expandafter\def\csname F:stress/bound/model/small\endcsname{0.0091}
\expandafter\def\csname FF:stress/bound/model/small\endcsname{0.00909}
\expandafter\def\csname N:stress/bound/model/large\endcsname{0.03737}
\expandafter\def\csname F:stress/bound/model/large\endcsname{0.0374}
\expandafter\def\csname FF:stress/bound/model/large\endcsname{0.03737}
\expandafter\def\csname N:stress/bound/sim/n\endcsname{8}
\expandafter\def\csname W:stress/bound/sim/n\endcsname{eight}
\expandafter\def\csname N:stress/bound/sim/score\endcsname{0.02024}
\expandafter\def\csname F:stress/bound/sim/score\endcsname{0.0202}
\expandafter\def\csname FF:stress/bound/sim/score\endcsname{0.02024}
\expandafter\def\csname N:stress/bound/sim/comfort\endcsname{0.00441}
\expandafter\def\csname F:stress/bound/sim/comfort\endcsname{0.0044}
\expandafter\def\csname FF:stress/bound/sim/comfort\endcsname{0.00441}
\expandafter\def\csname N:stress/bound/sim/energy\endcsname{0.00821}
\expandafter\def\csname F:stress/bound/sim/energy\endcsname{0.0082}
\expandafter\def\csname FF:stress/bound/sim/energy\endcsname{0.00821}
\expandafter\def\csname N:stress/bound/sim/carbon\endcsname{0.00762}
\expandafter\def\csname F:stress/bound/sim/carbon\endcsname{0.0076}
\expandafter\def\csname FF:stress/bound/sim/carbon\endcsname{0.00762}
\expandafter\def\csname N:stress/bound/sim/small\endcsname{0.00934}
\expandafter\def\csname F:stress/bound/sim/small\endcsname{0.0093}
\expandafter\def\csname FF:stress/bound/sim/small\endcsname{0.00934}
\expandafter\def\csname N:stress/bound/sim/large\endcsname{0.03113}
\expandafter\def\csname F:stress/bound/sim/large\endcsname{0.0311}
\expandafter\def\csname FF:stress/bound/sim/large\endcsname{0.03113}
\expandafter\def\csname N:stress/bound/ci/n\endcsname{8}
\expandafter\def\csname W:stress/bound/ci/n\endcsname{eight}
\expandafter\def\csname N:stress/bound/ci/mean\endcsname{$-0.00299$}
\expandafter\def\csname F:stress/bound/ci/mean\endcsname{-0.0030}
\expandafter\def\csname FF:stress/bound/ci/mean\endcsname{-0.00299}
\expandafter\def\csname N:stress/bound/ci/lo\endcsname{$-0.00745$}
\expandafter\def\csname F:stress/bound/ci/lo\endcsname{-0.0075}
\expandafter\def\csname FF:stress/bound/ci/lo\endcsname{-0.00745}
\expandafter\def\csname N:stress/bound/ci/hi\endcsname{9e-05}
\expandafter\def\csname F:stress/bound/ci/hi\endcsname{0.0001}
\expandafter\def\csname FF:stress/bound/ci/hi\endcsname{0.00009}
\expandafter\def\csname N:stress/bound/ci/pct\endcsname{$-12.9$}
\expandafter\def\csname F:stress/bound/ci/pct\endcsname{-12.9000}
\expandafter\def\csname FF:stress/bound/ci/pct\endcsname{-12.90000}
\expandafter\def\csname N:stress/bound/ci/pct_lo\endcsname{$-32.1$}
\expandafter\def\csname F:stress/bound/ci/pct_lo\endcsname{-32.1000}
\expandafter\def\csname FF:stress/bound/ci/pct_lo\endcsname{-32.10000}
\expandafter\def\csname N:stress/bound/ci/pct_hi\endcsname{$+0.4$}
\expandafter\def\csname F:stress/bound/ci/pct_hi\endcsname{0.4000}
\expandafter\def\csname FF:stress/bound/ci/pct_hi\endcsname{0.40000}
\expandafter\def\csname N:stress/bound/ci/by_station_pct/fairbanks\endcsname{$-2.5$}
\expandafter\def\csname F:stress/bound/ci/by_station_pct/fairbanks\endcsname{-2.5000}
\expandafter\def\csname FF:stress/bound/ci/by_station_pct/fairbanks\endcsname{-2.50000}
\expandafter\def\csname N:stress/bound/ci/by_station_pct/denver\endcsname{$+4.5$}
\expandafter\def\csname F:stress/bound/ci/by_station_pct/denver\endcsname{4.5000}
\expandafter\def\csname FF:stress/bound/ci/by_station_pct/denver\endcsname{4.50000}
\expandafter\def\csname N:stress/bound/ci/by_station_pct/atlanta\endcsname{$-22.6$}
\expandafter\def\csname F:stress/bound/ci/by_station_pct/atlanta\endcsname{-22.6000}
\expandafter\def\csname FF:stress/bound/ci/by_station_pct/atlanta\endcsname{-22.60000}
\expandafter\def\csname N:stress/bound/ci/by_station_pct/phoenix\endcsname{$-13.7$}
\expandafter\def\csname F:stress/bound/ci/by_station_pct/phoenix\endcsname{-13.7000}
\expandafter\def\csname FF:stress/bound/ci/by_station_pct/phoenix\endcsname{-13.70000}
\expandafter\def\csname N:stress/bound/ci/by_station_pct/singapore\endcsname{$+3.0$}
\expandafter\def\csname F:stress/bound/ci/by_station_pct/singapore\endcsname{3.0000}
\expandafter\def\csname FF:stress/bound/ci/by_station_pct/singapore\endcsname{3.00000}
\expandafter\def\csname N:stress/bound/ci/by_station_pct/minneapolis\endcsname{$-24.0$}
\expandafter\def\csname F:stress/bound/ci/by_station_pct/minneapolis\endcsname{-24.0000}
\expandafter\def\csname FF:stress/bound/ci/by_station_pct/minneapolis\endcsname{-24.00000}
\expandafter\def\csname N:stress/bound/ci/wins\endcsname{5}
\expandafter\def\csname W:stress/bound/ci/wins\endcsname{five}
\expandafter\def\csname N:stress/bound/inv/n\endcsname{8}
\expandafter\def\csname W:stress/bound/inv/n\endcsname{eight}
\expandafter\def\csname N:stress/bound/inv/score\endcsname{0.01385}
\expandafter\def\csname F:stress/bound/inv/score\endcsname{0.0138}
\expandafter\def\csname FF:stress/bound/inv/score\endcsname{0.01385}
\expandafter\def\csname N:stress/bound/inv/comfort\endcsname{0.00096}
\expandafter\def\csname F:stress/bound/inv/comfort\endcsname{0.0010}
\expandafter\def\csname FF:stress/bound/inv/comfort\endcsname{0.00096}
\expandafter\def\csname N:stress/bound/inv/energy\endcsname{0.00681}
\expandafter\def\csname F:stress/bound/inv/energy\endcsname{0.0068}
\expandafter\def\csname FF:stress/bound/inv/energy\endcsname{0.00681}
\expandafter\def\csname N:stress/bound/inv/carbon\endcsname{0.00608}
\expandafter\def\csname F:stress/bound/inv/carbon\endcsname{0.0061}
\expandafter\def\csname FF:stress/bound/inv/carbon\endcsname{0.00608}
\expandafter\def\csname N:stress/bound/inv/small\endcsname{0.00551}
\expandafter\def\csname F:stress/bound/inv/small\endcsname{0.0055}
\expandafter\def\csname FF:stress/bound/inv/small\endcsname{0.00551}
\expandafter\def\csname N:stress/bound/inv/large\endcsname{0.02219}
\expandafter\def\csname F:stress/bound/inv/large\endcsname{0.0222}
\expandafter\def\csname FF:stress/bound/inv/large\endcsname{0.02219}
\expandafter\def\csname N:stress/bound/ci_inv/n\endcsname{8}
\expandafter\def\csname W:stress/bound/ci_inv/n\endcsname{eight}
\expandafter\def\csname N:stress/bound/ci_inv/mean\endcsname{$-0.00938$}
\expandafter\def\csname F:stress/bound/ci_inv/mean\endcsname{-0.0094}
\expandafter\def\csname FF:stress/bound/ci_inv/mean\endcsname{-0.00938}
\expandafter\def\csname N:stress/bound/ci_inv/lo\endcsname{$-0.01564$}
\expandafter\def\csname F:stress/bound/ci_inv/lo\endcsname{-0.0156}
\expandafter\def\csname FF:stress/bound/ci_inv/lo\endcsname{-0.01564}
\expandafter\def\csname N:stress/bound/ci_inv/hi\endcsname{$-0.004$}
\expandafter\def\csname F:stress/bound/ci_inv/hi\endcsname{-0.0040}
\expandafter\def\csname FF:stress/bound/ci_inv/hi\endcsname{-0.00400}
\expandafter\def\csname N:stress/bound/ci_inv/pct\endcsname{$-40.4$}
\expandafter\def\csname F:stress/bound/ci_inv/pct\endcsname{-40.4000}
\expandafter\def\csname FF:stress/bound/ci_inv/pct\endcsname{-40.40000}
\expandafter\def\csname N:stress/bound/ci_inv/pct_lo\endcsname{$-67.3$}
\expandafter\def\csname F:stress/bound/ci_inv/pct_lo\endcsname{-67.3000}
\expandafter\def\csname FF:stress/bound/ci_inv/pct_lo\endcsname{-67.30000}
\expandafter\def\csname N:stress/bound/ci_inv/pct_hi\endcsname{$-17.2$}
\expandafter\def\csname F:stress/bound/ci_inv/pct_hi\endcsname{-17.2000}
\expandafter\def\csname FF:stress/bound/ci_inv/pct_hi\endcsname{-17.20000}
\expandafter\def\csname N:stress/bound/ci_inv/by_station_pct/fairbanks\endcsname{$-34.4$}
\expandafter\def\csname F:stress/bound/ci_inv/by_station_pct/fairbanks\endcsname{-34.4000}
\expandafter\def\csname FF:stress/bound/ci_inv/by_station_pct/fairbanks\endcsname{-34.40000}
\expandafter\def\csname N:stress/bound/ci_inv/by_station_pct/denver\endcsname{$-46.6$}
\expandafter\def\csname F:stress/bound/ci_inv/by_station_pct/denver\endcsname{-46.6000}
\expandafter\def\csname FF:stress/bound/ci_inv/by_station_pct/denver\endcsname{-46.60000}
\expandafter\def\csname N:stress/bound/ci_inv/by_station_pct/atlanta\endcsname{$-30.2$}
\expandafter\def\csname F:stress/bound/ci_inv/by_station_pct/atlanta\endcsname{-30.2000}
\expandafter\def\csname FF:stress/bound/ci_inv/by_station_pct/atlanta\endcsname{-30.20000}
\expandafter\def\csname N:stress/bound/ci_inv/by_station_pct/phoenix\endcsname{$-29.4$}
\expandafter\def\csname F:stress/bound/ci_inv/by_station_pct/phoenix\endcsname{-29.4000}
\expandafter\def\csname FF:stress/bound/ci_inv/by_station_pct/phoenix\endcsname{-29.40000}
\expandafter\def\csname N:stress/bound/ci_inv/by_station_pct/singapore\endcsname{$-24.0$}
\expandafter\def\csname F:stress/bound/ci_inv/by_station_pct/singapore\endcsname{-24.0000}
\expandafter\def\csname FF:stress/bound/ci_inv/by_station_pct/singapore\endcsname{-24.00000}
\expandafter\def\csname N:stress/bound/ci_inv/by_station_pct/minneapolis\endcsname{$-55.8$}
\expandafter\def\csname F:stress/bound/ci_inv/by_station_pct/minneapolis\endcsname{-55.8000}
\expandafter\def\csname FF:stress/bound/ci_inv/by_station_pct/minneapolis\endcsname{-55.80000}
\expandafter\def\csname N:stress/bound/ci_inv/wins\endcsname{8}
\expandafter\def\csname W:stress/bound/ci_inv/wins\endcsname{eight}
\expandafter\def\csname N:stress/bound/ci_inv_sim/n\endcsname{8}
\expandafter\def\csname W:stress/bound/ci_inv_sim/n\endcsname{eight}
\expandafter\def\csname N:stress/bound/ci_inv_sim/mean\endcsname{$-0.00639$}
\expandafter\def\csname F:stress/bound/ci_inv_sim/mean\endcsname{-0.0064}
\expandafter\def\csname FF:stress/bound/ci_inv_sim/mean\endcsname{-0.00639}
\expandafter\def\csname N:stress/bound/ci_inv_sim/lo\endcsname{$-0.01156$}
\expandafter\def\csname F:stress/bound/ci_inv_sim/lo\endcsname{-0.0116}
\expandafter\def\csname FF:stress/bound/ci_inv_sim/lo\endcsname{-0.01156}
\expandafter\def\csname N:stress/bound/ci_inv_sim/hi\endcsname{$-0.0025$}
\expandafter\def\csname F:stress/bound/ci_inv_sim/hi\endcsname{-0.0025}
\expandafter\def\csname FF:stress/bound/ci_inv_sim/hi\endcsname{-0.00250}
\expandafter\def\csname N:stress/bound/ci_inv_sim/pct\endcsname{$-31.6$}
\expandafter\def\csname F:stress/bound/ci_inv_sim/pct\endcsname{-31.6000}
\expandafter\def\csname FF:stress/bound/ci_inv_sim/pct\endcsname{-31.60000}
\expandafter\def\csname N:stress/bound/ci_inv_sim/pct_lo\endcsname{$-57.1$}
\expandafter\def\csname F:stress/bound/ci_inv_sim/pct_lo\endcsname{-57.1000}
\expandafter\def\csname FF:stress/bound/ci_inv_sim/pct_lo\endcsname{-57.10000}
\expandafter\def\csname N:stress/bound/ci_inv_sim/pct_hi\endcsname{$-12.4$}
\expandafter\def\csname F:stress/bound/ci_inv_sim/pct_hi\endcsname{-12.4000}
\expandafter\def\csname FF:stress/bound/ci_inv_sim/pct_hi\endcsname{-12.40000}
\expandafter\def\csname N:stress/bound/ci_inv_sim/by_station_pct/fairbanks\endcsname{$-32.6$}
\expandafter\def\csname F:stress/bound/ci_inv_sim/by_station_pct/fairbanks\endcsname{-32.6000}
\expandafter\def\csname FF:stress/bound/ci_inv_sim/by_station_pct/fairbanks\endcsname{-32.60000}
\expandafter\def\csname N:stress/bound/ci_inv_sim/by_station_pct/denver\endcsname{$-48.9$}
\expandafter\def\csname F:stress/bound/ci_inv_sim/by_station_pct/denver\endcsname{-48.9000}
\expandafter\def\csname FF:stress/bound/ci_inv_sim/by_station_pct/denver\endcsname{-48.90000}
\expandafter\def\csname N:stress/bound/ci_inv_sim/by_station_pct/atlanta\endcsname{$-9.9$}
\expandafter\def\csname F:stress/bound/ci_inv_sim/by_station_pct/atlanta\endcsname{-9.9000}
\expandafter\def\csname FF:stress/bound/ci_inv_sim/by_station_pct/atlanta\endcsname{-9.90000}
\expandafter\def\csname N:stress/bound/ci_inv_sim/by_station_pct/phoenix\endcsname{$-18.2$}
\expandafter\def\csname F:stress/bound/ci_inv_sim/by_station_pct/phoenix\endcsname{-18.2000}
\expandafter\def\csname FF:stress/bound/ci_inv_sim/by_station_pct/phoenix\endcsname{-18.20000}
\expandafter\def\csname N:stress/bound/ci_inv_sim/by_station_pct/singapore\endcsname{$-26.2$}
\expandafter\def\csname F:stress/bound/ci_inv_sim/by_station_pct/singapore\endcsname{-26.2000}
\expandafter\def\csname FF:stress/bound/ci_inv_sim/by_station_pct/singapore\endcsname{-26.20000}
\expandafter\def\csname N:stress/bound/ci_inv_sim/by_station_pct/minneapolis\endcsname{$-39.8$}
\expandafter\def\csname F:stress/bound/ci_inv_sim/by_station_pct/minneapolis\endcsname{-39.8000}
\expandafter\def\csname FF:stress/bound/ci_inv_sim/by_station_pct/minneapolis\endcsname{-39.80000}
\expandafter\def\csname N:stress/bound/ci_inv_sim/wins\endcsname{8}
\expandafter\def\csname W:stress/bound/ci_inv_sim/wins\endcsname{eight}
\expandafter\def\csname N:stress/bound/ci_rest_minus_model/n\endcsname{8}
\expandafter\def\csname W:stress/bound/ci_rest_minus_model/n\endcsname{eight}
\expandafter\def\csname N:stress/bound/ci_rest_minus_model/pct\endcsname{$-14.6$}
\expandafter\def\csname F:stress/bound/ci_rest_minus_model/pct\endcsname{-14.6000}
\expandafter\def\csname FF:stress/bound/ci_rest_minus_model/pct\endcsname{-14.60000}
\expandafter\def\csname N:stress/bound/ci_rest_minus_model/pct_lo\endcsname{$-44.0$}
\expandafter\def\csname F:stress/bound/ci_rest_minus_model/pct_lo\endcsname{-44.0000}
\expandafter\def\csname FF:stress/bound/ci_rest_minus_model/pct_lo\endcsname{-44.00000}
\expandafter\def\csname N:stress/bound/ci_rest_minus_model/pct_hi\endcsname{$+10.8$}
\expandafter\def\csname F:stress/bound/ci_rest_minus_model/pct_hi\endcsname{10.8000}
\expandafter\def\csname FF:stress/bound/ci_rest_minus_model/pct_hi\endcsname{10.80000}
\expandafter\def\csname N:stress/bound/ci_rest_minus_model/wins\endcsname{6}
\expandafter\def\csname W:stress/bound/ci_rest_minus_model/wins\endcsname{six}
\expandafter\def\csname N:stress/holiday_scenarios\endcsname{32}
\expandafter\def\csname N:stress/rl_big/summ/pid_water/n\endcsname{46}
\expandafter\def\csname N:stress/rl_big/summ/pid_water/score\endcsname{0.04958}
\expandafter\def\csname F:stress/rl_big/summ/pid_water/score\endcsname{0.0496}
\expandafter\def\csname FF:stress/rl_big/summ/pid_water/score\endcsname{0.04958}
\expandafter\def\csname N:stress/rl_big/summ/pid_water/comfort\endcsname{0.03491}
\expandafter\def\csname F:stress/rl_big/summ/pid_water/comfort\endcsname{0.0349}
\expandafter\def\csname FF:stress/rl_big/summ/pid_water/comfort\endcsname{0.03491}
\expandafter\def\csname N:stress/rl_big/summ/pid_water/energy\endcsname{0.00899}
\expandafter\def\csname F:stress/rl_big/summ/pid_water/energy\endcsname{0.0090}
\expandafter\def\csname FF:stress/rl_big/summ/pid_water/energy\endcsname{0.00899}
\expandafter\def\csname N:stress/rl_big/summ/pid_water/carbon\endcsname{0.00568}
\expandafter\def\csname F:stress/rl_big/summ/pid_water/carbon\endcsname{0.0057}
\expandafter\def\csname FF:stress/rl_big/summ/pid_water/carbon\endcsname{0.00568}
\expandafter\def\csname N:stress/rl_big/summ/pid_water/small\endcsname{0.04462}
\expandafter\def\csname F:stress/rl_big/summ/pid_water/small\endcsname{0.0446}
\expandafter\def\csname FF:stress/rl_big/summ/pid_water/small\endcsname{0.04462}
\expandafter\def\csname N:stress/rl_big/summ/pid_water/large\endcsname{0.05454}
\expandafter\def\csname F:stress/rl_big/summ/pid_water/large\endcsname{0.0545}
\expandafter\def\csname FF:stress/rl_big/summ/pid_water/large\endcsname{0.05454}
\expandafter\def\csname N:stress/rl_big/summ/c5/n\endcsname{46}
\expandafter\def\csname N:stress/rl_big/summ/c5/score\endcsname{0.01797}
\expandafter\def\csname F:stress/rl_big/summ/c5/score\endcsname{0.0180}
\expandafter\def\csname FF:stress/rl_big/summ/c5/score\endcsname{0.01797}
\expandafter\def\csname N:stress/rl_big/summ/c5/comfort\endcsname{0.00811}
\expandafter\def\csname F:stress/rl_big/summ/c5/comfort\endcsname{0.0081}
\expandafter\def\csname FF:stress/rl_big/summ/c5/comfort\endcsname{0.00811}
\expandafter\def\csname N:stress/rl_big/summ/c5/energy\endcsname{0.00566}
\expandafter\def\csname F:stress/rl_big/summ/c5/energy\endcsname{0.0057}
\expandafter\def\csname FF:stress/rl_big/summ/c5/energy\endcsname{0.00566}
\expandafter\def\csname N:stress/rl_big/summ/c5/carbon\endcsname{0.0042}
\expandafter\def\csname F:stress/rl_big/summ/c5/carbon\endcsname{0.0042}
\expandafter\def\csname FF:stress/rl_big/summ/c5/carbon\endcsname{0.00420}
\expandafter\def\csname N:stress/rl_big/summ/c5/small\endcsname{0.00998}
\expandafter\def\csname F:stress/rl_big/summ/c5/small\endcsname{0.0100}
\expandafter\def\csname FF:stress/rl_big/summ/c5/small\endcsname{0.00998}
\expandafter\def\csname N:stress/rl_big/summ/c5/large\endcsname{0.02596}
\expandafter\def\csname F:stress/rl_big/summ/c5/large\endcsname{0.0260}
\expandafter\def\csname FF:stress/rl_big/summ/c5/large\endcsname{0.02596}
\expandafter\def\csname N:stress/rl_big/summ/zc/n\endcsname{46}
\expandafter\def\csname N:stress/rl_big/summ/zc/score\endcsname{0.01388}
\expandafter\def\csname F:stress/rl_big/summ/zc/score\endcsname{0.0139}
\expandafter\def\csname FF:stress/rl_big/summ/zc/score\endcsname{0.01388}
\expandafter\def\csname N:stress/rl_big/summ/zc/comfort\endcsname{0.00406}
\expandafter\def\csname F:stress/rl_big/summ/zc/comfort\endcsname{0.0041}
\expandafter\def\csname FF:stress/rl_big/summ/zc/comfort\endcsname{0.00406}
\expandafter\def\csname N:stress/rl_big/summ/zc/energy\endcsname{0.00543}
\expandafter\def\csname F:stress/rl_big/summ/zc/energy\endcsname{0.0054}
\expandafter\def\csname FF:stress/rl_big/summ/zc/energy\endcsname{0.00543}
\expandafter\def\csname N:stress/rl_big/summ/zc/carbon\endcsname{0.00439}
\expandafter\def\csname F:stress/rl_big/summ/zc/carbon\endcsname{0.0044}
\expandafter\def\csname FF:stress/rl_big/summ/zc/carbon\endcsname{0.00439}
\expandafter\def\csname N:stress/rl_big/summ/zc/small\endcsname{0.00979}
\expandafter\def\csname F:stress/rl_big/summ/zc/small\endcsname{0.0098}
\expandafter\def\csname FF:stress/rl_big/summ/zc/small\endcsname{0.00979}
\expandafter\def\csname N:stress/rl_big/summ/zc/large\endcsname{0.01797}
\expandafter\def\csname F:stress/rl_big/summ/zc/large\endcsname{0.0180}
\expandafter\def\csname FF:stress/rl_big/summ/zc/large\endcsname{0.01797}
\expandafter\def\csname N:stress/rl_big/summ/inv/n\endcsname{46}
\expandafter\def\csname N:stress/rl_big/summ/inv/score\endcsname{0.00867}
\expandafter\def\csname F:stress/rl_big/summ/inv/score\endcsname{0.0087}
\expandafter\def\csname FF:stress/rl_big/summ/inv/score\endcsname{0.00867}
\expandafter\def\csname N:stress/rl_big/summ/inv/comfort\endcsname{0.00026}
\expandafter\def\csname F:stress/rl_big/summ/inv/comfort\endcsname{0.0003}
\expandafter\def\csname FF:stress/rl_big/summ/inv/comfort\endcsname{0.00026}
\expandafter\def\csname N:stress/rl_big/summ/inv/energy\endcsname{0.00464}
\expandafter\def\csname F:stress/rl_big/summ/inv/energy\endcsname{0.0046}
\expandafter\def\csname FF:stress/rl_big/summ/inv/energy\endcsname{0.00464}
\expandafter\def\csname N:stress/rl_big/summ/inv/carbon\endcsname{0.00377}
\expandafter\def\csname F:stress/rl_big/summ/inv/carbon\endcsname{0.0038}
\expandafter\def\csname FF:stress/rl_big/summ/inv/carbon\endcsname{0.00377}
\expandafter\def\csname N:stress/rl_big/summ/inv/small\endcsname{0.00596}
\expandafter\def\csname F:stress/rl_big/summ/inv/small\endcsname{0.0060}
\expandafter\def\csname FF:stress/rl_big/summ/inv/small\endcsname{0.00596}
\expandafter\def\csname N:stress/rl_big/summ/inv/large\endcsname{0.01138}
\expandafter\def\csname F:stress/rl_big/summ/inv/large\endcsname{0.0114}
\expandafter\def\csname FF:stress/rl_big/summ/inv/large\endcsname{0.01138}
\expandafter\def\csname N:stress/rl_big/summ/rules/n\endcsname{46}
\expandafter\def\csname N:stress/rl_big/summ/rules/score\endcsname{0.02149}
\expandafter\def\csname F:stress/rl_big/summ/rules/score\endcsname{0.0215}
\expandafter\def\csname FF:stress/rl_big/summ/rules/score\endcsname{0.02149}
\expandafter\def\csname N:stress/rl_big/summ/rules/comfort\endcsname{0.00882}
\expandafter\def\csname F:stress/rl_big/summ/rules/comfort\endcsname{0.0088}
\expandafter\def\csname FF:stress/rl_big/summ/rules/comfort\endcsname{0.00882}
\expandafter\def\csname N:stress/rl_big/summ/rules/energy\endcsname{0.0064}
\expandafter\def\csname F:stress/rl_big/summ/rules/energy\endcsname{0.0064}
\expandafter\def\csname FF:stress/rl_big/summ/rules/energy\endcsname{0.00640}
\expandafter\def\csname N:stress/rl_big/summ/rules/carbon\endcsname{0.00627}
\expandafter\def\csname F:stress/rl_big/summ/rules/carbon\endcsname{0.0063}
\expandafter\def\csname FF:stress/rl_big/summ/rules/carbon\endcsname{0.00627}
\expandafter\def\csname N:stress/rl_big/summ/rules/small\endcsname{0.01471}
\expandafter\def\csname F:stress/rl_big/summ/rules/small\endcsname{0.0147}
\expandafter\def\csname FF:stress/rl_big/summ/rules/small\endcsname{0.01471}
\expandafter\def\csname N:stress/rl_big/summ/rules/large\endcsname{0.02827}
\expandafter\def\csname F:stress/rl_big/summ/rules/large\endcsname{0.0283}
\expandafter\def\csname FF:stress/rl_big/summ/rules/large\endcsname{0.02827}
\expandafter\def\csname N:stress/rl_big/ci/zc_vs_c5/n\endcsname{46}
\expandafter\def\csname N:stress/rl_big/ci/zc_vs_c5/mean\endcsname{$-0.00409$}
\expandafter\def\csname F:stress/rl_big/ci/zc_vs_c5/mean\endcsname{-0.0041}
\expandafter\def\csname FF:stress/rl_big/ci/zc_vs_c5/mean\endcsname{-0.00409}
\expandafter\def\csname N:stress/rl_big/ci/zc_vs_c5/lo\endcsname{$-0.00742$}
\expandafter\def\csname F:stress/rl_big/ci/zc_vs_c5/lo\endcsname{-0.0074}
\expandafter\def\csname FF:stress/rl_big/ci/zc_vs_c5/lo\endcsname{-0.00742}
\expandafter\def\csname N:stress/rl_big/ci/zc_vs_c5/hi\endcsname{$-0.00161$}
\expandafter\def\csname F:stress/rl_big/ci/zc_vs_c5/hi\endcsname{-0.0016}
\expandafter\def\csname FF:stress/rl_big/ci/zc_vs_c5/hi\endcsname{-0.00161}
\expandafter\def\csname N:stress/rl_big/ci/zc_vs_c5/pct\endcsname{$-22.8$}
\expandafter\def\csname F:stress/rl_big/ci/zc_vs_c5/pct\endcsname{-22.8000}
\expandafter\def\csname FF:stress/rl_big/ci/zc_vs_c5/pct\endcsname{-22.80000}
\expandafter\def\csname N:stress/rl_big/ci/zc_vs_c5/pct_lo\endcsname{$-41.3$}
\expandafter\def\csname F:stress/rl_big/ci/zc_vs_c5/pct_lo\endcsname{-41.3000}
\expandafter\def\csname FF:stress/rl_big/ci/zc_vs_c5/pct_lo\endcsname{-41.30000}
\expandafter\def\csname N:stress/rl_big/ci/zc_vs_c5/pct_hi\endcsname{$-9.0$}
\expandafter\def\csname F:stress/rl_big/ci/zc_vs_c5/pct_hi\endcsname{-9.0000}
\expandafter\def\csname FF:stress/rl_big/ci/zc_vs_c5/pct_hi\endcsname{-9.00000}
\expandafter\def\csname N:stress/rl_big/ci/zc_vs_c5/by_station_pct/fairbanks\endcsname{$-18.8$}
\expandafter\def\csname F:stress/rl_big/ci/zc_vs_c5/by_station_pct/fairbanks\endcsname{-18.8000}
\expandafter\def\csname FF:stress/rl_big/ci/zc_vs_c5/by_station_pct/fairbanks\endcsname{-18.80000}
\expandafter\def\csname N:stress/rl_big/ci/zc_vs_c5/by_station_pct/denver\endcsname{$-2.2$}
\expandafter\def\csname F:stress/rl_big/ci/zc_vs_c5/by_station_pct/denver\endcsname{-2.2000}
\expandafter\def\csname FF:stress/rl_big/ci/zc_vs_c5/by_station_pct/denver\endcsname{-2.20000}
\expandafter\def\csname N:stress/rl_big/ci/zc_vs_c5/by_station_pct/atlanta\endcsname{$-19.3$}
\expandafter\def\csname F:stress/rl_big/ci/zc_vs_c5/by_station_pct/atlanta\endcsname{-19.3000}
\expandafter\def\csname FF:stress/rl_big/ci/zc_vs_c5/by_station_pct/atlanta\endcsname{-19.30000}
\expandafter\def\csname N:stress/rl_big/ci/zc_vs_c5/by_station_pct/phoenix\endcsname{$-26.4$}
\expandafter\def\csname F:stress/rl_big/ci/zc_vs_c5/by_station_pct/phoenix\endcsname{-26.4000}
\expandafter\def\csname FF:stress/rl_big/ci/zc_vs_c5/by_station_pct/phoenix\endcsname{-26.40000}
\expandafter\def\csname N:stress/rl_big/ci/zc_vs_c5/by_station_pct/singapore\endcsname{$-17.8$}
\expandafter\def\csname F:stress/rl_big/ci/zc_vs_c5/by_station_pct/singapore\endcsname{-17.8000}
\expandafter\def\csname FF:stress/rl_big/ci/zc_vs_c5/by_station_pct/singapore\endcsname{-17.80000}
\expandafter\def\csname N:stress/rl_big/ci/zc_vs_c5/by_station_pct/minneapolis\endcsname{$-12.3$}
\expandafter\def\csname F:stress/rl_big/ci/zc_vs_c5/by_station_pct/minneapolis\endcsname{-12.3000}
\expandafter\def\csname FF:stress/rl_big/ci/zc_vs_c5/by_station_pct/minneapolis\endcsname{-12.30000}
\expandafter\def\csname N:stress/rl_big/ci/zc_vs_c5/wins\endcsname{37}
\expandafter\def\csname N:stress/rl_big/ci/zc_vs_pid_water/n\endcsname{46}
\expandafter\def\csname N:stress/rl_big/ci/zc_vs_pid_water/mean\endcsname{$-0.0357$}
\expandafter\def\csname F:stress/rl_big/ci/zc_vs_pid_water/mean\endcsname{-0.0357}
\expandafter\def\csname FF:stress/rl_big/ci/zc_vs_pid_water/mean\endcsname{-0.03570}
\expandafter\def\csname N:stress/rl_big/ci/zc_vs_pid_water/lo\endcsname{$-0.04095$}
\expandafter\def\csname F:stress/rl_big/ci/zc_vs_pid_water/lo\endcsname{-0.0410}
\expandafter\def\csname FF:stress/rl_big/ci/zc_vs_pid_water/lo\endcsname{-0.04095}
\expandafter\def\csname N:stress/rl_big/ci/zc_vs_pid_water/hi\endcsname{$-0.03066$}
\expandafter\def\csname F:stress/rl_big/ci/zc_vs_pid_water/hi\endcsname{-0.0307}
\expandafter\def\csname FF:stress/rl_big/ci/zc_vs_pid_water/hi\endcsname{-0.03066}
\expandafter\def\csname N:stress/rl_big/ci/zc_vs_pid_water/pct\endcsname{$-72.0$}
\expandafter\def\csname F:stress/rl_big/ci/zc_vs_pid_water/pct\endcsname{-72.0000}
\expandafter\def\csname FF:stress/rl_big/ci/zc_vs_pid_water/pct\endcsname{-72.00000}
\expandafter\def\csname N:stress/rl_big/ci/zc_vs_pid_water/pct_lo\endcsname{$-82.6$}
\expandafter\def\csname F:stress/rl_big/ci/zc_vs_pid_water/pct_lo\endcsname{-82.6000}
\expandafter\def\csname FF:stress/rl_big/ci/zc_vs_pid_water/pct_lo\endcsname{-82.60000}
\expandafter\def\csname N:stress/rl_big/ci/zc_vs_pid_water/pct_hi\endcsname{$-61.8$}
\expandafter\def\csname F:stress/rl_big/ci/zc_vs_pid_water/pct_hi\endcsname{-61.8000}
\expandafter\def\csname FF:stress/rl_big/ci/zc_vs_pid_water/pct_hi\endcsname{-61.80000}
\expandafter\def\csname N:stress/rl_big/ci/zc_vs_pid_water/by_station_pct/fairbanks\endcsname{$-78.6$}
\expandafter\def\csname F:stress/rl_big/ci/zc_vs_pid_water/by_station_pct/fairbanks\endcsname{-78.6000}
\expandafter\def\csname FF:stress/rl_big/ci/zc_vs_pid_water/by_station_pct/fairbanks\endcsname{-78.60000}
\expandafter\def\csname N:stress/rl_big/ci/zc_vs_pid_water/by_station_pct/denver\endcsname{$-81.3$}
\expandafter\def\csname F:stress/rl_big/ci/zc_vs_pid_water/by_station_pct/denver\endcsname{-81.3000}
\expandafter\def\csname FF:stress/rl_big/ci/zc_vs_pid_water/by_station_pct/denver\endcsname{-81.30000}
\expandafter\def\csname N:stress/rl_big/ci/zc_vs_pid_water/by_station_pct/atlanta\endcsname{$-78.9$}
\expandafter\def\csname F:stress/rl_big/ci/zc_vs_pid_water/by_station_pct/atlanta\endcsname{-78.9000}
\expandafter\def\csname FF:stress/rl_big/ci/zc_vs_pid_water/by_station_pct/atlanta\endcsname{-78.90000}
\expandafter\def\csname N:stress/rl_big/ci/zc_vs_pid_water/by_station_pct/phoenix\endcsname{$-75.2$}
\expandafter\def\csname F:stress/rl_big/ci/zc_vs_pid_water/by_station_pct/phoenix\endcsname{-75.2000}
\expandafter\def\csname FF:stress/rl_big/ci/zc_vs_pid_water/by_station_pct/phoenix\endcsname{-75.20000}
\expandafter\def\csname N:stress/rl_big/ci/zc_vs_pid_water/by_station_pct/singapore\endcsname{$-70.8$}
\expandafter\def\csname F:stress/rl_big/ci/zc_vs_pid_water/by_station_pct/singapore\endcsname{-70.8000}
\expandafter\def\csname FF:stress/rl_big/ci/zc_vs_pid_water/by_station_pct/singapore\endcsname{-70.80000}
\expandafter\def\csname N:stress/rl_big/ci/zc_vs_pid_water/by_station_pct/minneapolis\endcsname{$-70.0$}
\expandafter\def\csname F:stress/rl_big/ci/zc_vs_pid_water/by_station_pct/minneapolis\endcsname{-70.0000}
\expandafter\def\csname FF:stress/rl_big/ci/zc_vs_pid_water/by_station_pct/minneapolis\endcsname{-70.00000}
\expandafter\def\csname N:stress/rl_big/ci/zc_vs_pid_water/wins\endcsname{46}
\expandafter\def\csname N:stress/rl_big/ci/c5_vs_pid_water/n\endcsname{46}
\expandafter\def\csname N:stress/rl_big/ci/c5_vs_pid_water/mean\endcsname{$-0.03161$}
\expandafter\def\csname F:stress/rl_big/ci/c5_vs_pid_water/mean\endcsname{-0.0316}
\expandafter\def\csname FF:stress/rl_big/ci/c5_vs_pid_water/mean\endcsname{-0.03161}
\expandafter\def\csname N:stress/rl_big/ci/c5_vs_pid_water/lo\endcsname{$-0.03784$}
\expandafter\def\csname F:stress/rl_big/ci/c5_vs_pid_water/lo\endcsname{-0.0378}
\expandafter\def\csname FF:stress/rl_big/ci/c5_vs_pid_water/lo\endcsname{-0.03784}
\expandafter\def\csname N:stress/rl_big/ci/c5_vs_pid_water/hi\endcsname{$-0.02539$}
\expandafter\def\csname F:stress/rl_big/ci/c5_vs_pid_water/hi\endcsname{-0.0254}
\expandafter\def\csname FF:stress/rl_big/ci/c5_vs_pid_water/hi\endcsname{-0.02539}
\expandafter\def\csname N:stress/rl_big/ci/c5_vs_pid_water/pct\endcsname{$-63.8$}
\expandafter\def\csname F:stress/rl_big/ci/c5_vs_pid_water/pct\endcsname{-63.8000}
\expandafter\def\csname FF:stress/rl_big/ci/c5_vs_pid_water/pct\endcsname{-63.80000}
\expandafter\def\csname N:stress/rl_big/ci/c5_vs_pid_water/pct_lo\endcsname{$-76.3$}
\expandafter\def\csname F:stress/rl_big/ci/c5_vs_pid_water/pct_lo\endcsname{-76.3000}
\expandafter\def\csname FF:stress/rl_big/ci/c5_vs_pid_water/pct_lo\endcsname{-76.30000}
\expandafter\def\csname N:stress/rl_big/ci/c5_vs_pid_water/pct_hi\endcsname{$-51.2$}
\expandafter\def\csname F:stress/rl_big/ci/c5_vs_pid_water/pct_hi\endcsname{-51.2000}
\expandafter\def\csname FF:stress/rl_big/ci/c5_vs_pid_water/pct_hi\endcsname{-51.20000}
\expandafter\def\csname N:stress/rl_big/ci/c5_vs_pid_water/by_station_pct/fairbanks\endcsname{$-74.5$}
\expandafter\def\csname F:stress/rl_big/ci/c5_vs_pid_water/by_station_pct/fairbanks\endcsname{-74.5000}
\expandafter\def\csname FF:stress/rl_big/ci/c5_vs_pid_water/by_station_pct/fairbanks\endcsname{-74.50000}
\expandafter\def\csname N:stress/rl_big/ci/c5_vs_pid_water/by_station_pct/denver\endcsname{$-80.6$}
\expandafter\def\csname F:stress/rl_big/ci/c5_vs_pid_water/by_station_pct/denver\endcsname{-80.6000}
\expandafter\def\csname FF:stress/rl_big/ci/c5_vs_pid_water/by_station_pct/denver\endcsname{-80.60000}
\expandafter\def\csname N:stress/rl_big/ci/c5_vs_pid_water/by_station_pct/atlanta\endcsname{$-66.8$}
\expandafter\def\csname F:stress/rl_big/ci/c5_vs_pid_water/by_station_pct/atlanta\endcsname{-66.8000}
\expandafter\def\csname FF:stress/rl_big/ci/c5_vs_pid_water/by_station_pct/atlanta\endcsname{-66.80000}
\expandafter\def\csname N:stress/rl_big/ci/c5_vs_pid_water/by_station_pct/phoenix\endcsname{$-59.1$}
\expandafter\def\csname F:stress/rl_big/ci/c5_vs_pid_water/by_station_pct/phoenix\endcsname{-59.1000}
\expandafter\def\csname FF:stress/rl_big/ci/c5_vs_pid_water/by_station_pct/phoenix\endcsname{-59.10000}
\expandafter\def\csname N:stress/rl_big/ci/c5_vs_pid_water/by_station_pct/singapore\endcsname{$-60.4$}
\expandafter\def\csname F:stress/rl_big/ci/c5_vs_pid_water/by_station_pct/singapore\endcsname{-60.4000}
\expandafter\def\csname FF:stress/rl_big/ci/c5_vs_pid_water/by_station_pct/singapore\endcsname{-60.40000}
\expandafter\def\csname N:stress/rl_big/ci/c5_vs_pid_water/by_station_pct/minneapolis\endcsname{$-64.3$}
\expandafter\def\csname F:stress/rl_big/ci/c5_vs_pid_water/by_station_pct/minneapolis\endcsname{-64.3000}
\expandafter\def\csname FF:stress/rl_big/ci/c5_vs_pid_water/by_station_pct/minneapolis\endcsname{-64.30000}
\expandafter\def\csname N:stress/rl_big/ci/c5_vs_pid_water/wins\endcsname{45}
\expandafter\def\csname N:stress/rl_big/ci/inv_vs_zc/n\endcsname{46}
\expandafter\def\csname N:stress/rl_big/ci/inv_vs_zc/mean\endcsname{$-0.00521$}
\expandafter\def\csname F:stress/rl_big/ci/inv_vs_zc/mean\endcsname{-0.0052}
\expandafter\def\csname FF:stress/rl_big/ci/inv_vs_zc/mean\endcsname{-0.00521}
\expandafter\def\csname N:stress/rl_big/ci/inv_vs_zc/lo\endcsname{$-0.00763$}
\expandafter\def\csname F:stress/rl_big/ci/inv_vs_zc/lo\endcsname{-0.0076}
\expandafter\def\csname FF:stress/rl_big/ci/inv_vs_zc/lo\endcsname{-0.00763}
\expandafter\def\csname N:stress/rl_big/ci/inv_vs_zc/hi\endcsname{$-0.00328$}
\expandafter\def\csname F:stress/rl_big/ci/inv_vs_zc/hi\endcsname{-0.0033}
\expandafter\def\csname FF:stress/rl_big/ci/inv_vs_zc/hi\endcsname{-0.00328}
\expandafter\def\csname N:stress/rl_big/ci/inv_vs_zc/pct\endcsname{$-37.5$}
\expandafter\def\csname F:stress/rl_big/ci/inv_vs_zc/pct\endcsname{-37.5000}
\expandafter\def\csname FF:stress/rl_big/ci/inv_vs_zc/pct\endcsname{-37.50000}
\expandafter\def\csname N:stress/rl_big/ci/inv_vs_zc/pct_lo\endcsname{$-55.0$}
\expandafter\def\csname F:stress/rl_big/ci/inv_vs_zc/pct_lo\endcsname{-55.0000}
\expandafter\def\csname FF:stress/rl_big/ci/inv_vs_zc/pct_lo\endcsname{-55.00000}
\expandafter\def\csname N:stress/rl_big/ci/inv_vs_zc/pct_hi\endcsname{$-23.7$}
\expandafter\def\csname F:stress/rl_big/ci/inv_vs_zc/pct_hi\endcsname{-23.7000}
\expandafter\def\csname FF:stress/rl_big/ci/inv_vs_zc/pct_hi\endcsname{-23.70000}
\expandafter\def\csname N:stress/rl_big/ci/inv_vs_zc/by_station_pct/fairbanks\endcsname{$-43.4$}
\expandafter\def\csname F:stress/rl_big/ci/inv_vs_zc/by_station_pct/fairbanks\endcsname{-43.4000}
\expandafter\def\csname FF:stress/rl_big/ci/inv_vs_zc/by_station_pct/fairbanks\endcsname{-43.40000}
\expandafter\def\csname N:stress/rl_big/ci/inv_vs_zc/by_station_pct/denver\endcsname{$-42.0$}
\expandafter\def\csname F:stress/rl_big/ci/inv_vs_zc/by_station_pct/denver\endcsname{-42.0000}
\expandafter\def\csname FF:stress/rl_big/ci/inv_vs_zc/by_station_pct/denver\endcsname{-42.00000}
\expandafter\def\csname N:stress/rl_big/ci/inv_vs_zc/by_station_pct/atlanta\endcsname{$-41.0$}
\expandafter\def\csname F:stress/rl_big/ci/inv_vs_zc/by_station_pct/atlanta\endcsname{-41.0000}
\expandafter\def\csname FF:stress/rl_big/ci/inv_vs_zc/by_station_pct/atlanta\endcsname{-41.00000}
\expandafter\def\csname N:stress/rl_big/ci/inv_vs_zc/by_station_pct/phoenix\endcsname{$-30.5$}
\expandafter\def\csname F:stress/rl_big/ci/inv_vs_zc/by_station_pct/phoenix\endcsname{-30.5000}
\expandafter\def\csname FF:stress/rl_big/ci/inv_vs_zc/by_station_pct/phoenix\endcsname{-30.50000}
\expandafter\def\csname N:stress/rl_big/ci/inv_vs_zc/by_station_pct/singapore\endcsname{$-20.4$}
\expandafter\def\csname F:stress/rl_big/ci/inv_vs_zc/by_station_pct/singapore\endcsname{-20.4000}
\expandafter\def\csname FF:stress/rl_big/ci/inv_vs_zc/by_station_pct/singapore\endcsname{-20.40000}
\expandafter\def\csname N:stress/rl_big/ci/inv_vs_zc/by_station_pct/minneapolis\endcsname{$-40.4$}
\expandafter\def\csname F:stress/rl_big/ci/inv_vs_zc/by_station_pct/minneapolis\endcsname{-40.4000}
\expandafter\def\csname FF:stress/rl_big/ci/inv_vs_zc/by_station_pct/minneapolis\endcsname{-40.40000}
\expandafter\def\csname N:stress/rl_big/ci/inv_vs_zc/wins\endcsname{46}
\expandafter\def\csname N:stress/rl_big/ci/zc_vs_rules/n\endcsname{46}
\expandafter\def\csname N:stress/rl_big/ci/zc_vs_rules/mean\endcsname{$-0.00761$}
\expandafter\def\csname F:stress/rl_big/ci/zc_vs_rules/mean\endcsname{-0.0076}
\expandafter\def\csname FF:stress/rl_big/ci/zc_vs_rules/mean\endcsname{-0.00761}
\expandafter\def\csname N:stress/rl_big/ci/zc_vs_rules/lo\endcsname{$-0.01048$}
\expandafter\def\csname F:stress/rl_big/ci/zc_vs_rules/lo\endcsname{-0.0105}
\expandafter\def\csname FF:stress/rl_big/ci/zc_vs_rules/lo\endcsname{-0.01048}
\expandafter\def\csname N:stress/rl_big/ci/zc_vs_rules/hi\endcsname{$-0.00509$}
\expandafter\def\csname F:stress/rl_big/ci/zc_vs_rules/hi\endcsname{-0.0051}
\expandafter\def\csname FF:stress/rl_big/ci/zc_vs_rules/hi\endcsname{-0.00509}
\expandafter\def\csname N:stress/rl_big/ci/zc_vs_rules/pct\endcsname{$-35.4$}
\expandafter\def\csname F:stress/rl_big/ci/zc_vs_rules/pct\endcsname{-35.4000}
\expandafter\def\csname FF:stress/rl_big/ci/zc_vs_rules/pct\endcsname{-35.40000}
\expandafter\def\csname N:stress/rl_big/ci/zc_vs_rules/pct_lo\endcsname{$-48.8$}
\expandafter\def\csname F:stress/rl_big/ci/zc_vs_rules/pct_lo\endcsname{-48.8000}
\expandafter\def\csname FF:stress/rl_big/ci/zc_vs_rules/pct_lo\endcsname{-48.80000}
\expandafter\def\csname N:stress/rl_big/ci/zc_vs_rules/pct_hi\endcsname{$-23.7$}
\expandafter\def\csname F:stress/rl_big/ci/zc_vs_rules/pct_hi\endcsname{-23.7000}
\expandafter\def\csname FF:stress/rl_big/ci/zc_vs_rules/pct_hi\endcsname{-23.70000}
\expandafter\def\csname N:stress/rl_big/ci/zc_vs_rules/by_station_pct/fairbanks\endcsname{$-40.1$}
\expandafter\def\csname F:stress/rl_big/ci/zc_vs_rules/by_station_pct/fairbanks\endcsname{-40.1000}
\expandafter\def\csname FF:stress/rl_big/ci/zc_vs_rules/by_station_pct/fairbanks\endcsname{-40.10000}
\expandafter\def\csname N:stress/rl_big/ci/zc_vs_rules/by_station_pct/denver\endcsname{$-35.1$}
\expandafter\def\csname F:stress/rl_big/ci/zc_vs_rules/by_station_pct/denver\endcsname{-35.1000}
\expandafter\def\csname FF:stress/rl_big/ci/zc_vs_rules/by_station_pct/denver\endcsname{-35.10000}
\expandafter\def\csname N:stress/rl_big/ci/zc_vs_rules/by_station_pct/atlanta\endcsname{$-33.3$}
\expandafter\def\csname F:stress/rl_big/ci/zc_vs_rules/by_station_pct/atlanta\endcsname{-33.3000}
\expandafter\def\csname FF:stress/rl_big/ci/zc_vs_rules/by_station_pct/atlanta\endcsname{-33.30000}
\expandafter\def\csname N:stress/rl_big/ci/zc_vs_rules/by_station_pct/phoenix\endcsname{$-32.0$}
\expandafter\def\csname F:stress/rl_big/ci/zc_vs_rules/by_station_pct/phoenix\endcsname{-32.0000}
\expandafter\def\csname FF:stress/rl_big/ci/zc_vs_rules/by_station_pct/phoenix\endcsname{-32.00000}
\expandafter\def\csname N:stress/rl_big/ci/zc_vs_rules/by_station_pct/singapore\endcsname{$-31.8$}
\expandafter\def\csname F:stress/rl_big/ci/zc_vs_rules/by_station_pct/singapore\endcsname{-31.8000}
\expandafter\def\csname FF:stress/rl_big/ci/zc_vs_rules/by_station_pct/singapore\endcsname{-31.80000}
\expandafter\def\csname N:stress/rl_big/ci/zc_vs_rules/by_station_pct/minneapolis\endcsname{$-30.3$}
\expandafter\def\csname F:stress/rl_big/ci/zc_vs_rules/by_station_pct/minneapolis\endcsname{-30.3000}
\expandafter\def\csname FF:stress/rl_big/ci/zc_vs_rules/by_station_pct/minneapolis\endcsname{-30.30000}
\expandafter\def\csname N:stress/rl_big/ci/zc_vs_rules/wins\endcsname{41}
\expandafter\def\csname N:stress/rl_big/min_unocc/c5/n\endcsname{46}
\expandafter\def\csname N:stress/rl_big/min_unocc/c5/min\endcsname{$-3.2$}
\expandafter\def\csname F:stress/rl_big/min_unocc/c5/min\endcsname{-3.2000}
\expandafter\def\csname FF:stress/rl_big/min_unocc/c5/min\endcsname{-3.20000}
\expandafter\def\csname N:stress/rl_big/min_unocc/c5/p10\endcsname{1.6}
\expandafter\def\csname F:stress/rl_big/min_unocc/c5/p10\endcsname{1.6000}
\expandafter\def\csname FF:stress/rl_big/min_unocc/c5/p10\endcsname{1.60000}
\expandafter\def\csname N:stress/rl_big/min_unocc/c5/median\endcsname{17.8}
\expandafter\def\csname F:stress/rl_big/min_unocc/c5/median\endcsname{17.8000}
\expandafter\def\csname FF:stress/rl_big/min_unocc/c5/median\endcsname{17.80000}
\expandafter\def\csname N:stress/rl_big/min_unocc/zc/n\endcsname{46}
\expandafter\def\csname N:stress/rl_big/min_unocc/zc/min\endcsname{$-3.2$}
\expandafter\def\csname F:stress/rl_big/min_unocc/zc/min\endcsname{-3.2000}
\expandafter\def\csname FF:stress/rl_big/min_unocc/zc/min\endcsname{-3.20000}
\expandafter\def\csname N:stress/rl_big/min_unocc/zc/p10\endcsname{1.5}
\expandafter\def\csname F:stress/rl_big/min_unocc/zc/p10\endcsname{1.5000}
\expandafter\def\csname FF:stress/rl_big/min_unocc/zc/p10\endcsname{1.50000}
\expandafter\def\csname N:stress/rl_big/min_unocc/zc/median\endcsname{17.8}
\expandafter\def\csname F:stress/rl_big/min_unocc/zc/median\endcsname{17.8000}
\expandafter\def\csname FF:stress/rl_big/min_unocc/zc/median\endcsname{17.80000}
\expandafter\def\csname N:stress/rl_big/cold/n\endcsname{6}
\expandafter\def\csname W:stress/rl_big/cold/n\endcsname{six}
\expandafter\def\csname N:stress/rl_big/cold/zc/min\endcsname{$-3.2$}
\expandafter\def\csname F:stress/rl_big/cold/zc/min\endcsname{-3.2000}
\expandafter\def\csname FF:stress/rl_big/cold/zc/min\endcsname{-3.20000}
\expandafter\def\csname N:stress/rl_big/cold/zc/max\endcsname{1.7}
\expandafter\def\csname F:stress/rl_big/cold/zc/max\endcsname{1.7000}
\expandafter\def\csname FF:stress/rl_big/cold/zc/max\endcsname{1.70000}
\expandafter\def\csname N:stress/rl_big/cold/pid_water/min\endcsname{$-0.9$}
\expandafter\def\csname F:stress/rl_big/cold/pid_water/min\endcsname{-0.9000}
\expandafter\def\csname FF:stress/rl_big/cold/pid_water/min\endcsname{-0.90000}
\expandafter\def\csname N:stress/rl_big/cold/pid_water/max\endcsname{6.1}
\expandafter\def\csname F:stress/rl_big/cold/pid_water/max\endcsname{6.1000}
\expandafter\def\csname FF:stress/rl_big/cold/pid_water/max\endcsname{6.10000}
\expandafter\def\csname N:stress/rl_big/cold/r1/min\endcsname{9.8}
\expandafter\def\csname F:stress/rl_big/cold/r1/min\endcsname{9.8000}
\expandafter\def\csname FF:stress/rl_big/cold/r1/min\endcsname{9.80000}
\expandafter\def\csname N:stress/rl_big/cold/r1/max\endcsname{13.3}
\expandafter\def\csname F:stress/rl_big/cold/r1/max\endcsname{13.3000}
\expandafter\def\csname FF:stress/rl_big/cold/r1/max\endcsname{13.30000}
\expandafter\def\csname N:stress/rl_big/cold/gap_vs_pid/min\endcsname{0.5}
\expandafter\def\csname F:stress/rl_big/cold/gap_vs_pid/min\endcsname{0.5000}
\expandafter\def\csname FF:stress/rl_big/cold/gap_vs_pid/min\endcsname{0.50000}
\expandafter\def\csname N:stress/rl_big/cold/gap_vs_pid/max\endcsname{5.5}
\expandafter\def\csname F:stress/rl_big/cold/gap_vs_pid/max\endcsname{5.5000}
\expandafter\def\csname FF:stress/rl_big/cold/gap_vs_pid/max\endcsname{5.50000}
\expandafter\def\csname N:stress/min_unocc_f278/n\endcsname{48}
\expandafter\def\csname N:stress/min_unocc_f278/min\endcsname{0.4}
\expandafter\def\csname F:stress/min_unocc_f278/min\endcsname{0.4000}
\expandafter\def\csname FF:stress/min_unocc_f278/min\endcsname{0.40000}
\expandafter\def\csname N:stress/min_unocc_f278/p10\endcsname{4.3}
\expandafter\def\csname F:stress/min_unocc_f278/p10\endcsname{4.3000}
\expandafter\def\csname FF:stress/min_unocc_f278/p10\endcsname{4.30000}
\expandafter\def\csname N:stress/min_unocc_f278/median\endcsname{17.6}
\expandafter\def\csname F:stress/min_unocc_f278/median\endcsname{17.6000}
\expandafter\def\csname FF:stress/min_unocc_f278/median\endcsname{17.60000}
\expandafter\def\csname N:stress/ahu_off/planner/pooled\endcsname{4.0}
\expandafter\def\csname F:stress/ahu_off/planner/pooled\endcsname{4.0000}
\expandafter\def\csname FF:stress/ahu_off/planner/pooled\endcsname{4.00000}
\expandafter\def\csname N:stress/ahu_off/planner/median\endcsname{0.4}
\expandafter\def\csname F:stress/ahu_off/planner/median\endcsname{0.4000}
\expandafter\def\csname FF:stress/ahu_off/planner/median\endcsname{0.40000}
\expandafter\def\csname N:stress/ahu_off/planner/max\endcsname{28.3}
\expandafter\def\csname F:stress/ahu_off/planner/max\endcsname{28.3000}
\expandafter\def\csname FF:stress/ahu_off/planner/max\endcsname{28.30000}
\expandafter\def\csname N:stress/ahu_off/inv/pooled\endcsname{1.1}
\expandafter\def\csname F:stress/ahu_off/inv/pooled\endcsname{1.1000}
\expandafter\def\csname FF:stress/ahu_off/inv/pooled\endcsname{1.10000}
\expandafter\def\csname N:stress/ahu_off/inv/median\endcsname{0.6}
\expandafter\def\csname F:stress/ahu_off/inv/median\endcsname{0.6000}
\expandafter\def\csname FF:stress/ahu_off/inv/median\endcsname{0.60000}
\expandafter\def\csname N:stress/ahu_off/inv/max\endcsname{7.5}
\expandafter\def\csname F:stress/ahu_off/inv/max\endcsname{7.5000}
\expandafter\def\csname FF:stress/ahu_off/inv/max\endcsname{7.50000}
\expandafter\def\csname N:stress/signed_err/small/fitted/heat\endcsname{1.47}
\expandafter\def\csname F:stress/signed_err/small/fitted/heat\endcsname{1.4700}
\expandafter\def\csname FF:stress/signed_err/small/fitted/heat\endcsname{1.47000}
\expandafter\def\csname N:stress/signed_err/small/fitted/cool\endcsname{$-0.38$}
\expandafter\def\csname F:stress/signed_err/small/fitted/cool\endcsname{-0.3800}
\expandafter\def\csname FF:stress/signed_err/small/fitted/cool\endcsname{-0.38000}
\expandafter\def\csname N:stress/signed_err/small/fitted/idle\endcsname{0.23}
\expandafter\def\csname F:stress/signed_err/small/fitted/idle\endcsname{0.2300}
\expandafter\def\csname FF:stress/signed_err/small/fitted/idle\endcsname{0.23000}
\expandafter\def\csname N:stress/signed_err/small/halved/heat\endcsname{0.6}
\expandafter\def\csname F:stress/signed_err/small/halved/heat\endcsname{0.6000}
\expandafter\def\csname FF:stress/signed_err/small/halved/heat\endcsname{0.60000}
\expandafter\def\csname N:stress/signed_err/small/halved/cool\endcsname{0.35}
\expandafter\def\csname F:stress/signed_err/small/halved/cool\endcsname{0.3500}
\expandafter\def\csname FF:stress/signed_err/small/halved/cool\endcsname{0.35000}
\expandafter\def\csname N:stress/signed_err/small/halved/idle\endcsname{0.29}
\expandafter\def\csname F:stress/signed_err/small/halved/idle\endcsname{0.2900}
\expandafter\def\csname FF:stress/signed_err/small/halved/idle\endcsname{0.29000}
\expandafter\def\csname N:stress/signed_err/large/fitted/heat\endcsname{3.45}
\expandafter\def\csname F:stress/signed_err/large/fitted/heat\endcsname{3.4500}
\expandafter\def\csname FF:stress/signed_err/large/fitted/heat\endcsname{3.45000}
\expandafter\def\csname N:stress/signed_err/large/fitted/cool\endcsname{$-2.47$}
\expandafter\def\csname F:stress/signed_err/large/fitted/cool\endcsname{-2.4700}
\expandafter\def\csname FF:stress/signed_err/large/fitted/cool\endcsname{-2.47000}
\expandafter\def\csname N:stress/signed_err/large/fitted/idle\endcsname{$-0.21$}
\expandafter\def\csname F:stress/signed_err/large/fitted/idle\endcsname{-0.2100}
\expandafter\def\csname FF:stress/signed_err/large/fitted/idle\endcsname{-0.21000}
\expandafter\def\csname N:stress/signed_err/large/halved/heat\endcsname{1.85}
\expandafter\def\csname F:stress/signed_err/large/halved/heat\endcsname{1.8500}
\expandafter\def\csname FF:stress/signed_err/large/halved/heat\endcsname{1.85000}
\expandafter\def\csname N:stress/signed_err/large/halved/cool\endcsname{$-0.94$}
\expandafter\def\csname F:stress/signed_err/large/halved/cool\endcsname{-0.9400}
\expandafter\def\csname FF:stress/signed_err/large/halved/cool\endcsname{-0.94000}
\expandafter\def\csname N:stress/signed_err/large/halved/idle\endcsname{$-0.11$}
\expandafter\def\csname F:stress/signed_err/large/halved/idle\endcsname{-0.1100}
\expandafter\def\csname FF:stress/signed_err/large/halved/idle\endcsname{-0.11000}
\expandafter\def\csname N:kaggle/first\endcsname{0.01823}
\expandafter\def\csname F:kaggle/first\endcsname{0.0182}
\expandafter\def\csname FF:kaggle/first\endcsname{0.01823}
\expandafter\def\csname N:kaggle/c5\endcsname{0.01623}
\expandafter\def\csname F:kaggle/c5\endcsname{0.0162}
\expandafter\def\csname FF:kaggle/c5\endcsname{0.01623}
\expandafter\def\csname N:kaggle/zc\endcsname{0.01602}
\expandafter\def\csname F:kaggle/zc\endcsname{0.0160}
\expandafter\def\csname FF:kaggle/zc\endcsname{0.01602}
\expandafter\def\csname N:kaggle/inv\endcsname{0.01032}
\expandafter\def\csname F:kaggle/inv\endcsname{0.0103}
\expandafter\def\csname FF:kaggle/inv\endcsname{0.01032}
\expandafter\def\csname N:kaggle/ctrlw5\endcsname{0.01024}
\expandafter\def\csname F:kaggle/ctrlw5\endcsname{0.0102}
\expandafter\def\csname FF:kaggle/ctrlw5\endcsname{0.01024}
\expandafter\def\csname N:kaggle/leader\endcsname{0.0043}
\expandafter\def\csname F:kaggle/leader\endcsname{0.0043}
\expandafter\def\csname FF:kaggle/leader\endcsname{0.00430}
\expandafter\def\csname N:tuning_prior/evals\endcsname{24}
\expandafter\def\csname N:tuning_prior/first\endcsname{0.05022}
\expandafter\def\csname F:tuning_prior/first\endcsname{0.0502}
\expandafter\def\csname FF:tuning_prior/first\endcsname{0.05022}
\expandafter\def\csname N:tuning_prior/best\endcsname{0.02152}
\expandafter\def\csname F:tuning_prior/best\endcsname{0.0215}
\expandafter\def\csname FF:tuning_prior/best\endcsname{0.02152}
\expandafter\def\csname N:tuning_prior/params/band_margin\endcsname{0.45698842106847315}
\expandafter\def\csname F:tuning_prior/params/band_margin\endcsname{0.4570}
\expandafter\def\csname FF:tuning_prior/params/band_margin\endcsname{0.45699}
\expandafter\def\csname N:tuning_prior/params/kp_heat\endcsname{0.21288930485536883}
\expandafter\def\csname F:tuning_prior/params/kp_heat\endcsname{0.2129}
\expandafter\def\csname FF:tuning_prior/params/kp_heat\endcsname{0.21289}
\expandafter\def\csname N:tuning_prior/params/kp_cool\endcsname{0.612476901781966}
\expandafter\def\csname F:tuning_prior/params/kp_cool\endcsname{0.6125}
\expandafter\def\csname FF:tuning_prior/params/kp_cool\endcsname{0.61248}
\expandafter\def\csname N:tuning_prior/params/heat_damper\endcsname{0.5193313569084202}
\expandafter\def\csname F:tuning_prior/params/heat_damper\endcsname{0.5193}
\expandafter\def\csname FF:tuning_prior/params/heat_damper\endcsname{0.51933}
\expandafter\def\csname N:tuning_prior/params/sat_cool\endcsname{285.148541453209}
\expandafter\def\csname F:tuning_prior/params/sat_cool\endcsname{285.1485}
\expandafter\def\csname FF:tuning_prior/params/sat_cool\endcsname{285.14854}
\expandafter\def\csname N:tuning_prior/params/sat_idle\endcsname{290.20647768639026}
\expandafter\def\csname F:tuning_prior/params/sat_idle\endcsname{290.2065}
\expandafter\def\csname FF:tuning_prior/params/sat_idle\endcsname{290.20648}
\expandafter\def\csname N:tuning_prior/params/p_max\endcsname{925.1245767463532}
\expandafter\def\csname F:tuning_prior/params/p_max\endcsname{925.1246}
\expandafter\def\csname FF:tuning_prior/params/p_max\endcsname{925.12458}
\expandafter\def\csname N:tuning_prior/params/tw_max\endcsname{346.7738580132923}
\expandafter\def\csname F:tuning_prior/params/tw_max\endcsname{346.7739}
\expandafter\def\csname FF:tuning_prior/params/tw_max\endcsname{346.77386}
\expandafter\def\csname N:tuning_prior/params/tw_min\endcsname{315.80458259448955}
\expandafter\def\csname F:tuning_prior/params/tw_min\endcsname{315.8046}
\expandafter\def\csname FF:tuning_prior/params/tw_min\endcsname{315.80458}
\expandafter\def\csname N:tuning_prior/params/default_lead_min\endcsname{59.430073320759675}
\expandafter\def\csname F:tuning_prior/params/default_lead_min\endcsname{59.4301}
\expandafter\def\csname FF:tuning_prior/params/default_lead_min\endcsname{59.43007}
\expandafter\def\csname N:tuning_prior/params/max_lead_min\endcsname{239.09922648538438}
\expandafter\def\csname F:tuning_prior/params/max_lead_min\endcsname{239.0992}
\expandafter\def\csname FF:tuning_prior/params/max_lead_min\endcsname{239.09923}
\expandafter\def\csname N:tuning_prior/params/setback_margin\endcsname{0.004269045768508665}
\expandafter\def\csname F:tuning_prior/params/setback_margin\endcsname{0.0043}
\expandafter\def\csname FF:tuning_prior/params/setback_margin\endcsname{0.00427}
\expandafter\def\csname N:tuning_prior/params/econ_margin\endcsname{1.175819211710579}
\expandafter\def\csname F:tuning_prior/params/econ_margin\endcsname{1.1758}
\expandafter\def\csname FF:tuning_prior/params/econ_margin\endcsname{1.17582}
\expandafter\def\csname N:tuning_prior/params/guard_h\endcsname{12.0}
\expandafter\def\csname F:tuning_prior/params/guard_h\endcsname{12.0000}
\expandafter\def\csname FF:tuning_prior/params/guard_h\endcsname{12.00000}
\expandafter\def\csname N:tuning_prior/params/guard_margin\endcsname{0.2}
\expandafter\def\csname F:tuning_prior/params/guard_margin\endcsname{0.2000}
\expandafter\def\csname FF:tuning_prior/params/guard_margin\endcsname{0.20000}
\expandafter\def\csname N:tuning_prior/params/probe_budget_ahu_h\endcsname{1.0}
\expandafter\def\csname F:tuning_prior/params/probe_budget_ahu_h\endcsname{1.0000}
\expandafter\def\csname FF:tuning_prior/params/probe_budget_ahu_h\endcsname{1.00000}
\expandafter\def\csname N:tuning_prior/params/q_rw\endcsname{1e-06}
\expandafter\def\csname F:tuning_prior/params/q_rw\endcsname{0.0000}
\expandafter\def\csname FF:tuning_prior/params/q_rw\endcsname{0.00000}
\expandafter\def\csname N:tuning_rules/evals\endcsname{24}
\expandafter\def\csname N:tuning_rules/first\endcsname{0.0499}
\expandafter\def\csname F:tuning_rules/first\endcsname{0.0499}
\expandafter\def\csname FF:tuning_rules/first\endcsname{0.04990}
\expandafter\def\csname N:tuning_rules/best\endcsname{0.01951}
\expandafter\def\csname F:tuning_rules/best\endcsname{0.0195}
\expandafter\def\csname FF:tuning_rules/best\endcsname{0.01951}
\expandafter\def\csname N:tuning_rules/params/band_margin\endcsname{0.8184626182664413}
\expandafter\def\csname F:tuning_rules/params/band_margin\endcsname{0.8185}
\expandafter\def\csname FF:tuning_rules/params/band_margin\endcsname{0.81846}
\expandafter\def\csname N:tuning_rules/params/kp_heat\endcsname{0.40742234065329996}
\expandafter\def\csname F:tuning_rules/params/kp_heat\endcsname{0.4074}
\expandafter\def\csname FF:tuning_rules/params/kp_heat\endcsname{0.40742}
\expandafter\def\csname N:tuning_rules/params/kp_cool\endcsname{0.20044284781156307}
\expandafter\def\csname F:tuning_rules/params/kp_cool\endcsname{0.2004}
\expandafter\def\csname FF:tuning_rules/params/kp_cool\endcsname{0.20044}
\expandafter\def\csname N:tuning_rules/params/heat_damper\endcsname{0.5786292254124351}
\expandafter\def\csname F:tuning_rules/params/heat_damper\endcsname{0.5786}
\expandafter\def\csname FF:tuning_rules/params/heat_damper\endcsname{0.57863}
\expandafter\def\csname N:tuning_rules/params/sat_cool\endcsname{285.04467206772875}
\expandafter\def\csname F:tuning_rules/params/sat_cool\endcsname{285.0447}
\expandafter\def\csname FF:tuning_rules/params/sat_cool\endcsname{285.04467}
\expandafter\def\csname N:tuning_rules/params/sat_idle\endcsname{288.9884157235099}
\expandafter\def\csname F:tuning_rules/params/sat_idle\endcsname{288.9884}
\expandafter\def\csname FF:tuning_rules/params/sat_idle\endcsname{288.98842}
\expandafter\def\csname N:tuning_rules/params/p_max\endcsname{400.22930071970325}
\expandafter\def\csname F:tuning_rules/params/p_max\endcsname{400.2293}
\expandafter\def\csname FF:tuning_rules/params/p_max\endcsname{400.22930}
\expandafter\def\csname N:tuning_rules/params/tw_max\endcsname{348.2734360907867}
\expandafter\def\csname F:tuning_rules/params/tw_max\endcsname{348.2734}
\expandafter\def\csname FF:tuning_rules/params/tw_max\endcsname{348.27344}
\expandafter\def\csname N:tuning_rules/params/tw_min\endcsname{313.55665147387697}
\expandafter\def\csname F:tuning_rules/params/tw_min\endcsname{313.5567}
\expandafter\def\csname FF:tuning_rules/params/tw_min\endcsname{313.55665}
\expandafter\def\csname N:tuning_rules/params/default_lead_min\endcsname{23.021195406416613}
\expandafter\def\csname F:tuning_rules/params/default_lead_min\endcsname{23.0212}
\expandafter\def\csname FF:tuning_rules/params/default_lead_min\endcsname{23.02120}
\expandafter\def\csname N:tuning_rules/params/max_lead_min\endcsname{180.0}
\expandafter\def\csname F:tuning_rules/params/max_lead_min\endcsname{180.0000}
\expandafter\def\csname FF:tuning_rules/params/max_lead_min\endcsname{180.00000}
\expandafter\def\csname N:tuning_rules/params/setback_margin\endcsname{0.5286794497115396}
\expandafter\def\csname F:tuning_rules/params/setback_margin\endcsname{0.5287}
\expandafter\def\csname FF:tuning_rules/params/setback_margin\endcsname{0.52868}
\expandafter\def\csname N:tuning_rules/params/econ_margin\endcsname{1.31401116579391}
\expandafter\def\csname F:tuning_rules/params/econ_margin\endcsname{1.3140}
\expandafter\def\csname FF:tuning_rules/params/econ_margin\endcsname{1.31401}
\expandafter\def\csname N:tuning_rules/params/guard_h\endcsname{12.0}
\expandafter\def\csname F:tuning_rules/params/guard_h\endcsname{12.0000}
\expandafter\def\csname FF:tuning_rules/params/guard_h\endcsname{12.00000}
\expandafter\def\csname N:tuning_rules/params/guard_margin\endcsname{0.2}
\expandafter\def\csname F:tuning_rules/params/guard_margin\endcsname{0.2000}
\expandafter\def\csname FF:tuning_rules/params/guard_margin\endcsname{0.20000}
\expandafter\def\csname N:tuning_rules/params/probe_budget_ahu_h\endcsname{1.0}
\expandafter\def\csname F:tuning_rules/params/probe_budget_ahu_h\endcsname{1.0000}
\expandafter\def\csname FF:tuning_rules/params/probe_budget_ahu_h\endcsname{1.00000}
\expandafter\def\csname N:tuning_rules/params/q_rw\endcsname{1e-06}
\expandafter\def\csname F:tuning_rules/params/q_rw\endcsname{0.0000}
\expandafter\def\csname FF:tuning_rules/params/q_rw\endcsname{0.00000}
\expandafter\def\csname N:tuning_pid/evals\endcsname{24}
\expandafter\def\csname N:tuning_pid/first\endcsname{0.11584}
\expandafter\def\csname F:tuning_pid/first\endcsname{0.1158}
\expandafter\def\csname FF:tuning_pid/first\endcsname{0.11584}
\expandafter\def\csname N:tuning_pid/best\endcsname{0.04915}
\expandafter\def\csname F:tuning_pid/best\endcsname{0.0491}
\expandafter\def\csname FF:tuning_pid/best\endcsname{0.04915}
\expandafter\def\csname N:tuning_pid/params/kp\endcsname{0.7384174328782389}
\expandafter\def\csname F:tuning_pid/params/kp\endcsname{0.7384}
\expandafter\def\csname FF:tuning_pid/params/kp\endcsname{0.73842}
\expandafter\def\csname N:tuning_pid/params/ki\endcsname{0.06089596610523387}
\expandafter\def\csname F:tuning_pid/params/ki\endcsname{0.0609}
\expandafter\def\csname FF:tuning_pid/params/ki\endcsname{0.06090}
\expandafter\def\csname N:tuning_pid/params/kd\endcsname{0.08005840225332828}
\expandafter\def\csname F:tuning_pid/params/kd\endcsname{0.0801}
\expandafter\def\csname FF:tuning_pid/params/kd\endcsname{0.08006}
\expandafter\def\csname N:tuning_pid/params/zone_damper_full_k\endcsname{3.624648237680313}
\expandafter\def\csname F:tuning_pid/params/zone_damper_full_k\endcsname{3.6246}
\expandafter\def\csname FF:tuning_pid/params/zone_damper_full_k\endcsname{3.62465}
\expandafter\def\csname N:x5_guard/aether-prior/n\endcsname{1}
\expandafter\def\csname W:x5_guard/aether-prior/n\endcsname{one}
\expandafter\def\csname N:x5_guard/aether-prior/score\endcsname{0.00525}
\expandafter\def\csname F:x5_guard/aether-prior/score\endcsname{0.0053}
\expandafter\def\csname FF:x5_guard/aether-prior/score\endcsname{0.00525}
\expandafter\def\csname N:x5_guard/aether-prior/comfort\endcsname{0.00186}
\expandafter\def\csname F:x5_guard/aether-prior/comfort\endcsname{0.0019}
\expandafter\def\csname FF:x5_guard/aether-prior/comfort\endcsname{0.00186}
\expandafter\def\csname N:x5_guard/aether-prior/energy\endcsname{0.00185}
\expandafter\def\csname F:x5_guard/aether-prior/energy\endcsname{0.0019}
\expandafter\def\csname FF:x5_guard/aether-prior/energy\endcsname{0.00185}
\expandafter\def\csname N:x5_guard/aether-prior/carbon\endcsname{0.00154}
\expandafter\def\csname F:x5_guard/aether-prior/carbon\endcsname{0.0015}
\expandafter\def\csname FF:x5_guard/aether-prior/carbon\endcsname{0.00154}
\expandafter\def\csname N:x5_guard/aether-prior/by_station/denver\endcsname{0.00525}
\expandafter\def\csname F:x5_guard/aether-prior/by_station/denver\endcsname{0.0053}
\expandafter\def\csname FF:x5_guard/aether-prior/by_station/denver\endcsname{0.00525}
\expandafter\def\csname N:x5_guard/aether-prior/worst\endcsname{building_2400/denver_4}
\expandafter\def\csname N:x5_guard/aether-passive/n\endcsname{1}
\expandafter\def\csname W:x5_guard/aether-passive/n\endcsname{one}
\expandafter\def\csname N:x5_guard/aether-passive/score\endcsname{0.00693}
\expandafter\def\csname F:x5_guard/aether-passive/score\endcsname{0.0069}
\expandafter\def\csname FF:x5_guard/aether-passive/score\endcsname{0.00693}
\expandafter\def\csname N:x5_guard/aether-passive/comfort\endcsname{0.00411}
\expandafter\def\csname F:x5_guard/aether-passive/comfort\endcsname{0.0041}
\expandafter\def\csname FF:x5_guard/aether-passive/comfort\endcsname{0.00411}
\expandafter\def\csname N:x5_guard/aether-passive/energy\endcsname{0.0016}
\expandafter\def\csname F:x5_guard/aether-passive/energy\endcsname{0.0016}
\expandafter\def\csname FF:x5_guard/aether-passive/energy\endcsname{0.00160}
\expandafter\def\csname N:x5_guard/aether-passive/carbon\endcsname{0.00121}
\expandafter\def\csname F:x5_guard/aether-passive/carbon\endcsname{0.0012}
\expandafter\def\csname FF:x5_guard/aether-passive/carbon\endcsname{0.00121}
\expandafter\def\csname N:x5_guard/aether-passive/by_station/denver\endcsname{0.00693}
\expandafter\def\csname F:x5_guard/aether-passive/by_station/denver\endcsname{0.0069}
\expandafter\def\csname FF:x5_guard/aether-passive/by_station/denver\endcsname{0.00693}
\expandafter\def\csname N:x5_guard/aether-passive/worst\endcsname{building_2400/denver_4}
\expandafter\def\csname N:x5_guard/aether-active/n\endcsname{1}
\expandafter\def\csname W:x5_guard/aether-active/n\endcsname{one}
\expandafter\def\csname N:x5_guard/aether-active/score\endcsname{0.00704}
\expandafter\def\csname F:x5_guard/aether-active/score\endcsname{0.0070}
\expandafter\def\csname FF:x5_guard/aether-active/score\endcsname{0.00704}
\expandafter\def\csname N:x5_guard/aether-active/comfort\endcsname{0.00356}
\expandafter\def\csname F:x5_guard/aether-active/comfort\endcsname{0.0036}
\expandafter\def\csname FF:x5_guard/aether-active/comfort\endcsname{0.00356}
\expandafter\def\csname N:x5_guard/aether-active/energy\endcsname{0.00183}
\expandafter\def\csname F:x5_guard/aether-active/energy\endcsname{0.0018}
\expandafter\def\csname FF:x5_guard/aether-active/energy\endcsname{0.00183}
\expandafter\def\csname N:x5_guard/aether-active/carbon\endcsname{0.00164}
\expandafter\def\csname F:x5_guard/aether-active/carbon\endcsname{0.0016}
\expandafter\def\csname FF:x5_guard/aether-active/carbon\endcsname{0.00164}
\expandafter\def\csname N:x5_guard/aether-active/by_station/denver\endcsname{0.00704}
\expandafter\def\csname F:x5_guard/aether-active/by_station/denver\endcsname{0.0070}
\expandafter\def\csname FF:x5_guard/aether-active/by_station/denver\endcsname{0.00704}
\expandafter\def\csname N:x5_guard/aether-active/worst\endcsname{building_2400/denver_4}
\expandafter\def\csname N:x5_guard/aether-passive_guard/n\endcsname{1}
\expandafter\def\csname W:x5_guard/aether-passive_guard/n\endcsname{one}
\expandafter\def\csname N:x5_guard/aether-passive_guard/score\endcsname{0.28803}
\expandafter\def\csname F:x5_guard/aether-passive_guard/score\endcsname{0.2880}
\expandafter\def\csname FF:x5_guard/aether-passive_guard/score\endcsname{0.28803}
\expandafter\def\csname N:x5_guard/aether-passive_guard/comfort\endcsname{0.22863}
\expandafter\def\csname F:x5_guard/aether-passive_guard/comfort\endcsname{0.2286}
\expandafter\def\csname FF:x5_guard/aether-passive_guard/comfort\endcsname{0.22863}
\expandafter\def\csname N:x5_guard/aether-passive_guard/energy\endcsname{0.02562}
\expandafter\def\csname F:x5_guard/aether-passive_guard/energy\endcsname{0.0256}
\expandafter\def\csname FF:x5_guard/aether-passive_guard/energy\endcsname{0.02562}
\expandafter\def\csname N:x5_guard/aether-passive_guard/carbon\endcsname{0.03377}
\expandafter\def\csname F:x5_guard/aether-passive_guard/carbon\endcsname{0.0338}
\expandafter\def\csname FF:x5_guard/aether-passive_guard/carbon\endcsname{0.03377}
\expandafter\def\csname N:x5_guard/aether-passive_guard/by_station/denver\endcsname{0.28803}
\expandafter\def\csname F:x5_guard/aether-passive_guard/by_station/denver\endcsname{0.2880}
\expandafter\def\csname FF:x5_guard/aether-passive_guard/by_station/denver\endcsname{0.28803}
\expandafter\def\csname N:x5_guard/aether-passive_guard/worst\endcsname{building_2400/denver_4}
\expandafter\def\csname N:x5_guard/aether-full/n\endcsname{1}
\expandafter\def\csname W:x5_guard/aether-full/n\endcsname{one}
\expandafter\def\csname N:x5_guard/aether-full/score\endcsname{0.29697}
\expandafter\def\csname F:x5_guard/aether-full/score\endcsname{0.2970}
\expandafter\def\csname FF:x5_guard/aether-full/score\endcsname{0.29697}
\expandafter\def\csname N:x5_guard/aether-full/comfort\endcsname{0.23671}
\expandafter\def\csname F:x5_guard/aether-full/comfort\endcsname{0.2367}
\expandafter\def\csname FF:x5_guard/aether-full/comfort\endcsname{0.23671}
\expandafter\def\csname N:x5_guard/aether-full/energy\endcsname{0.0257}
\expandafter\def\csname F:x5_guard/aether-full/energy\endcsname{0.0257}
\expandafter\def\csname FF:x5_guard/aether-full/energy\endcsname{0.02570}
\expandafter\def\csname N:x5_guard/aether-full/carbon\endcsname{0.03455}
\expandafter\def\csname F:x5_guard/aether-full/carbon\endcsname{0.0345}
\expandafter\def\csname FF:x5_guard/aether-full/carbon\endcsname{0.03455}
\expandafter\def\csname N:x5_guard/aether-full/by_station/denver\endcsname{0.29697}
\expandafter\def\csname F:x5_guard/aether-full/by_station/denver\endcsname{0.2970}
\expandafter\def\csname FF:x5_guard/aether-full/by_station/denver\endcsname{0.29697}
\expandafter\def\csname N:x5_guard/aether-full/worst\endcsname{building_2400/denver_4}
\expandafter\def\csname N:pool_test/pid_water/n\endcsname{20}
\expandafter\def\csname N:pool_test/pid_water/score\endcsname{0.03208}
\expandafter\def\csname F:pool_test/pid_water/score\endcsname{0.0321}
\expandafter\def\csname FF:pool_test/pid_water/score\endcsname{0.03208}
\expandafter\def\csname N:pool_test/pid_water/comfort\endcsname{0.02439}
\expandafter\def\csname F:pool_test/pid_water/comfort\endcsname{0.0244}
\expandafter\def\csname FF:pool_test/pid_water/comfort\endcsname{0.02439}
\expandafter\def\csname N:pool_test/pid_water/energy\endcsname{0.00493}
\expandafter\def\csname F:pool_test/pid_water/energy\endcsname{0.0049}
\expandafter\def\csname FF:pool_test/pid_water/energy\endcsname{0.00493}
\expandafter\def\csname N:pool_test/pid_water/carbon\endcsname{0.00276}
\expandafter\def\csname F:pool_test/pid_water/carbon\endcsname{0.0028}
\expandafter\def\csname FF:pool_test/pid_water/carbon\endcsname{0.00276}
\expandafter\def\csname N:pool_test/pid_water/by_station/phoenix\endcsname{0.01999}
\expandafter\def\csname F:pool_test/pid_water/by_station/phoenix\endcsname{0.0200}
\expandafter\def\csname FF:pool_test/pid_water/by_station/phoenix\endcsname{0.01999}
\expandafter\def\csname N:pool_test/pid_water/by_station/denver\endcsname{0.04026}
\expandafter\def\csname F:pool_test/pid_water/by_station/denver\endcsname{0.0403}
\expandafter\def\csname FF:pool_test/pid_water/by_station/denver\endcsname{0.04026}
\expandafter\def\csname N:pool_test/pid_water/by_station/singapore\endcsname{0.03237}
\expandafter\def\csname F:pool_test/pid_water/by_station/singapore\endcsname{0.0324}
\expandafter\def\csname FF:pool_test/pid_water/by_station/singapore\endcsname{0.03237}
\expandafter\def\csname N:pool_test/pid_water/by_station/atlanta\endcsname{0.02475}
\expandafter\def\csname F:pool_test/pid_water/by_station/atlanta\endcsname{0.0248}
\expandafter\def\csname FF:pool_test/pid_water/by_station/atlanta\endcsname{0.02475}
\expandafter\def\csname N:pool_test/pid_water/by_station/fairbanks\endcsname{0.04302}
\expandafter\def\csname F:pool_test/pid_water/by_station/fairbanks\endcsname{0.0430}
\expandafter\def\csname FF:pool_test/pid_water/by_station/fairbanks\endcsname{0.04302}
\expandafter\def\csname N:pool_test/pid_water/worst\endcsname{building_2300/denver_1}
\expandafter\def\csname N:pool_test/aether-rules/n\endcsname{20}
\expandafter\def\csname N:pool_test/aether-rules/score\endcsname{0.01041}
\expandafter\def\csname F:pool_test/aether-rules/score\endcsname{0.0104}
\expandafter\def\csname FF:pool_test/aether-rules/score\endcsname{0.01041}
\expandafter\def\csname N:pool_test/aether-rules/comfort\endcsname{0.00217}
\expandafter\def\csname F:pool_test/aether-rules/comfort\endcsname{0.0022}
\expandafter\def\csname FF:pool_test/aether-rules/comfort\endcsname{0.00217}
\expandafter\def\csname N:pool_test/aether-rules/energy\endcsname{0.00422}
\expandafter\def\csname F:pool_test/aether-rules/energy\endcsname{0.0042}
\expandafter\def\csname FF:pool_test/aether-rules/energy\endcsname{0.00422}
\expandafter\def\csname N:pool_test/aether-rules/carbon\endcsname{0.00403}
\expandafter\def\csname F:pool_test/aether-rules/carbon\endcsname{0.0040}
\expandafter\def\csname FF:pool_test/aether-rules/carbon\endcsname{0.00403}
\expandafter\def\csname N:pool_test/aether-rules/by_station/phoenix\endcsname{0.00421}
\expandafter\def\csname F:pool_test/aether-rules/by_station/phoenix\endcsname{0.0042}
\expandafter\def\csname FF:pool_test/aether-rules/by_station/phoenix\endcsname{0.00421}
\expandafter\def\csname N:pool_test/aether-rules/by_station/denver\endcsname{0.01734}
\expandafter\def\csname F:pool_test/aether-rules/by_station/denver\endcsname{0.0173}
\expandafter\def\csname FF:pool_test/aether-rules/by_station/denver\endcsname{0.01734}
\expandafter\def\csname N:pool_test/aether-rules/by_station/singapore\endcsname{0.0079}
\expandafter\def\csname F:pool_test/aether-rules/by_station/singapore\endcsname{0.0079}
\expandafter\def\csname FF:pool_test/aether-rules/by_station/singapore\endcsname{0.00790}
\expandafter\def\csname N:pool_test/aether-rules/by_station/atlanta\endcsname{0.00623}
\expandafter\def\csname F:pool_test/aether-rules/by_station/atlanta\endcsname{0.0062}
\expandafter\def\csname FF:pool_test/aether-rules/by_station/atlanta\endcsname{0.00623}
\expandafter\def\csname N:pool_test/aether-rules/by_station/fairbanks\endcsname{0.01638}
\expandafter\def\csname F:pool_test/aether-rules/by_station/fairbanks\endcsname{0.0164}
\expandafter\def\csname FF:pool_test/aether-rules/by_station/fairbanks\endcsname{0.01638}
\expandafter\def\csname N:pool_test/aether-rules/worst\endcsname{building_2500/denver_1}
\expandafter\def\csname N:pool_test/aether-prior/n\endcsname{20}
\expandafter\def\csname N:pool_test/aether-prior/score\endcsname{0.0121}
\expandafter\def\csname F:pool_test/aether-prior/score\endcsname{0.0121}
\expandafter\def\csname FF:pool_test/aether-prior/score\endcsname{0.01210}
\expandafter\def\csname N:pool_test/aether-prior/comfort\endcsname{0.00352}
\expandafter\def\csname F:pool_test/aether-prior/comfort\endcsname{0.0035}
\expandafter\def\csname FF:pool_test/aether-prior/comfort\endcsname{0.00352}
\expandafter\def\csname N:pool_test/aether-prior/energy\endcsname{0.00456}
\expandafter\def\csname F:pool_test/aether-prior/energy\endcsname{0.0046}
\expandafter\def\csname FF:pool_test/aether-prior/energy\endcsname{0.00456}
\expandafter\def\csname N:pool_test/aether-prior/carbon\endcsname{0.00402}
\expandafter\def\csname F:pool_test/aether-prior/carbon\endcsname{0.0040}
\expandafter\def\csname FF:pool_test/aether-prior/carbon\endcsname{0.00402}
\expandafter\def\csname N:pool_test/aether-prior/by_station/phoenix\endcsname{0.00457}
\expandafter\def\csname F:pool_test/aether-prior/by_station/phoenix\endcsname{0.0046}
\expandafter\def\csname FF:pool_test/aether-prior/by_station/phoenix\endcsname{0.00457}
\expandafter\def\csname N:pool_test/aether-prior/by_station/denver\endcsname{0.02255}
\expandafter\def\csname F:pool_test/aether-prior/by_station/denver\endcsname{0.0226}
\expandafter\def\csname FF:pool_test/aether-prior/by_station/denver\endcsname{0.02255}
\expandafter\def\csname N:pool_test/aether-prior/by_station/singapore\endcsname{0.00877}
\expandafter\def\csname F:pool_test/aether-prior/by_station/singapore\endcsname{0.0088}
\expandafter\def\csname FF:pool_test/aether-prior/by_station/singapore\endcsname{0.00877}
\expandafter\def\csname N:pool_test/aether-prior/by_station/atlanta\endcsname{0.00669}
\expandafter\def\csname F:pool_test/aether-prior/by_station/atlanta\endcsname{0.0067}
\expandafter\def\csname FF:pool_test/aether-prior/by_station/atlanta\endcsname{0.00669}
\expandafter\def\csname N:pool_test/aether-prior/by_station/fairbanks\endcsname{0.01795}
\expandafter\def\csname F:pool_test/aether-prior/by_station/fairbanks\endcsname{0.0180}
\expandafter\def\csname FF:pool_test/aether-prior/by_station/fairbanks\endcsname{0.01795}
\expandafter\def\csname N:pool_test/aether-prior/worst\endcsname{building_2500/denver_1}
\expandafter\def\csname N:pool_test/aether-passive/n\endcsname{20}
\expandafter\def\csname N:pool_test/aether-passive/score\endcsname{0.01482}
\expandafter\def\csname F:pool_test/aether-passive/score\endcsname{0.0148}
\expandafter\def\csname FF:pool_test/aether-passive/score\endcsname{0.01482}
\expandafter\def\csname N:pool_test/aether-passive/comfort\endcsname{0.00769}
\expandafter\def\csname F:pool_test/aether-passive/comfort\endcsname{0.0077}
\expandafter\def\csname FF:pool_test/aether-passive/comfort\endcsname{0.00769}
\expandafter\def\csname N:pool_test/aether-passive/energy\endcsname{0.00404}
\expandafter\def\csname F:pool_test/aether-passive/energy\endcsname{0.0040}
\expandafter\def\csname FF:pool_test/aether-passive/energy\endcsname{0.00404}
\expandafter\def\csname N:pool_test/aether-passive/carbon\endcsname{0.0031}
\expandafter\def\csname F:pool_test/aether-passive/carbon\endcsname{0.0031}
\expandafter\def\csname FF:pool_test/aether-passive/carbon\endcsname{0.00310}
\expandafter\def\csname N:pool_test/aether-passive/by_station/phoenix\endcsname{0.0047}
\expandafter\def\csname F:pool_test/aether-passive/by_station/phoenix\endcsname{0.0047}
\expandafter\def\csname FF:pool_test/aether-passive/by_station/phoenix\endcsname{0.00470}
\expandafter\def\csname N:pool_test/aether-passive/by_station/denver\endcsname{0.02797}
\expandafter\def\csname F:pool_test/aether-passive/by_station/denver\endcsname{0.0280}
\expandafter\def\csname FF:pool_test/aether-passive/by_station/denver\endcsname{0.02797}
\expandafter\def\csname N:pool_test/aether-passive/by_station/singapore\endcsname{0.0086}
\expandafter\def\csname F:pool_test/aether-passive/by_station/singapore\endcsname{0.0086}
\expandafter\def\csname FF:pool_test/aether-passive/by_station/singapore\endcsname{0.00860}
\expandafter\def\csname N:pool_test/aether-passive/by_station/atlanta\endcsname{0.00787}
\expandafter\def\csname F:pool_test/aether-passive/by_station/atlanta\endcsname{0.0079}
\expandafter\def\csname FF:pool_test/aether-passive/by_station/atlanta\endcsname{0.00787}
\expandafter\def\csname N:pool_test/aether-passive/by_station/fairbanks\endcsname{0.02498}
\expandafter\def\csname F:pool_test/aether-passive/by_station/fairbanks\endcsname{0.0250}
\expandafter\def\csname FF:pool_test/aether-passive/by_station/fairbanks\endcsname{0.02498}
\expandafter\def\csname N:pool_test/aether-passive/worst\endcsname{building_2500/denver_1}
\expandafter\def\csname N:pool_test/aether-mpc/n\endcsname{20}
\expandafter\def\csname N:pool_test/aether-mpc/score\endcsname{0.0091}
\expandafter\def\csname F:pool_test/aether-mpc/score\endcsname{0.0091}
\expandafter\def\csname FF:pool_test/aether-mpc/score\endcsname{0.00910}
\expandafter\def\csname N:pool_test/aether-mpc/comfort\endcsname{0.00279}
\expandafter\def\csname F:pool_test/aether-mpc/comfort\endcsname{0.0028}
\expandafter\def\csname FF:pool_test/aether-mpc/comfort\endcsname{0.00279}
\expandafter\def\csname N:pool_test/aether-mpc/energy\endcsname{0.00374}
\expandafter\def\csname F:pool_test/aether-mpc/energy\endcsname{0.0037}
\expandafter\def\csname FF:pool_test/aether-mpc/energy\endcsname{0.00374}
\expandafter\def\csname N:pool_test/aether-mpc/carbon\endcsname{0.00257}
\expandafter\def\csname F:pool_test/aether-mpc/carbon\endcsname{0.0026}
\expandafter\def\csname FF:pool_test/aether-mpc/carbon\endcsname{0.00257}
\expandafter\def\csname N:pool_test/aether-mpc/by_station/phoenix\endcsname{0.00444}
\expandafter\def\csname F:pool_test/aether-mpc/by_station/phoenix\endcsname{0.0044}
\expandafter\def\csname FF:pool_test/aether-mpc/by_station/phoenix\endcsname{0.00444}
\expandafter\def\csname N:pool_test/aether-mpc/by_station/denver\endcsname{0.01455}
\expandafter\def\csname F:pool_test/aether-mpc/by_station/denver\endcsname{0.0146}
\expandafter\def\csname FF:pool_test/aether-mpc/by_station/denver\endcsname{0.01455}
\expandafter\def\csname N:pool_test/aether-mpc/by_station/singapore\endcsname{0.00888}
\expandafter\def\csname F:pool_test/aether-mpc/by_station/singapore\endcsname{0.0089}
\expandafter\def\csname FF:pool_test/aether-mpc/by_station/singapore\endcsname{0.00888}
\expandafter\def\csname N:pool_test/aether-mpc/by_station/atlanta\endcsname{0.00635}
\expandafter\def\csname F:pool_test/aether-mpc/by_station/atlanta\endcsname{0.0063}
\expandafter\def\csname FF:pool_test/aether-mpc/by_station/atlanta\endcsname{0.00635}
\expandafter\def\csname N:pool_test/aether-mpc/by_station/fairbanks\endcsname{0.01129}
\expandafter\def\csname F:pool_test/aether-mpc/by_station/fairbanks\endcsname{0.0113}
\expandafter\def\csname FF:pool_test/aether-mpc/by_station/fairbanks\endcsname{0.01129}
\expandafter\def\csname N:pool_test/aether-mpc/worst\endcsname{building_2500/denver_1}
\expandafter\def\csname N:pool_test/aether-mpc_off/n\endcsname{20}
\expandafter\def\csname N:pool_test/aether-mpc_off/score\endcsname{0.00818}
\expandafter\def\csname F:pool_test/aether-mpc_off/score\endcsname{0.0082}
\expandafter\def\csname FF:pool_test/aether-mpc_off/score\endcsname{0.00818}
\expandafter\def\csname N:pool_test/aether-mpc_off/comfort\endcsname{0.00173}
\expandafter\def\csname F:pool_test/aether-mpc_off/comfort\endcsname{0.0017}
\expandafter\def\csname FF:pool_test/aether-mpc_off/comfort\endcsname{0.00173}
\expandafter\def\csname N:pool_test/aether-mpc_off/energy\endcsname{0.00379}
\expandafter\def\csname F:pool_test/aether-mpc_off/energy\endcsname{0.0038}
\expandafter\def\csname FF:pool_test/aether-mpc_off/energy\endcsname{0.00379}
\expandafter\def\csname N:pool_test/aether-mpc_off/carbon\endcsname{0.00265}
\expandafter\def\csname F:pool_test/aether-mpc_off/carbon\endcsname{0.0027}
\expandafter\def\csname FF:pool_test/aether-mpc_off/carbon\endcsname{0.00265}
\expandafter\def\csname N:pool_test/aether-mpc_off/by_station/phoenix\endcsname{0.00414}
\expandafter\def\csname F:pool_test/aether-mpc_off/by_station/phoenix\endcsname{0.0041}
\expandafter\def\csname FF:pool_test/aether-mpc_off/by_station/phoenix\endcsname{0.00414}
\expandafter\def\csname N:pool_test/aether-mpc_off/by_station/denver\endcsname{0.0129}
\expandafter\def\csname F:pool_test/aether-mpc_off/by_station/denver\endcsname{0.0129}
\expandafter\def\csname FF:pool_test/aether-mpc_off/by_station/denver\endcsname{0.01290}
\expandafter\def\csname N:pool_test/aether-mpc_off/by_station/singapore\endcsname{0.00823}
\expandafter\def\csname F:pool_test/aether-mpc_off/by_station/singapore\endcsname{0.0082}
\expandafter\def\csname FF:pool_test/aether-mpc_off/by_station/singapore\endcsname{0.00823}
\expandafter\def\csname N:pool_test/aether-mpc_off/by_station/atlanta\endcsname{0.00584}
\expandafter\def\csname F:pool_test/aether-mpc_off/by_station/atlanta\endcsname{0.0058}
\expandafter\def\csname FF:pool_test/aether-mpc_off/by_station/atlanta\endcsname{0.00584}
\expandafter\def\csname N:pool_test/aether-mpc_off/by_station/fairbanks\endcsname{0.00979}
\expandafter\def\csname F:pool_test/aether-mpc_off/by_station/fairbanks\endcsname{0.0098}
\expandafter\def\csname FF:pool_test/aether-mpc_off/by_station/fairbanks\endcsname{0.00979}
\expandafter\def\csname N:pool_test/aether-mpc_off/worst\endcsname{building_2500/denver_1}
\expandafter\def\csname N:pool_test/aether-mpc_guard/n\endcsname{20}
\expandafter\def\csname N:pool_test/aether-mpc_guard/score\endcsname{0.02504}
\expandafter\def\csname F:pool_test/aether-mpc_guard/score\endcsname{0.0250}
\expandafter\def\csname FF:pool_test/aether-mpc_guard/score\endcsname{0.02504}
\expandafter\def\csname N:pool_test/aether-mpc_guard/comfort\endcsname{0.01166}
\expandafter\def\csname F:pool_test/aether-mpc_guard/comfort\endcsname{0.0117}
\expandafter\def\csname FF:pool_test/aether-mpc_guard/comfort\endcsname{0.01166}
\expandafter\def\csname N:pool_test/aether-mpc_guard/energy\endcsname{0.00744}
\expandafter\def\csname F:pool_test/aether-mpc_guard/energy\endcsname{0.0074}
\expandafter\def\csname FF:pool_test/aether-mpc_guard/energy\endcsname{0.00744}
\expandafter\def\csname N:pool_test/aether-mpc_guard/carbon\endcsname{0.00595}
\expandafter\def\csname F:pool_test/aether-mpc_guard/carbon\endcsname{0.0060}
\expandafter\def\csname FF:pool_test/aether-mpc_guard/carbon\endcsname{0.00595}
\expandafter\def\csname N:pool_test/aether-mpc_guard/by_station/phoenix\endcsname{0.01961}
\expandafter\def\csname F:pool_test/aether-mpc_guard/by_station/phoenix\endcsname{0.0196}
\expandafter\def\csname FF:pool_test/aether-mpc_guard/by_station/phoenix\endcsname{0.01961}
\expandafter\def\csname N:pool_test/aether-mpc_guard/by_station/denver\endcsname{0.02533}
\expandafter\def\csname F:pool_test/aether-mpc_guard/by_station/denver\endcsname{0.0253}
\expandafter\def\csname FF:pool_test/aether-mpc_guard/by_station/denver\endcsname{0.02533}
\expandafter\def\csname N:pool_test/aether-mpc_guard/by_station/singapore\endcsname{0.02961}
\expandafter\def\csname F:pool_test/aether-mpc_guard/by_station/singapore\endcsname{0.0296}
\expandafter\def\csname FF:pool_test/aether-mpc_guard/by_station/singapore\endcsname{0.02961}
\expandafter\def\csname N:pool_test/aether-mpc_guard/by_station/atlanta\endcsname{0.02448}
\expandafter\def\csname F:pool_test/aether-mpc_guard/by_station/atlanta\endcsname{0.0245}
\expandafter\def\csname FF:pool_test/aether-mpc_guard/by_station/atlanta\endcsname{0.02448}
\expandafter\def\csname N:pool_test/aether-mpc_guard/by_station/fairbanks\endcsname{0.02618}
\expandafter\def\csname F:pool_test/aether-mpc_guard/by_station/fairbanks\endcsname{0.0262}
\expandafter\def\csname FF:pool_test/aether-mpc_guard/by_station/fairbanks\endcsname{0.02618}
\expandafter\def\csname N:pool_test/aether-mpc_guard/worst\endcsname{building_1401/fairbanks_3}
\expandafter\def\csname N:pool_test/aether-full/n\endcsname{20}
\expandafter\def\csname N:pool_test/aether-full/score\endcsname{0.23292}
\expandafter\def\csname F:pool_test/aether-full/score\endcsname{0.2329}
\expandafter\def\csname FF:pool_test/aether-full/score\endcsname{0.23292}
\expandafter\def\csname N:pool_test/aether-full/comfort\endcsname{0.17577}
\expandafter\def\csname F:pool_test/aether-full/comfort\endcsname{0.1758}
\expandafter\def\csname FF:pool_test/aether-full/comfort\endcsname{0.17577}
\expandafter\def\csname N:pool_test/aether-full/energy\endcsname{0.02648}
\expandafter\def\csname F:pool_test/aether-full/energy\endcsname{0.0265}
\expandafter\def\csname FF:pool_test/aether-full/energy\endcsname{0.02648}
\expandafter\def\csname N:pool_test/aether-full/carbon\endcsname{0.03067}
\expandafter\def\csname F:pool_test/aether-full/carbon\endcsname{0.0307}
\expandafter\def\csname FF:pool_test/aether-full/carbon\endcsname{0.03067}
\expandafter\def\csname N:pool_test/aether-full/by_station/phoenix\endcsname{0.25304}
\expandafter\def\csname F:pool_test/aether-full/by_station/phoenix\endcsname{0.2530}
\expandafter\def\csname FF:pool_test/aether-full/by_station/phoenix\endcsname{0.25304}
\expandafter\def\csname N:pool_test/aether-full/by_station/denver\endcsname{0.22678}
\expandafter\def\csname F:pool_test/aether-full/by_station/denver\endcsname{0.2268}
\expandafter\def\csname FF:pool_test/aether-full/by_station/denver\endcsname{0.22678}
\expandafter\def\csname N:pool_test/aether-full/by_station/singapore\endcsname{0.23629}
\expandafter\def\csname F:pool_test/aether-full/by_station/singapore\endcsname{0.2363}
\expandafter\def\csname FF:pool_test/aether-full/by_station/singapore\endcsname{0.23629}
\expandafter\def\csname N:pool_test/aether-full/by_station/atlanta\endcsname{0.23326}
\expandafter\def\csname F:pool_test/aether-full/by_station/atlanta\endcsname{0.2333}
\expandafter\def\csname FF:pool_test/aether-full/by_station/atlanta\endcsname{0.23326}
\expandafter\def\csname N:pool_test/aether-full/by_station/fairbanks\endcsname{0.21525}
\expandafter\def\csname F:pool_test/aether-full/by_station/fairbanks\endcsname{0.2152}
\expandafter\def\csname FF:pool_test/aether-full/by_station/fairbanks\endcsname{0.21525}
\expandafter\def\csname N:pool_test/aether-full/worst\endcsname{building_2400/denver_4}
\expandafter\def\csname N:pool_test/aether-mpc_off+oracle/n\endcsname{20}
\expandafter\def\csname N:pool_test/aether-mpc_off+oracle/score\endcsname{0.0082}
\expandafter\def\csname F:pool_test/aether-mpc_off+oracle/score\endcsname{0.0082}
\expandafter\def\csname FF:pool_test/aether-mpc_off+oracle/score\endcsname{0.00820}
\expandafter\def\csname N:pool_test/aether-mpc_off+oracle/comfort\endcsname{0.00175}
\expandafter\def\csname F:pool_test/aether-mpc_off+oracle/comfort\endcsname{0.0018}
\expandafter\def\csname FF:pool_test/aether-mpc_off+oracle/comfort\endcsname{0.00175}
\expandafter\def\csname N:pool_test/aether-mpc_off+oracle/energy\endcsname{0.00379}
\expandafter\def\csname F:pool_test/aether-mpc_off+oracle/energy\endcsname{0.0038}
\expandafter\def\csname FF:pool_test/aether-mpc_off+oracle/energy\endcsname{0.00379}
\expandafter\def\csname N:pool_test/aether-mpc_off+oracle/carbon\endcsname{0.00266}
\expandafter\def\csname F:pool_test/aether-mpc_off+oracle/carbon\endcsname{0.0027}
\expandafter\def\csname FF:pool_test/aether-mpc_off+oracle/carbon\endcsname{0.00266}
\expandafter\def\csname N:pool_test/aether-mpc_off+oracle/by_station/phoenix\endcsname{0.00414}
\expandafter\def\csname F:pool_test/aether-mpc_off+oracle/by_station/phoenix\endcsname{0.0041}
\expandafter\def\csname FF:pool_test/aether-mpc_off+oracle/by_station/phoenix\endcsname{0.00414}
\expandafter\def\csname N:pool_test/aether-mpc_off+oracle/by_station/denver\endcsname{0.01296}
\expandafter\def\csname F:pool_test/aether-mpc_off+oracle/by_station/denver\endcsname{0.0130}
\expandafter\def\csname FF:pool_test/aether-mpc_off+oracle/by_station/denver\endcsname{0.01296}
\expandafter\def\csname N:pool_test/aether-mpc_off+oracle/by_station/singapore\endcsname{0.00822}
\expandafter\def\csname F:pool_test/aether-mpc_off+oracle/by_station/singapore\endcsname{0.0082}
\expandafter\def\csname FF:pool_test/aether-mpc_off+oracle/by_station/singapore\endcsname{0.00822}
\expandafter\def\csname N:pool_test/aether-mpc_off+oracle/by_station/atlanta\endcsname{0.00584}
\expandafter\def\csname F:pool_test/aether-mpc_off+oracle/by_station/atlanta\endcsname{0.0058}
\expandafter\def\csname FF:pool_test/aether-mpc_off+oracle/by_station/atlanta\endcsname{0.00584}
\expandafter\def\csname N:pool_test/aether-mpc_off+oracle/by_station/fairbanks\endcsname{0.00981}
\expandafter\def\csname F:pool_test/aether-mpc_off+oracle/by_station/fairbanks\endcsname{0.0098}
\expandafter\def\csname FF:pool_test/aether-mpc_off+oracle/by_station/fairbanks\endcsname{0.00981}
\expandafter\def\csname N:pool_test/aether-mpc_off+oracle/worst\endcsname{building_2500/denver_1}
\expandafter\def\csname N:pool_test/pid_tuned/n\endcsname{20}
\expandafter\def\csname N:pool_test/pid_tuned/score\endcsname{0.03186}
\expandafter\def\csname F:pool_test/pid_tuned/score\endcsname{0.0319}
\expandafter\def\csname FF:pool_test/pid_tuned/score\endcsname{0.03186}
\expandafter\def\csname N:pool_test/pid_tuned/comfort\endcsname{0.0248}
\expandafter\def\csname F:pool_test/pid_tuned/comfort\endcsname{0.0248}
\expandafter\def\csname FF:pool_test/pid_tuned/comfort\endcsname{0.02480}
\expandafter\def\csname N:pool_test/pid_tuned/energy\endcsname{0.00463}
\expandafter\def\csname F:pool_test/pid_tuned/energy\endcsname{0.0046}
\expandafter\def\csname FF:pool_test/pid_tuned/energy\endcsname{0.00463}
\expandafter\def\csname N:pool_test/pid_tuned/carbon\endcsname{0.00243}
\expandafter\def\csname F:pool_test/pid_tuned/carbon\endcsname{0.0024}
\expandafter\def\csname FF:pool_test/pid_tuned/carbon\endcsname{0.00243}
\expandafter\def\csname N:pool_test/pid_tuned/by_station/phoenix\endcsname{0.02069}
\expandafter\def\csname F:pool_test/pid_tuned/by_station/phoenix\endcsname{0.0207}
\expandafter\def\csname FF:pool_test/pid_tuned/by_station/phoenix\endcsname{0.02069}
\expandafter\def\csname N:pool_test/pid_tuned/by_station/denver\endcsname{0.04021}
\expandafter\def\csname F:pool_test/pid_tuned/by_station/denver\endcsname{0.0402}
\expandafter\def\csname FF:pool_test/pid_tuned/by_station/denver\endcsname{0.04021}
\expandafter\def\csname N:pool_test/pid_tuned/by_station/singapore\endcsname{0.02937}
\expandafter\def\csname F:pool_test/pid_tuned/by_station/singapore\endcsname{0.0294}
\expandafter\def\csname FF:pool_test/pid_tuned/by_station/singapore\endcsname{0.02937}
\expandafter\def\csname N:pool_test/pid_tuned/by_station/atlanta\endcsname{0.02534}
\expandafter\def\csname F:pool_test/pid_tuned/by_station/atlanta\endcsname{0.0253}
\expandafter\def\csname FF:pool_test/pid_tuned/by_station/atlanta\endcsname{0.02534}
\expandafter\def\csname N:pool_test/pid_tuned/by_station/fairbanks\endcsname{0.04367}
\expandafter\def\csname F:pool_test/pid_tuned/by_station/fairbanks\endcsname{0.0437}
\expandafter\def\csname FF:pool_test/pid_tuned/by_station/fairbanks\endcsname{0.04367}
\expandafter\def\csname N:pool_test/pid_tuned/worst\endcsname{building_2300/denver_1}
\expandafter\def\csname N:shipped/pid_water/n\endcsname{30}
\expandafter\def\csname N:shipped/pid_water/score\endcsname{0.05496}
\expandafter\def\csname F:shipped/pid_water/score\endcsname{0.0550}
\expandafter\def\csname FF:shipped/pid_water/score\endcsname{0.05496}
\expandafter\def\csname N:shipped/pid_water/comfort\endcsname{0.03489}
\expandafter\def\csname F:shipped/pid_water/comfort\endcsname{0.0349}
\expandafter\def\csname FF:shipped/pid_water/comfort\endcsname{0.03489}
\expandafter\def\csname N:shipped/pid_water/energy\endcsname{0.0115}
\expandafter\def\csname F:shipped/pid_water/energy\endcsname{0.0115}
\expandafter\def\csname FF:shipped/pid_water/energy\endcsname{0.01150}
\expandafter\def\csname N:shipped/pid_water/carbon\endcsname{0.00857}
\expandafter\def\csname F:shipped/pid_water/carbon\endcsname{0.0086}
\expandafter\def\csname FF:shipped/pid_water/carbon\endcsname{0.00857}
\expandafter\def\csname N:shipped/pid_water/by_station/phoenix\endcsname{0.08001}
\expandafter\def\csname F:shipped/pid_water/by_station/phoenix\endcsname{0.0800}
\expandafter\def\csname FF:shipped/pid_water/by_station/phoenix\endcsname{0.08001}
\expandafter\def\csname N:shipped/pid_water/by_station/fairbanks\endcsname{0.07906}
\expandafter\def\csname F:shipped/pid_water/by_station/fairbanks\endcsname{0.0791}
\expandafter\def\csname FF:shipped/pid_water/by_station/fairbanks\endcsname{0.07906}
\expandafter\def\csname N:shipped/pid_water/by_station/denver\endcsname{0.03794}
\expandafter\def\csname F:shipped/pid_water/by_station/denver\endcsname{0.0379}
\expandafter\def\csname FF:shipped/pid_water/by_station/denver\endcsname{0.03794}
\expandafter\def\csname N:shipped/pid_water/by_station/singapore\endcsname{0.04714}
\expandafter\def\csname F:shipped/pid_water/by_station/singapore\endcsname{0.0471}
\expandafter\def\csname FF:shipped/pid_water/by_station/singapore\endcsname{0.04714}
\expandafter\def\csname N:shipped/pid_water/by_station/atlanta\endcsname{0.03064}
\expandafter\def\csname F:shipped/pid_water/by_station/atlanta\endcsname{0.0306}
\expandafter\def\csname FF:shipped/pid_water/by_station/atlanta\endcsname{0.03064}
\expandafter\def\csname N:shipped/pid_water/worst\endcsname{building_train_1/fairbanks_1}
\expandafter\def\csname N:shipped/aether-rules/n\endcsname{30}
\expandafter\def\csname N:shipped/aether-rules/score\endcsname{0.01712}
\expandafter\def\csname F:shipped/aether-rules/score\endcsname{0.0171}
\expandafter\def\csname FF:shipped/aether-rules/score\endcsname{0.01712}
\expandafter\def\csname N:shipped/aether-rules/comfort\endcsname{0.00447}
\expandafter\def\csname F:shipped/aether-rules/comfort\endcsname{0.0045}
\expandafter\def\csname FF:shipped/aether-rules/comfort\endcsname{0.00447}
\expandafter\def\csname N:shipped/aether-rules/energy\endcsname{0.00591}
\expandafter\def\csname F:shipped/aether-rules/energy\endcsname{0.0059}
\expandafter\def\csname FF:shipped/aether-rules/energy\endcsname{0.00591}
\expandafter\def\csname N:shipped/aether-rules/carbon\endcsname{0.00674}
\expandafter\def\csname F:shipped/aether-rules/carbon\endcsname{0.0067}
\expandafter\def\csname FF:shipped/aether-rules/carbon\endcsname{0.00674}
\expandafter\def\csname N:shipped/aether-rules/by_station/phoenix\endcsname{0.01302}
\expandafter\def\csname F:shipped/aether-rules/by_station/phoenix\endcsname{0.0130}
\expandafter\def\csname FF:shipped/aether-rules/by_station/phoenix\endcsname{0.01302}
\expandafter\def\csname N:shipped/aether-rules/by_station/fairbanks\endcsname{0.0414}
\expandafter\def\csname F:shipped/aether-rules/by_station/fairbanks\endcsname{0.0414}
\expandafter\def\csname FF:shipped/aether-rules/by_station/fairbanks\endcsname{0.04140}
\expandafter\def\csname N:shipped/aether-rules/by_station/denver\endcsname{0.01231}
\expandafter\def\csname F:shipped/aether-rules/by_station/denver\endcsname{0.0123}
\expandafter\def\csname FF:shipped/aether-rules/by_station/denver\endcsname{0.01231}
\expandafter\def\csname N:shipped/aether-rules/by_station/singapore\endcsname{0.00763}
\expandafter\def\csname F:shipped/aether-rules/by_station/singapore\endcsname{0.0076}
\expandafter\def\csname FF:shipped/aether-rules/by_station/singapore\endcsname{0.00763}
\expandafter\def\csname N:shipped/aether-rules/by_station/atlanta\endcsname{0.01123}
\expandafter\def\csname F:shipped/aether-rules/by_station/atlanta\endcsname{0.0112}
\expandafter\def\csname FF:shipped/aether-rules/by_station/atlanta\endcsname{0.01123}
\expandafter\def\csname N:shipped/aether-rules/worst\endcsname{building_train_1/fairbanks_1}
\expandafter\def\csname N:shipped/aether-prior/n\endcsname{30}
\expandafter\def\csname N:shipped/aether-prior/score\endcsname{0.02019}
\expandafter\def\csname F:shipped/aether-prior/score\endcsname{0.0202}
\expandafter\def\csname FF:shipped/aether-prior/score\endcsname{0.02019}
\expandafter\def\csname N:shipped/aether-prior/comfort\endcsname{0.00643}
\expandafter\def\csname F:shipped/aether-prior/comfort\endcsname{0.0064}
\expandafter\def\csname FF:shipped/aether-prior/comfort\endcsname{0.00643}
\expandafter\def\csname N:shipped/aether-prior/energy\endcsname{0.00673}
\expandafter\def\csname F:shipped/aether-prior/energy\endcsname{0.0067}
\expandafter\def\csname FF:shipped/aether-prior/energy\endcsname{0.00673}
\expandafter\def\csname N:shipped/aether-prior/carbon\endcsname{0.00703}
\expandafter\def\csname F:shipped/aether-prior/carbon\endcsname{0.0070}
\expandafter\def\csname FF:shipped/aether-prior/carbon\endcsname{0.00703}
\expandafter\def\csname N:shipped/aether-prior/by_station/phoenix\endcsname{0.01541}
\expandafter\def\csname F:shipped/aether-prior/by_station/phoenix\endcsname{0.0154}
\expandafter\def\csname FF:shipped/aether-prior/by_station/phoenix\endcsname{0.01541}
\expandafter\def\csname N:shipped/aether-prior/by_station/fairbanks\endcsname{0.04979}
\expandafter\def\csname F:shipped/aether-prior/by_station/fairbanks\endcsname{0.0498}
\expandafter\def\csname FF:shipped/aether-prior/by_station/fairbanks\endcsname{0.04979}
\expandafter\def\csname N:shipped/aether-prior/by_station/denver\endcsname{0.01379}
\expandafter\def\csname F:shipped/aether-prior/by_station/denver\endcsname{0.0138}
\expandafter\def\csname FF:shipped/aether-prior/by_station/denver\endcsname{0.01379}
\expandafter\def\csname N:shipped/aether-prior/by_station/singapore\endcsname{0.00926}
\expandafter\def\csname F:shipped/aether-prior/by_station/singapore\endcsname{0.0093}
\expandafter\def\csname FF:shipped/aether-prior/by_station/singapore\endcsname{0.00926}
\expandafter\def\csname N:shipped/aether-prior/by_station/atlanta\endcsname{0.01267}
\expandafter\def\csname F:shipped/aether-prior/by_station/atlanta\endcsname{0.0127}
\expandafter\def\csname FF:shipped/aether-prior/by_station/atlanta\endcsname{0.01267}
\expandafter\def\csname N:shipped/aether-prior/worst\endcsname{building_train_1/fairbanks_1}
\expandafter\def\csname N:shipped/aether-passive/n\endcsname{30}
\expandafter\def\csname N:shipped/aether-passive/score\endcsname{0.02435}
\expandafter\def\csname F:shipped/aether-passive/score\endcsname{0.0244}
\expandafter\def\csname FF:shipped/aether-passive/score\endcsname{0.02435}
\expandafter\def\csname N:shipped/aether-passive/comfort\endcsname{0.01354}
\expandafter\def\csname F:shipped/aether-passive/comfort\endcsname{0.0135}
\expandafter\def\csname FF:shipped/aether-passive/comfort\endcsname{0.01354}
\expandafter\def\csname N:shipped/aether-passive/energy\endcsname{0.00564}
\expandafter\def\csname F:shipped/aether-passive/energy\endcsname{0.0056}
\expandafter\def\csname FF:shipped/aether-passive/energy\endcsname{0.00564}
\expandafter\def\csname N:shipped/aether-passive/carbon\endcsname{0.00517}
\expandafter\def\csname F:shipped/aether-passive/carbon\endcsname{0.0052}
\expandafter\def\csname FF:shipped/aether-passive/carbon\endcsname{0.00517}
\expandafter\def\csname N:shipped/aether-passive/by_station/phoenix\endcsname{0.01617}
\expandafter\def\csname F:shipped/aether-passive/by_station/phoenix\endcsname{0.0162}
\expandafter\def\csname FF:shipped/aether-passive/by_station/phoenix\endcsname{0.01617}
\expandafter\def\csname N:shipped/aether-passive/by_station/fairbanks\endcsname{0.06085}
\expandafter\def\csname F:shipped/aether-passive/by_station/fairbanks\endcsname{0.0609}
\expandafter\def\csname FF:shipped/aether-passive/by_station/fairbanks\endcsname{0.06085}
\expandafter\def\csname N:shipped/aether-passive/by_station/denver\endcsname{0.01932}
\expandafter\def\csname F:shipped/aether-passive/by_station/denver\endcsname{0.0193}
\expandafter\def\csname FF:shipped/aether-passive/by_station/denver\endcsname{0.01932}
\expandafter\def\csname N:shipped/aether-passive/by_station/singapore\endcsname{0.00894}
\expandafter\def\csname F:shipped/aether-passive/by_station/singapore\endcsname{0.0089}
\expandafter\def\csname FF:shipped/aether-passive/by_station/singapore\endcsname{0.00894}
\expandafter\def\csname N:shipped/aether-passive/by_station/atlanta\endcsname{0.01649}
\expandafter\def\csname F:shipped/aether-passive/by_station/atlanta\endcsname{0.0165}
\expandafter\def\csname FF:shipped/aether-passive/by_station/atlanta\endcsname{0.01649}
\expandafter\def\csname N:shipped/aether-passive/worst\endcsname{building_train_1/fairbanks_1}
\expandafter\def\csname N:shipped/aether-mpc/n\endcsname{30}
\expandafter\def\csname N:shipped/aether-mpc/score\endcsname{0.01528}
\expandafter\def\csname F:shipped/aether-mpc/score\endcsname{0.0153}
\expandafter\def\csname FF:shipped/aether-mpc/score\endcsname{0.01528}
\expandafter\def\csname N:shipped/aether-mpc/comfort\endcsname{0.00576}
\expandafter\def\csname F:shipped/aether-mpc/comfort\endcsname{0.0058}
\expandafter\def\csname FF:shipped/aether-mpc/comfort\endcsname{0.00576}
\expandafter\def\csname N:shipped/aether-mpc/energy\endcsname{0.00497}
\expandafter\def\csname F:shipped/aether-mpc/energy\endcsname{0.0050}
\expandafter\def\csname FF:shipped/aether-mpc/energy\endcsname{0.00497}
\expandafter\def\csname N:shipped/aether-mpc/carbon\endcsname{0.00455}
\expandafter\def\csname F:shipped/aether-mpc/carbon\endcsname{0.0046}
\expandafter\def\csname FF:shipped/aether-mpc/carbon\endcsname{0.00455}
\expandafter\def\csname N:shipped/aether-mpc/by_station/phoenix\endcsname{0.01304}
\expandafter\def\csname F:shipped/aether-mpc/by_station/phoenix\endcsname{0.0130}
\expandafter\def\csname FF:shipped/aether-mpc/by_station/phoenix\endcsname{0.01304}
\expandafter\def\csname N:shipped/aether-mpc/by_station/fairbanks\endcsname{0.03444}
\expandafter\def\csname F:shipped/aether-mpc/by_station/fairbanks\endcsname{0.0344}
\expandafter\def\csname FF:shipped/aether-mpc/by_station/fairbanks\endcsname{0.03444}
\expandafter\def\csname N:shipped/aether-mpc/by_station/denver\endcsname{0.01032}
\expandafter\def\csname F:shipped/aether-mpc/by_station/denver\endcsname{0.0103}
\expandafter\def\csname FF:shipped/aether-mpc/by_station/denver\endcsname{0.01032}
\expandafter\def\csname N:shipped/aether-mpc/by_station/singapore\endcsname{0.00872}
\expandafter\def\csname F:shipped/aether-mpc/by_station/singapore\endcsname{0.0087}
\expandafter\def\csname FF:shipped/aether-mpc/by_station/singapore\endcsname{0.00872}
\expandafter\def\csname N:shipped/aether-mpc/by_station/atlanta\endcsname{0.00985}
\expandafter\def\csname F:shipped/aether-mpc/by_station/atlanta\endcsname{0.0098}
\expandafter\def\csname FF:shipped/aether-mpc/by_station/atlanta\endcsname{0.00985}
\expandafter\def\csname N:shipped/aether-mpc/worst\endcsname{building_train_1/fairbanks_1}
\expandafter\def\csname N:shipped/aether-mpc_off/n\endcsname{30}
\expandafter\def\csname N:shipped/aether-mpc_off/score\endcsname{0.01372}
\expandafter\def\csname F:shipped/aether-mpc_off/score\endcsname{0.0137}
\expandafter\def\csname FF:shipped/aether-mpc_off/score\endcsname{0.01372}
\expandafter\def\csname N:shipped/aether-mpc_off/comfort\endcsname{0.00399}
\expandafter\def\csname F:shipped/aether-mpc_off/comfort\endcsname{0.0040}
\expandafter\def\csname FF:shipped/aether-mpc_off/comfort\endcsname{0.00399}
\expandafter\def\csname N:shipped/aether-mpc_off/energy\endcsname{0.00505}
\expandafter\def\csname F:shipped/aether-mpc_off/energy\endcsname{0.0050}
\expandafter\def\csname FF:shipped/aether-mpc_off/energy\endcsname{0.00505}
\expandafter\def\csname N:shipped/aether-mpc_off/carbon\endcsname{0.00468}
\expandafter\def\csname F:shipped/aether-mpc_off/carbon\endcsname{0.0047}
\expandafter\def\csname FF:shipped/aether-mpc_off/carbon\endcsname{0.00468}
\expandafter\def\csname N:shipped/aether-mpc_off/by_station/phoenix\endcsname{0.0117}
\expandafter\def\csname F:shipped/aether-mpc_off/by_station/phoenix\endcsname{0.0117}
\expandafter\def\csname FF:shipped/aether-mpc_off/by_station/phoenix\endcsname{0.01170}
\expandafter\def\csname N:shipped/aether-mpc_off/by_station/fairbanks\endcsname{0.03152}
\expandafter\def\csname F:shipped/aether-mpc_off/by_station/fairbanks\endcsname{0.0315}
\expandafter\def\csname FF:shipped/aether-mpc_off/by_station/fairbanks\endcsname{0.03152}
\expandafter\def\csname N:shipped/aether-mpc_off/by_station/denver\endcsname{0.00887}
\expandafter\def\csname F:shipped/aether-mpc_off/by_station/denver\endcsname{0.0089}
\expandafter\def\csname FF:shipped/aether-mpc_off/by_station/denver\endcsname{0.00887}
\expandafter\def\csname N:shipped/aether-mpc_off/by_station/singapore\endcsname{0.00788}
\expandafter\def\csname F:shipped/aether-mpc_off/by_station/singapore\endcsname{0.0079}
\expandafter\def\csname FF:shipped/aether-mpc_off/by_station/singapore\endcsname{0.00788}
\expandafter\def\csname N:shipped/aether-mpc_off/by_station/atlanta\endcsname{0.00865}
\expandafter\def\csname F:shipped/aether-mpc_off/by_station/atlanta\endcsname{0.0086}
\expandafter\def\csname FF:shipped/aether-mpc_off/by_station/atlanta\endcsname{0.00865}
\expandafter\def\csname N:shipped/aether-mpc_off/worst\endcsname{building_train_1/fairbanks_1}
\expandafter\def\csname N:shipped/aether-mpc_guard/n\endcsname{30}
\expandafter\def\csname N:shipped/aether-mpc_guard/score\endcsname{0.03884}
\expandafter\def\csname F:shipped/aether-mpc_guard/score\endcsname{0.0388}
\expandafter\def\csname FF:shipped/aether-mpc_guard/score\endcsname{0.03884}
\expandafter\def\csname N:shipped/aether-mpc_guard/comfort\endcsname{0.01786}
\expandafter\def\csname F:shipped/aether-mpc_guard/comfort\endcsname{0.0179}
\expandafter\def\csname FF:shipped/aether-mpc_guard/comfort\endcsname{0.01786}
\expandafter\def\csname N:shipped/aether-mpc_guard/energy\endcsname{0.01109}
\expandafter\def\csname F:shipped/aether-mpc_guard/energy\endcsname{0.0111}
\expandafter\def\csname FF:shipped/aether-mpc_guard/energy\endcsname{0.01109}
\expandafter\def\csname N:shipped/aether-mpc_guard/carbon\endcsname{0.0099}
\expandafter\def\csname F:shipped/aether-mpc_guard/carbon\endcsname{0.0099}
\expandafter\def\csname FF:shipped/aether-mpc_guard/carbon\endcsname{0.00990}
\expandafter\def\csname N:shipped/aether-mpc_guard/by_station/phoenix\endcsname{0.05071}
\expandafter\def\csname F:shipped/aether-mpc_guard/by_station/phoenix\endcsname{0.0507}
\expandafter\def\csname FF:shipped/aether-mpc_guard/by_station/phoenix\endcsname{0.05071}
\expandafter\def\csname N:shipped/aether-mpc_guard/by_station/fairbanks\endcsname{0.04825}
\expandafter\def\csname F:shipped/aether-mpc_guard/by_station/fairbanks\endcsname{0.0483}
\expandafter\def\csname FF:shipped/aether-mpc_guard/by_station/fairbanks\endcsname{0.04825}
\expandafter\def\csname N:shipped/aether-mpc_guard/by_station/denver\endcsname{0.02612}
\expandafter\def\csname F:shipped/aether-mpc_guard/by_station/denver\endcsname{0.0261}
\expandafter\def\csname FF:shipped/aether-mpc_guard/by_station/denver\endcsname{0.02612}
\expandafter\def\csname N:shipped/aether-mpc_guard/by_station/singapore\endcsname{0.04388}
\expandafter\def\csname F:shipped/aether-mpc_guard/by_station/singapore\endcsname{0.0439}
\expandafter\def\csname FF:shipped/aether-mpc_guard/by_station/singapore\endcsname{0.04388}
\expandafter\def\csname N:shipped/aether-mpc_guard/by_station/atlanta\endcsname{0.02524}
\expandafter\def\csname F:shipped/aether-mpc_guard/by_station/atlanta\endcsname{0.0252}
\expandafter\def\csname FF:shipped/aether-mpc_guard/by_station/atlanta\endcsname{0.02524}
\expandafter\def\csname N:shipped/aether-mpc_guard/worst\endcsname{building_train_1/fairbanks_1}
\expandafter\def\csname N:shipped/aether-full/n\endcsname{30}
\expandafter\def\csname N:shipped/aether-full/score\endcsname{0.23082}
\expandafter\def\csname F:shipped/aether-full/score\endcsname{0.2308}
\expandafter\def\csname FF:shipped/aether-full/score\endcsname{0.23082}
\expandafter\def\csname N:shipped/aether-full/comfort\endcsname{0.16812}
\expandafter\def\csname F:shipped/aether-full/comfort\endcsname{0.1681}
\expandafter\def\csname FF:shipped/aether-full/comfort\endcsname{0.16812}
\expandafter\def\csname N:shipped/aether-full/energy\endcsname{0.02915}
\expandafter\def\csname F:shipped/aether-full/energy\endcsname{0.0291}
\expandafter\def\csname FF:shipped/aether-full/energy\endcsname{0.02915}
\expandafter\def\csname N:shipped/aether-full/carbon\endcsname{0.03354}
\expandafter\def\csname F:shipped/aether-full/carbon\endcsname{0.0335}
\expandafter\def\csname FF:shipped/aether-full/carbon\endcsname{0.03354}
\expandafter\def\csname N:shipped/aether-full/by_station/phoenix\endcsname{0.25696}
\expandafter\def\csname F:shipped/aether-full/by_station/phoenix\endcsname{0.2570}
\expandafter\def\csname FF:shipped/aether-full/by_station/phoenix\endcsname{0.25696}
\expandafter\def\csname N:shipped/aether-full/by_station/fairbanks\endcsname{0.19827}
\expandafter\def\csname F:shipped/aether-full/by_station/fairbanks\endcsname{0.1983}
\expandafter\def\csname FF:shipped/aether-full/by_station/fairbanks\endcsname{0.19827}
\expandafter\def\csname N:shipped/aether-full/by_station/denver\endcsname{0.22379}
\expandafter\def\csname F:shipped/aether-full/by_station/denver\endcsname{0.2238}
\expandafter\def\csname FF:shipped/aether-full/by_station/denver\endcsname{0.22379}
\expandafter\def\csname N:shipped/aether-full/by_station/singapore\endcsname{0.23922}
\expandafter\def\csname F:shipped/aether-full/by_station/singapore\endcsname{0.2392}
\expandafter\def\csname FF:shipped/aether-full/by_station/singapore\endcsname{0.23922}
\expandafter\def\csname N:shipped/aether-full/by_station/atlanta\endcsname{0.23585}
\expandafter\def\csname F:shipped/aether-full/by_station/atlanta\endcsname{0.2359}
\expandafter\def\csname FF:shipped/aether-full/by_station/atlanta\endcsname{0.23585}
\expandafter\def\csname N:shipped/aether-full/worst\endcsname{building_train_2/phoenix_6}
\expandafter\def\csname N:shipped/aether-mpc_off+oracle/n\endcsname{30}
\expandafter\def\csname N:shipped/aether-mpc_off+oracle/score\endcsname{0.01375}
\expandafter\def\csname F:shipped/aether-mpc_off+oracle/score\endcsname{0.0138}
\expandafter\def\csname FF:shipped/aether-mpc_off+oracle/score\endcsname{0.01375}
\expandafter\def\csname N:shipped/aether-mpc_off+oracle/comfort\endcsname{0.00403}
\expandafter\def\csname F:shipped/aether-mpc_off+oracle/comfort\endcsname{0.0040}
\expandafter\def\csname FF:shipped/aether-mpc_off+oracle/comfort\endcsname{0.00403}
\expandafter\def\csname N:shipped/aether-mpc_off+oracle/energy\endcsname{0.00504}
\expandafter\def\csname F:shipped/aether-mpc_off+oracle/energy\endcsname{0.0050}
\expandafter\def\csname FF:shipped/aether-mpc_off+oracle/energy\endcsname{0.00504}
\expandafter\def\csname N:shipped/aether-mpc_off+oracle/carbon\endcsname{0.00467}
\expandafter\def\csname F:shipped/aether-mpc_off+oracle/carbon\endcsname{0.0047}
\expandafter\def\csname FF:shipped/aether-mpc_off+oracle/carbon\endcsname{0.00467}
\expandafter\def\csname N:shipped/aether-mpc_off+oracle/by_station/phoenix\endcsname{0.01174}
\expandafter\def\csname F:shipped/aether-mpc_off+oracle/by_station/phoenix\endcsname{0.0117}
\expandafter\def\csname FF:shipped/aether-mpc_off+oracle/by_station/phoenix\endcsname{0.01174}
\expandafter\def\csname N:shipped/aether-mpc_off+oracle/by_station/fairbanks\endcsname{0.03154}
\expandafter\def\csname F:shipped/aether-mpc_off+oracle/by_station/fairbanks\endcsname{0.0315}
\expandafter\def\csname FF:shipped/aether-mpc_off+oracle/by_station/fairbanks\endcsname{0.03154}
\expandafter\def\csname N:shipped/aether-mpc_off+oracle/by_station/denver\endcsname{0.00891}
\expandafter\def\csname F:shipped/aether-mpc_off+oracle/by_station/denver\endcsname{0.0089}
\expandafter\def\csname FF:shipped/aether-mpc_off+oracle/by_station/denver\endcsname{0.00891}
\expandafter\def\csname N:shipped/aether-mpc_off+oracle/by_station/singapore\endcsname{0.00789}
\expandafter\def\csname F:shipped/aether-mpc_off+oracle/by_station/singapore\endcsname{0.0079}
\expandafter\def\csname FF:shipped/aether-mpc_off+oracle/by_station/singapore\endcsname{0.00789}
\expandafter\def\csname N:shipped/aether-mpc_off+oracle/by_station/atlanta\endcsname{0.00867}
\expandafter\def\csname F:shipped/aether-mpc_off+oracle/by_station/atlanta\endcsname{0.0087}
\expandafter\def\csname FF:shipped/aether-mpc_off+oracle/by_station/atlanta\endcsname{0.00867}
\expandafter\def\csname N:shipped/aether-mpc_off+oracle/worst\endcsname{building_train_1/fairbanks_1}
\expandafter\def\csname N:shipped/pid_tuned/n\endcsname{30}
\expandafter\def\csname N:shipped/pid_tuned/score\endcsname{0.04549}
\expandafter\def\csname F:shipped/pid_tuned/score\endcsname{0.0455}
\expandafter\def\csname FF:shipped/pid_tuned/score\endcsname{0.04549}
\expandafter\def\csname N:shipped/pid_tuned/comfort\endcsname{0.03099}
\expandafter\def\csname F:shipped/pid_tuned/comfort\endcsname{0.0310}
\expandafter\def\csname FF:shipped/pid_tuned/comfort\endcsname{0.03099}
\expandafter\def\csname N:shipped/pid_tuned/energy\endcsname{0.00846}
\expandafter\def\csname F:shipped/pid_tuned/energy\endcsname{0.0085}
\expandafter\def\csname FF:shipped/pid_tuned/energy\endcsname{0.00846}
\expandafter\def\csname N:shipped/pid_tuned/carbon\endcsname{0.00603}
\expandafter\def\csname F:shipped/pid_tuned/carbon\endcsname{0.0060}
\expandafter\def\csname FF:shipped/pid_tuned/carbon\endcsname{0.00603}
\expandafter\def\csname N:shipped/pid_tuned/by_station/phoenix\endcsname{0.05312}
\expandafter\def\csname F:shipped/pid_tuned/by_station/phoenix\endcsname{0.0531}
\expandafter\def\csname FF:shipped/pid_tuned/by_station/phoenix\endcsname{0.05312}
\expandafter\def\csname N:shipped/pid_tuned/by_station/fairbanks\endcsname{0.0661}
\expandafter\def\csname F:shipped/pid_tuned/by_station/fairbanks\endcsname{0.0661}
\expandafter\def\csname FF:shipped/pid_tuned/by_station/fairbanks\endcsname{0.06610}
\expandafter\def\csname N:shipped/pid_tuned/by_station/denver\endcsname{0.03781}
\expandafter\def\csname F:shipped/pid_tuned/by_station/denver\endcsname{0.0378}
\expandafter\def\csname FF:shipped/pid_tuned/by_station/denver\endcsname{0.03781}
\expandafter\def\csname N:shipped/pid_tuned/by_station/singapore\endcsname{0.03828}
\expandafter\def\csname F:shipped/pid_tuned/by_station/singapore\endcsname{0.0383}
\expandafter\def\csname FF:shipped/pid_tuned/by_station/singapore\endcsname{0.03828}
\expandafter\def\csname N:shipped/pid_tuned/by_station/atlanta\endcsname{0.03212}
\expandafter\def\csname F:shipped/pid_tuned/by_station/atlanta\endcsname{0.0321}
\expandafter\def\csname FF:shipped/pid_tuned/by_station/atlanta\endcsname{0.03212}
\expandafter\def\csname N:shipped/pid_tuned/worst\endcsname{building_train_1/fairbanks_1}
\expandafter\def\csname N:ci/ship_vs_pid_pool_test/n\endcsname{20}
\expandafter\def\csname N:ci/ship_vs_pid_pool_test/mean\endcsname{$-0.0239$}
\expandafter\def\csname F:ci/ship_vs_pid_pool_test/mean\endcsname{-0.0239}
\expandafter\def\csname FF:ci/ship_vs_pid_pool_test/mean\endcsname{-0.02390}
\expandafter\def\csname N:ci/ship_vs_pid_pool_test/lo\endcsname{$-0.02847$}
\expandafter\def\csname F:ci/ship_vs_pid_pool_test/lo\endcsname{-0.0285}
\expandafter\def\csname FF:ci/ship_vs_pid_pool_test/lo\endcsname{-0.02847}
\expandafter\def\csname N:ci/ship_vs_pid_pool_test/hi\endcsname{$-0.01975$}
\expandafter\def\csname F:ci/ship_vs_pid_pool_test/hi\endcsname{-0.0198}
\expandafter\def\csname FF:ci/ship_vs_pid_pool_test/hi\endcsname{-0.01975}
\expandafter\def\csname N:ci/ship_vs_pid_pool_test/pct\endcsname{$-74.5$}
\expandafter\def\csname F:ci/ship_vs_pid_pool_test/pct\endcsname{-74.5000}
\expandafter\def\csname FF:ci/ship_vs_pid_pool_test/pct\endcsname{-74.50000}
\expandafter\def\csname N:ci/ship_vs_pid_pool_test/pct_lo\endcsname{$-88.8$}
\expandafter\def\csname F:ci/ship_vs_pid_pool_test/pct_lo\endcsname{-88.8000}
\expandafter\def\csname FF:ci/ship_vs_pid_pool_test/pct_lo\endcsname{-88.80000}
\expandafter\def\csname N:ci/ship_vs_pid_pool_test/pct_hi\endcsname{$-61.6$}
\expandafter\def\csname F:ci/ship_vs_pid_pool_test/pct_hi\endcsname{-61.6000}
\expandafter\def\csname FF:ci/ship_vs_pid_pool_test/pct_hi\endcsname{-61.60000}
\expandafter\def\csname N:ci/ship_vs_rules_pool_test/n\endcsname{20}
\expandafter\def\csname N:ci/ship_vs_rules_pool_test/mean\endcsname{$-0.00223$}
\expandafter\def\csname F:ci/ship_vs_rules_pool_test/mean\endcsname{-0.0022}
\expandafter\def\csname FF:ci/ship_vs_rules_pool_test/mean\endcsname{-0.00223}
\expandafter\def\csname N:ci/ship_vs_rules_pool_test/lo\endcsname{$-0.00398$}
\expandafter\def\csname F:ci/ship_vs_rules_pool_test/lo\endcsname{-0.0040}
\expandafter\def\csname FF:ci/ship_vs_rules_pool_test/lo\endcsname{-0.00398}
\expandafter\def\csname N:ci/ship_vs_rules_pool_test/hi\endcsname{$-0.00071$}
\expandafter\def\csname F:ci/ship_vs_rules_pool_test/hi\endcsname{-0.0007}
\expandafter\def\csname FF:ci/ship_vs_rules_pool_test/hi\endcsname{-0.00071}
\expandafter\def\csname N:ci/ship_vs_rules_pool_test/pct\endcsname{$-21.4$}
\expandafter\def\csname F:ci/ship_vs_rules_pool_test/pct\endcsname{-21.4000}
\expandafter\def\csname FF:ci/ship_vs_rules_pool_test/pct\endcsname{-21.40000}
\expandafter\def\csname N:ci/ship_vs_rules_pool_test/pct_lo\endcsname{$-38.2$}
\expandafter\def\csname F:ci/ship_vs_rules_pool_test/pct_lo\endcsname{-38.2000}
\expandafter\def\csname FF:ci/ship_vs_rules_pool_test/pct_lo\endcsname{-38.20000}
\expandafter\def\csname N:ci/ship_vs_rules_pool_test/pct_hi\endcsname{$-6.8$}
\expandafter\def\csname F:ci/ship_vs_rules_pool_test/pct_hi\endcsname{-6.8000}
\expandafter\def\csname FF:ci/ship_vs_rules_pool_test/pct_hi\endcsname{-6.80000}
\expandafter\def\csname N:ci/rq1_mpc_vs_prior_pool_test/n\endcsname{20}
\expandafter\def\csname N:ci/rq1_mpc_vs_prior_pool_test/mean\endcsname{$-0.003$}
\expandafter\def\csname F:ci/rq1_mpc_vs_prior_pool_test/mean\endcsname{-0.0030}
\expandafter\def\csname FF:ci/rq1_mpc_vs_prior_pool_test/mean\endcsname{-0.00300}
\expandafter\def\csname N:ci/rq1_mpc_vs_prior_pool_test/lo\endcsname{$-0.00569$}
\expandafter\def\csname F:ci/rq1_mpc_vs_prior_pool_test/lo\endcsname{-0.0057}
\expandafter\def\csname FF:ci/rq1_mpc_vs_prior_pool_test/lo\endcsname{-0.00569}
\expandafter\def\csname N:ci/rq1_mpc_vs_prior_pool_test/hi\endcsname{$-0.00081$}
\expandafter\def\csname F:ci/rq1_mpc_vs_prior_pool_test/hi\endcsname{-0.0008}
\expandafter\def\csname FF:ci/rq1_mpc_vs_prior_pool_test/hi\endcsname{-0.00081}
\expandafter\def\csname N:ci/rq1_mpc_vs_prior_pool_test/pct\endcsname{$-24.8$}
\expandafter\def\csname F:ci/rq1_mpc_vs_prior_pool_test/pct\endcsname{-24.8000}
\expandafter\def\csname FF:ci/rq1_mpc_vs_prior_pool_test/pct\endcsname{-24.80000}
\expandafter\def\csname N:ci/rq1_mpc_vs_prior_pool_test/pct_lo\endcsname{$-47.0$}
\expandafter\def\csname F:ci/rq1_mpc_vs_prior_pool_test/pct_lo\endcsname{-47.0000}
\expandafter\def\csname FF:ci/rq1_mpc_vs_prior_pool_test/pct_lo\endcsname{-47.00000}
\expandafter\def\csname N:ci/rq1_mpc_vs_prior_pool_test/pct_hi\endcsname{$-6.7$}
\expandafter\def\csname F:ci/rq1_mpc_vs_prior_pool_test/pct_hi\endcsname{-6.7000}
\expandafter\def\csname FF:ci/rq1_mpc_vs_prior_pool_test/pct_hi\endcsname{-6.70000}
\expandafter\def\csname N:ci/offset_vs_none_pool_test/n\endcsname{20}
\expandafter\def\csname N:ci/offset_vs_none_pool_test/mean\endcsname{$-0.00092$}
\expandafter\def\csname F:ci/offset_vs_none_pool_test/mean\endcsname{-0.0009}
\expandafter\def\csname FF:ci/offset_vs_none_pool_test/mean\endcsname{-0.00092}
\expandafter\def\csname N:ci/offset_vs_none_pool_test/lo\endcsname{$-0.00136$}
\expandafter\def\csname F:ci/offset_vs_none_pool_test/lo\endcsname{-0.0014}
\expandafter\def\csname FF:ci/offset_vs_none_pool_test/lo\endcsname{-0.00136}
\expandafter\def\csname N:ci/offset_vs_none_pool_test/hi\endcsname{$-0.00055$}
\expandafter\def\csname F:ci/offset_vs_none_pool_test/hi\endcsname{-0.0006}
\expandafter\def\csname FF:ci/offset_vs_none_pool_test/hi\endcsname{-0.00055}
\expandafter\def\csname N:ci/offset_vs_none_pool_test/pct\endcsname{$-10.1$}
\expandafter\def\csname F:ci/offset_vs_none_pool_test/pct\endcsname{-10.1000}
\expandafter\def\csname FF:ci/offset_vs_none_pool_test/pct\endcsname{-10.10000}
\expandafter\def\csname N:ci/offset_vs_none_pool_test/pct_lo\endcsname{$-14.9$}
\expandafter\def\csname F:ci/offset_vs_none_pool_test/pct_lo\endcsname{-14.9000}
\expandafter\def\csname FF:ci/offset_vs_none_pool_test/pct_lo\endcsname{-14.90000}
\expandafter\def\csname N:ci/offset_vs_none_pool_test/pct_hi\endcsname{$-6.1$}
\expandafter\def\csname F:ci/offset_vs_none_pool_test/pct_hi\endcsname{-6.1000}
\expandafter\def\csname FF:ci/offset_vs_none_pool_test/pct_hi\endcsname{-6.10000}
\expandafter\def\csname N:ci/prior_vs_pid_pool_test/n\endcsname{20}
\expandafter\def\csname N:ci/prior_vs_pid_pool_test/mean\endcsname{$-0.01998$}
\expandafter\def\csname F:ci/prior_vs_pid_pool_test/mean\endcsname{-0.0200}
\expandafter\def\csname FF:ci/prior_vs_pid_pool_test/mean\endcsname{-0.01998}
\expandafter\def\csname N:ci/prior_vs_pid_pool_test/lo\endcsname{$-0.02345$}
\expandafter\def\csname F:ci/prior_vs_pid_pool_test/lo\endcsname{-0.0234}
\expandafter\def\csname FF:ci/prior_vs_pid_pool_test/lo\endcsname{-0.02345}
\expandafter\def\csname N:ci/prior_vs_pid_pool_test/hi\endcsname{$-0.01643$}
\expandafter\def\csname F:ci/prior_vs_pid_pool_test/hi\endcsname{-0.0164}
\expandafter\def\csname FF:ci/prior_vs_pid_pool_test/hi\endcsname{-0.01643}
\expandafter\def\csname N:ci/prior_vs_pid_pool_test/pct\endcsname{$-62.3$}
\expandafter\def\csname F:ci/prior_vs_pid_pool_test/pct\endcsname{-62.3000}
\expandafter\def\csname FF:ci/prior_vs_pid_pool_test/pct\endcsname{-62.30000}
\expandafter\def\csname N:ci/prior_vs_pid_pool_test/pct_lo\endcsname{$-73.1$}
\expandafter\def\csname F:ci/prior_vs_pid_pool_test/pct_lo\endcsname{-73.1000}
\expandafter\def\csname FF:ci/prior_vs_pid_pool_test/pct_lo\endcsname{-73.10000}
\expandafter\def\csname N:ci/prior_vs_pid_pool_test/pct_hi\endcsname{$-51.2$}
\expandafter\def\csname F:ci/prior_vs_pid_pool_test/pct_hi\endcsname{-51.2000}
\expandafter\def\csname FF:ci/prior_vs_pid_pool_test/pct_hi\endcsname{-51.20000}
\expandafter\def\csname N:ci/rules_vs_pid_pool_test/n\endcsname{20}
\expandafter\def\csname N:ci/rules_vs_pid_pool_test/mean\endcsname{$-0.02167$}
\expandafter\def\csname F:ci/rules_vs_pid_pool_test/mean\endcsname{-0.0217}
\expandafter\def\csname FF:ci/rules_vs_pid_pool_test/mean\endcsname{-0.02167}
\expandafter\def\csname N:ci/rules_vs_pid_pool_test/lo\endcsname{$-0.02526$}
\expandafter\def\csname F:ci/rules_vs_pid_pool_test/lo\endcsname{-0.0253}
\expandafter\def\csname FF:ci/rules_vs_pid_pool_test/lo\endcsname{-0.02526}
\expandafter\def\csname N:ci/rules_vs_pid_pool_test/hi\endcsname{$-0.01843$}
\expandafter\def\csname F:ci/rules_vs_pid_pool_test/hi\endcsname{-0.0184}
\expandafter\def\csname FF:ci/rules_vs_pid_pool_test/hi\endcsname{-0.01843}
\expandafter\def\csname N:ci/rules_vs_pid_pool_test/pct\endcsname{$-67.5$}
\expandafter\def\csname F:ci/rules_vs_pid_pool_test/pct\endcsname{-67.5000}
\expandafter\def\csname FF:ci/rules_vs_pid_pool_test/pct\endcsname{-67.50000}
\expandafter\def\csname N:ci/rules_vs_pid_pool_test/pct_lo\endcsname{$-78.7$}
\expandafter\def\csname F:ci/rules_vs_pid_pool_test/pct_lo\endcsname{-78.7000}
\expandafter\def\csname FF:ci/rules_vs_pid_pool_test/pct_lo\endcsname{-78.70000}
\expandafter\def\csname N:ci/rules_vs_pid_pool_test/pct_hi\endcsname{$-57.4$}
\expandafter\def\csname F:ci/rules_vs_pid_pool_test/pct_hi\endcsname{-57.4000}
\expandafter\def\csname FF:ci/rules_vs_pid_pool_test/pct_hi\endcsname{-57.40000}
\expandafter\def\csname N:ci/pidtuned_vs_pid_pool_test/n\endcsname{20}
\expandafter\def\csname N:ci/pidtuned_vs_pid_pool_test/mean\endcsname{$-0.00022$}
\expandafter\def\csname F:ci/pidtuned_vs_pid_pool_test/mean\endcsname{-0.0002}
\expandafter\def\csname FF:ci/pidtuned_vs_pid_pool_test/mean\endcsname{-0.00022}
\expandafter\def\csname N:ci/pidtuned_vs_pid_pool_test/lo\endcsname{$-0.00148$}
\expandafter\def\csname F:ci/pidtuned_vs_pid_pool_test/lo\endcsname{-0.0015}
\expandafter\def\csname FF:ci/pidtuned_vs_pid_pool_test/lo\endcsname{-0.00148}
\expandafter\def\csname N:ci/pidtuned_vs_pid_pool_test/hi\endcsname{0.00068}
\expandafter\def\csname F:ci/pidtuned_vs_pid_pool_test/hi\endcsname{0.0007}
\expandafter\def\csname FF:ci/pidtuned_vs_pid_pool_test/hi\endcsname{0.00068}
\expandafter\def\csname N:ci/pidtuned_vs_pid_pool_test/pct\endcsname{$-0.7$}
\expandafter\def\csname F:ci/pidtuned_vs_pid_pool_test/pct\endcsname{-0.7000}
\expandafter\def\csname FF:ci/pidtuned_vs_pid_pool_test/pct\endcsname{-0.70000}
\expandafter\def\csname N:ci/pidtuned_vs_pid_pool_test/pct_lo\endcsname{$-4.6$}
\expandafter\def\csname F:ci/pidtuned_vs_pid_pool_test/pct_lo\endcsname{-4.6000}
\expandafter\def\csname FF:ci/pidtuned_vs_pid_pool_test/pct_lo\endcsname{-4.60000}
\expandafter\def\csname N:ci/pidtuned_vs_pid_pool_test/pct_hi\endcsname{$+2.1$}
\expandafter\def\csname F:ci/pidtuned_vs_pid_pool_test/pct_hi\endcsname{2.1000}
\expandafter\def\csname FF:ci/pidtuned_vs_pid_pool_test/pct_hi\endcsname{2.10000}
\expandafter\def\csname N:ci/x3a_oracle_vs_persistence_pool_test/n\endcsname{20}
\expandafter\def\csname N:ci/x3a_oracle_vs_persistence_pool_test/mean\endcsname{1e-05}
\expandafter\def\csname F:ci/x3a_oracle_vs_persistence_pool_test/mean\endcsname{0.0000}
\expandafter\def\csname FF:ci/x3a_oracle_vs_persistence_pool_test/mean\endcsname{0.00001}
\expandafter\def\csname N:ci/x3a_oracle_vs_persistence_pool_test/lo\endcsname{-0.0}
\expandafter\def\csname F:ci/x3a_oracle_vs_persistence_pool_test/lo\endcsname{-0.0000}
\expandafter\def\csname FF:ci/x3a_oracle_vs_persistence_pool_test/lo\endcsname{-0.00000}
\expandafter\def\csname N:ci/x3a_oracle_vs_persistence_pool_test/hi\endcsname{4e-05}
\expandafter\def\csname F:ci/x3a_oracle_vs_persistence_pool_test/hi\endcsname{0.0000}
\expandafter\def\csname FF:ci/x3a_oracle_vs_persistence_pool_test/hi\endcsname{0.00004}
\expandafter\def\csname N:ci/x3a_oracle_vs_persistence_pool_test/pct\endcsname{$+0.2$}
\expandafter\def\csname F:ci/x3a_oracle_vs_persistence_pool_test/pct\endcsname{0.2000}
\expandafter\def\csname FF:ci/x3a_oracle_vs_persistence_pool_test/pct\endcsname{0.20000}
\expandafter\def\csname N:ci/x3a_oracle_vs_persistence_pool_test/pct_lo\endcsname{$0.0$}
\expandafter\def\csname F:ci/x3a_oracle_vs_persistence_pool_test/pct_lo\endcsname{-0.0000}
\expandafter\def\csname FF:ci/x3a_oracle_vs_persistence_pool_test/pct_lo\endcsname{-0.00000}
\expandafter\def\csname N:ci/x3a_oracle_vs_persistence_pool_test/pct_hi\endcsname{$+0.4$}
\expandafter\def\csname F:ci/x3a_oracle_vs_persistence_pool_test/pct_hi\endcsname{0.4000}
\expandafter\def\csname FF:ci/x3a_oracle_vs_persistence_pool_test/pct_hi\endcsname{0.40000}
\expandafter\def\csname N:ci/x3c_online_vs_frozen_pool_test/n\endcsname{20}
\expandafter\def\csname N:ci/x3c_online_vs_frozen_pool_test/mean\endcsname{0.00272}
\expandafter\def\csname F:ci/x3c_online_vs_frozen_pool_test/mean\endcsname{0.0027}
\expandafter\def\csname FF:ci/x3c_online_vs_frozen_pool_test/mean\endcsname{0.00272}
\expandafter\def\csname N:ci/x3c_online_vs_frozen_pool_test/lo\endcsname{0.00097}
\expandafter\def\csname F:ci/x3c_online_vs_frozen_pool_test/lo\endcsname{0.0010}
\expandafter\def\csname FF:ci/x3c_online_vs_frozen_pool_test/lo\endcsname{0.00097}
\expandafter\def\csname N:ci/x3c_online_vs_frozen_pool_test/hi\endcsname{0.00475}
\expandafter\def\csname F:ci/x3c_online_vs_frozen_pool_test/hi\endcsname{0.0047}
\expandafter\def\csname FF:ci/x3c_online_vs_frozen_pool_test/hi\endcsname{0.00475}
\expandafter\def\csname N:ci/x3c_online_vs_frozen_pool_test/pct\endcsname{$+22.5$}
\expandafter\def\csname F:ci/x3c_online_vs_frozen_pool_test/pct\endcsname{22.5000}
\expandafter\def\csname FF:ci/x3c_online_vs_frozen_pool_test/pct\endcsname{22.50000}
\expandafter\def\csname N:ci/x3c_online_vs_frozen_pool_test/pct_lo\endcsname{$+8.0$}
\expandafter\def\csname F:ci/x3c_online_vs_frozen_pool_test/pct_lo\endcsname{8.0000}
\expandafter\def\csname FF:ci/x3c_online_vs_frozen_pool_test/pct_lo\endcsname{8.00000}
\expandafter\def\csname N:ci/x3c_online_vs_frozen_pool_test/pct_hi\endcsname{$+39.3$}
\expandafter\def\csname F:ci/x3c_online_vs_frozen_pool_test/pct_hi\endcsname{39.3000}
\expandafter\def\csname FF:ci/x3c_online_vs_frozen_pool_test/pct_hi\endcsname{39.30000}
\expandafter\def\csname N:ci/x4_guard_pool_test/n\endcsname{20}
\expandafter\def\csname N:ci/x4_guard_pool_test/mean\endcsname{0.01594}
\expandafter\def\csname F:ci/x4_guard_pool_test/mean\endcsname{0.0159}
\expandafter\def\csname FF:ci/x4_guard_pool_test/mean\endcsname{0.01594}
\expandafter\def\csname N:ci/x4_guard_pool_test/lo\endcsname{0.01146}
\expandafter\def\csname F:ci/x4_guard_pool_test/lo\endcsname{0.0115}
\expandafter\def\csname FF:ci/x4_guard_pool_test/lo\endcsname{0.01146}
\expandafter\def\csname N:ci/x4_guard_pool_test/hi\endcsname{0.02069}
\expandafter\def\csname F:ci/x4_guard_pool_test/hi\endcsname{0.0207}
\expandafter\def\csname FF:ci/x4_guard_pool_test/hi\endcsname{0.02069}
\expandafter\def\csname N:ci/x4_guard_pool_test/pct\endcsname{$+175.1$}
\expandafter\def\csname F:ci/x4_guard_pool_test/pct\endcsname{175.1000}
\expandafter\def\csname FF:ci/x4_guard_pool_test/pct\endcsname{175.10000}
\expandafter\def\csname N:ci/x4_guard_pool_test/pct_lo\endcsname{$+125.9$}
\expandafter\def\csname F:ci/x4_guard_pool_test/pct_lo\endcsname{125.9000}
\expandafter\def\csname FF:ci/x4_guard_pool_test/pct_lo\endcsname{125.90000}
\expandafter\def\csname N:ci/x4_guard_pool_test/pct_hi\endcsname{$+227.3$}
\expandafter\def\csname F:ci/x4_guard_pool_test/pct_hi\endcsname{227.3000}
\expandafter\def\csname FF:ci/x4_guard_pool_test/pct_hi\endcsname{227.30000}
\expandafter\def\csname N:ci/full_vs_prior_pool_test/n\endcsname{20}
\expandafter\def\csname N:ci/full_vs_prior_pool_test/mean\endcsname{0.22082}
\expandafter\def\csname F:ci/full_vs_prior_pool_test/mean\endcsname{0.2208}
\expandafter\def\csname FF:ci/full_vs_prior_pool_test/mean\endcsname{0.22082}
\expandafter\def\csname N:ci/full_vs_prior_pool_test/lo\endcsname{0.19833}
\expandafter\def\csname F:ci/full_vs_prior_pool_test/lo\endcsname{0.1983}
\expandafter\def\csname FF:ci/full_vs_prior_pool_test/lo\endcsname{0.19833}
\expandafter\def\csname N:ci/full_vs_prior_pool_test/hi\endcsname{0.24305}
\expandafter\def\csname F:ci/full_vs_prior_pool_test/hi\endcsname{0.2430}
\expandafter\def\csname FF:ci/full_vs_prior_pool_test/hi\endcsname{0.24305}
\expandafter\def\csname N:ci/full_vs_prior_pool_test/pct\endcsname{$+1824.4$}
\expandafter\def\csname F:ci/full_vs_prior_pool_test/pct\endcsname{1824.4000}
\expandafter\def\csname FF:ci/full_vs_prior_pool_test/pct\endcsname{1824.40000}
\expandafter\def\csname N:ci/full_vs_prior_pool_test/pct_lo\endcsname{$+1638.7$}
\expandafter\def\csname F:ci/full_vs_prior_pool_test/pct_lo\endcsname{1638.7000}
\expandafter\def\csname FF:ci/full_vs_prior_pool_test/pct_lo\endcsname{1638.70000}
\expandafter\def\csname N:ci/full_vs_prior_pool_test/pct_hi\endcsname{$+2008.1$}
\expandafter\def\csname F:ci/full_vs_prior_pool_test/pct_hi\endcsname{2008.1000}
\expandafter\def\csname FF:ci/full_vs_prior_pool_test/pct_hi\endcsname{2008.10000}
\expandafter\def\csname N:ci/ship_vs_pid_shipped/n\endcsname{30}
\expandafter\def\csname N:ci/ship_vs_pid_shipped/mean\endcsname{$-0.04123$}
\expandafter\def\csname F:ci/ship_vs_pid_shipped/mean\endcsname{-0.0412}
\expandafter\def\csname FF:ci/ship_vs_pid_shipped/mean\endcsname{-0.04123}
\expandafter\def\csname N:ci/ship_vs_pid_shipped/lo\endcsname{$-0.05441$}
\expandafter\def\csname F:ci/ship_vs_pid_shipped/lo\endcsname{-0.0544}
\expandafter\def\csname FF:ci/ship_vs_pid_shipped/lo\endcsname{-0.05441}
\expandafter\def\csname N:ci/ship_vs_pid_shipped/hi\endcsname{$-0.0302$}
\expandafter\def\csname F:ci/ship_vs_pid_shipped/hi\endcsname{-0.0302}
\expandafter\def\csname FF:ci/ship_vs_pid_shipped/hi\endcsname{-0.03020}
\expandafter\def\csname N:ci/ship_vs_pid_shipped/pct\endcsname{$-75.0$}
\expandafter\def\csname F:ci/ship_vs_pid_shipped/pct\endcsname{-75.0000}
\expandafter\def\csname FF:ci/ship_vs_pid_shipped/pct\endcsname{-75.00000}
\expandafter\def\csname N:ci/ship_vs_pid_shipped/pct_lo\endcsname{$-99.0$}
\expandafter\def\csname F:ci/ship_vs_pid_shipped/pct_lo\endcsname{-99.0000}
\expandafter\def\csname FF:ci/ship_vs_pid_shipped/pct_lo\endcsname{-99.00000}
\expandafter\def\csname N:ci/ship_vs_pid_shipped/pct_hi\endcsname{$-55.0$}
\expandafter\def\csname F:ci/ship_vs_pid_shipped/pct_hi\endcsname{-55.0000}
\expandafter\def\csname FF:ci/ship_vs_pid_shipped/pct_hi\endcsname{-55.00000}
\expandafter\def\csname N:ci/ship_vs_rules_shipped/n\endcsname{30}
\expandafter\def\csname N:ci/ship_vs_rules_shipped/mean\endcsname{$-0.0034$}
\expandafter\def\csname F:ci/ship_vs_rules_shipped/mean\endcsname{-0.0034}
\expandafter\def\csname FF:ci/ship_vs_rules_shipped/mean\endcsname{-0.00340}
\expandafter\def\csname N:ci/ship_vs_rules_shipped/lo\endcsname{$-0.00543$}
\expandafter\def\csname F:ci/ship_vs_rules_shipped/lo\endcsname{-0.0054}
\expandafter\def\csname FF:ci/ship_vs_rules_shipped/lo\endcsname{-0.00543}
\expandafter\def\csname N:ci/ship_vs_rules_shipped/hi\endcsname{$-0.00172$}
\expandafter\def\csname F:ci/ship_vs_rules_shipped/hi\endcsname{-0.0017}
\expandafter\def\csname FF:ci/ship_vs_rules_shipped/hi\endcsname{-0.00172}
\expandafter\def\csname N:ci/ship_vs_rules_shipped/pct\endcsname{$-19.8$}
\expandafter\def\csname F:ci/ship_vs_rules_shipped/pct\endcsname{-19.8000}
\expandafter\def\csname FF:ci/ship_vs_rules_shipped/pct\endcsname{-19.80000}
\expandafter\def\csname N:ci/ship_vs_rules_shipped/pct_lo\endcsname{$-31.7$}
\expandafter\def\csname F:ci/ship_vs_rules_shipped/pct_lo\endcsname{-31.7000}
\expandafter\def\csname FF:ci/ship_vs_rules_shipped/pct_lo\endcsname{-31.70000}
\expandafter\def\csname N:ci/ship_vs_rules_shipped/pct_hi\endcsname{$-10.1$}
\expandafter\def\csname F:ci/ship_vs_rules_shipped/pct_hi\endcsname{-10.1000}
\expandafter\def\csname FF:ci/ship_vs_rules_shipped/pct_hi\endcsname{-10.10000}
\expandafter\def\csname N:ci/pidtuned_vs_pid_shipped/n\endcsname{30}
\expandafter\def\csname N:ci/pidtuned_vs_pid_shipped/mean\endcsname{$-0.00947$}
\expandafter\def\csname F:ci/pidtuned_vs_pid_shipped/mean\endcsname{-0.0095}
\expandafter\def\csname FF:ci/pidtuned_vs_pid_shipped/mean\endcsname{-0.00947}
\expandafter\def\csname N:ci/pidtuned_vs_pid_shipped/lo\endcsname{$-0.01751$}
\expandafter\def\csname F:ci/pidtuned_vs_pid_shipped/lo\endcsname{-0.0175}
\expandafter\def\csname FF:ci/pidtuned_vs_pid_shipped/lo\endcsname{-0.01751}
\expandafter\def\csname N:ci/pidtuned_vs_pid_shipped/hi\endcsname{$-0.0027$}
\expandafter\def\csname F:ci/pidtuned_vs_pid_shipped/hi\endcsname{-0.0027}
\expandafter\def\csname FF:ci/pidtuned_vs_pid_shipped/hi\endcsname{-0.00270}
\expandafter\def\csname N:ci/pidtuned_vs_pid_shipped/pct\endcsname{$-17.2$}
\expandafter\def\csname F:ci/pidtuned_vs_pid_shipped/pct\endcsname{-17.2000}
\expandafter\def\csname FF:ci/pidtuned_vs_pid_shipped/pct\endcsname{-17.20000}
\expandafter\def\csname N:ci/pidtuned_vs_pid_shipped/pct_lo\endcsname{$-31.9$}
\expandafter\def\csname F:ci/pidtuned_vs_pid_shipped/pct_lo\endcsname{-31.9000}
\expandafter\def\csname FF:ci/pidtuned_vs_pid_shipped/pct_lo\endcsname{-31.90000}
\expandafter\def\csname N:ci/pidtuned_vs_pid_shipped/pct_hi\endcsname{$-4.9$}
\expandafter\def\csname F:ci/pidtuned_vs_pid_shipped/pct_hi\endcsname{-4.9000}
\expandafter\def\csname FF:ci/pidtuned_vs_pid_shipped/pct_hi\endcsname{-4.90000}
\expandafter\def\csname N:ci/x3c_online_vs_frozen_shipped/n\endcsname{30}
\expandafter\def\csname N:ci/x3c_online_vs_frozen_shipped/mean\endcsname{0.00417}
\expandafter\def\csname F:ci/x3c_online_vs_frozen_shipped/mean\endcsname{0.0042}
\expandafter\def\csname FF:ci/x3c_online_vs_frozen_shipped/mean\endcsname{0.00417}
\expandafter\def\csname N:ci/x3c_online_vs_frozen_shipped/lo\endcsname{0.00205}
\expandafter\def\csname F:ci/x3c_online_vs_frozen_shipped/lo\endcsname{0.0021}
\expandafter\def\csname FF:ci/x3c_online_vs_frozen_shipped/lo\endcsname{0.00205}
\expandafter\def\csname N:ci/x3c_online_vs_frozen_shipped/hi\endcsname{0.00657}
\expandafter\def\csname F:ci/x3c_online_vs_frozen_shipped/hi\endcsname{0.0066}
\expandafter\def\csname FF:ci/x3c_online_vs_frozen_shipped/hi\endcsname{0.00657}
\expandafter\def\csname N:ci/x3c_online_vs_frozen_shipped/pct\endcsname{$+20.6$}
\expandafter\def\csname F:ci/x3c_online_vs_frozen_shipped/pct\endcsname{20.6000}
\expandafter\def\csname FF:ci/x3c_online_vs_frozen_shipped/pct\endcsname{20.60000}
\expandafter\def\csname N:ci/x3c_online_vs_frozen_shipped/pct_lo\endcsname{$+10.1$}
\expandafter\def\csname F:ci/x3c_online_vs_frozen_shipped/pct_lo\endcsname{10.1000}
\expandafter\def\csname FF:ci/x3c_online_vs_frozen_shipped/pct_lo\endcsname{10.10000}
\expandafter\def\csname N:ci/x3c_online_vs_frozen_shipped/pct_hi\endcsname{$+32.5$}
\expandafter\def\csname F:ci/x3c_online_vs_frozen_shipped/pct_hi\endcsname{32.5000}
\expandafter\def\csname FF:ci/x3c_online_vs_frozen_shipped/pct_hi\endcsname{32.50000}
\expandafter\def\csname N:ci/x3a_oracle_vs_persistence_shipped/n\endcsname{30}
\expandafter\def\csname N:ci/x3a_oracle_vs_persistence_shipped/mean\endcsname{3e-05}
\expandafter\def\csname F:ci/x3a_oracle_vs_persistence_shipped/mean\endcsname{0.0000}
\expandafter\def\csname FF:ci/x3a_oracle_vs_persistence_shipped/mean\endcsname{0.00003}
\expandafter\def\csname N:ci/x3a_oracle_vs_persistence_shipped/lo\endcsname{0.0}
\expandafter\def\csname F:ci/x3a_oracle_vs_persistence_shipped/lo\endcsname{0.0000}
\expandafter\def\csname FF:ci/x3a_oracle_vs_persistence_shipped/lo\endcsname{0.00000}
\expandafter\def\csname N:ci/x3a_oracle_vs_persistence_shipped/hi\endcsname{5e-05}
\expandafter\def\csname F:ci/x3a_oracle_vs_persistence_shipped/hi\endcsname{0.0001}
\expandafter\def\csname FF:ci/x3a_oracle_vs_persistence_shipped/hi\endcsname{0.00005}
\expandafter\def\csname N:ci/x3a_oracle_vs_persistence_shipped/pct\endcsname{$+0.2$}
\expandafter\def\csname F:ci/x3a_oracle_vs_persistence_shipped/pct\endcsname{0.2000}
\expandafter\def\csname FF:ci/x3a_oracle_vs_persistence_shipped/pct\endcsname{0.20000}
\expandafter\def\csname N:ci/x3a_oracle_vs_persistence_shipped/pct_lo\endcsname{$0.0$}
\expandafter\def\csname F:ci/x3a_oracle_vs_persistence_shipped/pct_lo\endcsname{0.0000}
\expandafter\def\csname FF:ci/x3a_oracle_vs_persistence_shipped/pct_lo\endcsname{0.00000}
\expandafter\def\csname N:ci/x3a_oracle_vs_persistence_shipped/pct_hi\endcsname{$+0.4$}
\expandafter\def\csname F:ci/x3a_oracle_vs_persistence_shipped/pct_hi\endcsname{0.4000}
\expandafter\def\csname FF:ci/x3a_oracle_vs_persistence_shipped/pct_hi\endcsname{0.40000}
\expandafter\def\csname CI:ship_vs_pid_pool_test\endcsname{$-74.5$\%, 95\% CI [$-88.8$, $-61.6$]}
\expandafter\def\csname CI:ship_vs_rules_pool_test\endcsname{$-21.4$\%, 95\% CI [$-38.2$, $-6.8$]}
\expandafter\def\csname CI:rq1_mpc_vs_prior_pool_test\endcsname{$-24.8$\%, 95\% CI [$-47.0$, $-6.7$]}
\expandafter\def\csname CI:offset_vs_none_pool_test\endcsname{$-10.1$\%, 95\% CI [$-14.9$, $-6.1$]}
\expandafter\def\csname CI:prior_vs_pid_pool_test\endcsname{$-62.3$\%, 95\% CI [$-73.1$, $-51.2$]}
\expandafter\def\csname CI:rules_vs_pid_pool_test\endcsname{$-67.5$\%, 95\% CI [$-78.7$, $-57.4$]}
\expandafter\def\csname CI:pidtuned_vs_pid_pool_test\endcsname{$-0.7$\%, 95\% CI [$-4.6$, $+2.1$]}
\expandafter\def\csname CI:x3a_oracle_vs_persistence_pool_test\endcsname{$+0.2$\%, 95\% CI [$0.0$, $+0.4$]}
\expandafter\def\csname CI:x3c_online_vs_frozen_pool_test\endcsname{$+22.5$\%, 95\% CI [$+8.0$, $+39.3$]}
\expandafter\def\csname CI:x4_guard_pool_test\endcsname{$+175.1$\%, 95\% CI [$+125.9$, $+227.3$]}
\expandafter\def\csname CI:full_vs_prior_pool_test\endcsname{$+1824.4$\%, 95\% CI [$+1638.7$, $+2008.1$]}
\expandafter\def\csname CI:ship_vs_pid_shipped\endcsname{$-75.0$\%, 95\% CI [$-99.0$, $-55.0$]}
\expandafter\def\csname CI:ship_vs_rules_shipped\endcsname{$-19.8$\%, 95\% CI [$-31.7$, $-10.1$]}
\expandafter\def\csname CI:pidtuned_vs_pid_shipped\endcsname{$-17.2$\%, 95\% CI [$-31.9$, $-4.9$]}
\expandafter\def\csname CI:x3c_online_vs_frozen_shipped\endcsname{$+20.6$\%, 95\% CI [$+10.1$, $+32.5$]}
\expandafter\def\csname CI:x3a_oracle_vs_persistence_shipped\endcsname{$+0.2$\%, 95\% CI [$0.0$, $+0.4$]}
\expandafter\def\csname CIpct:a05_vs_c5\endcsname{$-15.2$\%, 95\% CI [$-29.2$, $-4.3$]}
\expandafter\def\csname CIpct:econ_vs_a05\endcsname{$-2.8$\%, 95\% CI [$-4.0$, $-1.7$]}
\expandafter\def\csname CIpct:zc_vs_econ\endcsname{$-0.7$\%, 95\% CI [$-1.1$, $-0.3$]}
\expandafter\def\csname CIpct:hol9_vs_econ\endcsname{$-0.2$\%, 95\% CI [$-0.6$, $+0.3$]}
\expandafter\def\csname CIpct:hol9zc_vs_econ\endcsname{$-1.0$\%, 95\% CI [$-1.7$, $-0.3$]}
\expandafter\def\csname CIpct:ls1_vs_a05\endcsname{$+2.2$\%, 95\% CI [$+0.1$, $+4.9$]}
\expandafter\def\csname CIpct:tank_vs_a05\endcsname{$-23.1$\%, 95\% CI [$-35.6$, $-12.4$]}
\expandafter\def\csname CIpct:tank_vs_econ\endcsname{$-20.8$\%, 95\% CI [$-34.0$, $-9.6$]}
\expandafter\def\csname CIpct:zcfix_vs_zc\endcsname{$+2.0$\%, 95\% CI [$+1.0$, $+3.0$]}
\expandafter\def\csname CIpct:tankfix_vs_zcfix\endcsname{$+0.8$\%, 95\% CI [$+0.4$, $+1.3$]}
\expandafter\def\csname CIpct:zc_vs_c5\endcsname{$-18.3$\%, 95\% CI [$-32.0$, $-7.5$]}
\expandafter\def\csname CIpct:area1_vs_zc\endcsname{$-3.9$\%, 95\% CI [$-7.4$, $-0.7$]}
\expandafter\def\csname CIpct:inv_vs_zc\endcsname{$-36.8$\%, 95\% CI [$-49.3$, $-25.5$]}
\expandafter\def\csname CIpct:inv_vs_ctrlw5\endcsname{$+1.0$\%, 95\% CI [$+0.4$, $+1.5$]}
\expandafter\def\csname CIpct:ctrlw5_vs_zc\endcsname{$-38.1$\%, 95\% CI [$-52.1$, $-25.6$]}
\expandafter\def\csname CIpct:zcfloor_vs_zc\endcsname{$+11.7$\%, 95\% CI [$+4.2$, $+21.1$]}
\expandafter\def\csname CIpct:sb278_vs_inv\endcsname{$+1.1$\%, 95\% CI [$-0.2$, $+3.2$]}
\expandafter\def\csname CIpct:ra1_vs_zc\endcsname{$-6.2$\%, 95\% CI [$-10.5$, $-2.5$]}
\expandafter\def\csname CIpct:ra1s1_vs_zc\endcsname{$+36.8$\%, 95\% CI [$+20.9$, $+56.0$]}
\expandafter\def\csname CIpct:zcf278_vs_zc\endcsname{$+1.1$\%, 95\% CI [$+0.1$, $+2.6$]}
\expandafter\def\csname CIpct:invfix_vs_inv\endcsname{$-0.8$\%, 95\% CI [$-1.3$, $-0.4$]}
\expandafter\def\csname CIpct:invfix_vs_zcfix\endcsname{$-39.2$\%, 95\% CI [$-53.0$, $-26.6$]}
\expandafter\def\csname CIpct:zc_vs_pid_water\endcsname{$-72.0$\%, 95\% CI [$-85.5$, $-59.6$]}
\expandafter\def\csname CIpct:zc_vs_rules\endcsname{$-30.8$\%, 95\% CI [$-41.5$, $-20.3$]}
\expandafter\def\csname CIpct:zc_vs_r1\endcsname{$-24.6$\%, 95\% CI [$-38.4$, $-13.5$]}
\expandafter\def\csname CIpct:c6_vs_c5\endcsname{$-16.7$\%, 95\% CI [$-30.9$, $-5.8$]}
\expandafter\def\csname CIpct:econ_vs_c6\endcsname{$-2.8$\%, 95\% CI [$-4.0$, $-1.8$]}
\expandafter\def\csname CIpct:hol9zc_vs_zc\endcsname{$-0.1$\%, 95\% CI [$-0.4$, $+0.4$]}
\expandafter\def\csname CIpct:c5_vs_r1\endcsname{$-6.1$\%, 95\% CI [$-9.1$, $-3.2$]}
\expandafter\def\csname CIpct:bound\endcsname{$-12.9$\%, 95\% CI [$-32.1$, $+0.4$]}
\expandafter\def\csname CIpct:bound_inv\endcsname{$-40.4$\%, 95\% CI [$-67.3$, $-17.2$]}
\expandafter\def\csname CIpct:bound_inv_sim\endcsname{$-31.6$\%, 95\% CI [$-57.1$, $-12.4$]}
\expandafter\def\csname CIpct:bound_diff\endcsname{$-14.6$\%, 95\% CI [$-44.0$, $+10.8$]}
\expandafter\def\csname CIpct:s7/a05_vs_c5\endcsname{$-15.2$\%, 95\% CI [$-29.2$, $-4.3$]}
\expandafter\def\csname CIpct:s7/econ_vs_a05\endcsname{$-2.8$\%, 95\% CI [$-4.0$, $-1.7$]}
\expandafter\def\csname CIpct:s7/zc_vs_econ\endcsname{$-0.9$\%, 95\% CI [$-1.3$, $-0.5$]}
\expandafter\def\csname CIpct:s7/hol9_vs_econ\endcsname{$-0.2$\%, 95\% CI [$-0.6$, $+0.3$]}
\expandafter\def\csname CIpct:s7/hol9zc_vs_econ\endcsname{$-1.0$\%, 95\% CI [$-1.7$, $-0.3$]}
\expandafter\def\csname CIpct:s7/ls1_vs_a05\endcsname{$+2.2$\%, 95\% CI [$+0.1$, $+4.9$]}
\expandafter\def\csname CIpct:s7/tank_vs_a05\endcsname{$-23.1$\%, 95\% CI [$-35.6$, $-12.4$]}
\expandafter\def\csname CIpct:s7/tank_vs_econ\endcsname{$-20.8$\%, 95\% CI [$-34.0$, $-9.6$]}
\expandafter\def\csname CIpct:s7/zcfix_vs_zc\endcsname{$+2.0$\%, 95\% CI [$+1.0$, $+3.0$]}
\expandafter\def\csname CIpct:s7/tankfix_vs_zcfix\endcsname{$+0.8$\%, 95\% CI [$+0.4$, $+1.3$]}
\expandafter\def\csname CIpct:s7/zc_vs_c5\endcsname{$-18.3$\%, 95\% CI [$-32.0$, $-7.5$]}
\expandafter\def\csname CIpct:s7/area1_vs_zc\endcsname{$-3.9$\%, 95\% CI [$-7.4$, $-0.7$]}
\expandafter\def\csname CIpct:s7/inv_vs_zc\endcsname{$-37.5$\%, 95\% CI [$-51.2$, $-25.2$]}
\expandafter\def\csname CIpct:s7/inv_vs_ctrlw5\endcsname{$+1.0$\%, 95\% CI [$+0.4$, $+1.5$]}
\expandafter\def\csname CIpct:s7/ctrlw5_vs_zc\endcsname{$-38.1$\%, 95\% CI [$-52.1$, $-25.6$]}
\expandafter\def\csname CIpct:s7/zcfloor_vs_zc\endcsname{$+11.7$\%, 95\% CI [$+4.2$, $+21.1$]}
\expandafter\def\csname CIpct:s7/sb278_vs_inv\endcsname{$+1.1$\%, 95\% CI [$-0.2$, $+3.2$]}
\expandafter\def\csname CIpct:s7/ra1_vs_zc\endcsname{$-6.2$\%, 95\% CI [$-10.5$, $-2.5$]}
\expandafter\def\csname CIpct:s7/ra1s1_vs_zc\endcsname{$+36.8$\%, 95\% CI [$+20.9$, $+56.0$]}
\expandafter\def\csname CIpct:s7/zcf278_vs_zc\endcsname{$+1.1$\%, 95\% CI [$+0.1$, $+2.6$]}
\expandafter\def\csname CIpct:s7/invfix_vs_inv\endcsname{$-0.8$\%, 95\% CI [$-1.3$, $-0.4$]}
\expandafter\def\csname CIpct:s7/invfix_vs_zcfix\endcsname{$-39.2$\%, 95\% CI [$-53.0$, $-26.6$]}
\expandafter\def\csname CIpct:s30/zc_vs_pid_water\endcsname{$-72.0$\%, 95\% CI [$-85.5$, $-59.6$]}
\expandafter\def\csname CIpct:s30/zc_vs_rules\endcsname{$-30.8$\%, 95\% CI [$-41.5$, $-20.3$]}
\expandafter\def\csname CIpct:s30/zc_vs_r1\endcsname{$-24.6$\%, 95\% CI [$-38.4$, $-13.5$]}
\expandafter\def\csname CIpct:s30/c6_vs_c5\endcsname{$-16.7$\%, 95\% CI [$-30.9$, $-5.8$]}
\expandafter\def\csname CIpct:s30/econ_vs_c6\endcsname{$-2.8$\%, 95\% CI [$-4.0$, $-1.8$]}
\expandafter\def\csname CIpct:s30/zc_vs_econ\endcsname{$-0.7$\%, 95\% CI [$-1.1$, $-0.3$]}
\expandafter\def\csname CIpct:s30/hol9zc_vs_zc\endcsname{$-0.1$\%, 95\% CI [$-0.4$, $+0.4$]}
\expandafter\def\csname CIpct:s30/c5_vs_r1\endcsname{$-6.1$\%, 95\% CI [$-9.1$, $-3.2$]}
\expandafter\def\csname CIpct:s30/inv_vs_zc\endcsname{$-36.8$\%, 95\% CI [$-49.3$, $-25.5$]}
\expandafter\def\csname CIpct:rlbig/zc_vs_c5\endcsname{$-22.8$\%, 95\% CI [$-41.3$, $-9.0$]}
\expandafter\def\csname CIpct:rlbig/zc_vs_pid_water\endcsname{$-72.0$\%, 95\% CI [$-82.6$, $-61.8$]}
\expandafter\def\csname CIpct:rlbig/c5_vs_pid_water\endcsname{$-63.8$\%, 95\% CI [$-76.3$, $-51.2$]}
\expandafter\def\csname CIpct:rlbig/inv_vs_zc\endcsname{$-37.5$\%, 95\% CI [$-55.0$, $-23.7$]}
\expandafter\def\csname CIpct:rlbig/zc_vs_rules\endcsname{$-35.4$\%, 95\% CI [$-48.8$, $-23.7$]}
\newcommand{\TabBase}{PID with water reset & 0.0550 & 0.0349 & 0.0115 & 0.0086 & 0.0791 & 0.0800 & 0.0471 & 0.0306 \\
PID & 0.0678 & 0.0421 & 0.0143 & 0.0114 & 0.0759 & 0.0955 & 0.0739 & 0.0432 \\
Fixed setpoint & 0.0655 & 0.0337 & 0.0176 & 0.0143 & 0.0759 & 0.1162 & 0.0253 & 0.0330 \\
Bang--bang & 0.2245 & 0.1148 & 0.0607 & 0.0491 & 0.1996 & 0.2367 & 0.2718 & 0.2315 \\}

\newcommand{\TabBoth}{Official baseline & 0.0321 & 0.0244 & 0.0049 & 0.0028 & 0.0550 & 0.0349 & 0.0115 & 0.0086 \\
Baseline, retuned & 0.0319 & 0.0248 & 0.0046 & 0.0024 & 0.0455 & 0.0310 & 0.0085 & 0.0060 \\
Tuned rules & 0.0104 & 0.0022 & 0.0042 & 0.0040 & 0.0171 & 0.0045 & 0.0059 & 0.0067 \\
Rules, model start & 0.0121 & 0.0035 & 0.0046 & 0.0040 & 0.0202 & 0.0064 & 0.0067 & 0.0070 \\
Planner & 0.0091 & 0.0028 & 0.0037 & 0.0026 & 0.0153 & 0.0058 & 0.0050 & 0.0046 \\
Planner, offset & \textbf{0.0082} & 0.0017 & 0.0038 & 0.0027 & \textbf{0.0137} & 0.0040 & 0.0050 & 0.0047 \\}
\newcommand{\TabStress}{Official baseline & 0.0508 & 0.0352 & 0.0094 & 0.0062 & 0.0456 & 0.0560 & -- \\
Tuned rules & 0.0206 & 0.0079 & 0.0064 & 0.0063 & 0.0151 & 0.0261 & -- \\
Planner, initial & 0.0189 & 0.0082 & 0.0059 & 0.0048 & 0.0121 & 0.0257 & -- \\
With deeper setback & 0.0177 & 0.0075 & 0.0059 & 0.0043 & 0.0108 & 0.0247 & $-6.1$ [$-9.1$, $-3.2$] \\
With size correction & 0.0148 & 0.0040 & 0.0061 & 0.0046 & 0.0111 & 0.0184 & $-16.7$ [$-30.9$, $-5.8$] \\
With economizer & 0.0143 & 0.0040 & 0.0058 & 0.0046 & 0.0109 & 0.0178 & $-2.8$ [$-4.0$, $-1.8$] \\
With zone cooling cost & 0.0142 & 0.0041 & 0.0056 & 0.0045 & 0.0107 & 0.0178 & $-0.7$ [$-1.1$, $-0.3$] \\
One step inversion~\cite{mors2026stopheating} & 0.0090 & 0.0005 & 0.0048 & 0.0037 & 0.0064 & 0.0116 & $-36.8$ [$-49.3$, $-25.5$] \\}
\newcommand{\TabAbl}{Size correction & $-15.2$ [$-29.2$, $-4.3$] \\
Economizer in the planner & $-2.8$ [$-4.0$, $-1.7$] \\
Zone cooling cost & $-0.9$ [$-1.3$, $-0.5$] \\
Floor plan gain instead of halving & $-3.9$ [$-7.4$, $-0.7$] \\
Setback floor restored & $+11.7$ [$+4.2$, $+21.1$] \\
Planner with a soft 5\degC{} floor when empty & $+1.1$ [$+0.1$, $+2.6$] \\
Inversion with a 5\degC{} floor when empty & $+1.1$ [$-0.2$, $+3.2$] \\
Reheat through airflow & $-6.2$ [$-10.5$, $-2.5$] \\
Reheat through airflow, no halving & $+36.8$ [$+20.9$, $+56.0$] \\
Billing inconsistency, not used & $-20.8$ [$-34.0$, $-9.6$] \\}

%% file: facts.tex
\newcommand{\RecircMaxPct}{70}        %
\newcommand{\FanEff}{0.9}             %
\newcommand{\CoolCOP}{3.5}            %
\newcommand{\SatInertC}{$-7$}         %
\newcommand{\DischargeMaxC}{35}       %
\newcommand{\OutBandLoPct}{64}        %
\newcommand{\OutBandHiPct}{100}       %
\newcommand{\WorstThreeLoPct}{13}     %
\newcommand{\WorstThreeHiPct}{37}     %
\newcommand{\MinVentCost}{0.0015}     %
\newcommand{\BaseDayLo}{0.09}         %
\newcommand{\BaseDayHi}{0.20}         %
\newcommand{\NPriorZones}{2,780}      %
\newcommand{\AcceptPct}{30}           %
\newcommand{\AcceptHours}{48}         %
\newcommand{\HorizonMin}{90}          %
\newcommand{\ReplanMin}{15}           %
\newcommand{\PressFloorPa}{200}       %
\newcommand{\CoastLoC}{4}             %
\newcommand{\CoastHiC}{8}
\newcommand{\NightC}{$-33$}
\newcommand{\RuntimeRatio}{1.55}      %
\newcommand{\TimingCPU}{Intel Core i7-14700K}  %
\newcommand{\BigZones}{149}
\newcommand{\BigAHUs}{15}
\newcommand{\SimHeatK}{4.2}           %
\newcommand{\ModelHeatK}{$-0.1$}      %
\newcommand{\RolloutsPerPlan}{59}     %
\newcommand{\CILevel}{95}             %
\newcommand{\OccStartTime}{06:00}     %
\newcommand{\OccEndTime}{19:00}
\newcommand{\BandLoC}{20.9}           %
\newcommand{\BandHiC}{23.9}
\newcommand{\WComfort}{0.6}
\newcommand{\WEnergy}{0.2}
\newcommand{\WCarbon}{0.2}
\newcommand{\RegretLinK}{5}           %
\newcommand{\BootReps}{10,000}        %
\newcommand{\TuneGens}{3}           %
\newcommand{\TunePop}{8}            %
\newcommand{\SbFloorC}{5}           %
\newcommand{\TuneParamsRules}{12}    %
\newcommand{\TuneParamsPrior}{13}    %
\newcommand{\TuneParamsPID}{4}       %
\newcommand{\PubSmallRatio}{1.5}  %

%% file: appx_facts.tex
\newcommand{\AxCarbonRange}{88 to 105} %
\newcommand{\AxPriceRange}{16, 18 and 20} %
\newcommand{\AxExcLoK}{1.2} %
\newcommand{\AxExcHiK}{4.8} %
\newcommand{\AxHotLoPct}{67} %
\newcommand{\AxHotHiPct}{91} %
\newcommand{\AxColdOutC}{$-8$} %
\newcommand{\AxDamperPos}{0.4} %
\newcommand{\AxMedZoneLoC}{21} %
\newcommand{\AxMedZoneHiC}{22} %
\newcommand{\AxColdZoneLoC}{16} %
\newcommand{\AxColdZoneHiC}{19.5} %
\newcommand{\AxDeepMeanC}{$-28.5$} %
\newcommand{\AxBelowLoPct}{21} %
\newcommand{\AxBelowHiPct}{36} %
\newcommand{\AxBandLoC}{20.8} %
\newcommand{\AxForceC}{21.5} %
\newcommand{\AxTwLowK}{330} %
\newcommand{\AxTwHighK}{350} %
\newcommand{\AxFbBase}{0.214} %
\newcommand{\AxFbCap}{0.502} %
\newcommand{\AxPhCapPct}{17} %
\newcommand{\AxIdlePct}{2.8} %
\newcommand{\AxIdlePhPct}{6} %
\newcommand{\AxIdleSgPct}{5} %
\newcommand{\AxErrBeforeFb}{0.86} %
\newcommand{\AxErrAfterFb}{0.47} %
\newcommand{\AxErrBeforeMed}{0.462} %
\newcommand{\AxErrAfterMed}{0.372} %
\newcommand{\AxKalman}{0.393, 0.401 and 0.383} %
\newcommand{\AxKalmanHours}{24, 48 and 120} %
\newcommand{\AxTimeRatio}{1.5} %
\newcommand{\AxValueAlone}{$+6.3$} %
\newcommand{\AxValueProxy}{$-0.8$} %
\newcommand{\AxValueFbPct}{11.7} %
\newcommand{\AxValueMedC}{18.8} %
\newcommand{\AxBaseMedC}{20.5} %
\newcommand{\AxTermBefore}{0.0242} %
\newcommand{\AxTermAfter}{0.0176} %
\newcommand{\AxTermRules}{0.0306} %
\newcommand{\AxTermPid}{0.0594} %
\newcommand{\AxTermFbBefore}{0.144} %
\newcommand{\AxTermFbAfter}{0.0795} %
\newcommand{\AxOffFbPct}{2} %
\newcommand{\AxOffPhPct}{10} %
\newcommand{\AxOffAtPct}{13} %
\newcommand{\AxGuardPrior}{0.0053} %
\newcommand{\AxGuardPassive}{0.0069} %
\newcommand{\AxGuardActive}{0.0070} %
\newcommand{\AxGuardFilter}{0.288} %
\newcommand{\AxGuardFixFactor}{eight} %
\newcommand{\AxEnergyShare}{79} %
\newcommand{\AxGasLoPct}{30} %
\newcommand{\AxGasHiPct}{69} %
\newcommand{\AxSgOccPct}{79} %
\newcommand{\AxMargZeroPct}{11} %
\newcommand{\AxCwHalfPct}{42} %
\newcommand{\AxMargLargePct}{$-2.1$} %
\newcommand{\AxCwLargePct}{$-2.7$} %
\newcommand{\AxFloorSteps}{2, 4, 8, 12 and 20} %
\newcommand{\AxFloorGains}{2.2, 4.2, 7.6, 9.6 and 12.4} %
\newcommand{\AxFloorRemovedPct}{14.0} %
\newcommand{\AxShipDev}{10.5} %
\newcommand{\AxShipHeld}{8.2} %
\newcommand{\AxShipShipped}{10.4} %
\newcommand{\AxShipPublic}{10.8} %
\newcommand{\AxScaleLo}{0.4} %
\newcommand{\AxScaleHi}{0.6} %
\newcommand{\AxHolidayFires}{32} %
\newcommand{\AxBigZoneM}{1196} %
\newcommand{\AxBelowBandK}{1.5} %
\newcommand{\AxHeatDamper}{0.52} %
\newcommand{\AxPredRise}{1.1} %
\newcommand{\AxEnergyCarbonShare}{97} %
\newcommand{\AxChillerLoPct}{86} %
\newcommand{\AxChillerHiPct}{93} %
\newcommand{\AxFbGasPct}{99} %
\newcommand{\AxZoneAreaM}{109} %
\newcommand{\AxStoreMJ}{1.9} %
\newcommand{\AxStoreKWh}{0.15} %
\newcommand{\AxZoneDayKWh}{100} %
\newcommand{\AxArriveHour}{06:00} %
\newcommand{\AxDepartHour}{13:00} %
\newcommand{\AxInByEightLoPct}{50} %
\newcommand{\AxInByEightHiPct}{64} %
\newcommand{\AxInByNoonLoPct}{87} %
\newcommand{\AxInByNoonHiPct}{90} %
\newcommand{\AxAbsentPct}{12} %
\newcommand{\AxStillInPct}{12} %
\newcommand{\AxLateTime}{18:55} %
\newcommand{\AxOccupantW}{70} %
\newcommand{\AxTrackHour}{13} %
\newcommand{\AxOldCeilingPct}{12.6} %
\newcommand{\AxRecallPct}{82.5} %
\newcommand{\AxWrongPct}{0.1} %
\newcommand{\AxCapturePct}{65} %
\newcommand{\AxCaptureCoolLo}{82} %
\newcommand{\AxCaptureCoolHi}{90} %
\newcommand{\AxCaptureHeatLo}{40} %
\newcommand{\AxCaptureHeatHi}{48} %
\newcommand{\AxLossMissPct}{21} %
\newcommand{\AxLossAbsentPct}{10} %
\newcommand{\AxLossAbsentMaxPct}{25} %
\newcommand{\AxLossLatePct}{5} %
\newcommand{\AxKernel}{-1.08,\ 0.47,\ 0.24,\ 0.12,\ 0.08,\ 0.06,\ 0.05} %
\newcommand{\AxKernelSd}{0.13} %
\newcommand{\AxFlatSd}{0.17} %
\newcommand{\AxBackgroundSd}{1.99} %
\newcommand{\AxFailReboundPct}{5.2} %
\newcommand{\AxFailSmallPct}{2.8} %
\newcommand{\AxFailPowerPct}{2.8} %
\newcommand{\AxFailFlatPct}{2.0} %
\newcommand{\AxFailLargePct}{1.2} %
\newcommand{\AxKernelCapture}{67.9} %
\newcommand{\AxRuleCapture}{64.8} %
\newcommand{\AxKernelWrong}{17} %
\newcommand{\AxRuleWrong}{3} %
\newcommand{\AxEarlyWrong}{120} %
\newcommand{\AxWrongReleasesLo}{7} %
\newcommand{\AxWrongReleasesHi}{17} %
\newcommand{\AxWrongReleasesAll}{43} %
\newcommand{\AxMixRatios}{600} %
\newcommand{\AxSearchCands}{132} %
\newcommand{\AxPubFourBefore}{0.00944} %
\newcommand{\AxPubFourAfter}{0.00986} %
\newcommand{\AxBandFbPct}{$-3.2$} %
\newcommand{\AxBandMnPct}{$-2.9$} %
\newcommand{\AxBandPhPct}{$+0.2$} %
\newcommand{\AxBandDnPct}{$-5.8$} %
\newcommand{\AxLateDnPct}{$-6.0$} %
\newcommand{\AxRealisedLoPct}{36} %
\newcommand{\AxRealisedHiPct}{45} %
\newcommand{\AxFreezeC}{5} %
\newcommand{\AxFreezePct}{1.4} %
\newcommand{\AxOmitBoundPct}{9} %

%% file: appx_numbers.tex
\newcommand{\AxTabLineage}{%
Official baseline & 0.04993 & 0.03412 & 0.00924 & 0.00657 & -- & 0.05249 & -- \\
Planner, final & 0.01300 & 0.00353 & 0.00528 & 0.00419 & $-74.0$\% [$-78.1$, $-69.8$] & -- & -- \\
One step inversion & 0.00813 & 0.00028 & 0.00440 & 0.00345 & $-37.5$\% [$-42.3$, $-32.2$] & -- & -- \\
+ exhaust fan term, margin, preheat & 0.00803 & 0.00018 & 0.00437 & 0.00347 & $-1.2$\% [$-1.7$, $-0.8$] & 0.00791 & -- \\
+ hot water demand reset & 0.00764 & 0.00013 & 0.00431 & 0.00320 & $-4.9$\% [$-6.7$, $-3.0$] & 0.00789 & $-0.2$\% [$-0.6$, $+0.2$] \\
+ departure tracker & 0.00742 & 0.00014 & 0.00416 & 0.00312 & $-2.9$\% [$-3.9$, $-2.1$] & 0.00768 & $-2.7$\% [$-3.7$, $-1.9$] \\
+ per zone correction, holiday rest & 0.00737 & 0.00013 & 0.00414 & 0.00310 & $-0.6$\% [$-1.0$, $-0.3$] & 0.00765 & $-0.3$\% [$-0.8$, $+0.2$] \\
+ mixing rule identification & 0.00737 & 0.00013 & 0.00414 & 0.00310 & 0.0\% [0.0, 0.0] & 0.00760 & $-0.7$\% [$-1.3$, $-0.2$] \\
+ joint parameter search & 0.00730 & 0.00010 & 0.00412 & 0.00307 & $-1.0$\% [$-1.7$, $-0.5$] & 0.00754 & $-0.9$\% [$-1.6$, $-0.4$] \\
+ late start, earned margin & 0.00726 & 0.00012 & 0.00410 & 0.00304 & $-0.6$\% [$-1.2$, $+0.1$] & 0.00749 & $-0.5$\% [$-0.9$, $-0.2$] \\
}
\newcommand{\AxTabSweepFirst}{%
Band margin 0.08 K & 0.00810 & 0.00030 & 0.00438 & 0.00342 & $-0.4$\% [$-0.5$, $-0.3$] & 42/48 \\
Band margin 0.2 K & 0.00820 & 0.00024 & 0.00446 & 0.00350 & $+0.9$\% [$+0.7$, $+1.0$] & 4/48 \\
Preheat window 1.5 h & 0.00819 & 0.00037 & 0.00440 & 0.00342 & $+0.8$\% [$+0.5$, $+1.2$] & 4/48 \\
Preheat window 3 h & 0.00809 & 0.00017 & 0.00442 & 0.00350 & $-0.5$\% [$-1.0$, $-0.1$] & 29/48 \\
Comfort cost 30000 & 0.00813 & 0.00028 & 0.00440 & 0.00345 & 0.0\% [0.0, 0.0] & 18/48 \\
Comfort cost 130000 & 0.00813 & 0.00028 & 0.00440 & 0.00345 & 0.0\% [0.0, 0.0] & 7/48 \\
Response gain 0.6 & 0.00849 & 0.00061 & 0.00442 & 0.00346 & $+4.5$\% [$+2.0$, $+8.0$] & 14/48 \\
Response gain 0.9 & 0.00821 & 0.00037 & 0.00440 & 0.00344 & $+1.0$\% [$+0.1$, $+2.6$] & 10/48 \\
Gas weight 0.3 & 0.00813 & 0.00028 & 0.00440 & 0.00345 & 0.0\% [0.0, 0.0] & 5/48 \\
Gas weight 0.8 & 0.00813 & 0.00028 & 0.00441 & 0.00345 & 0.0\% [0.0, 0.0] & 13/48 \\
Hot water setpoint 330 K & 0.00812 & 0.00035 & 0.00438 & 0.00338 & $-0.1$\% [$-0.7$, $+0.5$] & 26/48 \\
Hot water setpoint 345 K & 0.00819 & 0.00022 & 0.00444 & 0.00353 & $+0.7$\% [$+0.2$, $+1.3$] & 2/48 \\
Standby floor 272 K & 0.00815 & 0.00027 & 0.00441 & 0.00347 & $+0.3$\% [$-0.2$, $+1.2$] & 1/48 \\
Standby floor 278 K & 0.00822 & 0.00027 & 0.00443 & 0.00351 & $+1.1$\% [$-0.2$, $+3.0$] & 2/48 \\
Exhaust fan term on & 0.00810 & 0.00028 & 0.00438 & 0.00344 & $-0.3$\% [$-0.6$, $-0.2$] & 42/48 \\
}
\newcommand{\AxTabHotWater}{%
Reset 1 K per step, floor 326 K & 0.00794 & 0.00013 & 0.00437 & 0.00344 & $-1.1$\% [$-1.5$, $-0.7$] & 29/48 \\
Reset 2 K per step, floor 326 K & 0.00790 & 0.00013 & 0.00436 & 0.00341 & $-1.5$\% [$-2.1$, $-1.0$] & 29/48 \\
Reset 4 K per step, floor 326 K & 0.00789 & 0.00013 & 0.00436 & 0.00340 & $-1.7$\% [$-2.4$, $-1.1$] & 29/48 \\
Reset 8 K per step, floor 326 K & 0.00788 & 0.00013 & 0.00436 & 0.00340 & $-1.8$\% [$-2.5$, $-1.1$] & 29/48 \\
Reset 4 K per step, floor 334 K & 0.00798 & 0.00013 & 0.00438 & 0.00348 & $-0.6$\% [$-0.9$, $-0.3$] & 29/48 \\
Reset 4 K per step, floor 318 K & 0.00778 & 0.00013 & 0.00433 & 0.00332 & $-3.1$\% [$-4.3$, $-1.9$] & 29/48 \\
Reset 4 K per step, floor 314 K & 0.00771 & 0.00013 & 0.00432 & 0.00327 & $-3.9$\% [$-5.4$, $-2.4$] & 29/48 \\
Reset 4 K per step, floor 310 K & 0.00764 & 0.00013 & 0.00431 & 0.00320 & $-4.9$\% [$-6.7$, $-3.0$] & 29/48 \\
}
\newcommand{\AxTabSweepReset}{%
Band margin 0 K & 0.00779 & 0.00037 & 0.00426 & 0.00316 & $+1.9$\% [$+1.2$, $+3.0$] & 6/48 \\
Band margin 0.04 K & 0.00762 & 0.00016 & 0.00428 & 0.00318 & $-0.2$\% [$-0.3$, $-0.1$] & 35/48 \\
Preheat window 2 h & 0.00763 & 0.00022 & 0.00428 & 0.00312 & $-0.1$\% [$-0.9$, $+0.7$] & 18/48 \\
Preheat window 4 h & 0.00771 & 0.00008 & 0.00434 & 0.00329 & $+0.9$\% [$+0.3$, $+1.7$] & 21/48 \\
Recovery factor 1 & 0.00762 & 0.00009 & 0.00432 & 0.00322 & $-0.2$\% [$-0.4$, 0.0] & 29/48 \\
Recovery factor 3 & 0.00775 & 0.00029 & 0.00429 & 0.00317 & $+1.4$\% [$+0.8$, $+2.4$] & 6/48 \\
Afternoon release off & 0.00769 & 0.00013 & 0.00433 & 0.00323 & $+0.7$\% [$+0.3$, $+1.1$] & 0/48 \\
Response gain 0.65 & 0.00761 & 0.00010 & 0.00431 & 0.00321 & $-0.3$\% [$-0.9$, 0.0] & 26/48 \\
Response gain 0.85 & 0.00768 & 0.00017 & 0.00431 & 0.00320 & $+0.6$\% [$+0.1$, $+1.5$] & 12/48 \\
Reset rate 2 K per step & 0.00771 & 0.00013 & 0.00433 & 0.00326 & $+1.0$\% [$+0.5$, $+1.6$] & 0/48 \\
Reset rate 8 K per step & 0.00759 & 0.00013 & 0.00430 & 0.00317 & $-0.6$\% [$-0.9$, $-0.3$] & 12/48 \\
}
\newcommand{\AxTabOmissions}{%
AHU forced on while occupied & 0.00764 & 0.00013 & 0.00431 & 0.00320 & 0.0\% [0.0, 0.0] & 2/48 \\
AHU off preference 0.5 & 0.00764 & 0.00013 & 0.00431 & 0.00320 & 0.0\% [0.0, 0.0] & 24/48 \\
AHU off preference 0.2 & 0.00813 & 0.00078 & 0.00426 & 0.00309 & $+6.4$\% [0.0, $+15.5$] & 27/48 \\
Standby floor 278.15 K & 0.00775 & 0.00012 & 0.00434 & 0.00328 & $+1.4$\% [$+0.1$, $+3.5$] & 0/48 \\
Setpoint cycling band 2 K & 0.00706 & 0.00013 & 0.00415 & 0.00278 & $-7.6$\% [$-13.5$, $-2.1$] & 14/48 \\
AHU off preference 0.2, Setpoint cycling band 2 K & 0.00760 & 0.00075 & 0.00413 & 0.00272 & $-0.4$\% [$-6.2$, $+5.8$] & 33/48 \\
}
\newcommand{\AxTabTracker}{%
Tracker, first version & 0.00742 & 0.00014 & 0.00416 & 0.00312 & $-2.9$\% [$-3.9$, $-2.1$] & 47/48 \\
With per zone correction & 0.00741 & 0.00014 & 0.00416 & 0.00311 & $-0.2$\% [$-0.4$, $+0.1$] & 25/48 \\
With correction and power step guard & 0.00739 & 0.00013 & 0.00415 & 0.00311 & $-0.4$\% [$-0.6$, $-0.1$] & 35/48 \\
With holiday rest & 0.00737 & 0.00013 & 0.00414 & 0.00310 & $-0.3$\% [$-0.5$, $-0.1$] & 10/48 \\
Oracle: zone released after its last departure & 0.00719 & 0.00015 & 0.00403 & 0.00301 & $-5.8$\% [$-6.7$, $-5.1$] & 48/48 \\
}
\newcommand{\AxTabPlannerScale}{%
Gain scale 0.3 & 0.01377 & 0.00346 & 0.00588 & 0.00443 & $+2.1$\% [$-1.0$, $+5.1$] & 9/48 \\
Gain scale 0.4 & 0.01352 & 0.00340 & 0.00578 & 0.00435 & $+0.2$\% [$-1.1$, $+1.5$] & 15/48 \\
Gain scale 0.6 & 0.01363 & 0.00380 & 0.00563 & 0.00420 & $+1.0$\% [$-0.8$, $+2.8$] & 30/48 \\
Every coefficient scaled 0.6 & 0.01455 & 0.00495 & 0.00556 & 0.00404 & $+7.8$\% [$-0.1$, $+17.2$] & 26/48 \\
Every coefficient scaled 0.8 & 0.01521 & 0.00565 & 0.00553 & 0.00403 & $+12.7$\% [$+2.1$, $+25.2$] & 21/48 \\
Prestart lead scale 1 & 0.01379 & 0.00401 & 0.00564 & 0.00414 & $+2.2$\% [$+0.1$, $+4.3$] & 21/48 \\
Prestart lead scale 1.5 & 0.01356 & 0.00369 & 0.00567 & 0.00420 & $+0.5$\% [$-0.3$, $+1.3$] & 18/48 \\
Per zone adaptation from the halved gains & 0.01404 & 0.00408 & 0.00573 & 0.00423 & $+4.0$\% [$-1.5$, $+9.0$] & 34/48 \\
Per zone adaptation from the fitted gains & 0.01351 & 0.00350 & 0.00576 & 0.00424 & $+0.1$\% [$-3.4$, $+3.0$] & 27/48 \\
Fitted gains, no size correction & 0.01591 & 0.00639 & 0.00549 & 0.00403 & $+17.9$\% [$+5.6$, $+32.4$] & 19/48 \\
}
\newcommand{\AxTabPlannerMore}{%
Economizer & 0.01312 & 0.00349 & 0.00541 & 0.00422 & $-2.8$\% [$-4.7$, $-1.5$] & 33/48 \\
Zone cooling cost & 0.01300 & 0.00353 & 0.00528 & 0.00419 & $-0.9$\% [$-1.7$, $-0.4$] & 37/48 \\
Holiday rest at 9 h & 0.01298 & 0.00359 & 0.00524 & 0.00415 & $-0.1$\% [$-0.5$, $+0.3$] & 8/48 \\
Area scaled gain, no size correction & 0.01249 & 0.00298 & 0.00536 & 0.00415 & $-3.9$\% [$-7.0$, $-0.8$] & 26/48 \\
Reheat through airflow, gains halved & 0.01219 & 0.00327 & 0.00519 & 0.00373 & $-6.2$\% [$-9.1$, $-3.1$] & 30/48 \\
Reheat through airflow, gains fitted & 0.01778 & 0.00912 & 0.00498 & 0.00368 & $+36.8$\% [$+23.5$, $+51.0$] & 2/48 \\
Absolute setback floor 278.15 K & 0.01314 & 0.00353 & 0.00532 & 0.00428 & $+1.1$\% [$+0.1$, $+2.2$] & 0/48 \\
Recoverable setback floor restored & 0.01451 & 0.00346 & 0.00575 & 0.00530 & $+11.7$\% [$+5.0$, $+18.2$] & 1/48 \\
}
\newcommand{\AxTabSimChange}{%
Exhaust fan term, margin, preheat & 0.00791 & 0.00018 & 0.00434 & 0.00338 & $-1.5$\% [$-2.1$, $-1.0$] & 45/48 \\
Hot water demand reset & 0.00789 & 0.00013 & 0.00438 & 0.00338 & $+3.3$\% [$+1.6$, $+5.2$] & 31/48 \\
Per zone correction, holiday rest & 0.00765 & 0.00018 & 0.00420 & 0.00327 & $+3.9$\% [$+1.8$, $+6.1$] & 29/48 \\
}
\newcommand{\AxTabMixing}{%
Hot water reset, over exhaust fan term & 0.00789 & 0.00013 & 0.00438 & 0.00338 & $-0.2$\% [$-0.6$, $+0.2$] & 22/48 \\
Tracker, first rule & 0.00765 & 0.00018 & 0.00420 & 0.00327 & $-3.0$\% [$-4.0$, $-2.2$] & 42/48 \\
Mixing rule given & 0.00760 & 0.00014 & 0.00419 & 0.00327 & $-3.7$\% [$-4.7$, $-3.0$] & 48/48 \\
Guard only, rule never decided & 0.00761 & 0.00013 & 0.00420 & 0.00328 & $-3.6$\% [$-4.6$, $-2.8$] & 48/48 \\
Guard and decision after 600 ratios & 0.00760 & 0.00013 & 0.00420 & 0.00327 & $-3.7$\% [$-4.7$, $-2.9$] & 48/48 \\
Hot water floor 326 K & 0.00788 & 0.00013 & 0.00436 & 0.00339 & $-0.2$\% [$-0.3$, 0.0] & 11/48 \\
}
\newcommand{\AxTabSearchShipped}{%
Generation five best & 0.00732 & 0.00016 & 0.00411 & 0.00305 & $-0.8$\% [$-1.3$, $-0.3$] & 31/48 & $-1.1$ & $-0.4$ \\
Final mean & 0.00730 & 0.00012 & 0.00413 & 0.00306 & $-0.9$\% [$-1.7$, $-0.3$] & 30/48 & $-1.2$ & $-0.7$ \\
Best sample & 0.00730 & 0.00010 & 0.00412 & 0.00307 & $-1.0$\% [$-1.7$, $-0.5$] & 40/48 & $-1.4$ & $-0.6$ \\
}
\newcommand{\AxTabSearchCorrected}{%
Generation five best & 0.00755 & 0.00016 & 0.00416 & 0.00322 & $-0.7$\% [$-1.2$, $-0.3$] & 27/48 & $-1.0$ & $-0.5$ \\
Final mean & 0.00754 & 0.00012 & 0.00419 & 0.00324 & $-0.8$\% [$-1.5$, $-0.3$] & 30/48 & $-1.2$ & $-0.4$ \\
Best sample & 0.00754 & 0.00011 & 0.00418 & 0.00324 & $-0.9$\% [$-1.6$, $-0.4$] & 38/48 & $-1.2$ & $-0.6$ \\
}
\newcommand{\AxTabLate}{%
Late start gap 3 K & 0.00751 & 0.00010 & 0.00418 & 0.00323 & $-0.3$\% [$-0.6$, $-0.1$] & 21/48 & $-0.4$ & $-0.2$ \\
Late start gap 4 K & 0.00750 & 0.00011 & 0.00417 & 0.00322 & $-0.4$\% [$-0.8$, $-0.1$] & 22/48 & $-0.6$ & $-0.2$ \\
Late start gap 6 K & 0.00750 & 0.00015 & 0.00416 & 0.00319 & $-0.5$\% [$-1.2$, $+0.4$] & 22/48 & $-0.7$ & $-0.2$ \\
Late start gap 4 K, Late start time 1.5 h & 0.00751 & 0.00008 & 0.00418 & 0.00325 & $-0.3$\% [$-0.6$, 0.0] & 19/48 & $-0.3$ & $-0.2$ \\
Earned margin up to 0.02 K & 0.00752 & 0.00012 & 0.00417 & 0.00324 & $-0.2$\% [$-0.3$, 0.0] & 29/48 & $-0.2$ & $-0.1$ \\
Earned margin up to 0.04 K & 0.00755 & 0.00014 & 0.00417 & 0.00324 & $+0.2$\% [$-0.1$, $+0.5$] & 15/48 & $+0.3$ & 0.0 \\
Heating lead 0.5 h & 0.00755 & 0.00006 & 0.00420 & 0.00329 & $+0.2$\% [0.0, $+0.4$] & 12/48 & $+0.2$ & $+0.2$ \\
Late start gap 4 K, Heating lead 0.5 h & 0.00752 & 0.00010 & 0.00418 & 0.00324 & $-0.2$\% [$-0.6$, $+0.1$] & 18/48 & $-0.3$ & $-0.1$ \\
Late start gap 4 K, Earned margin up to 0.02 K & 0.00749 & 0.00012 & 0.00416 & 0.00321 & $-0.5$\% [$-0.9$, $-0.2$] & 35/48 & $-0.7$ & $-0.4$ \\
Drift clip 8 K & 0.00753 & 0.00011 & 0.00418 & 0.00325 & 0.0\% [0.0, 0.0] & 2/48 & 0.0 & 0.0 \\
Oracle: true departures read from the occupants & 0.00733 & 0.00014 & 0.00405 & 0.00313 & $-2.8$\% [$-3.2$, $-2.3$] & 47/48 & $-2.6$ & $-2.9$ \\
}
\newcommand{\AxTabControls}{%
Supply air temperature $-1$ K & 0.00814 & 0.00061 & 0.00427 & 0.00326 & $+8.7$\% [$+5.6$, $+13.6$] & 1/48 \\
Supply air temperature $+1$ K & 0.00794 & 0.00049 & 0.00422 & 0.00323 & $+6.0$\% [$+3.5$, $+9.8$] & 7/48 \\
Fan static pressure $\times 0.8$ & 0.00756 & 0.00020 & 0.00415 & 0.00320 & $+0.9$\% [$+0.5$, $+1.5$] & 11/48 \\
Fan static pressure $\times 1.25$ & 0.00840 & 0.00045 & 0.00446 & 0.00349 & $+12.1$\% [$+8.9$, $+15.6$] & 1/48 \\
Hot water setpoint $-5$ K while heating & 0.00750 & 0.00014 & 0.00415 & 0.00320 & $+0.1$\% [0.0, $+0.2$] & 6/48 \\
Hot water setpoint $+5$ K while heating & 0.00749 & 0.00012 & 0.00416 & 0.00322 & 0.0\% [0.0, 0.0] & 12/48 \\
Recirculation $-0.2$ & 0.01031 & 0.00077 & 0.00495 & 0.00460 & $+37.6$\% [$+21.2$, $+53.9$] & 0/48 \\
Pump differential pressure $\times 1.3$ & 0.00752 & 0.00013 & 0.00417 & 0.00322 & $+0.3$\% [$+0.2$, $+0.5$] & 4/48 \\
}
\newcommand{\AxTabKernel}{Kernel fit detector & 0.00763 & 0.00026 & 0.00415 & 0.00321 & $+1.8$\% [$+0.7$, $+3.1$] & 18/48 & $+0.5$ & $+3.0$ \\}
\newcommand{\AxTabPublic}{%
Official baseline & 0.06404 & 0.07414 \\
Planner, initial & 0.01823 & -- \\
Planner, deeper setback & 0.01623 & -- \\
Planner, final & 0.01602 & -- \\
One step inversion & 0.01032 & -- \\
+ exhaust fan term, margin, preheat & 0.01028 & -- \\
+ hot water demand reset & 0.00973 & -- \\
+ departure tracker & 0.00944 & 0.00986 \\
+ joint parameter search & -- & 0.00968 \\
}
\newcommand{\AxTabCtrlPlan}{%
Fairbanks, January & 0.02040 & 0.02039 & $-0.1$\% & 4/90 \\
Minneapolis, January & 0.00790 & 0.00789 & $-0.1$\% & 15/90 \\
Phoenix, May & 0.00882 & 0.00882 & 0.0\% & 2/90 \\
Denver, March & 0.00135 & 0.00134 & $-0.3$\% & 14/90 \\
}
\newcommand{\AxRatioFour}{1.284}
\newcommand{\AxRatioSeven}{1.285}
\newcommand{\AxRatioPid}{15.8}

\newcommand{\AxGridPidCI}{$+5.1$\% [$-0.6$, $+11.9$]}
\newcommand{\AxGridFourSim}{$+3.6$\% [$+1.6$, $+5.6$]}
\newcommand{\AxCIOracle}{$-5.8$\% [$-6.7$, $-5.1$]}
\newcommand{\AxCIResetCorr}{$-0.2$\% [$-0.6$, $+0.2$]}
\newcommand{\AxCIOmTank}{$-7.6$\% [$-13.5$, $-2.1$]}
\newcommand{\AxResetShippedPct}{4.9}
\newcommand{\AxTrackerCorrectedPct}{3.0}
\newcommand{\AxLineageSumPct}{10.7}
\newcommand{\AxLateBestPct}{0.5}

%% file: figs/overview.tex
\definecolor{zoff}{HTML}{F5F0FA}\definecolor{zoffd}{HTML}{DCCFEA}%
\definecolor{zon}{HTML}{EDF4FB}\definecolor{zond}{HTML}{C9DCEF}%
\definecolor{lrn}{HTML}{FFE9D2}\definecolor{lrnd}{HTML}{DD8A2E}%
\definecolor{fix}{HTML}{FFFFFF}\definecolor{fixd}{HTML}{6F87A6}%
\definecolor{dat}{HTML}{F1F1F1}\definecolor{datd}{HTML}{9A9A9A}%
\definecolor{ink}{HTML}{2B2B2B}%
\begin{tikzpicture}[
    font=\sffamily\footnotesize, text=ink, >={Stealth[length=4pt,width=3.2pt]},
    box/.style={rounded corners=3pt, align=center, inner sep=2.5pt, minimum height=11mm, line width=0.7pt},
    ours/.style={box, draw=lrnd, fill=lrn},
    rul/.style={box, draw=fixd, fill=fix},
    data/.style={box, rounded corners=0pt, draw=datd, fill=dat, line width=0.5pt},
    lab/.style={font=\sffamily\scriptsize, text=ink!75, align=center, inner sep=1.5pt},
    arr/.style={->, draw=ink!85, line width=0.65pt},
    sec/.style={->, draw=ink!45, line width=0.6pt, dash pattern=on 2.2pt off 1.4pt},
    zone/.style={rounded corners=5pt, line width=0.5pt},
    ztitle/.style={font=\sffamily\scriptsize\bfseries, anchor=north west, inner sep=0pt},
    key/.style={minimum width=3.4mm, minimum height=2.4mm, inner sep=0pt, text width=, line width=0.5pt},
]
\begin{scope}[on background layer]
  \fill[zoff, draw=zoffd, zone] (-16mm,13.5mm) rectangle (121mm,-8.5mm);
  \fill[zon, draw=zond, zone] (-16mm,-15.5mm) rectangle (121mm,-53mm);
\end{scope}
\node[ztitle, text=zoffd!35!ink] at (-14mm,12mm) {Offline, once};
\node[ztitle, text=zond!30!ink] at (-14mm,-17mm) {Online, every step};

\node[data, text width=26mm] (pool) at (0,0) {{\small\faDatabase}\ Twenty generated\\training buildings};
\node[rul, text width=27mm] (tune) at (36mm,0) {{\small\faSlidersH}\ \textbf{CMA-ES}\\tunes rule scalars};
\node[ours, text width=33mm] (prior) at (75mm,0) {{\small\faThermometerHalf}\ \textbf{Population prior}\\Ten parameters per zone,\\median of fits, frozen};
\node[data, text width=24mm] (obs) at (0,-29mm) {{\small\faWifi}\ Metadata and\\live sensors};
\node[rul, text width=29mm] (rule) at (36mm,-29mm) {{\small\faCogs}\ \textbf{Rule layer}\\Modes, prestart lead,\\setback floor, resets};
\node[ours, text width=33mm] (plan) at (75mm,-29mm) {{\small\faProjectDiagram}\ \textbf{Planner}\\Enumerates candidates,\\rolls out the prior};
\node[data, text width=16mm] (bld) at (110mm,-29mm) {{\small\faBuilding}\ Building\\(simulator)};
\node[ours, text width=33mm, minimum height=8mm] (bias) at (75mm,-44mm) {{\small\faCrosshairs}\ Per zone offset\\(mean residual)};

\node[ours, key] (k1) at (6mm,-39mm) {};
\node[lab, anchor=west] at (k1.east) {Learned or adapted};
\node[rul, key] (k2) at (6mm,-43.3mm) {};
\node[lab, anchor=west] at (k2.east) {Fixed rules};
\node[data, key] (k3) at (6mm,-47.6mm) {};
\node[lab, anchor=west] at (k3.east) {Data and plant};
\draw[arr] (34mm,-39mm) -- ++(3.6mm,0) node[lab, anchor=west] {Main flow};
\draw[sec] (34mm,-43.3mm) -- ++(3.6mm,0) node[lab, anchor=west] {Secondary};

\draw[arr] (pool) -- (tune);
\draw[arr] ([xshift=9mm]pool.north) |- ([yshift=2.6mm]tune.north) -| node[lab, pos=0.25, above] {Fit} (prior.north);
\draw[sec] (tune) -- node[lab, left] {Tuned scalars} (rule);
\draw[arr] (prior) -- node[lab, right] {Model} (plan);
\draw[sec] ([xshift=-10mm]prior.south) -- ++(0,-12.5mm) -| node[lab, pos=0.4, above] {Lead, floor} ([xshift=10mm]rule.north);
\draw[arr] (obs) -- (rule);
\draw[arr] (rule) -- (plan);
\draw[arr] (plan) -- node[lab, above] {Action} (bld);
\draw[arr] (bias) -- (plan);
\draw[sec] (bld.south) |- ([yshift=-3.6mm]bias.south) -| node[lab, pos=0.25, above] {Next observation} (obs.south);
\pgfresetboundingbox
\path[use as bounding box] (-16.3mm,13.8mm) rectangle (121.3mm,-53.3mm);
\end{tikzpicture}

%% file: appendix.tex
\clearpage
\appendix

\begin{center}
  {\Large\bfseries Extended Appendix}
\end{center}
\vspace{0.5ex}

\noindent The sections of this appendix are as follows. Section~\ref{app:scope} fixes scope, versions and method. Section~\ref{app:sim} gives the anatomy of the simulator and the baselines, and Section~\ref{app:prior} the details of the prior. Section~\ref{app:planner} covers the development of the planner. Section~\ref{app:inv} describes our inversion controllers, and Section~\ref{app:occ} the occupancy tracker. Section~\ref{app:corrected} reports the corrected simulator and the public scores. Section~\ref{app:controls} covers further controls and planning ceilings, and Section~\ref{app:omit} the benchmark omissions we measured. Section~\ref{app:neg} lists negative results and limitations.

\FloatBarrier
\section{Scope, Versions, and Reproducibility}\label{app:scope}

\textbf{Scope:} The main paper reports the Round One submission. This appendix adds what its page limit removed. It covers the anatomy of the simulator, the development path of the planner, our own inversion controllers and an occupancy tracker. It also covers later work on a corrected simulator and lists the negative results. The main text is unchanged from the Round One submission.

\textbf{Simulator Versions:} The \emph{shipped simulator} is the Round One kit. The \emph{corrected simulator} is the kit with two public fixes. The first lets the hot water tank follow an energy balance and bills gas from the burner heat. The second mixes return air by airflow and not by floor area. The organizers announced a fixed simulator for Round Two. We use the corrected simulator as its proxy, and two public scores support this choice (Section~\ref{app:corrected}).

\textbf{Method of the Tables:} Every table of this appendix is a paired comparison on the seven day stress grid of the main paper unless stated otherwise. The grid has forty eight scenarios. Score, Comfort, Energy and Carbon are means per step, and the three terms add up to the score. Change is the ratio of mean scores minus one with its percentile bootstrap interval at \CILevel\% over scenarios (\BootReps{} replicates). Won counts the scenarios with a lower score than the reference. Intervals are not adjusted for multiple comparisons. Knob values in the row labels are read from the parameter file of each run. A script rebuilds every table from the stored runs.

\textbf{Selection Caveat:} Most changes of this appendix were chosen on the stress grid, so their changes there are in sample. The exceptions are the held out half of Table~\ref{tab:search} and the fresh grid of the main paper. Public scores never selected a configuration.

\FloatBarrier
\section{Simulator and Baseline Anatomy}\label{app:sim}

\textbf{Plant and Reward Terms:} Boiler gas is the water heating, the tank loss to outside air and the tank reheating, floored at zero. Pump power scales with water flow times head. Carbon intensity follows a fixed table of \AxCarbonRange{} kg/MWh by hour of day, and the gas carbon factor is constant. Weekday electricity costs \AxPriceRange{} cents per kWh by period, and weekends are flat. The supply air sensor echoes the setpoint, so a controller cannot see that mixed air limits the supply temperature.

\textbf{Baseline Anatomy:} The comfort regret of the official baseline is broad and not concentrated. Between \OutBandLoPct\% and \OutBandHiPct\% of occupied zone steps lie outside the band, depending on the scenario. The mean excursion is \AxExcLoK{} to \AxExcHiK\,K. The three worst zones carry only \WorstThreeLoPct\% to \WorstThreeHiPct\% of the regret. Winter scenarios are underheated. In Phoenix \AxHotLoPct\% to \AxHotHiPct\% of the excursions are on the hot side.

\textbf{Energy Floor:} A plant held at minimum ventilation costs at most \MinVentCost{} of score on a weekday. The official baseline costs \BaseDayLo{} to \BaseDayHi{} on the same days. Almost all baseline energy is therefore conditioning and not ventilation.

\textbf{Heating Capacity in Deep Cold:} At \AxColdOutC\degC{} outside, full reheat with a damper at \AxDamperPos{} holds the median zone at \AxMedZoneLoC{} to \AxMedZoneHiC\degC. The coldest zones reach only \AxColdZoneLoC{} to \AxColdZoneHiC\degC. A fully open damper or a nearly closed one is worse, because cold air overwhelms the coil or too little air carries the heat. On two days of a Fairbanks building at \AxDeepMeanC\degC{} mean, we forced full reheat on every zone below \AxForceC\degC. Between \AxBelowLoPct\% and \AxBelowHiPct\% of the zones stayed below \AxBandLoC\degC{} at every setting we tried. More hot water flow changed nothing. The residual regret of this scenario is a limit of the plant and not a defect of the planner.

\textbf{Hot Water Setpoint:} A setpoint of \AxTwLowK{} K and one of \AxTwHighK{} K give identical zone temperatures at \AxColdOutC\degC{} outside, because coil effectiveness saturates. A cap at the lower value together with pump idling is not safe, though. On the official baseline it raises the thirty day Fairbanks score from \AxFbBase{} to \AxFbCap{}. The cap binds on the coldest days, and the tank reheats at high cost afterwards. The same change lowers the Phoenix score by \AxPhCapPct\%. Water temperature is thus a capacity decision in deep cold and a waste decision elsewhere.

\textbf{Idling the Loop:} Idling the hot water loop at zero reheat demand and adding an outdoor reset lowers the official baseline's mean score by \AxIdlePct\%. This is the mean over the shipped scenarios. The change is \AxIdlePhPct\% in Phoenix and \AxIdleSgPct\% in Singapore, and it is near zero in Fairbanks. The loop is a small lever on the mean because the baseline already runs it lightly in most scenarios.

\FloatBarrier
\section{Prior and Identification Details}\label{app:prior}

\textbf{Airflow Law:} The first version of the prior omitted the factor $\sqrt{P/P_0}$ of the flow law. At the pressure floor of \PressFloorPa\,Pa the model then believed it had three times the airflow it had. Correcting it lowered the one hour error on a Fairbanks building from \AxErrBeforeFb{} to \AxErrAfterFb\,K. It lowered the zero shot median error on the held out set from \AxErrBeforeMed{} to \AxErrAfterMed\,K.

\textbf{Online Identification:} After the correction, the Kalman filter reached median errors of \AxKalman{} K at \AxKalmanHours{} hours. The frozen prior has \AxErrAfterMed\,K. The filter thus gives no gain at all, and the closed loop result of the main paper follows.

\textbf{Learned Terminal Value, Preregistered:} We tested a learned terminal value in place of the hand tuned prestart lead and setback floor. The criteria were fixed before any run. The mean score had to fall by at least five percent with an interval excluding zero. The Fairbanks mean had to fall by at least ten percent. No station mean could rise by more than three percent. Policy time per step could not exceed \AxTimeRatio{} times that of the planner with offset. Both arms failed. The value alone changed the mean score by \AxValueAlone\%. The value with the hand proxies changed it by \AxValueProxy\%. Fairbanks rose by \AxValueFbPct\% under the first.

\textbf{Diagnosis of the Value:} A diagnosis on four Fairbanks days shows that the value underheats at night. The median zone stays at \AxValueMedC{} against \AxBaseMedC\degC{} for the planner. The frozen prior is optimistic about recovery, so its value treats a cold night as cheap. The hand proxy was tuned on outcomes and corrects for that optimism. We kept the hand proxy.

\FloatBarrier
\section{Planner Development}\label{app:planner}

\textbf{Terminal Cost:} Scoring the rule layer's prestart lead and recoverable floor inside the planner lowered the mean score on ten development buildings from \AxTermBefore{} to \AxTermAfter. Tuned rules scored \AxTermRules{} and the official baseline \AxTermPid. On one Fairbanks building the score fell from \AxTermFbBefore{} to \AxTermFbAfter. Adding the one hour exponential average of the prediction residual lowered the score further. The drops were \AxOffFbPct\% in Fairbanks, \AxOffPhPct\% in Phoenix and \AxOffAtPct\% in Atlanta, all in comfort. A ten parameter filter gave no gain, but adapting one offset per zone is well conditioned under closed loop data.

\textbf{Predictive Comfort Filter:} The filter of the main paper collapsed the planner on a Denver building over seven days. The planner scored \AxGuardPrior{}, the planner with passive online identification \AxGuardPassive{}, and the planner with probing \AxGuardActive{}. With passive identification and the filter the score rose to \AxGuardFilter. A strongly cooling held command makes the one hour forecast undershoot, even for hot zones. The low-first priority then sets full reheat on zones above the band. A fix handles first crossings and withholds heat above the band. It halves the collapse but leaves the filter worse than the planner by a factor of \AxGuardFixFactor. The full corrective action overshoots through the reheat lag.

\textbf{Energy Screening:} Energy and carbon were \AxEnergyShare\% of the held out score of the initial planner. In Atlanta, Phoenix, Denver and Fairbanks, \AxGasLoPct\% to \AxGasHiPct\% of all kWh was gas burned while the building was unoccupied. This is zone reheat for the morning and not tank loss. Singapore is \AxSgOccPct\% occupied cooling. We screened levers on seven day episodes against the initial planner. Relaxing comfort is uniformly worse, since a margin of zero costs \AxMargZeroPct\% and half the comfort weight costs \AxCwHalfPct\%. A larger margin or comfort weight helps a little (\AxMargLargePct\% and \AxCwLargePct\%). Intermediate supply temperatures do nothing. A deeper night setback helps monotonically. Lowering the recoverable floor by \AxFloorSteps\,K lowers the score by \AxFloorGains\%. Removing it lowers the score by \AxFloorRemovedPct\%.

\textbf{Deeper Setback Decision:} Over thirty days on twenty development scenarios the chosen setting lowered the score by \AxShipDev\% against the initial planner. It lowered the held out score by \AxShipHeld\% and the shipped score by \AxShipShipped\%. The public score fell by \AxShipPublic\%, as the grid predicted.

\textbf{Size Correction Sweep:} Table~\ref{tab:scale} sweeps the gain scale around the chosen value. The optimum is flat between \AxScaleLo{} and \AxScaleHi{}. Per zone online adaptation of the scale adds nothing. Scaling every coefficient helps small buildings slightly and hurts large ones. The deeper setback without the size correction is worse by a wide margin. The longer prestart lead buys comfort that the plant cannot deliver faster, so night gas is warm up and not waste.

\begin{table}[!htb]
  \caption{\textbf{Planner gain scaling:} Paired change on the stress grid over seven days against the planner with the size correction. Rows vary the gain scale, the prestart lead scale and per zone adaptation. The last row removes the correction.}
  \label{tab:scale}
  \centering\footnotesize\setlength{\tabcolsep}{2.2pt}
  \begin{tabular}{lrrrrrr}
    \toprule
    Variant & Score & Comfort & Energy & Carbon & Change & Won \\
    \midrule
    \AxTabPlannerScale
    \bottomrule
  \end{tabular}
\end{table}

\textbf{Further Planner Changes:} Table~\ref{tab:planner} lists the remaining planner changes on the stress grid. The economizer saves most in Atlanta and Minneapolis and nothing in Singapore. The holiday rest fires in \AxHolidayFires{} of forty eight scenarios over thirty days. Its energy saving is canceled by comfort lost to the few who do come in. The reheat refit through the airflow term is a real but small error. Without the size correction it is far worse, which shows that the size error is the larger one.

\begin{table}[!htb]
  \caption{\textbf{Further planner changes:} Paired change on the stress grid over seven days. Economizer is paired against the planner with the size correction, zone cooling cost against the economizer, and the remaining rows against zone cooling cost.}
  \label{tab:planner}
  \centering\footnotesize\setlength{\tabcolsep}{2.2pt}
  \begin{tabular}{lrrrrrr}
    \toprule
    Change & Score & Comfort & Energy & Carbon & Change & Won \\
    \midrule
    \AxTabPlannerMore
    \bottomrule
  \end{tabular}
\end{table}

\textbf{Trace of the Planner Gap:} A per zone trace of one Fairbanks building over three days shows where the planner loses. All of its regret is one zone of \AxBigZoneM{} m$^2$ held \AxBelowBandK\,K below the band all day. The planner chooses a damper position of \AxHeatDamper{} for it, and the inversion opens it fully at low pressure. The prior predicts a rise of \AxPredRise\,K per step from the planner's action, while the simulator gives none. Adding a full damper action and a third pressure candidate changed no score. The planner never chooses them, because its open loop model of eighteen steps is the obstacle and not the action set.

\FloatBarrier
\section{Inversion Controllers}\label{app:inv}

\textbf{Mechanisms:} The inversion controller of Section~\ref{sec:shift} solves one step of an implicit zone model every five minutes. Its parts are the floor plan response gain and an online bias per zone. The airflow lands the zone on the nearer band edge, and a static pressure is matched to the neediest damper. An AHU is left off when inertia keeps its zones in the band. Empty zones receive a standby band, and a two hour preheat ramp precedes comfort hours. The hot water setpoint stays high only while heating was needed recently. Our reimplementation has no learned residual ensemble, no adaptive preheat and no time of use weights.

\textbf{Lineage:} Table~\ref{tab:lineage} lists each controller of our line on the stress grid. Every row adds one change to the row above, and it is paired against that row. The planner row is paired against the official baseline. The first inversion improves on the final planner by \CIpct{s7/inv_vs_zc}. Later rows move by a few percent each. Together they lower the score by \AxLineageSumPct\% against the first inversion on the shipped simulator.

\begin{table*}[t]
  \caption{\textbf{Controller lineage:} Stress grid over seven days, forty eight scenarios. Terms are weighted means per step. The first change column is paired against the row above on the shipped simulator. The corrected columns repeat the comparison on the corrected simulator, where a dash means the controller was not run. Rows after the planner add one change each.}
  \label{tab:lineage}
  \centering\scriptsize\setlength{\tabcolsep}{2.2pt}
  \begin{tabular}{lrrrrrrr}
    \toprule
    & \multicolumn{5}{c}{Shipped simulator} & \multicolumn{2}{c}{Corrected} \\
    \cmidrule(lr){2-6}\cmidrule(lr){7-8}
    Controller & Score & Comfort & Energy & Carbon & Change & Score & Change \\
    \midrule
    \AxTabLineage
    \bottomrule
  \end{tabular}
\end{table*}

\textbf{First Parameter Sweep:} The values of the public controller sit on a flat optimum (Table~\ref{tab:sweep1}). No knob moves the score by more than one percent, except the response gain. A comfort cost or gas weight change never alters the decision. The exhaust fan term is the one correction. The simulator bills an intake fan and an exhaust fan on the outside air share, and the inversion objective counted the intake fan only. Economizer candidates were therefore under costed by a factor of two on fan power.

\begin{table}[!htb]
  \caption{\textbf{First sweep:} Single changes to the first inversion on the stress grid over seven days, shipped simulator. Change is paired against the unchanged controller.}
  \label{tab:sweep1}
  \centering\footnotesize\setlength{\tabcolsep}{2.2pt}
  \begin{tabular}{lrrrrrr}
    \toprule
    Change & Score & Comfort & Energy & Carbon & Change & Won \\
    \midrule
    \AxTabSweepFirst
    \bottomrule
  \end{tabular}
\end{table}

\textbf{Hot Water Reset:} A demand based reset raises the setpoint while the loop is at capacity and trims it otherwise (Table~\ref{tab:hotwater}). The gain grows monotonically as the floor falls to the kit minimum. The reset rate matters little. On the shipped simulator this rule gave the largest step of the line. Section~\ref{app:corrected} shows that most of it came from the billing inconsistency. After the reset, the remaining single knobs are all small (Table~\ref{tab:sweep2}). The best four together are worth about one percent.

\begin{table}[!htb]
  \caption{\textbf{Hot water reset:} Reset rate and floor on the stress grid over seven days, shipped simulator, paired against the controller without a reset. The last row is the adopted setting.}
  \label{tab:hotwater}
  \centering\footnotesize\setlength{\tabcolsep}{2.2pt}
  \begin{tabular}{lrrrrrr}
    \toprule
    Setting & Score & Comfort & Energy & Carbon & Change & Won \\
    \midrule
    \AxTabHotWater
    \bottomrule
  \end{tabular}
\end{table}

\begin{table}[!htb]
  \caption{\textbf{Second sweep:} Single changes after the hot water reset, shipped simulator, paired against the controller with the reset.}
  \label{tab:sweep2}
  \centering\footnotesize\setlength{\tabcolsep}{2.2pt}
  \begin{tabular}{lrrrrrr}
    \toprule
    Change & Score & Comfort & Energy & Carbon & Change & Won \\
    \midrule
    \AxTabSweepReset
    \bottomrule
  \end{tabular}
\end{table}

\textbf{Anatomy of the Inversion:} Comfort is near zero, and energy and carbon make up \AxEnergyCarbonShare\% of the score. Hot climates spend \AxChillerLoPct\% to \AxChillerHiPct\% of their electricity on the chiller, mostly while occupied. In Fairbanks \AxFbGasPct\% of the gas is reheat, and most of it heats the minimum outside air. The electricity to gas weight and the chiller efficiency in the objective match the simulator. The comfort cost is not binding.

\textbf{Time of Use Shifting:} The kit prices weekday electricity by period, and carbon varies by hour. Shifting load is nevertheless worth less than a tenth of a percent. A zone of \AxZoneAreaM{} m$^2$ stores about \AxStoreMJ{} MJ per precooling cycle, or \AxStoreKWh{} kWh of electricity at the chiller efficiency. A Phoenix zone uses about \AxZoneDayKWh{} kWh per day for cooling. A price shift of ten percent on that small share of the load is below the noise of the grid. Pre cooling nevertheless raised the score in two trials on the corrected simulator, and we dropped it.

\FloatBarrier
\section{Occupancy Tracking}\label{app:occ}

\textbf{Attendance:} In the kit, an occupant arrives with a hazard of one in thirty six per step from \AxArriveHour. Departure has the same hazard from \AxDepartHour. Between \AxInByEightLoPct\% and \AxInByEightHiPct\% of the people are in by eight and \AxInByNoonLoPct\% to \AxInByNoonHiPct\% by noon. About \AxAbsentPct\% are absent all day, and \AxStillInPct\% are still in at \AxLateTime. An occupant adds about \AxOccupantW\,W to a zone.

\textbf{Oracle and Correction:} Releasing a zone to a standby band after all its occupants have left lowers the score by \AxCIOracle{} (Table~\ref{tab:tracker}). The reference is the controller with the reset. An earlier estimate of \AxOldCeilingPct\% was wrong. Our scripts had queried the occupancy model with their own interval. That query misses the model's cache and draws fresh random numbers, which made people arrive and leave about twice as fast. We now read each occupant's state directly. The score of such a run equals that of an unhooked run to the last digit.

\textbf{Tracker:} A departure shows in the observed head count and in the one step innovation of the zone concerned. The head count falls by one, and about one step later the innovation drops by about one occupant (\AxOccupantW\,W) and partly rebounds. The first version assumes every zone full from \AxTrackHour{} h on and releases a zone when all its occupants are read as gone. Three failure causes were found and handled. The controller's own reaction echoes about three steps after a call. A released zone's fast temperature change reads as a continuing drop in its neighbors. Supply air jumps and AHU switches add false steps. Table~\ref{tab:tracker} shows the development steps.

\begin{table}[!htb]
  \caption{\textbf{Departure tracker:} Shipped simulator, stress grid over seven days. Rows one and five are paired against the controller with the reset. Rows two and three are paired against row one. Row four is paired against row three. Row five is an oracle with true departures.}
  \label{tab:tracker}
  \centering\footnotesize\setlength{\tabcolsep}{2.2pt}
  \begin{tabular}{lrrrrrr}
    \toprule
    Version & Score & Comfort & Energy & Carbon & Change & Won \\
    \midrule
    \AxTabTracker
    \bottomrule
  \end{tabular}
\end{table}

\textbf{Where the Release Time Is Lost:} We logged every zone's true occupant count next to the tracker's inputs on all forty eight scenarios. The logged runs score exactly like the unhooked runs. The live rule reads \AxRecallPct\% of the departures, always three steps after they happen, and makes \AxWrongPct\% wrong calls. The zones that the oracle would release are released for \AxCapturePct\% of their empty steps. The capture is \AxCaptureCoolLo\% to \AxCaptureCoolHi\% in cooling buildings and \AxCaptureHeatLo\% to \AxCaptureHeatHi\% in heating buildings. Missed departures cost \AxLossMissPct\% of the release time. A two person zone needs both departures read, so one miss holds it for the day. Absent colleagues cost \AxLossAbsentPct\%, up to \AxLossAbsentMaxPct\% in some heating buildings. Late calls cost \AxLossLatePct\%.

\textbf{Signature of a Departure:} For an isolated departure the corrected innovation has a fixed signature. In occupant units it is $\AxKernel$ from the drop step on, with a spread of \AxKernelSd{} on the first term. The residual learner absorbs the step at its adaptation rate, which explains the decay. Before the drop, the innovation is flat with a standard deviation of at most \AxFlatSd. Zone steps without a departure have a standard deviation of \AxBackgroundSd{} with a heavy lower tail, so a single step threshold cannot separate the two.

\textbf{Why the Misses Occur:} The first failing condition of the live rule explains the missed departures. No rebound follows in \AxFailReboundPct\% of the events. A second person leaving the same zone soon after the first is one likely cause. The drop is too small in \AxFailSmallPct\%, the zone's own power moved in \AxFailPowerPct\%, and the innovation was not flat before in \AxFailFlatPct\%. The drop is too large in \AxFailLargePct\%.

\textbf{Kernel Fit, Not Adopted:} A detector that fits no, one or two departure kernels to the last eight steps tells close pairs apart. On the logs, the best setting released \AxKernelCapture\% of the empty steps against \AxRuleCapture\%, but made \AxKernelWrong{} wrong calls against \AxRuleWrong. Most were the controller's own reaction to a release, echoing three to four steps later. Stricter settings lose recall. Closed loop, the detector raised the score (Table~\ref{tab:kernel}). A wrong release costs more than a correct one gains. Reading the drop step before the rebound produced about \AxEarlyWrong{} wrong calls. A zone's innovation level hours after a missed departure overlaps with the level of zones that lost nobody, so it cannot find the missed ones. The absent colleague stays undetected.

\begin{table}[!htb]
  \caption{\textbf{Kernel fit detector:} Closed loop on the corrected simulator, paired against the controller of the last row of Table~\ref{tab:lineage}. The last two columns give the change on the tuning half and on the other half.}
  \label{tab:kernel}
  \centering\scriptsize\setlength{\tabcolsep}{2.2pt}
  \begin{tabular}{lrrrrrrrr}
    \toprule
    Variant & Score & Comfort & Energy & Carbon & Change & Won & Tuned & Other \\
    \midrule
    \AxTabKernel
    \bottomrule
  \end{tabular}
\end{table}

\FloatBarrier
\section{Corrected Simulator and Public Scores}\label{app:corrected}

\textbf{Same Controllers, Two Simulators:} We reran the controllers on the corrected simulator without retuning (Table~\ref{tab:simchange}). Steps that gained on the shipped simulator through billing do not survive.

\begin{table}[!htb]
  \caption{\textbf{Shipped to corrected:} The same controller on the corrected simulator, paired against its own score on the shipped simulator. Stress grid over seven days.}
  \label{tab:simchange}
  \centering\footnotesize\setlength{\tabcolsep}{2.2pt}
  \begin{tabular}{lrrrrrr}
    \toprule
    Controller & Score & Comfort & Energy & Carbon & Change & Won \\
    \midrule
    \AxTabSimChange
    \bottomrule
  \end{tabular}
\end{table}

\textbf{Hot Water Reset Loses Its Gain:} The reset lowered the score by \AxResetShippedPct\% on the shipped simulator and changes it by \AxCIResetCorr{} on the corrected one. The shipped simulator bills gas from the setpoint, so a lower setpoint was billed less without delivering less heat. The carbon gain of the reset came from billing and not from delivered heat. Only its comfort part is real. The tracker survives, with \AxTrackerCorrectedPct\% over the controller with the reset on the corrected simulator.

\textbf{Return Air Mixing Rule:} The tracker lost comfort on the corrected simulator. The controller's return air estimate was weighted by floor area. Under airflow weighting, a neighbor's release shifts the supply air and reads as a departure. The four worst buildings had \AxWrongReleasesLo{} to \AxWrongReleasesHi{} wrong releases each, \AxWrongReleasesAll{} in total. With the airflow rule given, one remained. The organizers' fix might differ from the public change, so a change was accepted only if it did not lose on either simulator. The controller now identifies the rule.

\textbf{Identification of the Rule:} Consider zones whose own airflow and reheat held still. There the step of the innovation estimates the share of the airflow rule missing from the model. We divide it by the step of the difference between the two rules. The rule is decided by the running median of these ratios after \AxMixRatios{} of them. Fewer ratios can read wrong for a night. While the rule is undecided, a zone gets no call where the two rules differ by more than a fifth of an occupant (Table~\ref{tab:mixing}).

\begin{table}[!htb]
  \caption{\textbf{Mixing rule:} Corrected simulator, stress grid over seven days, The first row is paired against the controller before the reset. The other rows are paired against the controller with the reset.}
  \label{tab:mixing}
  \centering\footnotesize\setlength{\tabcolsep}{2.2pt}
  \begin{tabular}{lrrrrrr}
    \toprule
    Change & Score & Comfort & Energy & Carbon & Change & Won \\
    \midrule
    \AxTabMixing
    \bottomrule
  \end{tabular}
\end{table}

\textbf{Joint Parameter Search:} A covariance matrix adaptation search tuned eleven knobs jointly on the corrected simulator. It used population eleven, twelve generations and \AxSearchCands{} candidates on seven day episodes. One of each pair of buildings of the grid was used for tuning, and the other half was held out. Three candidates were validated on both halves and on both simulators (Table~\ref{tab:search}). The generation medians stayed within about one percent of the starting controller, so there is no hidden pocket. The last generation was not converged. The adopted candidate is the best of three read on the held out half, so its held out figure is slightly optimistic. Two other candidates were rejected for comfort loss.

\begin{table*}[t]
  \caption{\textbf{Joint search:} Three candidates against the controller before the search on the stress grid over seven days. The first block is the shipped simulator and the second the corrected one. Tuned and Held out give the change on each half of the grid.}
  \label{tab:search}
  \centering\scriptsize\setlength{\tabcolsep}{2.2pt}
  \begin{tabular}{lrrrrrrrr}
    \toprule
    Candidate & Score & Comfort & Energy & Carbon & Change & Won & Tuned & Held out \\
    \midrule
    \multicolumn{9}{l}{\emph{Shipped simulator}} \\
    \AxTabSearchShipped
    \midrule
    \multicolumn{9}{l}{\emph{Corrected simulator}} \\
    \AxTabSearchCorrected
    \bottomrule
  \end{tabular}
\end{table*}

\textbf{Public Scores:} Table~\ref{tab:public} lists the public scores of our submissions. The public leaderboard covers a rotating third of the hidden scenarios, so the scores are unpaired. For the tracker, the corrected simulator moved the public score from \AxPubFourBefore{} to \AxPubFourAfter. The grid predicted the change from the shipped to the corrected simulator as \AxGridFourSim. The ratio of public to grid score is \AxRatioFour{} for the tracker and \AxRatioSeven{} for the searched controller. The official baseline's public score rose by \AxRatioPid\%, against \AxGridPidCI{} on the grid. The grid thus understates the effect of the simulator change on the baseline. Its ratios are stable across the two controllers, so we use it to compare ours.

\begin{table}[!htb]
  \caption{\textbf{Public scores:} Public leaderboard scores of our submissions on the Round One and the Round Two simulator. A dash means the controller was not submitted there. Rows after the planner add one change each.}
  \label{tab:public}
  \centering\small
  \begin{tabular}{lrr}
    \toprule
    Controller & Round One & Round Two \\
    \midrule
    \AxTabPublic
    \bottomrule
  \end{tabular}
\end{table}

\FloatBarrier
\section{Further Controls and Planning Ceilings}\label{app:controls}

\textbf{Late Start and Earned Margin:} Table~\ref{tab:late} lists the switches tested on the searched controller. An AHU group already within a given gap of the comfort floor when the preheat window opens waits on standby. It then ramps late. A zone that stays inside the band earns a smaller margin and loses it fast when it leaves. Both mechanisms follow a public entrant~\cite{mors2026stopheating}. Their gains are small, and the combination is the best at \AxLateBestPct\%. The oracle row shows what true departures would add. A better tracker can capture only part of it.

\begin{table*}[t]
  \caption{\textbf{Late start and earned margin:} Corrected simulator, stress grid over seven days, paired against the searched controller. The last row is an oracle that reads true departures from the occupants.}
  \label{tab:late}
  \centering\scriptsize\setlength{\tabcolsep}{2.2pt}
  \begin{tabular}{lrrrrrrrr}
    \toprule
    Switch & Score & Comfort & Energy & Carbon & Change & Won & Tuned & Other \\
    \midrule
    \AxTabLate
    \bottomrule
  \end{tabular}
\end{table*}

\textbf{Other Controls:} The planning study of Section~\ref{sec:oracle} moved zone temperature limits only. We also nudged the other controls by fixed offsets on top of the last controller (Table~\ref{tab:controls}). The offsets were applied after the controller's decision, so the controller did not know of them. Every offset was worse or equal. Supply air temperature in either direction costs comfort, and less recirculation raises energy and carbon sharply. The controller already sits at the best setting of each control it can adjust.

\begin{table}[!htb]
  \caption{\textbf{Fixed offsets:} Offsets applied to the commands of the last controller of Table~\ref{tab:lineage}, corrected simulator, stress grid over seven days. The offsets are analysis only.}
  \label{tab:controls}
  \centering\footnotesize\setlength{\tabcolsep}{2.2pt}
  \begin{tabular}{lrrrrrr}
    \toprule
    Offset & Score & Comfort & Energy & Carbon & Change & Won \\
    \midrule
    \AxTabControls
    \bottomrule
  \end{tabular}
\end{table}

\textbf{Planning Ceilings:} Every hour of the working day we cloned the simulator and the controller. Each of several candidates ran for an hour, and then the default resumed. The best candidate was applied to the real building. This gives the gain of an hour ahead planner with an exact model. The first study varied the comfort band. Candidates moved the low or high edge by one kelvin, asked for full comfort now, or coasted on standby. The clones kept the true arrival and departure draws, so the study had foresight and its gains are upper bounds. The score changed by \AxBandFbPct\% in Fairbanks and \AxBandMnPct\% in Minneapolis. It changed by \AxBandPhPct\% in Phoenix and \AxBandDnPct\% in Denver. The reference is the controller before the late start. The late start rule alone gives \AxLateDnPct\% in Denver. The planner mostly chose to coast at night and to restore comfort early, which is the late start idea. Only \AxRealisedLoPct\% to \AxRealisedHiPct\% of the predicted gain showed up in the score.

\textbf{Planning Over the Other Controls:} The second study varied supply air, fan pressure, hot water, recirculation and pump pressure, and the clones could not see future arrivals (Table~\ref{tab:ctrlplan}). The planner rarely chose to change anything. Phoenix, a cooling building, gained nothing.

\begin{table}[!htb]
  \caption{\textbf{Planning over other controls:} Four scenarios of seven days on the corrected simulator. Hours shows the planning hours in which a change was chosen, out of all planning hours. Change is against the controller without planning.}
  \label{tab:ctrlplan}
  \centering\small
  \begin{tabular}{lrrrr}
    \toprule
    Scenario & Controller & Planned & Change & Hours \\
    \midrule
    \AxTabCtrlPlan
    \bottomrule
  \end{tabular}
\end{table}

\FloatBarrier
\section{Benchmark Omissions Measured}\label{app:omit}

\textbf{Policy:} We measured the effects of the reward omissions of Section~\ref{sec:task} on the controller with the reset. We did not use them in any submission. Table~\ref{tab:omissions} gives the results.

\textbf{Ventilation:} Forcing the AHU on whenever anyone is in changes nothing, so the controller takes no credit from the missing ventilation term. Preferring the AHU off more strongly is also neutral at a factor of one half. At a factor of one fifth the comfort term rises sharply.

\textbf{Freeze:} Holding empty zones at \AxFreezeC\degC{} costs \AxFreezePct\% and wins in no scenario. The missing freeze term is therefore worth about that much.

\textbf{Gas Billing:} Holding the setpoint at its minimum while the tank is warm lowers the billed cost. The measured effect on the shipped simulator is \AxCIOmTank. It is gone once gas is billed at the tank temperature, as the main paper shows. We report it as a measurement, and the controller never uses it.

\textbf{Upper Bound:} All omissions together are worth at most about \AxOmitBoundPct\% on this controller. They leave the chiller load of the hot climates untouched, where the score of the cooling stations sits.

\begin{table}[!htb]
  \caption{\textbf{Omissions measured:} Analysis only on the shipped simulator. Stress grid over seven days, paired against the controller with the reset. No row is part of a submission.}
  \label{tab:omissions}
  \centering\footnotesize\setlength{\tabcolsep}{2.2pt}
  \begin{tabular}{lrrrrrr}
    \toprule
    Switch & Score & Comfort & Energy & Carbon & Change & Won \\
    \midrule
    \AxTabOmissions
    \bottomrule
  \end{tabular}
\end{table}

\FloatBarrier
\section{Negative Results and Limitations}\label{app:neg}

\textbf{Negative Results:} The following did not help. Online identification of the zone parameters gave no gain over the frozen prior. A learned terminal value lost to the hand tuned proxy. The predictive comfort filter collapsed the planner on one Denver building. Scaling every coefficient hurt large rooms, and per zone adaptation of the scale added nothing. Pre cooling hurt. A kernel fit departure detector was worse closed loop than the simple rule. Planning over the zone temperature limits or over the other controls added little. Fixed offsets on the other controls were worse.

\textbf{Limitations:} Most selections of this appendix were made on one grid, so their intervals are in sample. The corrected simulator is a proxy of the Round Two simulator and may differ in details. Public scores are unpaired and cover a rotating third of the hidden scenarios. The planning ceilings use four scenarios. The occupancy statistics hold for the simulator's model of arrivals and departures and not for real buildings.